\documentclass[10pt,twocolumn,letterpaper]{article}

\usepackage[pagenumbers]{wacv}      

\newcommand{\ours}{SCORP\xspace}
\newcommand{\Ours}{SCORP\xspace}

\colorlet{colorFst}{red!20}         
\colorlet{colorSnd}{orange!30}      
\colorlet{colorTrd}{yellow!30}      
\colorlet{colorLow}{darkgray!30}    

\newcommand{\best}{\cellcolor{colorFst}}
\newcommand{\sbest}{\cellcolor{colorSnd}}
\newcommand{\tbest}{\cellcolor{colorTrd}}
\newcommand{\false}{\cellcolor{colorLow}}

\usepackage{nicefrac}       
\usepackage{multirow}       
\usepackage{array}          
\usepackage[table]{xcolor}
\usepackage{adjustbox}
\usepackage{diagbox}

\usepackage{mathrsfs}

\usepackage{algorithm}
\usepackage{algorithmic}

\definecolor{wacvblue}{rgb}{0.21,0.49,0.74}
\usepackage[pagebackref,breaklinks,colorlinks,allcolors=wacvblue]{hyperref}

\def\wacvPaperID{2015} 
\def\confName{WACV}
\def\confYear{2026}

\title{SCORP: Scene-Consistent Object Refinement via Proxy Generation and Tuning}

\author{%
  Ziwei Chen\textsuperscript{1}\footnotemark[1] \quad
  Ziling Liu\textsuperscript{1}\footnotemark[1] \quad
  Zitong Huang\textsuperscript{1}\footnotemark[1] \quad
  Mingqi Gao\textsuperscript{1,2} \quad
  Feng Zheng\textsuperscript{1,3}\footnotemark[2] \\
  {\tt\small
    \textsuperscript{1}Southern University of Science and Technology
    \textsuperscript{2}University of Sheffield
    \textsuperscript{3}Spatialtemporal AI
  } \\
  {\tt\small\{chenzw2023, liuzl2024, huangzt2024\}@sustech.edu.cn,} \\ 
  {\tt\small mingqi.gao@outlook.com, f.zheng@ieee.org}
}

\begin{document}

\maketitle

\footnotetext[1]{Equal contribution} 
\footnotetext[2]{Corresponding authors} 

\begin{abstract}

Viewpoint missing of objects is common in scene reconstruction, as camera paths typically prioritize capturing the overall scene structure rather than individual objects. This makes it highly challenging to achieve high-fidelity object-level modeling while maintaining accurate scene-level representation.
Addressing this issue is critical for advancing downstream tasks requiring high-fidelity object reconstruction. In this paper, we introduce \underline{S}cene-\underline{C}onsistent \underline{O}bject \underline{R}efinement via \underline{P}roxy Generation and Tuning (\ours), a novel 3D enhancement framework that leverages 3D generative priors to recover fine-grained object geometry and appearance under missing views. 
Starting with proxy generation by substituting degraded objects using a 3D generation model, \ours then progressively refines geometry and texture by aligning each proxy to its degraded counterpart in 7-DoF pose, followed by correcting spatial and appearance inconsistencies through registration-constrained enhancement. 
This two-stage proxy tuning ensures the high-fidelity geometry and appearance of the original object in unseen views while maintaining consistency in spatial positioning, observed geometry, and appearance.
Across challenging benchmarks, \ours achieves consistent gains over recent state-of-the-art baselines on both novel view synthesis and geometry completion tasks. 
\ours is available at \url{https://github.com/PolySummit/SCORP}.

\end{abstract}    
\section{Introduction}
\label{sec:intro}

Multi-view 3D reconstruction is a core computer vision task, vital for autonomous vehicles, AR, and embodied AI. 
Recent advances in differentiable 3D representations, such as NeRF~\cite{nerf} and 3D Gaussian Splatting (3DGS)~\cite{3dgs}, have reshaped the novel view synthesis (NVS) and geometric reconstruction, making them unprecedentedly efficient and accessible to photorealistic NVS and accurate surface reconstruction solely from images. 
However, these methods rely heavily on dense input views for sufficient coverage. 
In real-world captures, camera trajectories typically prioritize large-scale, strongly structured areas of the overall scene~\cite{nbv,scannet,replica}, leaving objects with sparse and uneven views (\ie, they occupy few pixels, are often occluded or not captured), which limits object visibility.
Consequently, while scene-level reconstruction appears globally consistent, it remains suboptimal for individual objects.
Such degradation hinders downstream applications (\eg, simulation and interaction)~\cite{mahler2017dex,iwase2025zerograsp} that depend on high-quality object modeling. Although scanning objects individually can yield high-fidelity models, it is labor- and hardware-intensive~\cite{BigBird,GSO}, and compromises spatial realism by discarding scene-consistent information, including pose and shape of the original object and environment impact on the object, impairing downstream usability. 
Thus, a significant challenge is to refine object-level details in reconstructed scenes while preserving the integrity of the original context.

\begin{figure*}[!ht]
    \vspace{-1em}
    \centering
    \includegraphics[width=1\textwidth]{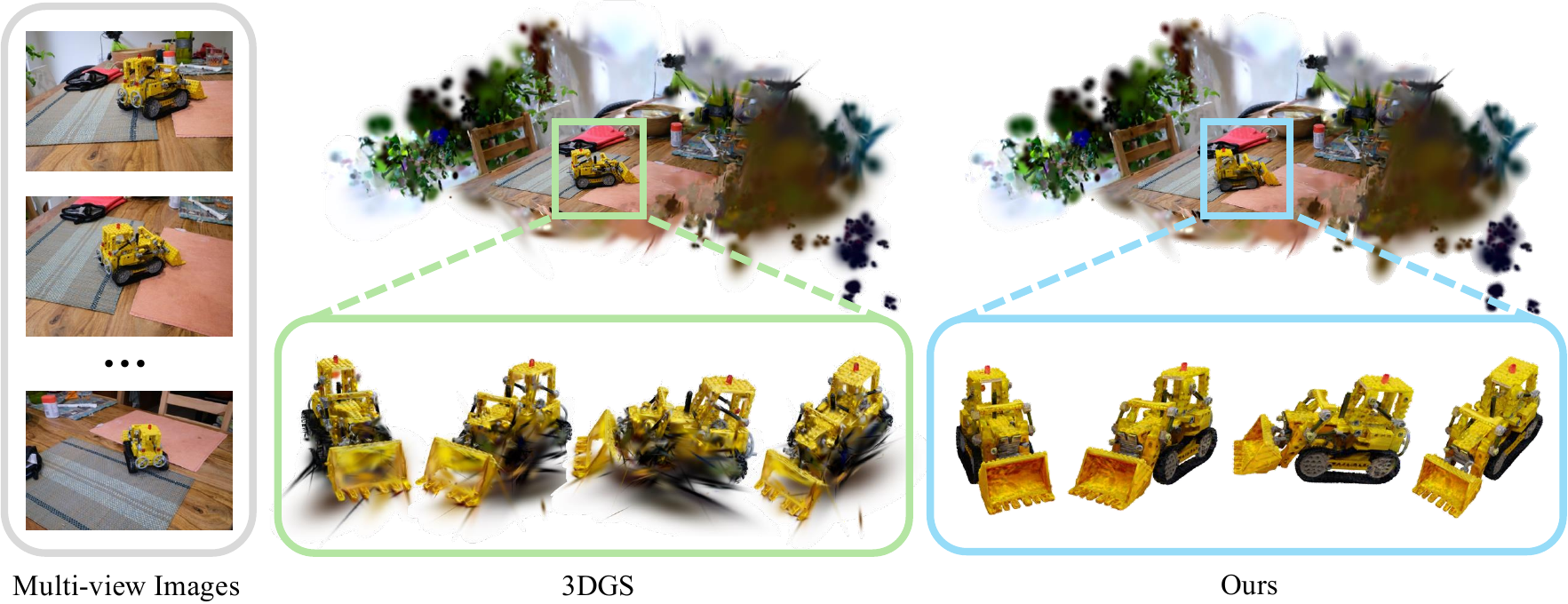}
    \caption{
    Left: limited-view images. Middle: 3DGS~\cite{3dgs} yields satisfactory scene-level reconstructions but poor object-level fidelity. Right: \Ours refines the object under scene constraints, improving texture and geometry while keeping the scene consistency.
}
    \label{fig:example}
    \vspace{-1em}
\end{figure*}
Intuitively, existing point cloud completion and 3D enhancement methods might serve as solutions, yet each comes with inherent limitations. Although point cloud completion methods~\cite{Pointr, AdaPointr,pcn,ComPC,RFNet, qi2017pointnet, xiang2021snowflakenet, yan2025symmcompletion} can recover the geometric structure of partial objects, they cannot reconstruct their appearance details. Existing 3D enhancement methods~\cite{diffusionerf, 3dgs-enhancer, wu2025difix3d+, genfusion} can simultaneously refine object geometry and appearance, which leverage 2D diffusion priors to interpolate novel views between input images and distill the enhanced results back into 3D representations via differentiable rendering. Although these methods enhance reconstruction quality, they remain fundamentally limited in completing invisible object parts due to the lack of explicit 3D understanding in 2D diffusion models.

Motivated by these gaps, we propose \ours, a 3D enhancement framework that leverages 3D generative capabilities to refine degraded objects within a reconstructed scene, simultaneously completing unseen regions and improving overall fidelity while preserving consistency (Fig.~\ref{fig:example}).
At a high level, our approach employs a grounding model with text prompts to obtain 2D object masks, aggregates them via a gradient voting strategy to segment degraded objects in 3D, then synthesizes proxy objects with a 3D generative model guided by a view selection scheme that balances completeness and diversity, and finally aligns proxy objects to the originals through a novel two-stage tuning process.
In the proposed two-stage proxy tuning, we first introduce an object registration module that combines Iterative Closest Point (ICP)~\cite{besl1992method} with a matching-based adjustment to align the proxy objects' seven-degree-of-freedom (7-DoF) poses with the original degraded objects. To address scale discrepancies and texture inconsistencies of proxy objects while preserving the already established alignment, we propose two refinement techniques: Scale-Undistorted Shape Refinement and Pose-Constrained Appearance Refinement. The former independently adjusts the proxy object's shape along the three coordinate axes to ensure shape consistency, while the latter refines the object’s appearance to align with visible views. These procedures enable seamless integration of proxy objects into the original scene and enhance the quality of objects in the scene without altering the scene structure.
Our contributions are summarized as follows:
\begin{itemize}
    \item We introduce a novel 3D object enhancement framework that leverages native 3D generative capabilities, which bridges the gap between scene-level and object-level reconstruction under limited views.  
    \item We propose a progressive registration and refinement strategy to ensure alignment between proxy objects and the original degraded counterparts, through initial pose registration and subsequent fine-grained adjustments in geometric scale and appearance.
    \item Across various limited-view regimes, \ours surpasses state-of-the-art approaches on object-level NVS and geometry completion, demonstrating superior performance in handling complex details of objects, yielding better detail fidelity and global scene consistency.
\end{itemize}

\section{Related Work}
\label{sec:relate}

\paragraph{Point Cloud Completion.}

Reconstructing full 3D shapes from partial scans has progressed rapidly. Early voxel or distance-field representations trained with 3D CNNs achieved robustness but incurred heavy memory and compute costs~\cite{3dShapeNets,GRNet,VoxNet,DeepSDF,PCC_Survey}. Operating directly on raw, unstructured point sets, PointNet and successors introduced permutation-invariant encoders with shared multilayer perceptrons and global pooling, establishing the basis for point cloud learning~\cite{qi2017pointnet,qi2017pointnetplusplus,qian2022pointnext,qi2017frustum}. PCN combined an encoder-decoder with FoldingNet to produce coarse yet plausible completions~\cite{pcn,yang2018foldingnet}. Later methods improve fine geometry by exploiting structural priors such as recursive point growth, skeleton guidance, or patch seeds~\cite{xiang2021snowflakenet,tang2022lake,zhou2022seedformer}, by using transformer-based set translation with adaptive queries~\cite{Pointr,yu2022adapointr}, and by semantic-aware completion with a semantic-prototype variational transformer~\cite{spovt}. 
Zero-shot methods such as SDS-Complete exploit pretrained priors to hallucinate missing regions~\cite{kasten2023point}, and ComPC applies 3DGS–inspired expansion from a single view to achieve completion without human input~\cite{ComPC}.
Despite strong geometric plausibility, most methods ignore appearance and color, limiting photorealistic scene integration.

\paragraph{Generative 3D Enhancement.} 

Generative priors help mitigate degradation in radiance-field or Gaussian representations under sparse or sub-optimal data. DiffusioNeRF injects a learned RGB-D prior via denoising diffusion to guide NeRF training~\cite{diffusionerf}. 3DGS-Enhancer and GenFusion employ video diffusion to refine novel views and distill them back to the underlying 3D representation~\cite{3dgs-enhancer,genfusion}. DIFIX3D+ and 3DENHANCER adapt 2D diffusion with reference views to improve multi-view quality and consistency~\cite{wu2025difix3d+,luo20243denhancer}. DP-RECON optimizes neural fields under novel views using score distillation from text prompts, yet text ambiguity often yields inconsistent appearance~\cite{dprecon,sds}. Although these approaches enhance images, 2D diffusion lacks explicit 3D understanding, which challenges spatial consistency and completion of unseen parts. Nerfbusters trains a local 3D diffusion prior to regularize NeRF geometry, but the query window is small and texture cannot be hallucinated~\cite{nerfbusters}. To address these issues, we propose the first 3D object enhancement framework based on a native 3D generative model \cite{trellis}.
\section{Preliminaries}

\paragraph{3D Gaussian Splatting.} We densely represent the scene using 3DGS~\cite{3dgs}, an explicit 3D representation that supports differentiable and efficient rendering. It represents the scenes with 3D Gaussian primitives which can be parameterized by mean position $\mu \in \mathbb{R}^3$,  opacity  $\alpha \in \mathbb{R}$, covariance $\Sigma \in \mathbb{R}^{3\times3}$ and spherical harmonics (SH).  The distribution of the $i$th primitive can be represented as: 
\begin{equation}
    g_i(x) = \alpha_i \cdot e^{-\frac{1}{2}(x-\mu_i)^\top\Sigma_i^{-1}(x-\mu_i)}.
\end{equation}
For rendering, 3D Gaussians are projected onto the 2D image plane using a camera extrinsic with a local affine approximation of the projective mapping. The rendered color and depth of pixel $x_p$ can be obtained via $\alpha$-blending over the $N$ overlapping projected 2D Gaussian distribution $\widetilde{g}$:
\begin{align}
    C(x_p) = \sum_{i \in N}{c_i\widetilde{g}_i\prod_{j=1}^{i-1}(1-\widetilde{g}_j)},\label{eq:gs_rgb}\\
    D(x_p) = \sum_{i \in N}{d_i\widetilde{g}_i\prod_{j=1}^{i-1}(1-\widetilde{g}_j)},\label{eq:gs_depth}
\end{align}
where $c_i$ is the color represented via SH and $d_i$ is the depth of the $i$th projected 2D Gaussian primitive. Guided by the photometric loss between the rendered image and the ground truth, the parameters of the 3D Gaussian primitives are optimized via gradient descent.

\begin{figure*}[t]
    \vspace{-1em}
    \centering
    \includegraphics[width=\textwidth]{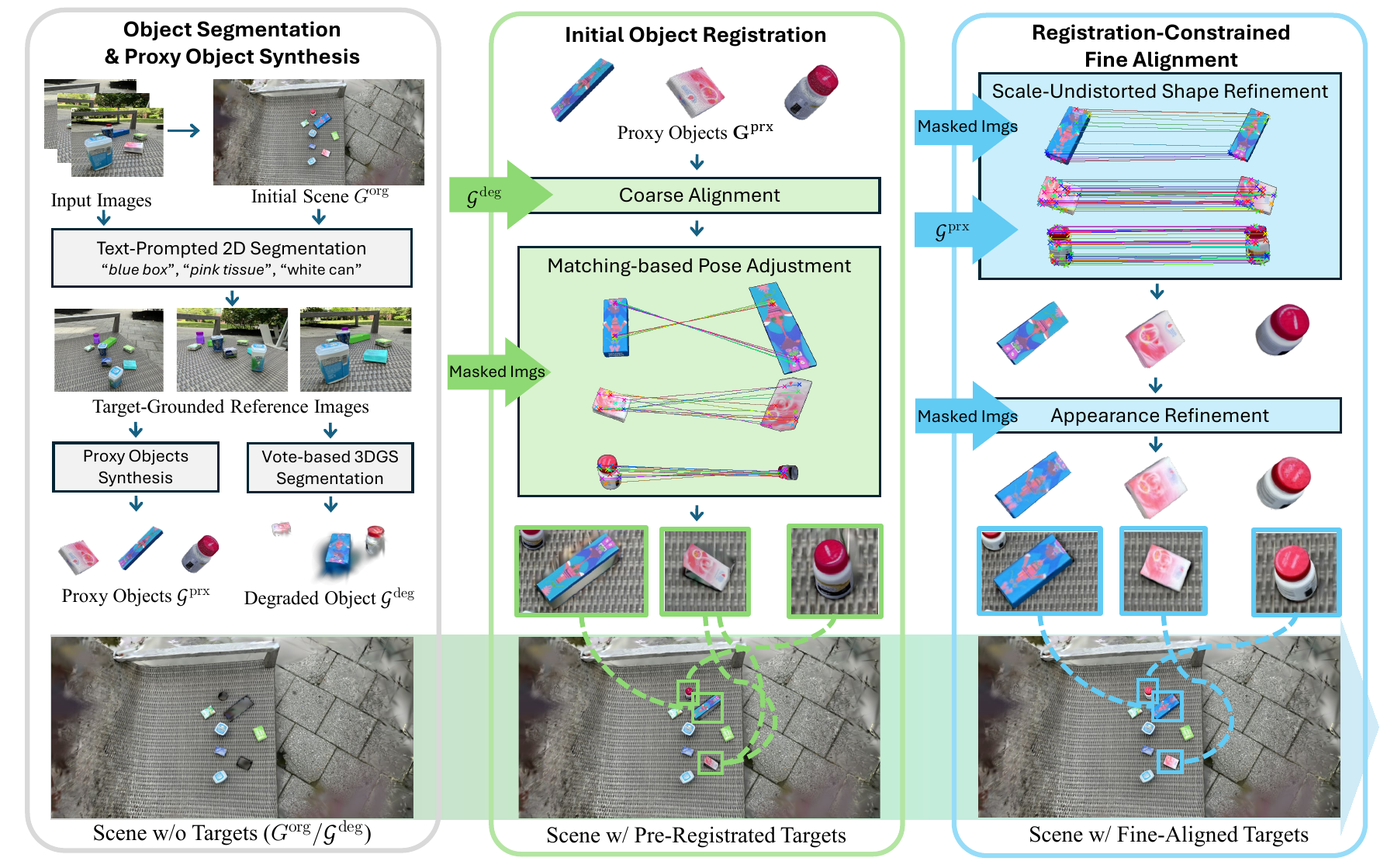}
    \vspace{-2em}
    \caption{
    Overview of \ours. Left: Object segmentation \& proxy synthesis. From multi-view inputs, we reconstruct an initial 3DGS scene, obtain text-prompted object masks, select informative views, and synthesize a proxy object for each target object with a generative model. Middle: Initial object registration. Each proxy is inserted into the scene via coarse alignment and matching-based pose adjustment using origin–proxy correspondences, yielding a scene with pre-registered targets. Right: Registration-constrained fine alignment. We perform scale-undistorted shape refinement followed by appearance refinement under pose constraints, producing fine-aligned objects that recover geometry and textures while preserving scene-level consistency. Bottom rows show the rendered scene after each stage.
    }
    \vspace{-1em}
    \label{fig:pipeline}
\end{figure*}

\section{Method}

This section details \ours, a three-stage framework: (1) Object Segmentation and Proxy Object Synthesis, (2) Initial Object Registration, and (3) Registration-constrained Fine Alignment. Stage (1) performs proxy generation, while stages (2) and (3) together constitute proxy tuning in a coarse-to-fine manner.
As shown in Fig.~\ref{fig:pipeline}, given $N_c$ RGB images $\mathcal{I}=\{I_i\}_{i=1}^{N_c}$ with camera intrinsic $K$ and extrinsic $\mathcal{T}=\{T_i\}_{i=1}^{N_c}$, the 3D scene $G^\text{org}$ reconstructed upon those, and text prompts indicating the objects of interest, \ours aims to enhance objects in both appearance and geometry quality, while keeping their original textures and poses. 

Specifically, we first (Sec.~\ref{sec:preprocessing}) employ text-driven visual grounding and tracking models to localize target objects in input images. Upon those, we segment degraded target objects $\mathcal{ G}^{\text{deg}}=\{G_i^{\text{deg}}\}_{i=1}^{N_o}$ from the reconstructed scene.
Given grounded input images, we generate proxy objects $\mathcal{G}^{\text{prx}}=\{G_i^{\text{prx}}\}_{i=1}^{N_o}$ for the incomplete counterparts using a 3D generative model. Next (Sec.~\ref{sec:coarse_refine}), we perform 7-DoF pose alignment to register $\mathcal{G}^{\text{prx}}$ coarsely. Finally (Sec.~\ref{sec:delicate_refine}), to address the limitations in generative fidelity, we introduce a fine refinement step that further aligns the anisotropic scale and visual attributes of the proxy objects with those of the real objects. This refinement ensures that the proxy objects can reliably replace the incomplete ones with high geometric and visual scene consistency.

\subsection{Object Segmentation \& Proxy Object Synthesis}
\label{sec:preprocessing}
At this stage, we segment the degraded object within the 3DGS scene and synthesize its corresponding proxy object.

\paragraph{Object Localization and Segmentation.} For object segmentation, text prompts describing the objects can either be generated automatically using a Visual Language Model (VLM) \cite{bai2025qwen2} or provided by users. To segment a target object and obtain its original 3DGS representation (despite incompleteness) as the reference for the proxy object, we first extract a binary mask on a reference image $I$, denoted as $M_I$, of the target object using a grounding model \cite{liu2023grounding} with the given prompt. The mask is then propagated across the entire image sequence with a tracking model \cite{ravi2024sam} to acquire all corresponding masks, denoted as $\mathcal{M}=\{M_{I_i}\}_{i=1}^{N_c}$.

With the 2D masks obtained, we can extract the degraded objects $\mathcal{G}^{\text{deg}}=\{G_i^{\text{deg}}\}_{i=1}^{N_o}$ using a gradient voting strategy \cite{joseph2024gradient} from the original 3DGS scene. Specifically, the derivative of the color attributes of each Gaussian primitive is computed with respect to the rendered pixel colors, which is $\widetilde{g}_i\prod_{j=1}^{i-1}(1-\widetilde{g}_j)$ in Eq.~\eqref{eq:gs_rgb} and considered as the vote weight for assigning Gaussian primitives to the target object. Pixels inside the object mask cast positive votes, while pixels in its complement cast negative votes. Statistically summing up these votes, Gaussian primitives with a positive final vote are assigned to the object.

\paragraph{Proxy Objects Synthesis.} 
To recover the missing geometry and appearance information, we leverage a 3D generative model to synthesize proxy objects $\mathcal{G}^{\text{prx}}=\{G_i^{\text{prx}}\}_{i=1}^{N_o}$ by referencing images captured from visible viewpoints. The generated proxies serve as seamless substitutes for the original objects. To identify more informative inputs, we apply a heuristic view selection scheme to choose few images from visible views of the object as input to the generative model.
Specifically, starting from an initially selected view, we iteratively expand the input set by evaluating the remaining candidate views and selecting the one that best balances object completeness within the image and overall viewpoint diversity. As for a target object in a given image, we quantitatively assess the geometric completeness of its mask $M$ by calculating the area ratio $Q_{\text{shape}}$  between the foreground area $\lvert M > 0 \rvert$ and the area of the convex hull of the foreground area.
Moreover, we incorporate viewpoint diversity score $Q_{\text{view}}$ by jointly considering positional and orientational differences, and compute the final image selection score $Q$ using the linear combination of $Q_{\text{shape}}$ and $Q_{\text{view}}$. Detailed formulation is shown in the supplementary material. The image with the highest score will be selected as the next input image set in each iteration. The selected images are then fed into the 3D generative model TRELLIS~\cite{trellis}, generating a proxy object represented with 3DGS.

\subsection{Initial Object Registration}
\label{sec:coarse_refine}

Although the generated proxy object demonstrates improvements in geometry and appearance, it still exhibits discrepancies in pose, shape, and visual attributes compared to the degraded object in the scene. Therefore, the goal of this stage is to align the generated proxy object with the 7-DoF pose of the degraded object to achieve an initial registration.

\paragraph{Coarse Alignment.} 
We first calculate the 3D bounding boxes for both the degraded object $G^\text{deg}$ and the proxy object $G^\text{prx}$. The scaling factor is then determined by taking the cube root of the ratio of the volumes of these two bounding boxes. Next, the centroids of $G^\text{deg}$ and $G^\text{prx}$ are computed as the centroids of their respective convex hulls \cite{barber1996quickhull}, constructed from the 3D positions of Gaussian primitives in 3DGS. Subsequently, we scale $G^\text{prx}$ to match the size of $G^\text{deg}$ and then translate it to align with the centroid of the degraded object $G^\text{deg}$. 
To align the pose of proxy object $G^\text{prx}$ with that of the degraded object $G^\text{deg}$, we employ the Iterative Closest Point (ICP) \cite{besl1992method} to adjust the pose of $G^\text{prx}$, utilizing the 3D positions of Gaussian primitives of two 3DGS models. 
Finally, the pose of the proxy object $G^\text{prx}$ and the degraded object $G^\text{deg}$ tend to be roughly aligned.

\newcommand{\matchwidth}{0.161}
\begin{figure*}[!t]
    \centering
    \addtolength{\tabcolsep}{-6.5pt}
    \footnotesize{
        \setlength{\tabcolsep}{1pt} 
        \begin{tabular}{cccccc}
            
            \includegraphics[width=\matchwidth\textwidth]{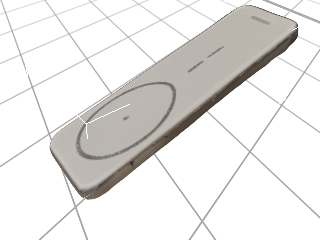} &
            \includegraphics[width=\matchwidth\textwidth]{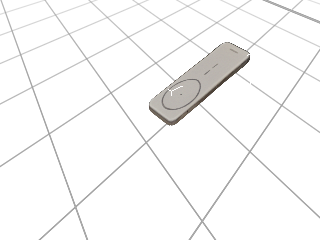} &
            \includegraphics[width=\matchwidth\textwidth]{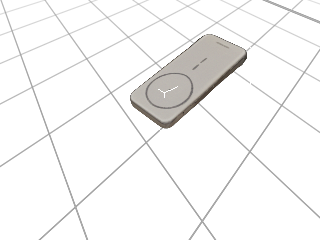} &
            \includegraphics[width=\matchwidth\textwidth]{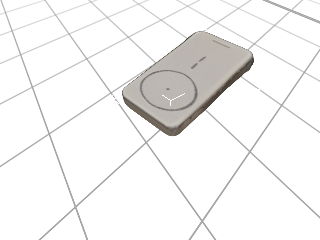} &
            \includegraphics[width=\matchwidth\textwidth]{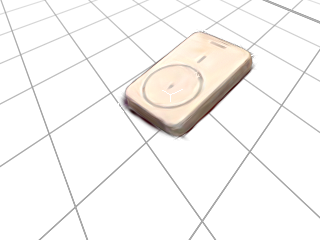} &
            \includegraphics[width=\matchwidth\textwidth]{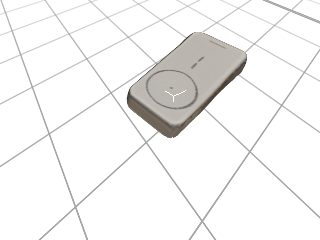} \\
            \includegraphics[width=\matchwidth\textwidth]{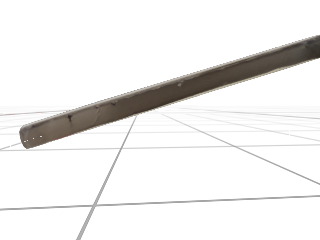} &
            \includegraphics[width=\matchwidth\textwidth]{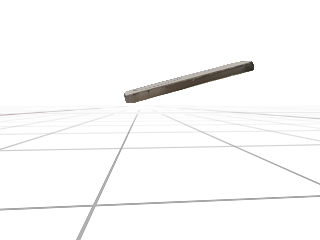} &
            \includegraphics[width=\matchwidth\textwidth]{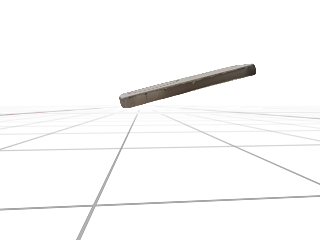} &
            \includegraphics[width=\matchwidth\textwidth]{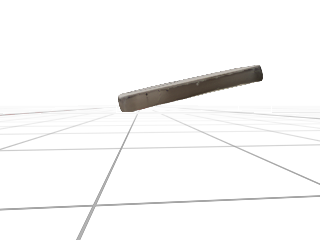} &
            \includegraphics[width=\matchwidth\textwidth]{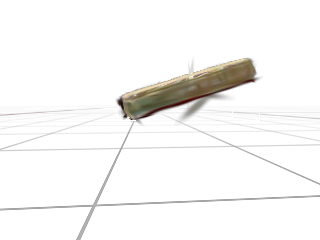} &
            \includegraphics[width=\matchwidth\textwidth]{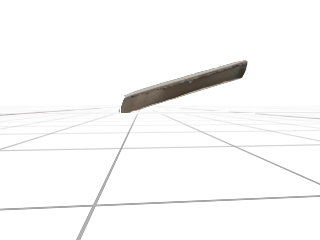} \\
            
            Init. & Iter. 3 & Iter. 5 & Iter. 7 & 2DGS & SVD \cite{chatrasingh2023generalized} Iter. 7 \\
        \end{tabular}
    }
    \caption{Qualitative ablations on Matching-Based Pose Adjustment and Scale-Undistorted Shape Refinement for a real-world object ``wireless charger''. ``Init.'' symbolizes the initial proxy object from Coarse Alignment, while ``Iter. $i$'' indicates the outcome after the $i$th iteration of ours. 2DGS, trained on all captured views, provides near-ground-truth reference images. The final column, “SVD Iter. 7,” presents the result of \citet{chatrasingh2023generalized}.
    Each of the two rows displays rendered images and bounding boxes from two viewpoints for every ablation method or iterative stage. The experiment comprises 7 iterations: the first 3 apply Pose Adjustment, and the last 4 apply Shape Refinement. The first row demonstrates that Pose Adjustment progressively corrects global object scale, whereas Shape Refinement then rectifies object shape, although the initial geometry is highly distorted. The second row highlights that during the Shape Refinement stage, the shape steadily converges toward a realistic, undistorted form. In contrast, \citet{chatrasingh2023generalized}'s result approximates the real shape but exhibits pronounced distortion.
    }
    \label{fig:ablation_demo_partial}
    \vspace{-1em}
\end{figure*}
\paragraph{Matching-Based Pose Adjustment.} 
So far, the proxy object $G^\text{prx}$ and the degraded object $G^\text{deg}$ may still exhibit pose discrepancies. To address this, for each camera view $T^c=(R^c,t^c)$, we calculate the 2D features $(F^\text{prx},F^\text{deg})$ of the rendered image $I^\text{prx}$ and the target image $I$ on the masked area $M_I$ of the object using MASt3R~\cite{leroy2024grounding}. Then we extract $N_m$ matched 2D-2D correspondences $\{(u^\text{prx}_i,u^\text{deg}_i)\}_{i=1}^{N_m}$ between these two images, ensuring mutual nearest neighbor~\cite{leroy2024grounding} relationship of each other.
We then unproject the depth value of matched 2D points to get their corresponding 3D positions $\{(P^\text{prx}_i,P^\text{deg}_i)\}_{i=1}^{N_m}$ in world space using:
\begin{equation}
    P=(R^c)^\top (K^{-1}[u^\top,1]^\top D[u]-t^c),\label{eq:world_points}
\end{equation}
where depth maps $D^\text{prx}$ and $D^\text{deg}$ are rendered from Eq.~\eqref{eq:gs_depth}.
With the 3D-3D correspondences for $N_c$ camera views $\mathcal{P}^\text{prx}, \mathcal{P}^\text{deg}\in \mathbb{R}^{3\times N_c\cdot N_m}$ in 3D world coordinate, the 7-DoF transformation between the proxy object $G^\text{prx}$ and the degraded object $G^\text{deg}$ can be estimated by Umeyama \cite{umeyama1991least}:
\begin{equation}
(\hat{R},\hat{t},\hat{s})=\underset{R,t,s}{\operatorname{argmin}}\sum_{i}\left\|sRP_i^\text{prx}+t-P_i^\text{deg}\right\|^2,\label{eq:7dof}
\end{equation}
where $\hat{R}\in\mathrm{SO}(3),\hat{t}\in \mathbb{R}^3,\hat{s}\in\mathbb{R}$. The proxy object $G^\text{prx}$ is then transformed by the estimated transformation $(\hat{R},\hat{t},\hat{s})$ to align with the degraded object $G^\text{deg}$ in the image space.

\subsection{Registration-Constrained Fine Refinement}
\label{sec:delicate_refine}

Although pose and scale are coarsely aligned through Initial Object Registration, the proxy object may still exhibit geometric and appearance deviations from the degraded object, as shown in ``Iter. 3'' in Fig.~\ref{fig:ablation_demo_partial}. 
We therefore introduce a shape refinement strategy with anisotropic scaling, allowing the proxy object $G^\text{prx}$ to better conform to the geometry of $G^\text{deg}$ along each spatial axis. Additionally, we refine the appearance of the proxy object $G^\text{prx}$ to keep the proxy visually consistent under seen views.

\begin{table*}[!ht]
    \caption{
        Quantitative comparisons on appearance on average in all scenes. The \colorbox{colorFst}{best}, the \colorbox{colorSnd}{second best}, and the \colorbox{colorTrd}{third best} are highlighted.
    }
    \vspace{-1em}
    \center
    \resizebox{0.9\linewidth}{!}
    {
    \begin{tabular}{l|cccc|cccc|cccc}
        \toprule
        \multirow{2}{*}{Method} & \multicolumn{4}{c|}{SSIM$\uparrow$} & \multicolumn{4}{c|}{PSNR$\uparrow$} & \multicolumn{4}{c}{LPIPS$\downarrow$} \\
        \addlinespace[1pt]\cline{2-13}\addlinespace[1pt]
        & $\nicefrac{2}{3}$ & $\nicefrac{4}{5}$ & $\nicefrac{6}{7}$ & Avg.& $\nicefrac{2}{3}$ & $\nicefrac{4}{5}$ & $\nicefrac{6}{7}$ & Avg.& $\nicefrac{2}{3}$ & $\nicefrac{4}{5}$ & $\nicefrac{6}{7}$ & Avg.\\
        \midrule
        3DGS~\cite{3dgs}
        & 0.902 & 0.896 & 0.884 & 0.894
        & 20.830 & 20.574 & 19.666 & 20.357
        & 0.107 & 0.112 & 0.122 & 0.114
        \\
        2DGS~\cite{2dgs}
        & \best{0.929} & \sbest{0.915} & \sbest{0.908} & \sbest{0.917}
        & \best{23.057} & \best{21.744} & \sbest{21.428} & \best{22.076}
        & \best{0.081} & \sbest{0.093} & \sbest{0.099} & \sbest{0.091}
        \\
        DNGaussian~\cite{dngaussian}
        & \tbest{0.911} & \tbest{0.903} & \tbest{0.897} & \tbest{0.904}
        & \sbest{21.757} & \tbest{21.057} & \tbest{20.771} & \tbest{21.195}
        & \tbest{0.099} & \tbest{0.107} & \tbest{0.112} & \tbest{0.106}
        \\
        GenFusion~\cite{genfusion}
        & 0.882 & 0.866 & 0.854 & 0.867
        & 20.330 & 19.465 & 18.630 & 19.475
        & 0.134 & 0.148 & 0.159 & 0.147
        \\
        \textbf{Ours}
        & \sbest{0.922} & \best{0.926} & \best{0.922} & \best{0.923}
        & \tbest{21.309} & \sbest{21.596} & \best{21.542} & \sbest{21.482}
        & \sbest{0.082} & \best{0.079} & \best{0.084} & \best{0.082} \\
        \midrule
        \multirow{2}{*}{Method} & \multicolumn{4}{c|}{MUSIQ$\uparrow$} & \multicolumn{4}{c|}{CLIPS$\uparrow$} & \multicolumn{4}{c}{mIoU$\uparrow$}\\
        \addlinespace[1pt]\cline{2-13}\addlinespace[2pt]
        & $\nicefrac{2}{3}$ & $\nicefrac{4}{5}$ & $\nicefrac{6}{7}$ & Avg.& $\nicefrac{2}{3}$ & $\nicefrac{4}{5}$ & $\nicefrac{6}{7}$ & Avg.& $\nicefrac{2}{3}$ & $\nicefrac{4}{5}$ & $\nicefrac{6}{7}$ & Avg.\\
        \midrule
        3DGS~\cite{3dgs}
        & \tbest{41.848} & \tbest{41.280} & \tbest{40.791} & \tbest{41.306}
        & 0.870 & 0.853 & 0.849 & 0.857
        & 0.531 & 0.511 & 0.474 & 0.505
        \\
        2DGS~\cite{2dgs}
        & \sbest{47.549} & \sbest{46.769} & \sbest{45.682} & \sbest{46.666}
        & \sbest{0.895} & \sbest{0.874} & \sbest{0.867} & \sbest{0.879}
        & \best{0.649} & \sbest{0.618} & \sbest{0.587} & \sbest{0.618}
        \\
        DNGaussian~\cite{dngaussian}
        & 41.778 & 40.623 & 40.413 & 40.938
        & \tbest{0.877} & \tbest{0.856} & \tbest{0.850} & \tbest{0.861}
        & \tbest{0.590} & \tbest{0.569} & \tbest{0.554} & \tbest{0.571}
        \\
        GenFusion~\cite{genfusion}
        & 28.673 & 29.651 & 29.050 & 29.125
        & 0.838 & 0.828 & 0.818 & 0.828
        & 0.512 & 0.453 & 0.428 & 0.465
        \\
        \textbf{Ours}
        & \best{56.317} & \best{56.033} & \best{54.534} & \best{55.628}
        & \best{0.906} & \best{0.907} & \best{0.902} & \best{0.905}
        & \sbest{0.614} & \best{0.636} & \best{0.620} & \best{0.623} \\
        \bottomrule
    \end{tabular}
    }
    \label{tab:quantitative-app}
    \vspace{-1em}
\end{table*}

\paragraph{Scale-Undistorted Shape Refinement.} 
To achieve shape alignment, we replace the isotropic scale parameter $s$, a scalar, in Eq.~\eqref{eq:7dof} of Pose Adjustment with an anisotropic scale $S=\operatorname{diag}(s_1,s_2,s_3)$, a diagonal matrix, where $s_1,s_2,s_3\in \mathbb{R}^+$. Consequently, the optimization objective is reformulated accordingly to accommodate directional scale flexibility:
\begin{equation}
    (\hat{R},\hat{t},\hat{S})=\underset{R,t,S}{\operatorname{argmin}}\sum_{i}\left\|RSP_i^\text{prx}+t-P_i^\text{deg}\right\|^2.\label{eq:9dof}
\end{equation}
\citet{chatrasingh2023generalized} addressed this optimization problem using singular value decomposition (SVD), following a procedure similar to the Umeyama \cite{umeyama1991least} algorithm. However, this solution is prone to the following two problems: (1) Instability occurs when the 3D-3D correspondences are inaccurate, as described in Sec.~\ref{sec:ablation}; (2) Directly optimizing the anisotropic scale $S$ can easily lead to undesirable distortions in the shape of the proxy object, as described in Fig.~\ref{fig:ablation_demo_partial}. 
Thus, we introduce a novel optimization objective that extends and regularizes Eq.~\eqref{eq:9dof} as follows:
\begin{align}
\underset{R,t,S,R'}{\operatorname{argmin}}&\;\Bigl\{\sum_{i}\left\|RR'^\top SR'P_i^\text{prx}+t-P_i^\text{deg}\right\|^2\nonumber\\
&+\lambda_R\mathcal{L}_R+\lambda_S\mathcal{L}_S\Bigl\},\label{eq:9dof_reg}\\
\mathcal{L}_R =& \tfrac12\|\log R\|_F^2,\quad \mathcal{L}_S = \left\|S-\tfrac{1}{3}\operatorname{Tr}(S)I\right\|^2_F,
\end{align}
where $\lambda_R$ and $\lambda_S$ are positive regularization coefficients. $\mathcal{L}_R$ and $\mathcal{L}_S$ are regularization terms to prevent unstablity.
Specifically, $\lambda_R$ constrains the rotation by penalizing a large rotation angle, effectively encouraging the solution to stay close to the initial pose and preventing unstable or excessive rotations during optimization. $\mathcal{L}_S$ regularizes the anisotropic scale matrix $S$ by minimizing the variance among its diagonal elements.
$R' \in \mathrm{SO}(3)$ denotes the rotation matrix that parameterizes a local axis frame for anisotropic scaling, thereby enabling the shape scaling of the proxy object in a new rotated frame rather than the definite contemporary frame, to avoid distortions occurring in optimization progress. The optimization objective is then solved by a gradient-based optimization \cite{kingma2014adam} method.

\paragraph{Pose-Constrained Appearance Refinement.} 
While the aligned proxy object already closely resembles the degraded object in appearance, some texture drift in the reference views is inevitable due to the hallucination of generation. To ensure high-fidelity visual consistency with the original object in the visible views, we further refine the appearance of the proxy object using visible training views.
Specifically, we freeze the position, covariance, and opacity attributes of the proxy object $G^\text{prx}$ and only adjust the spherical harmonics by minimizing the loss: 
\begin{equation}
    \mathcal{L}(M_I\odot I, I^{\text{deg}}) = (1 - \lambda)\mathcal{L}_1 + \lambda\mathcal{L}_{\text{D-SSIM}},
    \label{eq:refine_loss}
\end{equation}
between the rendered images $\mathcal{I}^\text{prx}$ of the proxy object and the target images $\mathcal{I}$ of the degraded object $G^\text{deg}$ from the training image set, according to \citet{3dgs}.

\section{Experiments}

\subsection{Experimental Setting}
\label{sec:experimental_setting}

\paragraph{Dataset.} 
Unlike traditional sparse-view reconstruction tasks, which aim to reconstruct the entire scene from a few input images, \ours focuses on refining degraded objects due to occlusions or insufficient viewpoint coverage, which is highly important for real-world applications, while current datasets do not provide sufficient support for evaluating performance in this aspect. 

To simulate this setting, we constructed a dedicated dataset to evaluate the appearance refinement performance of methods. We selected 11 scenes from Mip360 \cite{mip360}, ToyDesk \cite{toydesk}, LERF \cite{lerf}, and 3DGS-CD \cite{gs-cd} to evaluate methods on appearance refinement. In each scene, we pick a random start view and drop the $n$ closest views in joint position–orientation space, simulating missing observations. To assess robustness under varying degrees of missing viewpoints, we constructed three difficulty levels: easy, medium, and hard by removing $\nicefrac{2}{3}$, $\nicefrac{4}{5}$, and $\nicefrac{6}{7}$ of the images from image set as test views, using the remaining views for reconstruction.
For evaluation requiring geometry ground truth, we follow the approach of \citet{AdaPointr} to generate benchmark scenes. This allows access to complete object geometry information that is typically unavailable in real-world datasets such as ScanNet~\cite{scannet}. Specifically, each scene is composed of five objects randomly selected from the YCB-Video~\cite{PoseCNN} benchmark. We assign collision shapes to these objects and simulate physical interactions, falling onto a tabletop, to mimic complex scenarios that may arise in real environments. RGBD images are rendered from cameras positioned around the scene center, including a subset of views removed following the setups of different settings. As a result, 50 scenarios, with geometry ground truth, are constructed for geometric evaluation.

\newcommand{\renderwidth}{0.161}
\begin{figure*}[!htb]
    \vspace{-1em}
    \centering
    \addtolength{\tabcolsep}{-6.5pt}
    \footnotesize{
        \setlength{\tabcolsep}{1pt} 
        \begin{tabular}{cccccc}
            \includegraphics[width=\renderwidth\textwidth]{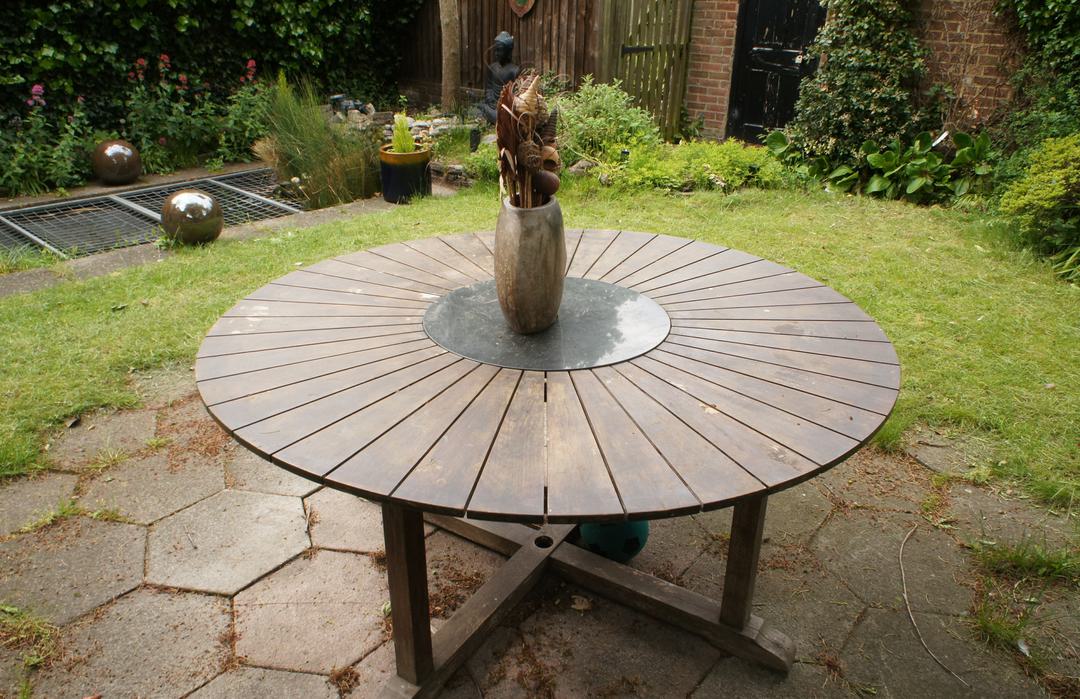} &
            \includegraphics[width=\renderwidth\textwidth]{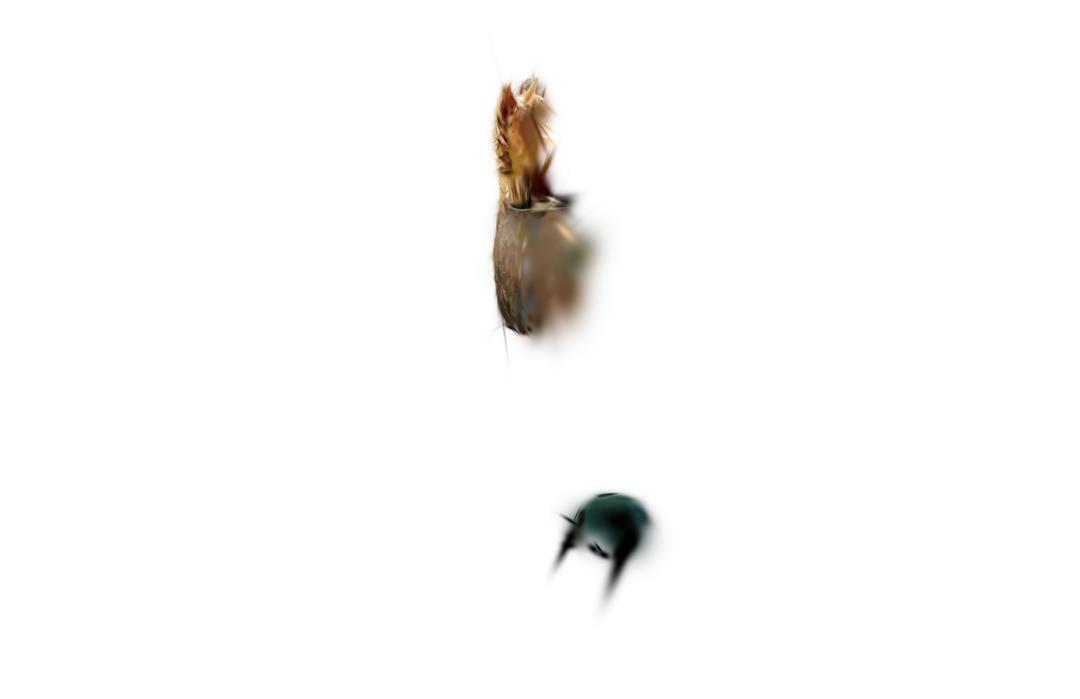} &
            \includegraphics[width=\renderwidth\textwidth]{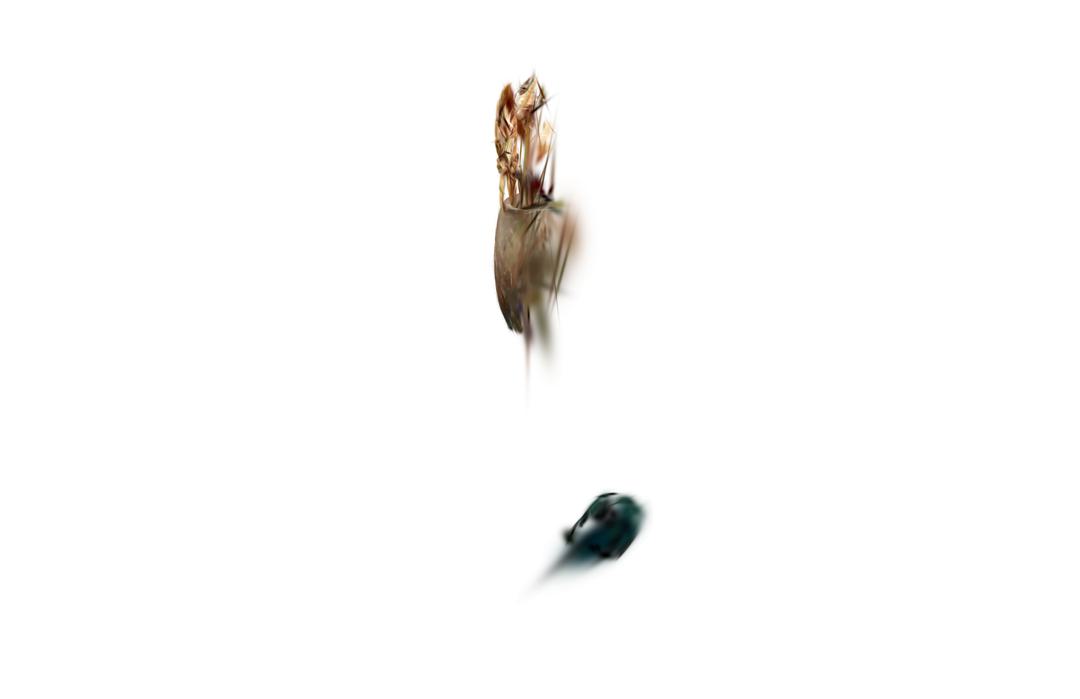} &
            \includegraphics[width=\renderwidth\textwidth]{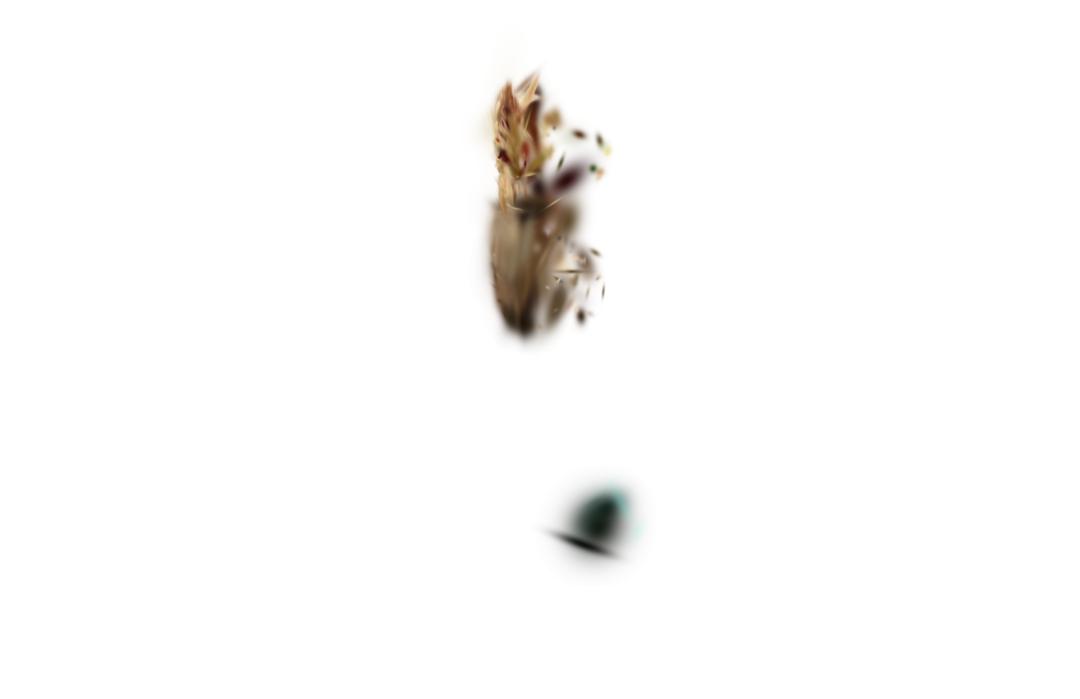} &
            \includegraphics[width=\renderwidth\textwidth]{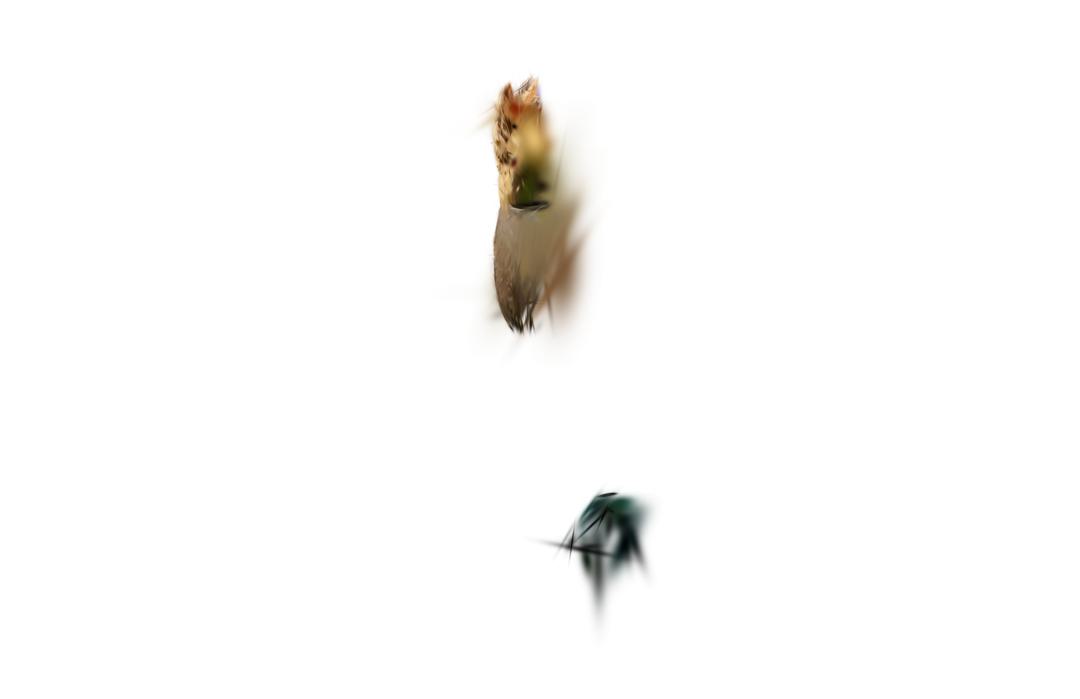} &
            \includegraphics[width=\renderwidth\textwidth]{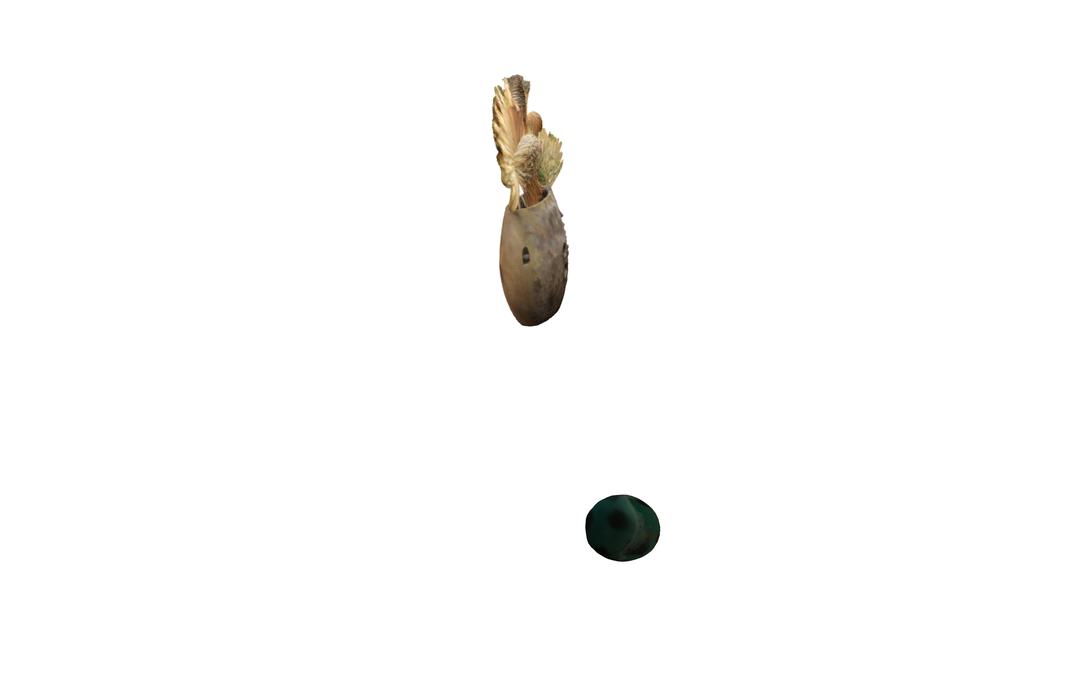} \\
            
            \includegraphics[width=\renderwidth\textwidth]{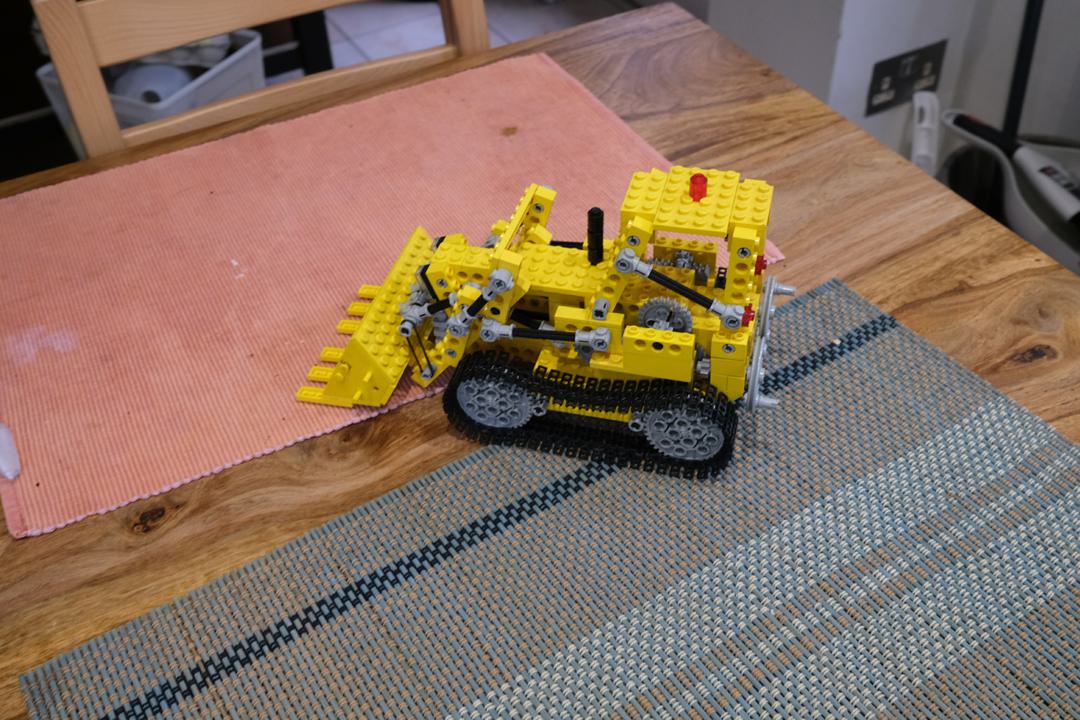} &
            \includegraphics[width=\renderwidth\textwidth]{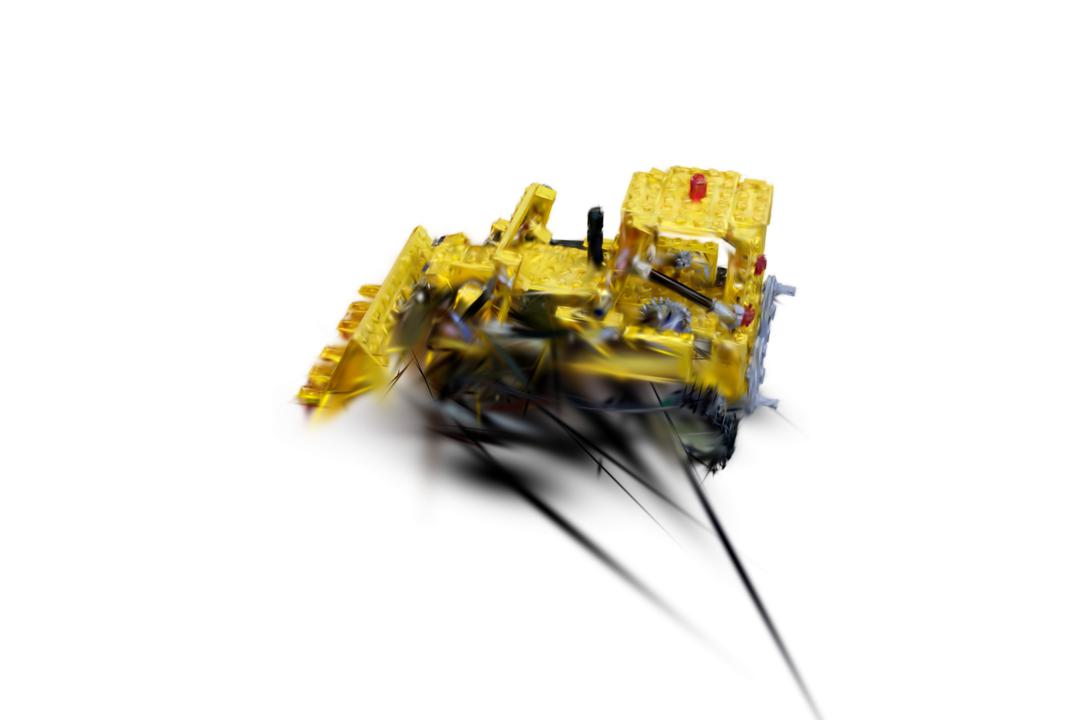} &
            \includegraphics[width=\renderwidth\textwidth]{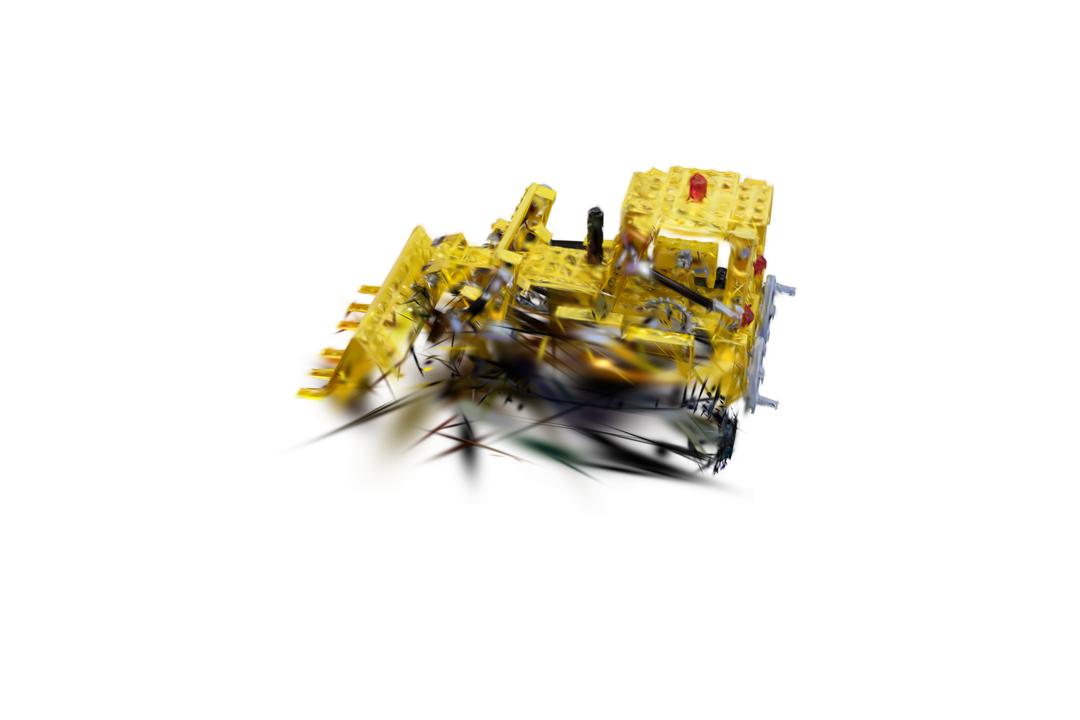} &
            \includegraphics[width=\renderwidth\textwidth]{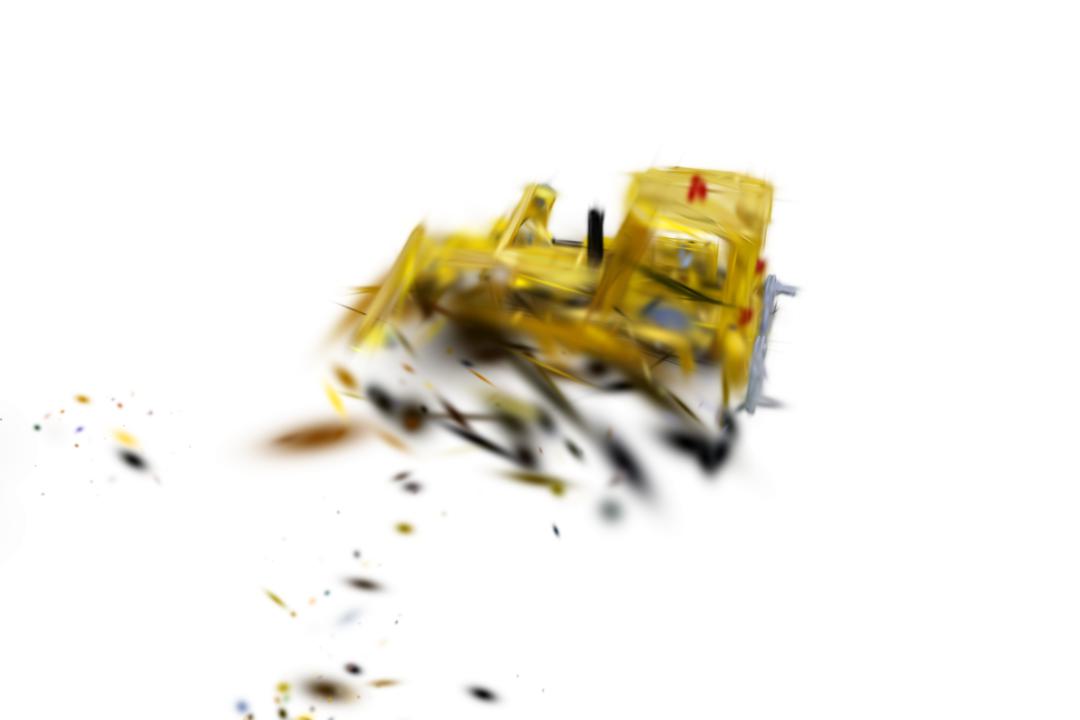} &
            \includegraphics[width=\renderwidth\textwidth]{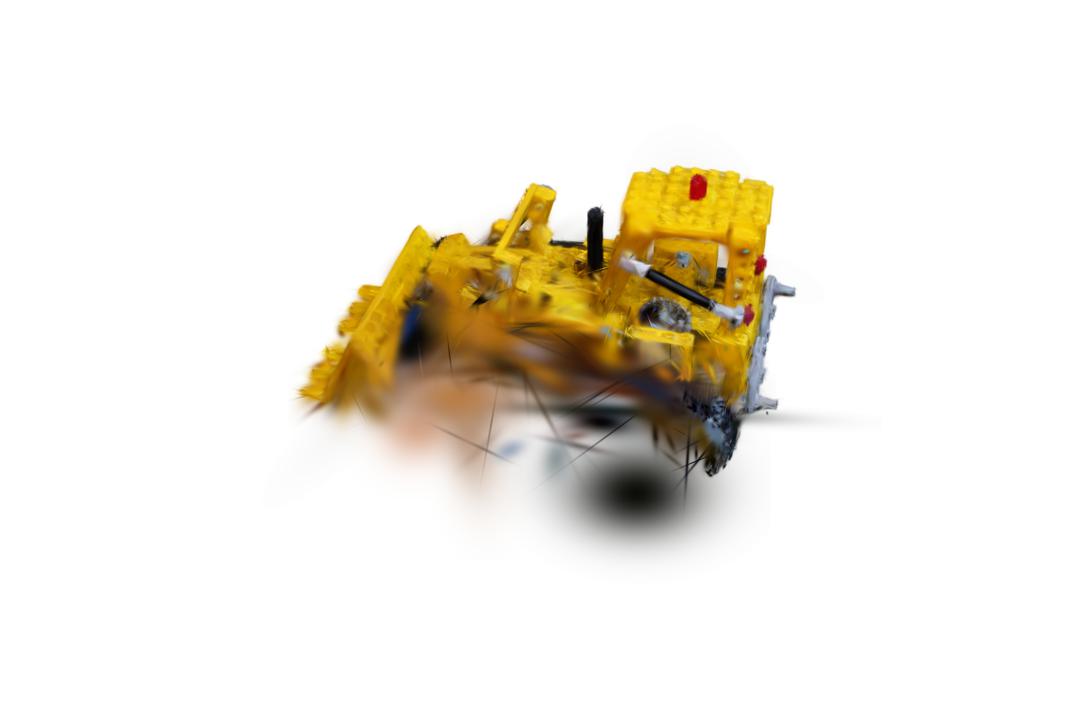} &
            \includegraphics[width=\renderwidth\textwidth]{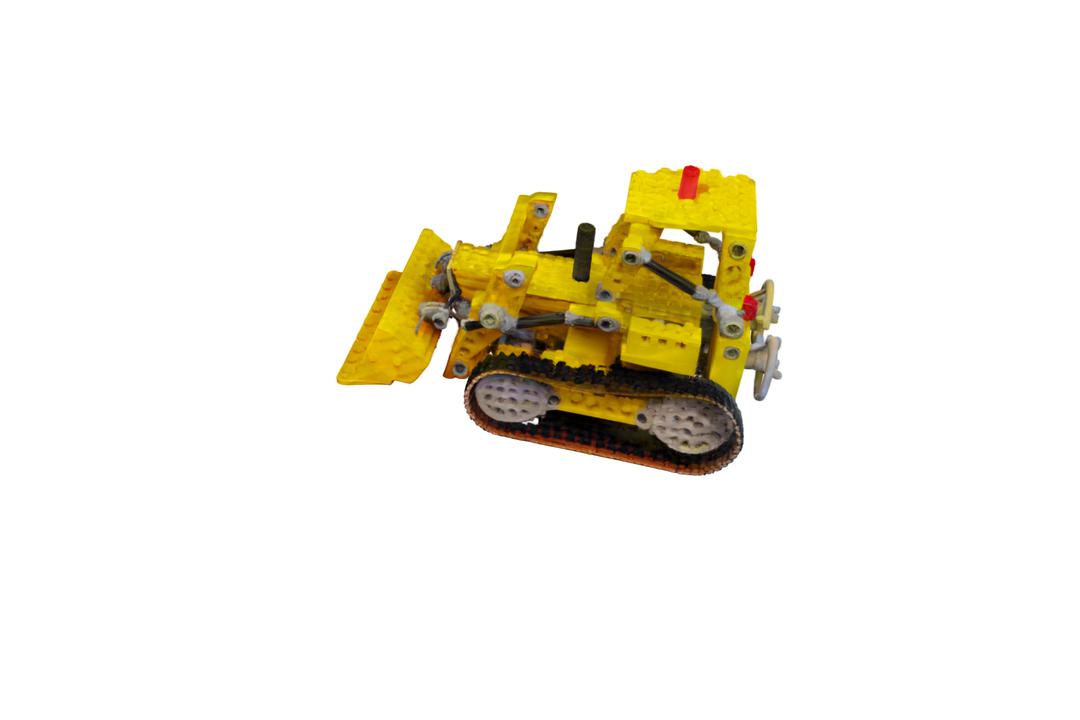} \\

            \includegraphics[width=\renderwidth\textwidth]{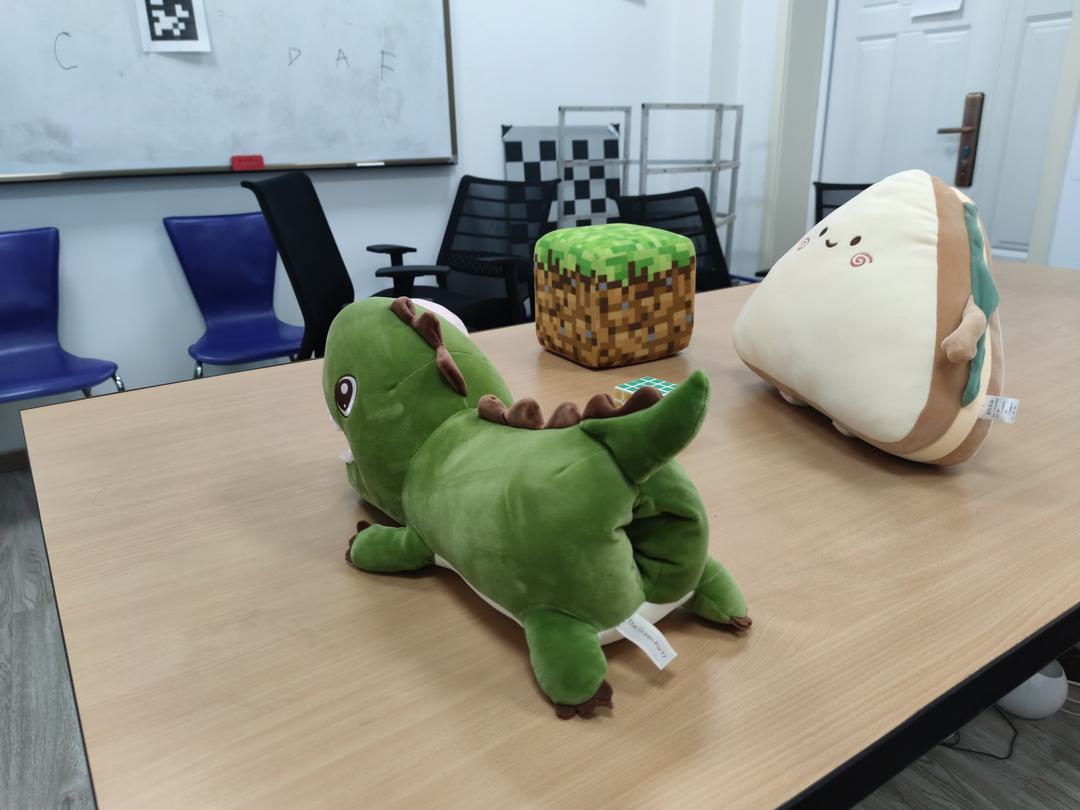} &
            \includegraphics[width=\renderwidth\textwidth]{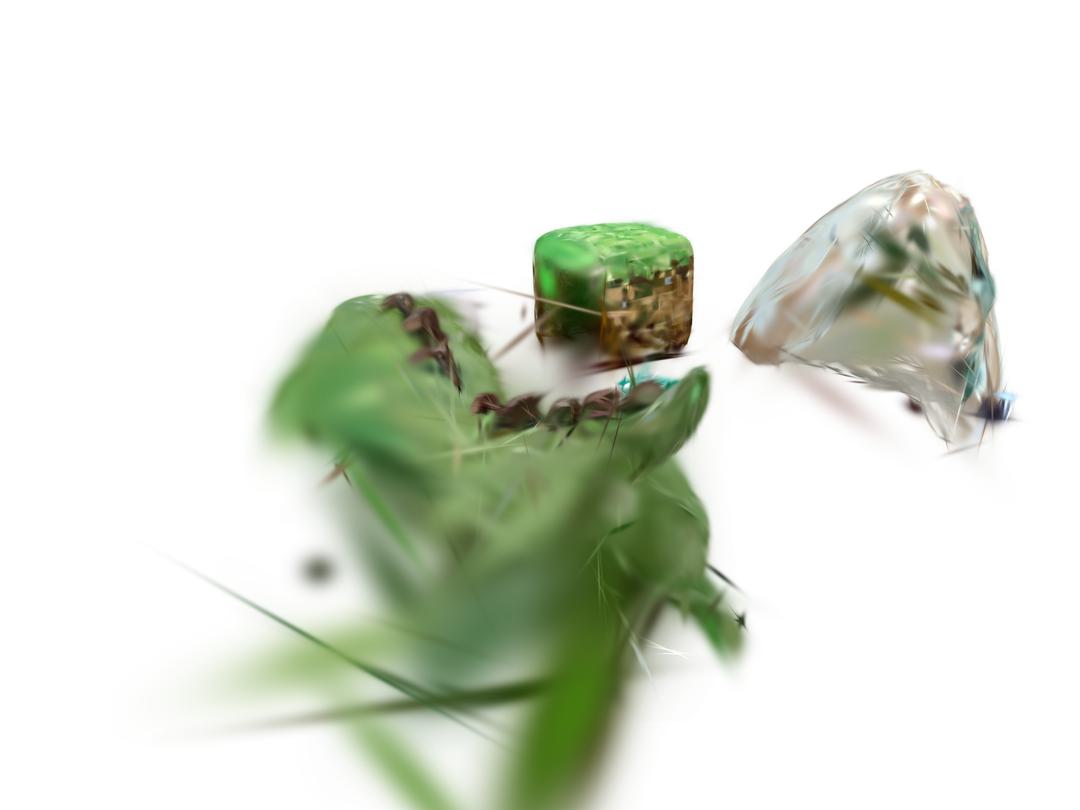} &
            \includegraphics[width=\renderwidth\textwidth]{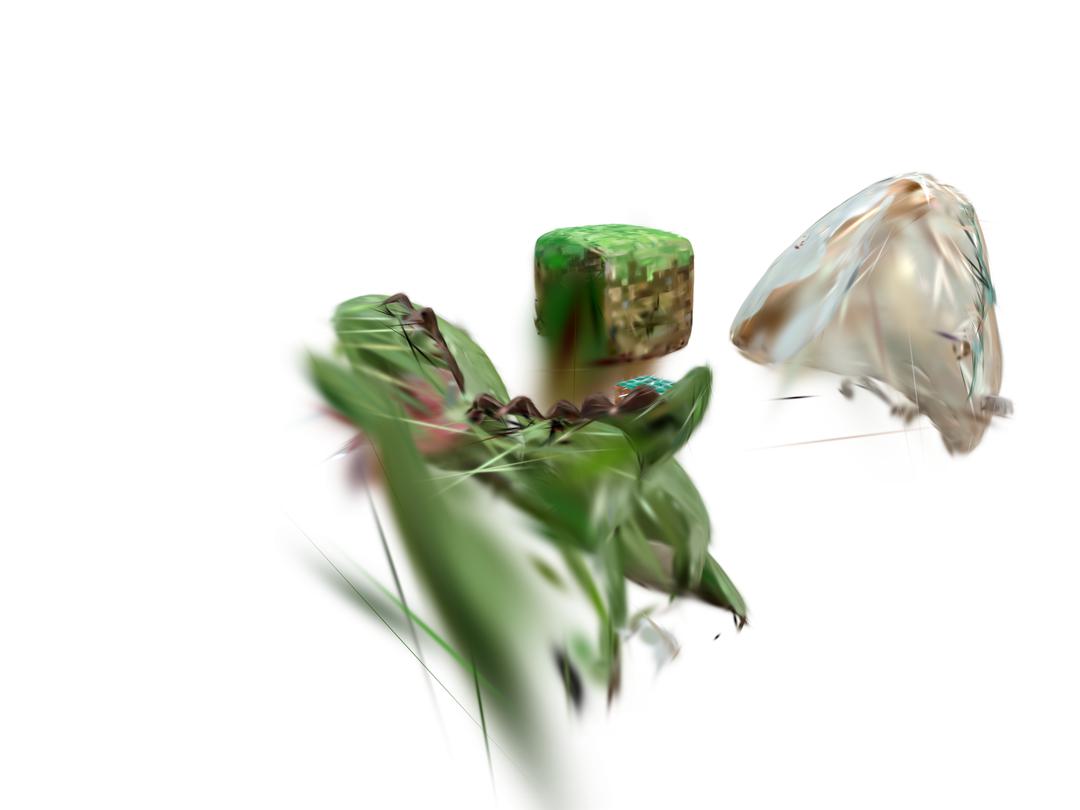} &
            \includegraphics[width=\renderwidth\textwidth]{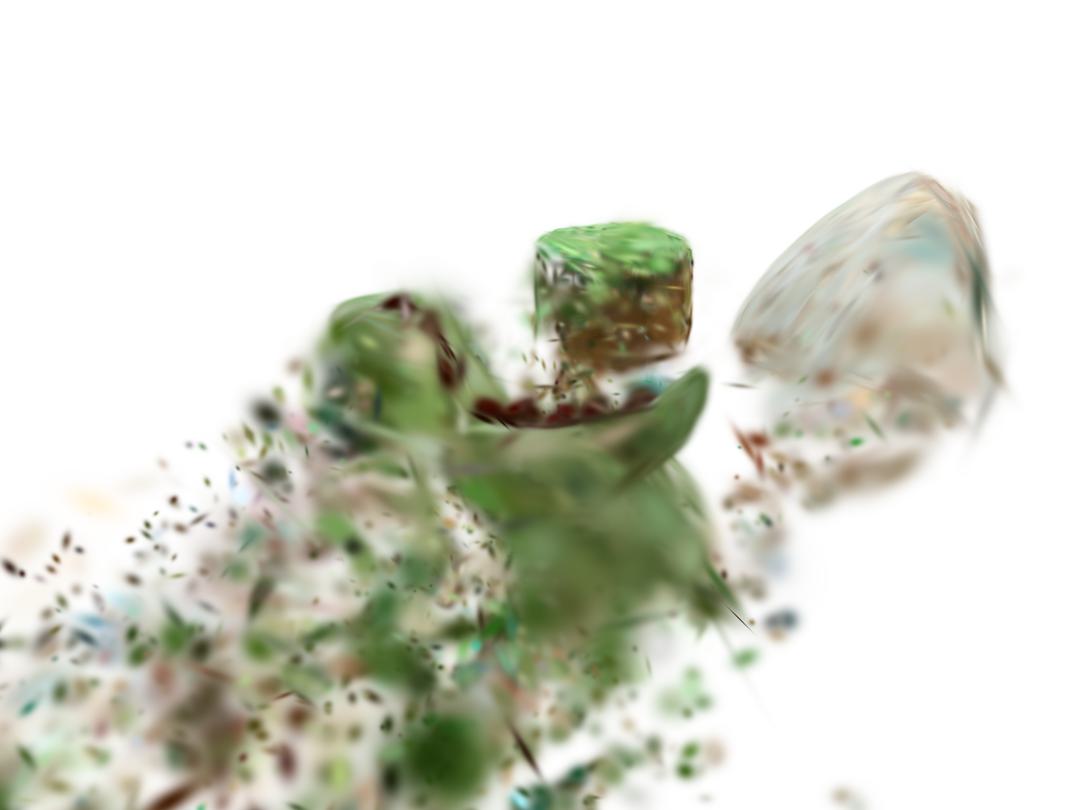} &
            \includegraphics[width=\renderwidth\textwidth]{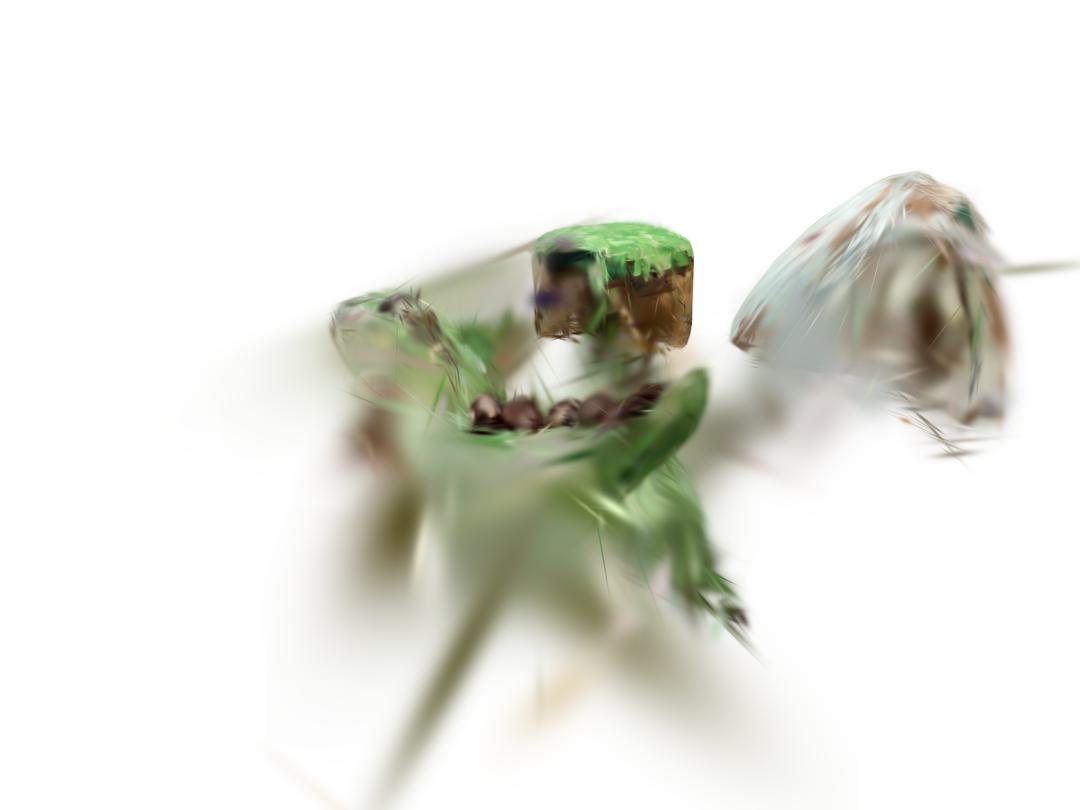} &
            \includegraphics[width=\renderwidth\textwidth]{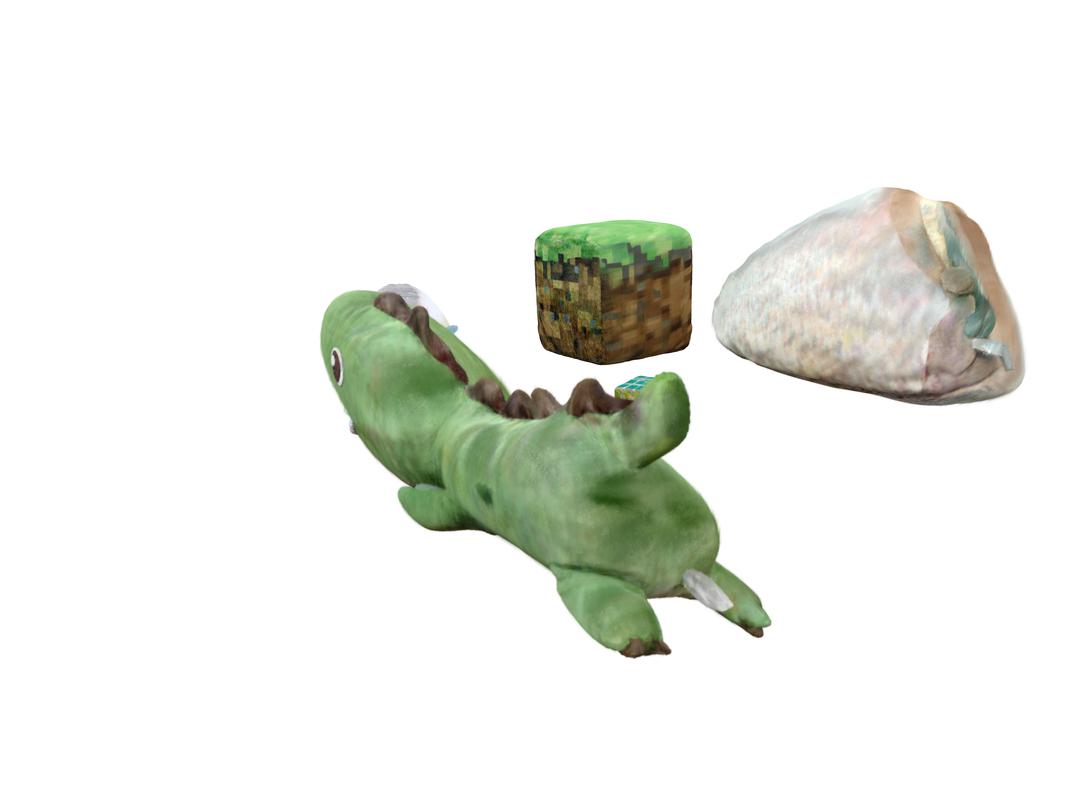} \\

            \includegraphics[width=\renderwidth\textwidth]{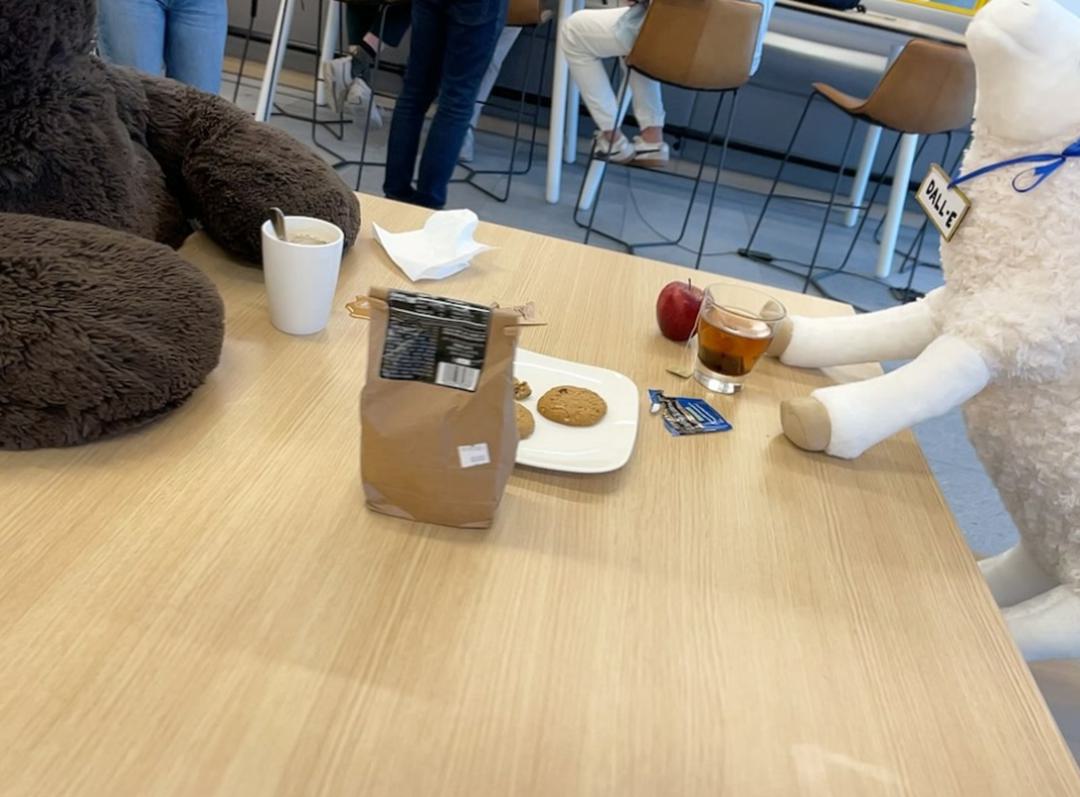} &
            \includegraphics[width=\renderwidth\textwidth]{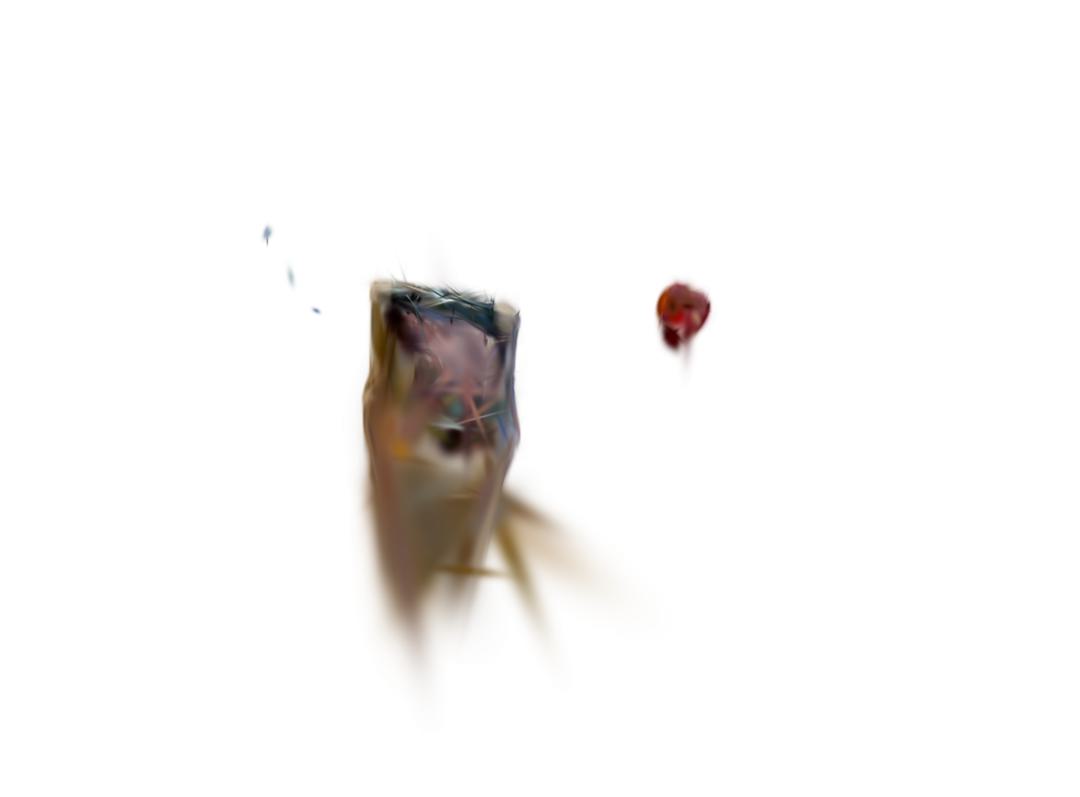} &
            \includegraphics[width=\renderwidth\textwidth]{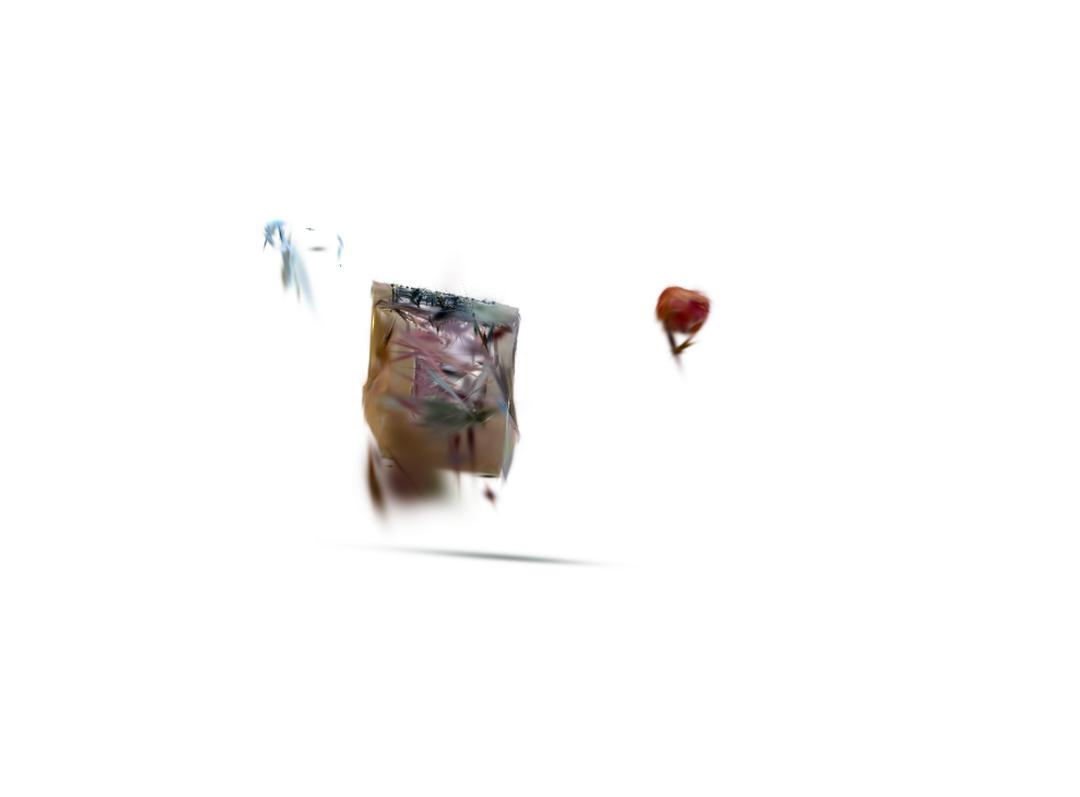} &
            \includegraphics[width=\renderwidth\textwidth]{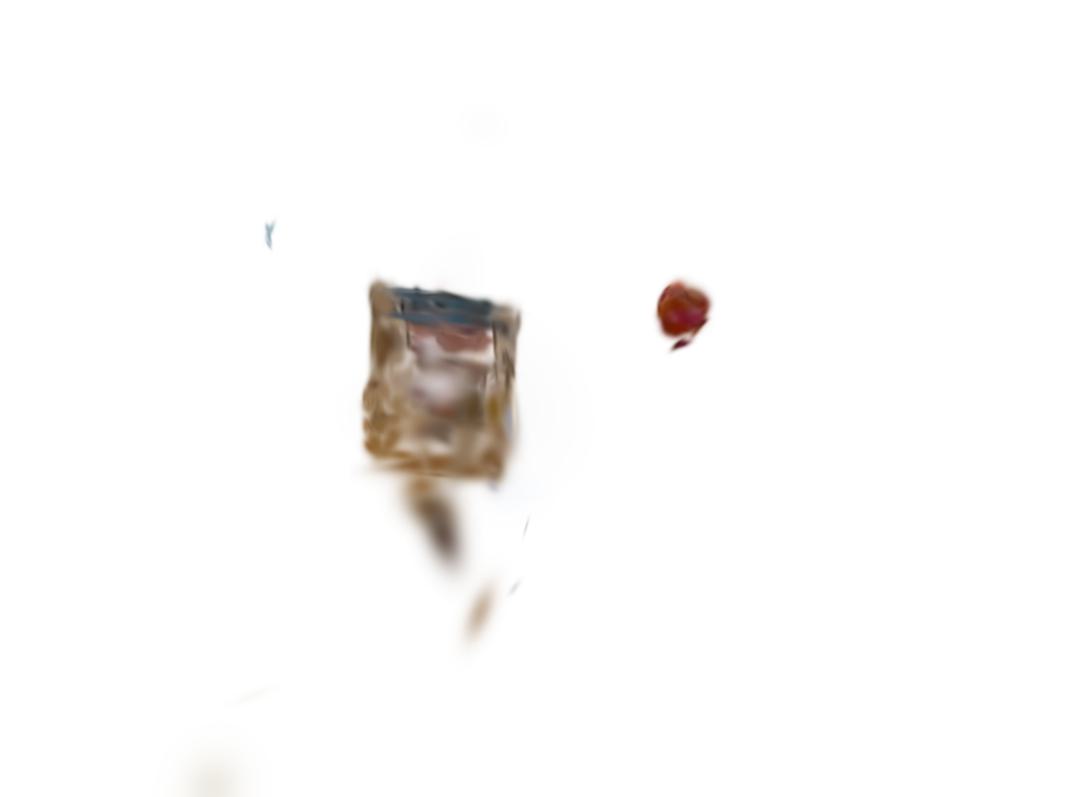} &
            \includegraphics[width=\renderwidth\textwidth]{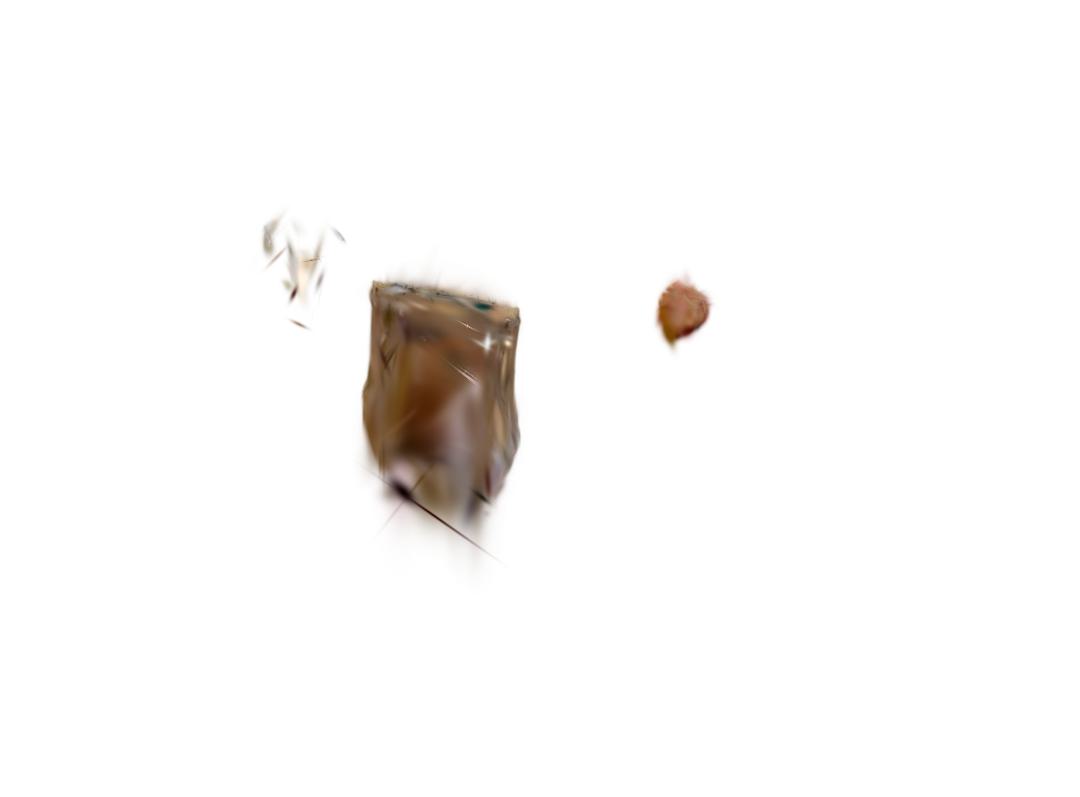} &
            \includegraphics[width=\renderwidth\textwidth]{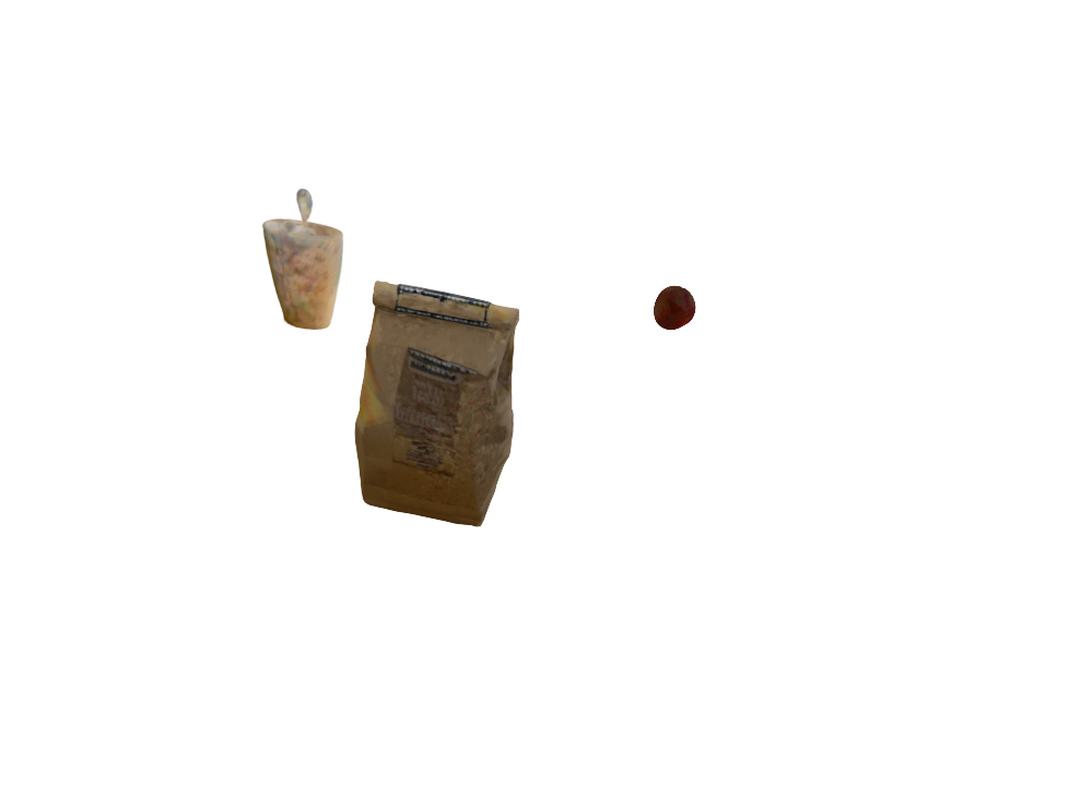} \\

            \includegraphics[width=\renderwidth\textwidth]{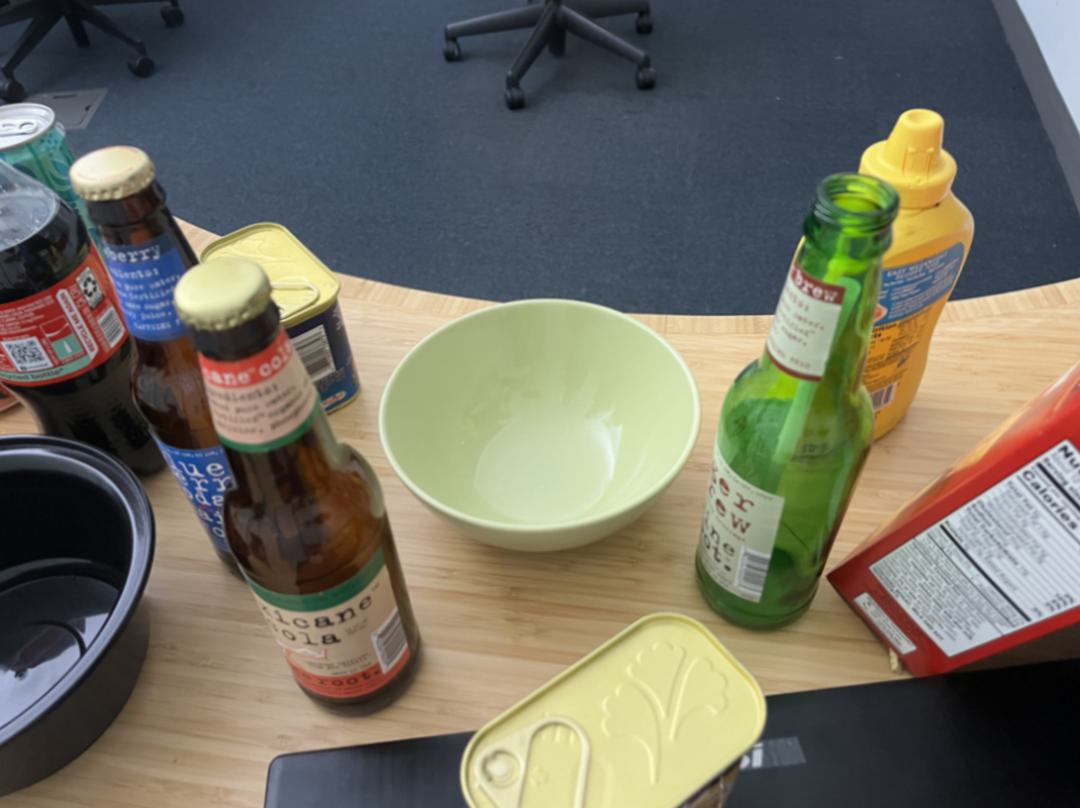} &
            \includegraphics[width=\renderwidth\textwidth]{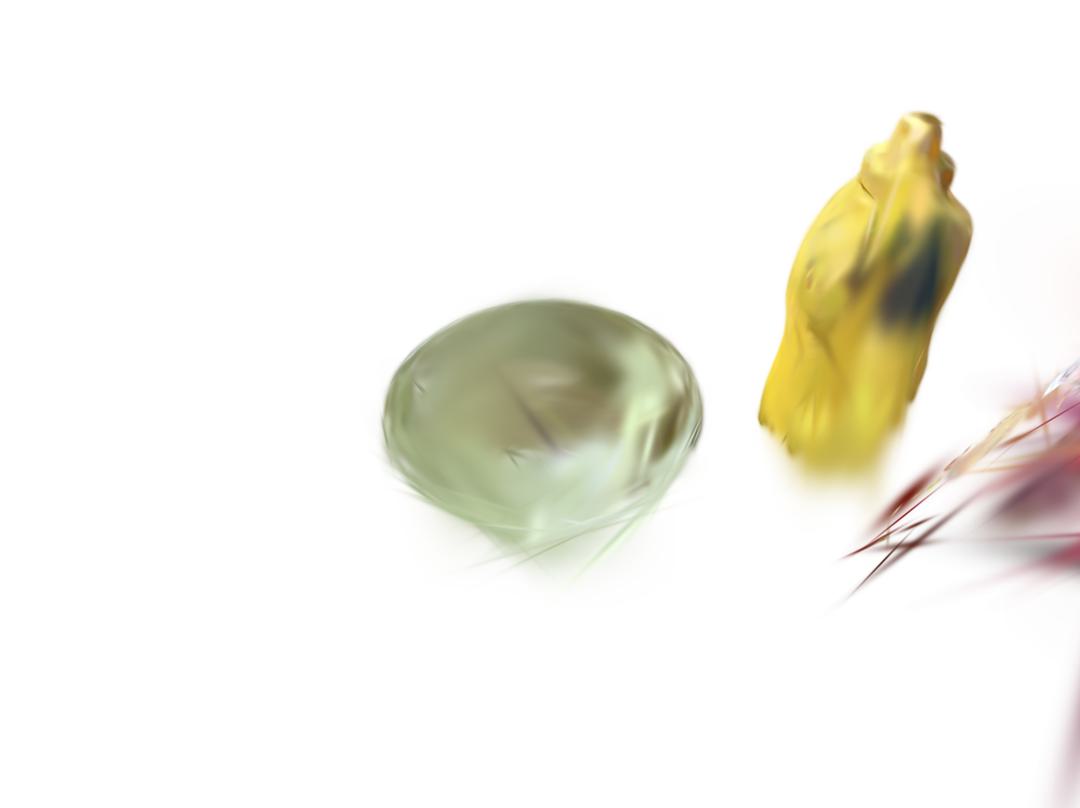} &
            \includegraphics[width=\renderwidth\textwidth]{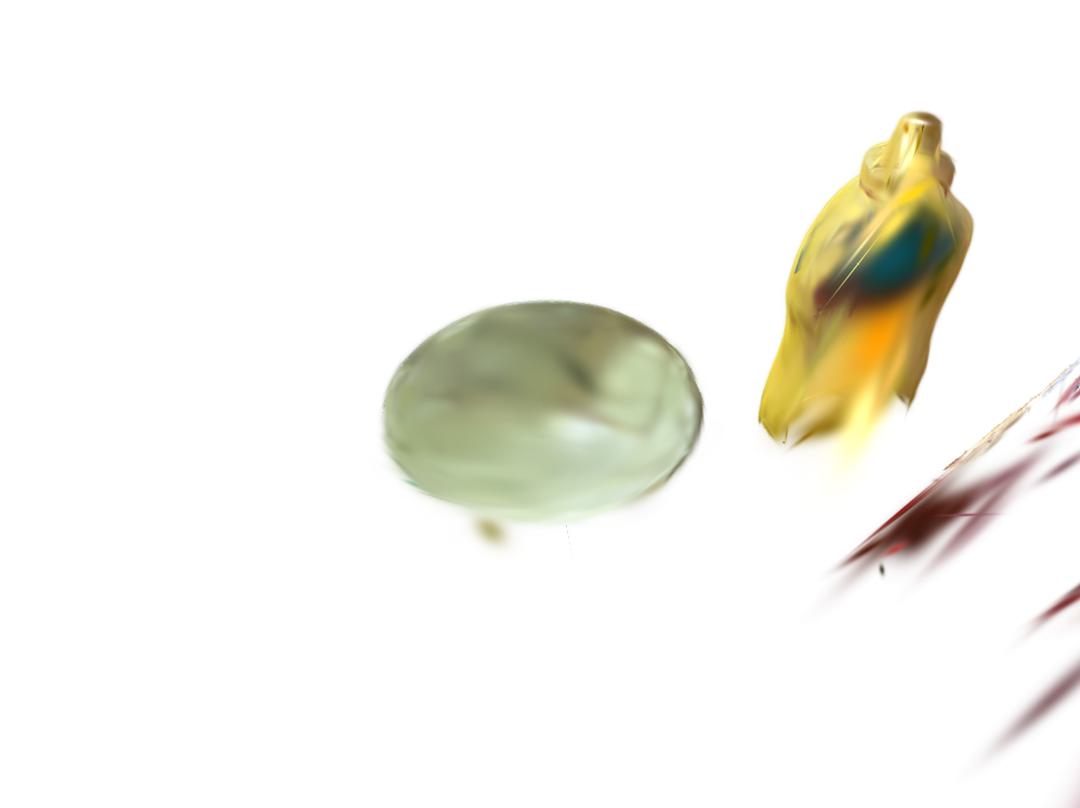} &
            \includegraphics[width=\renderwidth\textwidth]{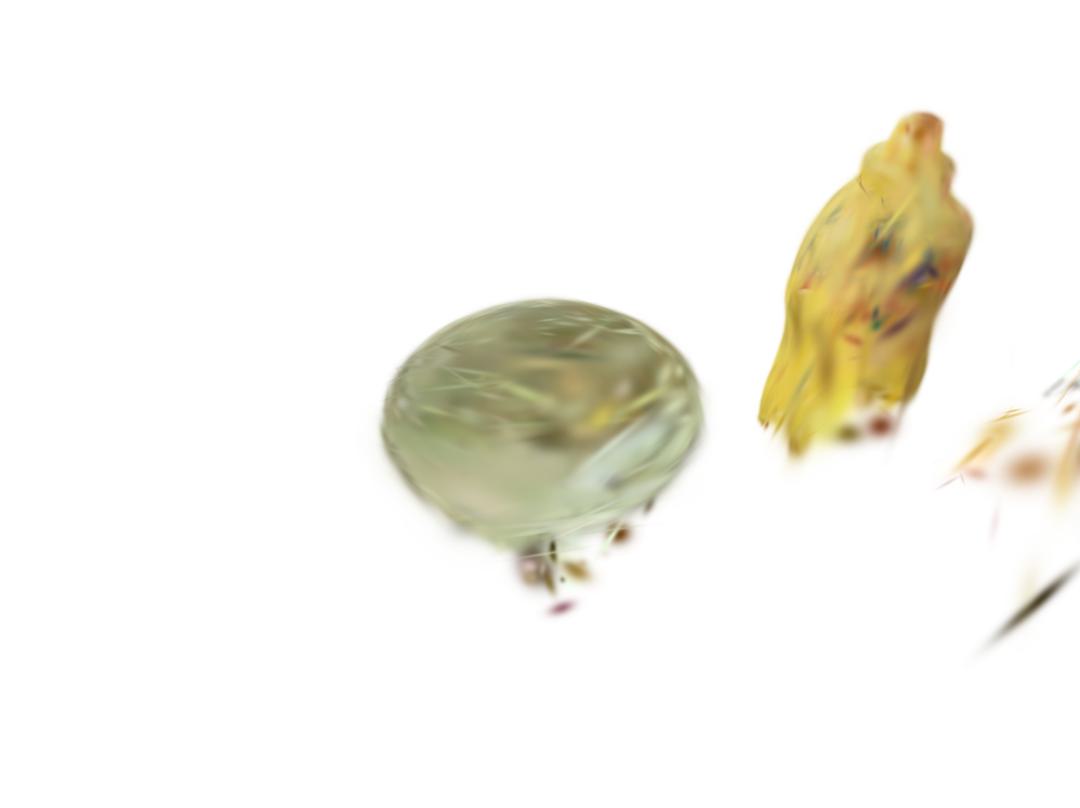} &
            \includegraphics[width=\renderwidth\textwidth]{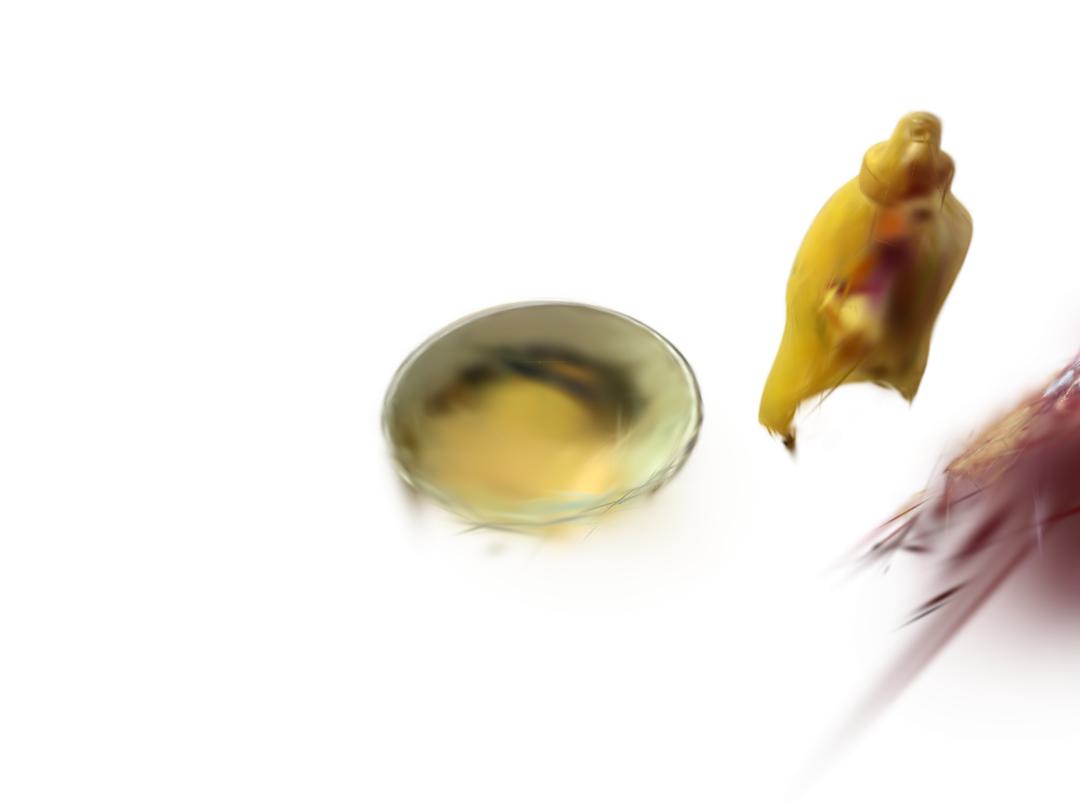} &
            \includegraphics[width=\renderwidth\textwidth]{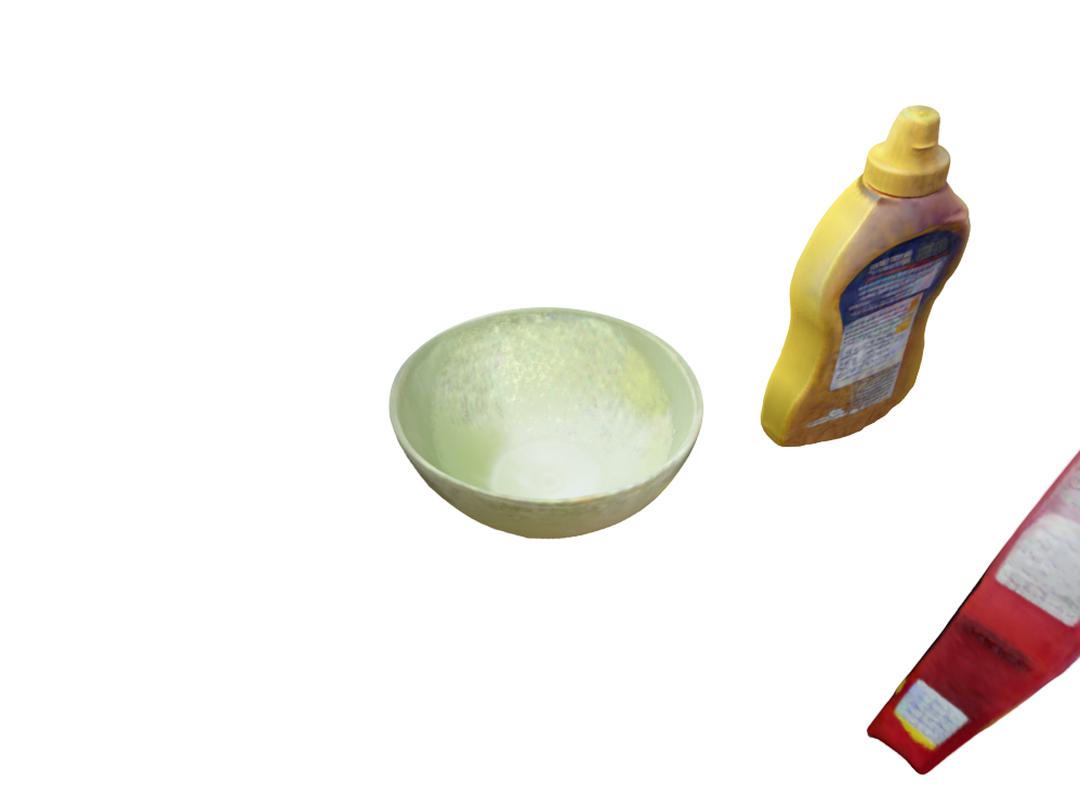} \\
            
            \includegraphics[width=\renderwidth\textwidth]{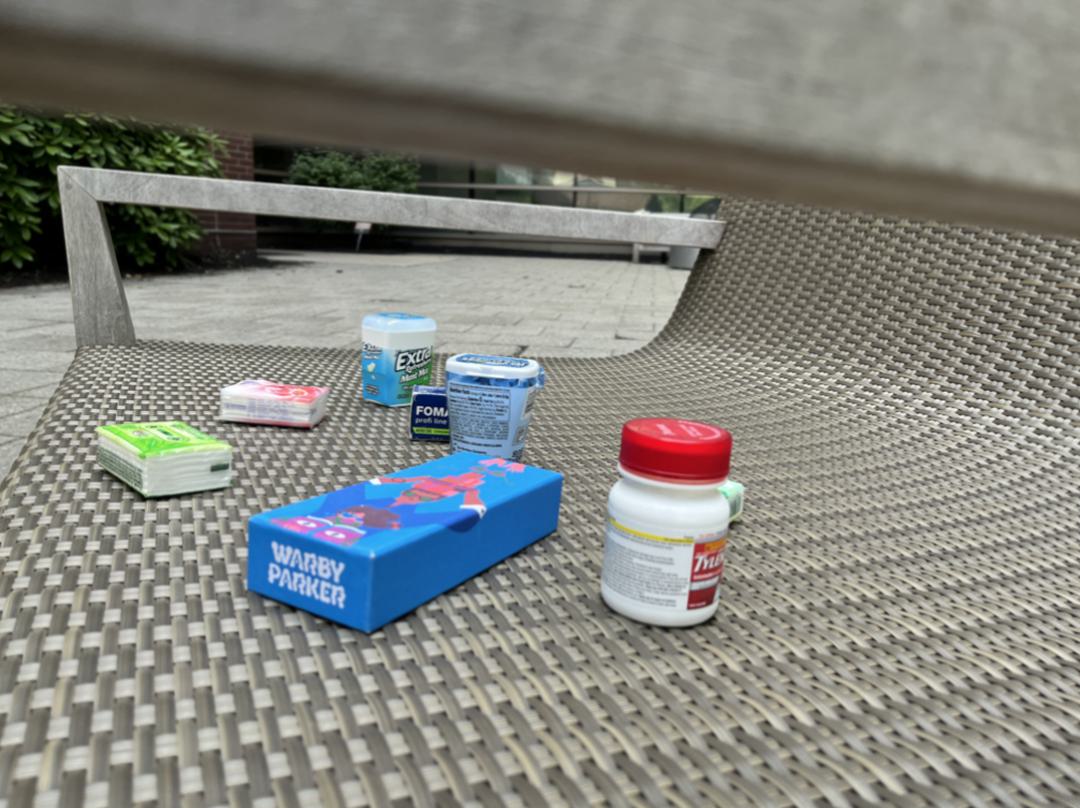} &
            \includegraphics[width=\renderwidth\textwidth]{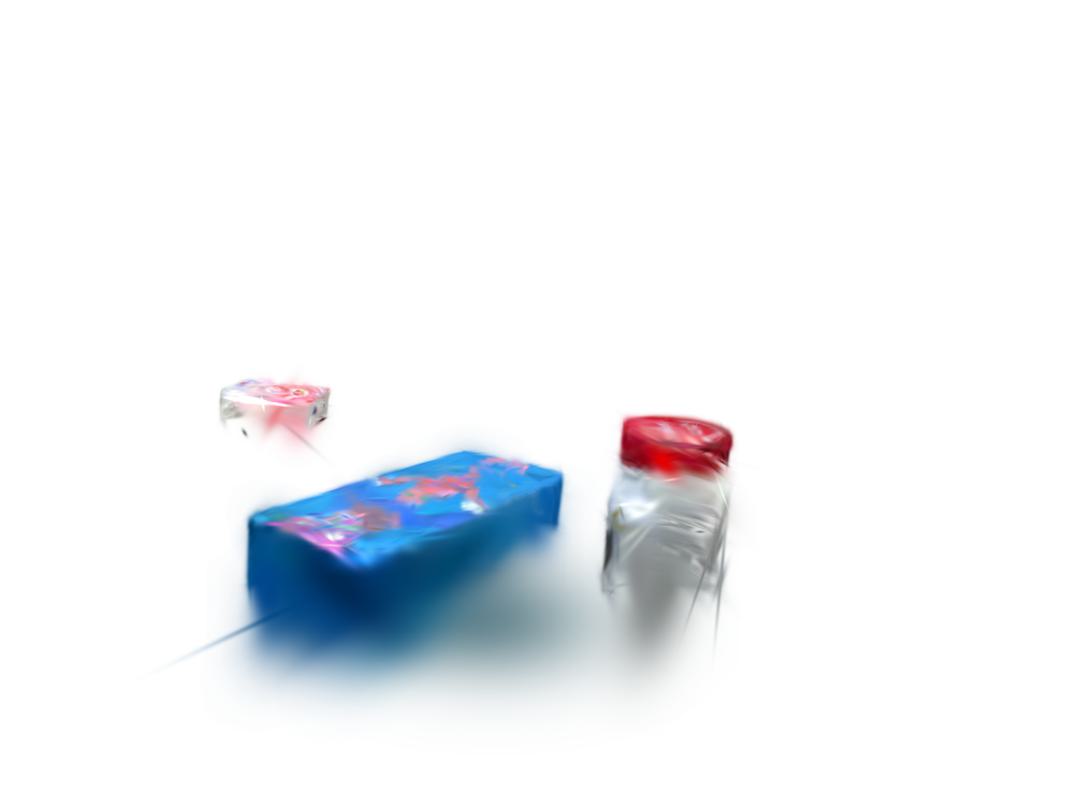} &
            \includegraphics[width=\renderwidth\textwidth]{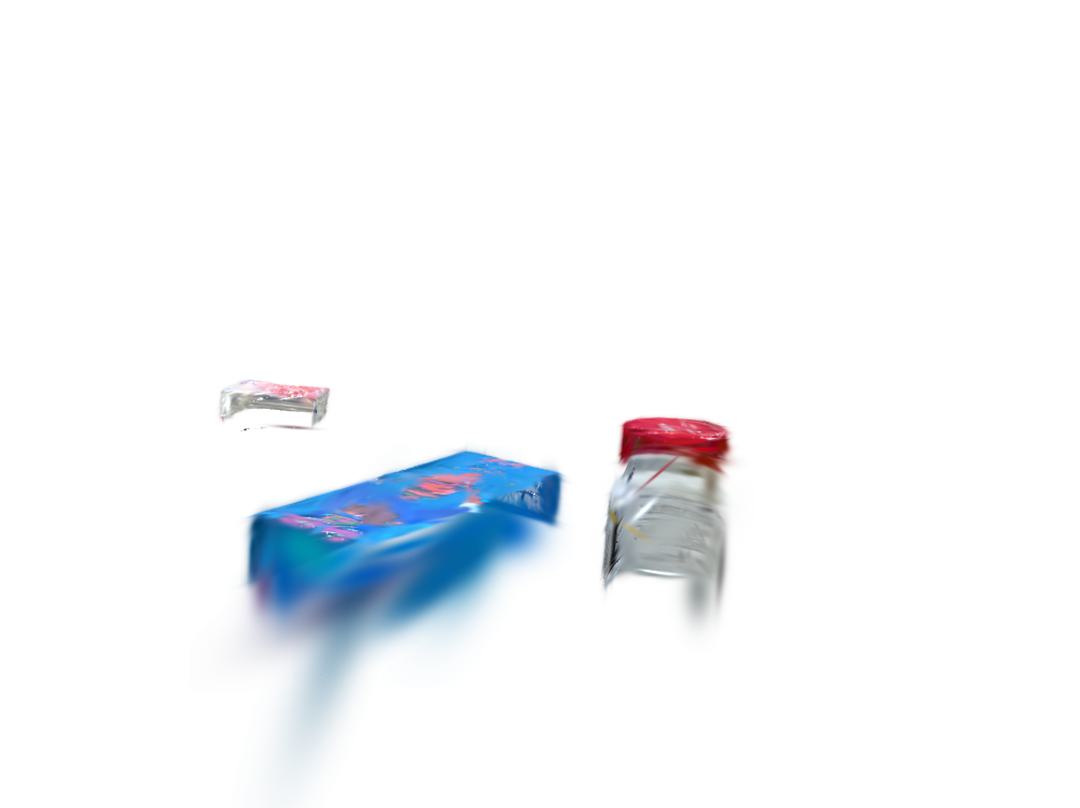} &
            \includegraphics[width=\renderwidth\textwidth]{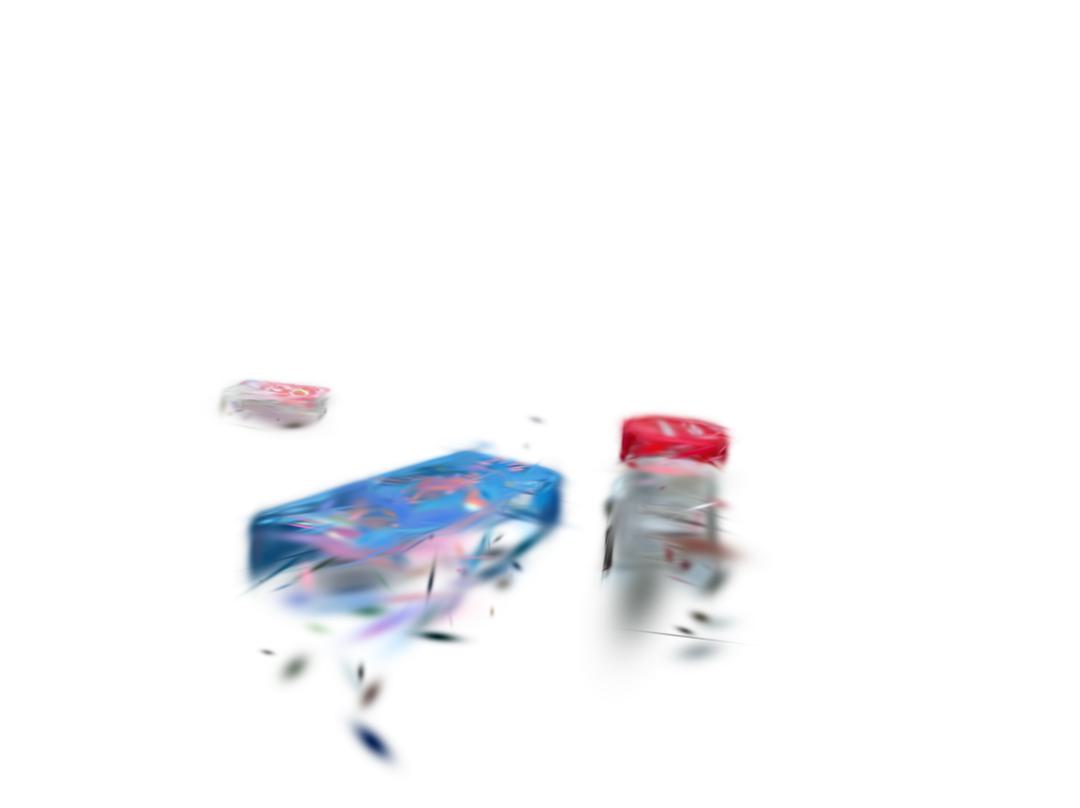} &
            \includegraphics[width=\renderwidth\textwidth]{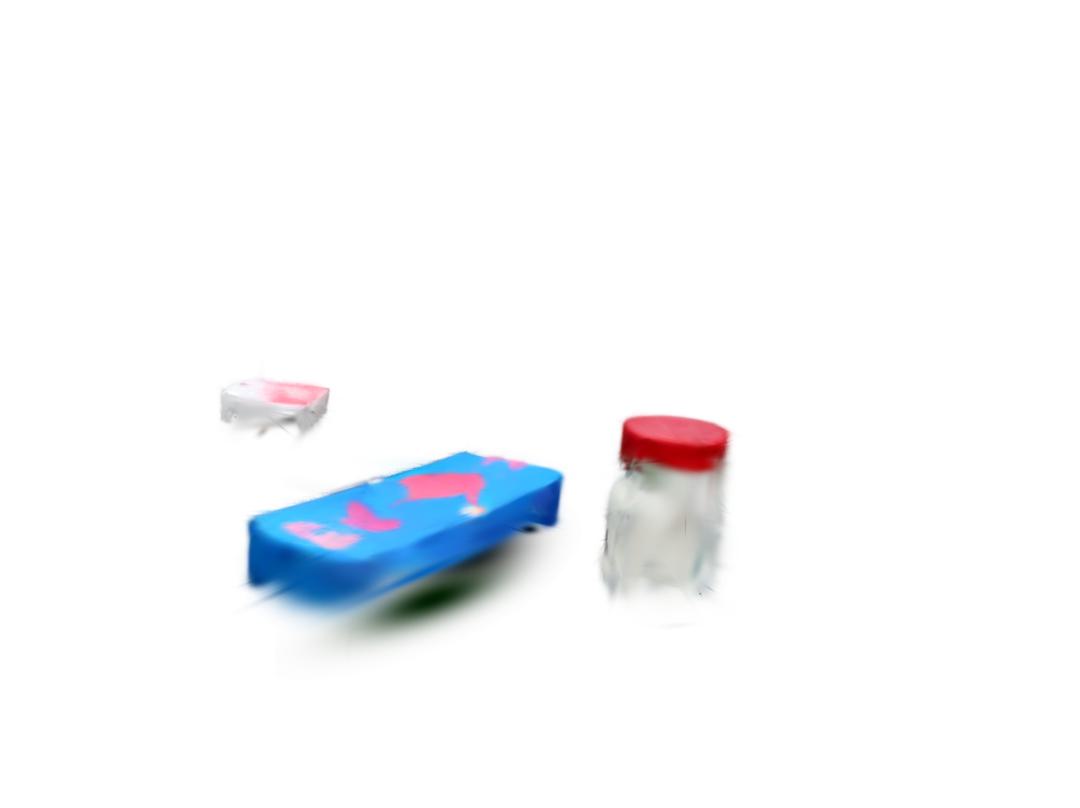} &
            \includegraphics[width=\renderwidth\textwidth]{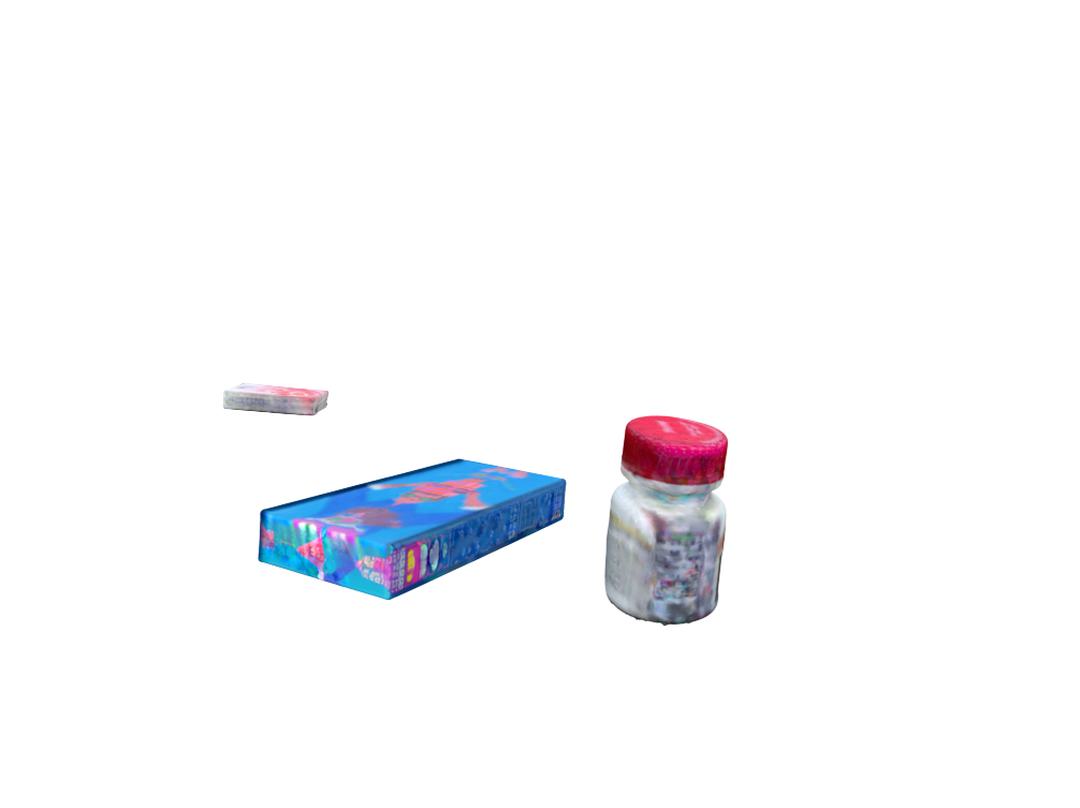} \\
            
            GT & 3DGS~\cite{3dgs} & 2DGS~\cite{2dgs} & DNGaussian~\cite{dngaussian} & GenFusion~\cite{genfusion} & Ours \\
        \end{tabular}
    }
    \vspace{-1em}
    \caption{
    Qualitative results of comparison of rendered images of target objects on unseen views under the medium difficulty setting. From top to bottom, images are chosen from Mip360-\textit{garden} \texttt{DSC08018}, Mip360-\textit{kitchen} \texttt{DSCF0860}, ToyDesk-\textit{scene1} \texttt{0056}, LERF-\textit{teatime} \texttt{frame\_00110}, 3DGS-CD-\textit{Desk} \texttt{frame\_00162}, and 3DGS-CD-\textit{Bench} \texttt{frame\_00110}, respectively.
    }
    \label{fig:qualitative_compare}
    \vspace{-2em}
\end{figure*}

\paragraph{Baselines. } 
In terms of appearance, we compare four methods: vanilla 3DGS \cite{3dgs}, 2DGS~\cite{2dgs}, depth-regularized sparse view reconstruction method DNGaussian \cite{dngaussian}, and generative reconstruction method GenFusion~\cite{genfusion}. We use SAM2~\cite{ravi2024sam} to segment target objects in test images, producing RGBA object-only images as ground truth for evaluation. We then follow Sec.~\ref{sec:preprocessing} to segment objects from the scene and render them from all test viewpoints. 

As for geometry, we also compare our results with GenFusion \cite{genfusion}, DNGaussian \cite{dngaussian}, and an additional state-of-the-art point cloud completion method, ComPC~\cite{ComPC}, to evaluate our geometry completion performance. All incomplete point clouds input to ComPC are standardized according to \citet{pcn}.  Following the appearance refinement protocol, we extract objects from the reconstructed scene as Sec.~\ref{sec:preprocessing} and evaluate reconstruction accuracy.

\paragraph{Metrics.}
To evaluate appearance quality, we adopt a comprehensive set of metrics. PSNR, SSIM, and LPIPS are used to assess the perceptual fidelity of the rendered object images. To evaluate perceptual quality, MUSIQ~\cite{musiq}, which captures overall image aesthetics, is included. Additionally, CLIP~\cite{clip} Similarity (CLIPS) is employed to measure semantic alignment between the rendered views and their corresponding real images. Furthermore, we compute the mean Intersection over Union (mIoU) between the predicted object masks and the ground-truth masks to evaluate the consistency of object position, pose, and overall shape. As for geometry, we used Chamfer Distance (CD) and Earth-Mover Distance (EMD) to quantify the geometric accuracy following the settings of \citet{ComPC}. 

\subsection{Results}
\begin{figure*}[!htb]
    \vspace{-1em}
    \centering
    \footnotesize
    \setlength{\tabcolsep}{1pt}
    \begin{tabular}{@{}ccccc@{}}
        

        \adjustbox{valign=m}{\includegraphics[trim=20pt 10pt 20pt 10pt, clip, width=0.18\linewidth]{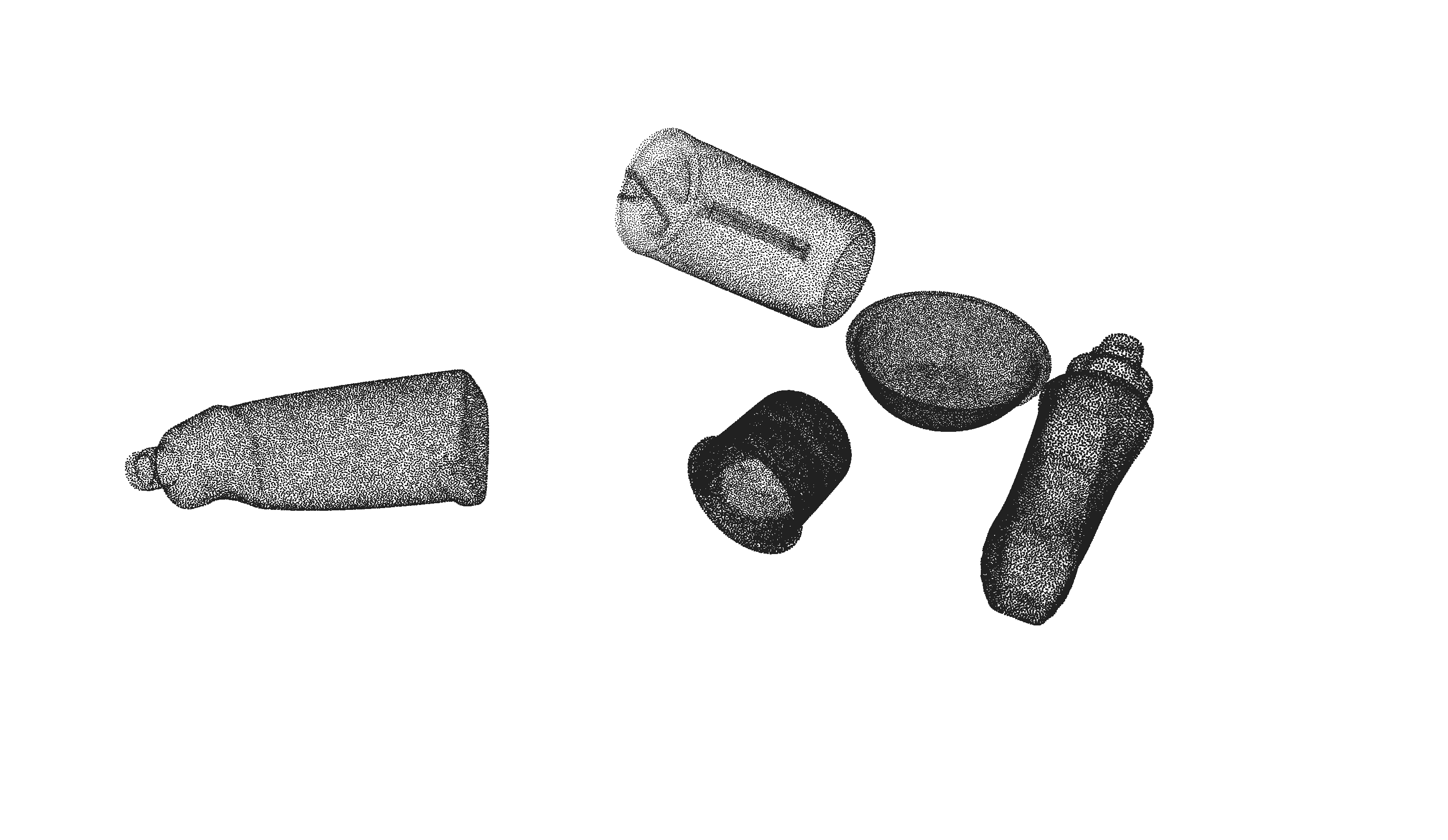}} &
        \adjustbox{valign=m}{\includegraphics[trim=20pt 10pt 20pt 10pt, clip, width=0.18\linewidth]{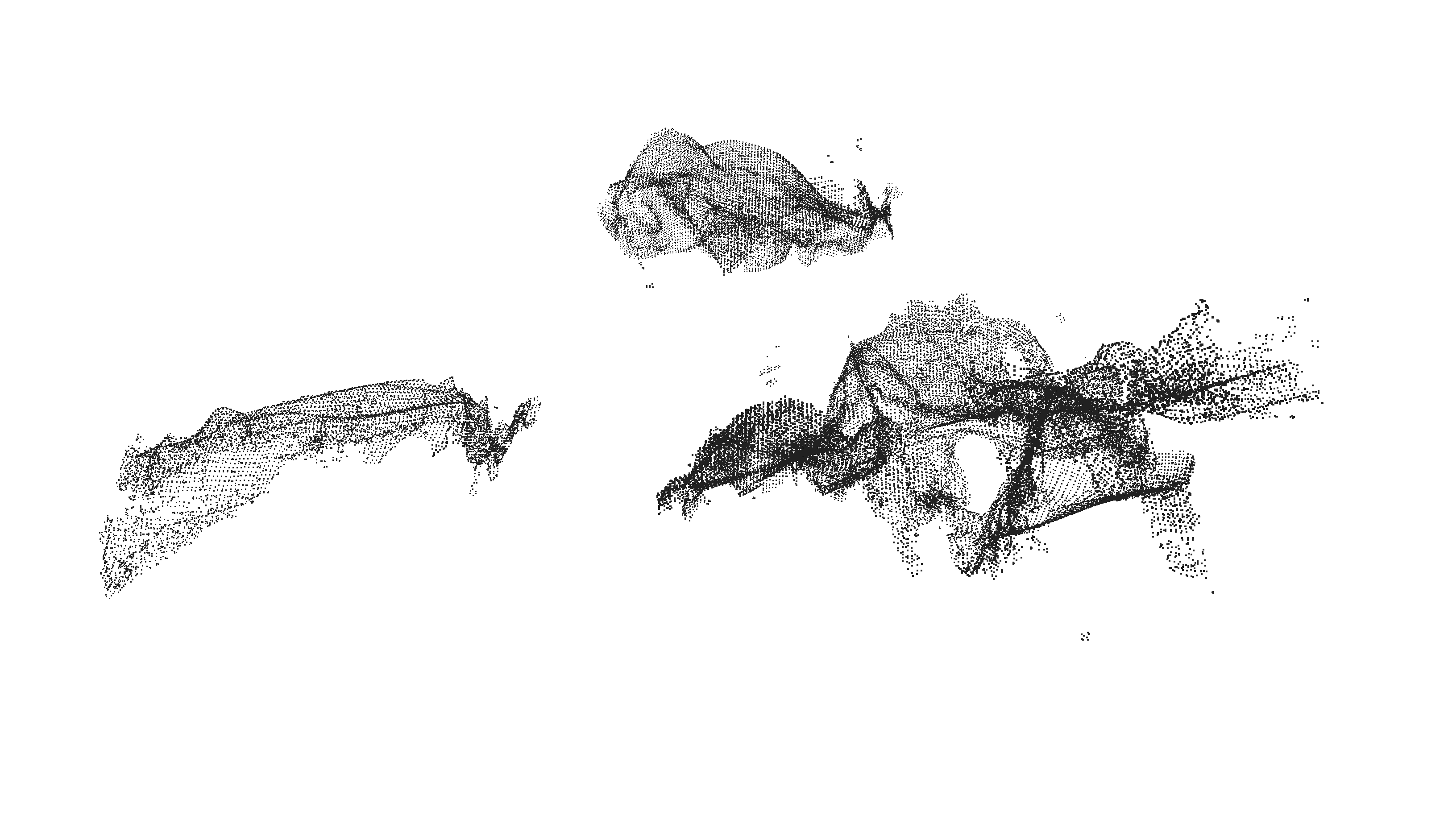}} &
        \adjustbox{valign=m}{\includegraphics[trim=20pt 10pt 20pt 10pt, clip, width=0.18\linewidth]{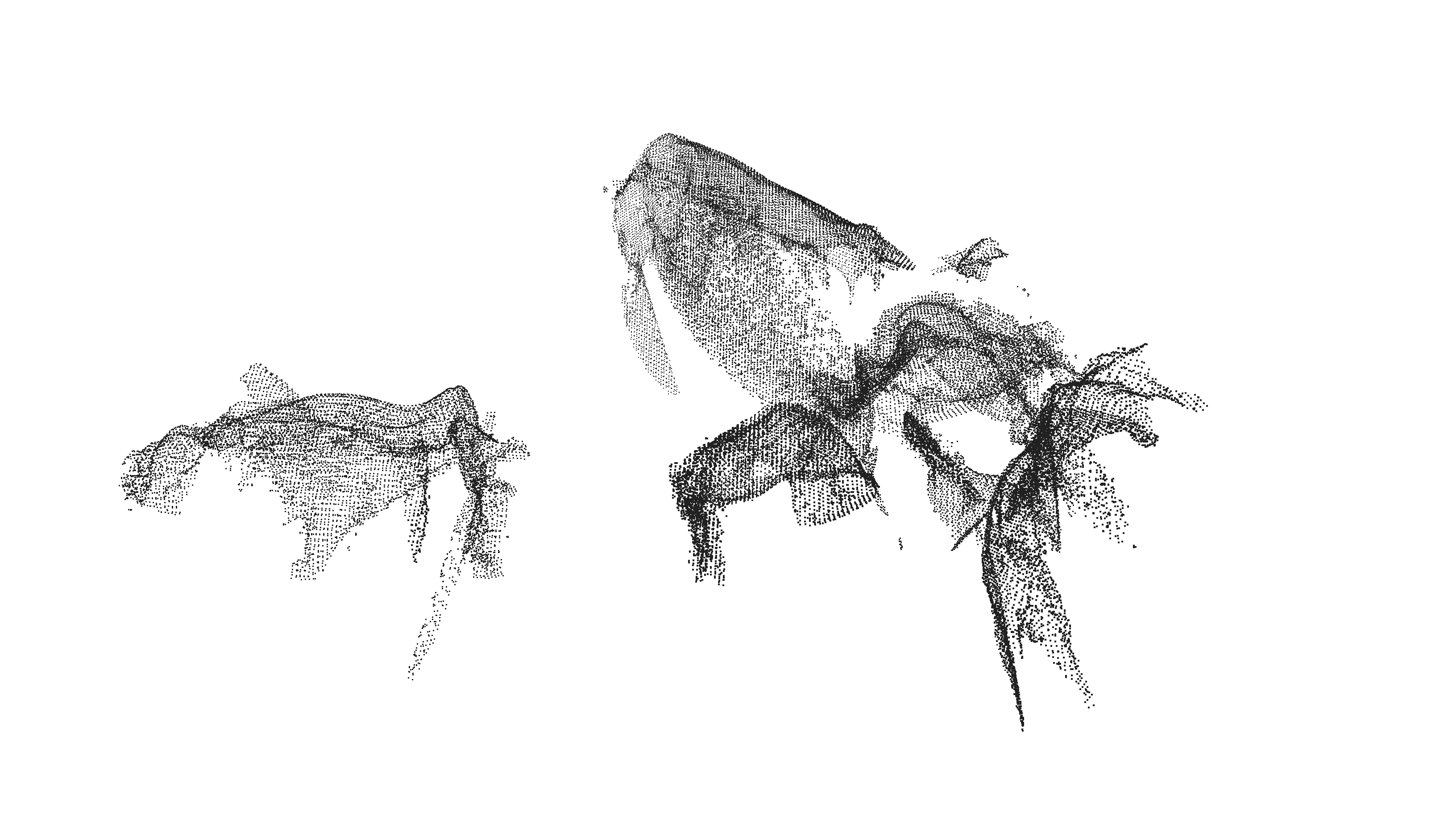}} &
        \adjustbox{valign=m}{\includegraphics[trim=20pt 10pt 20pt 10pt, clip, width=0.18\linewidth]{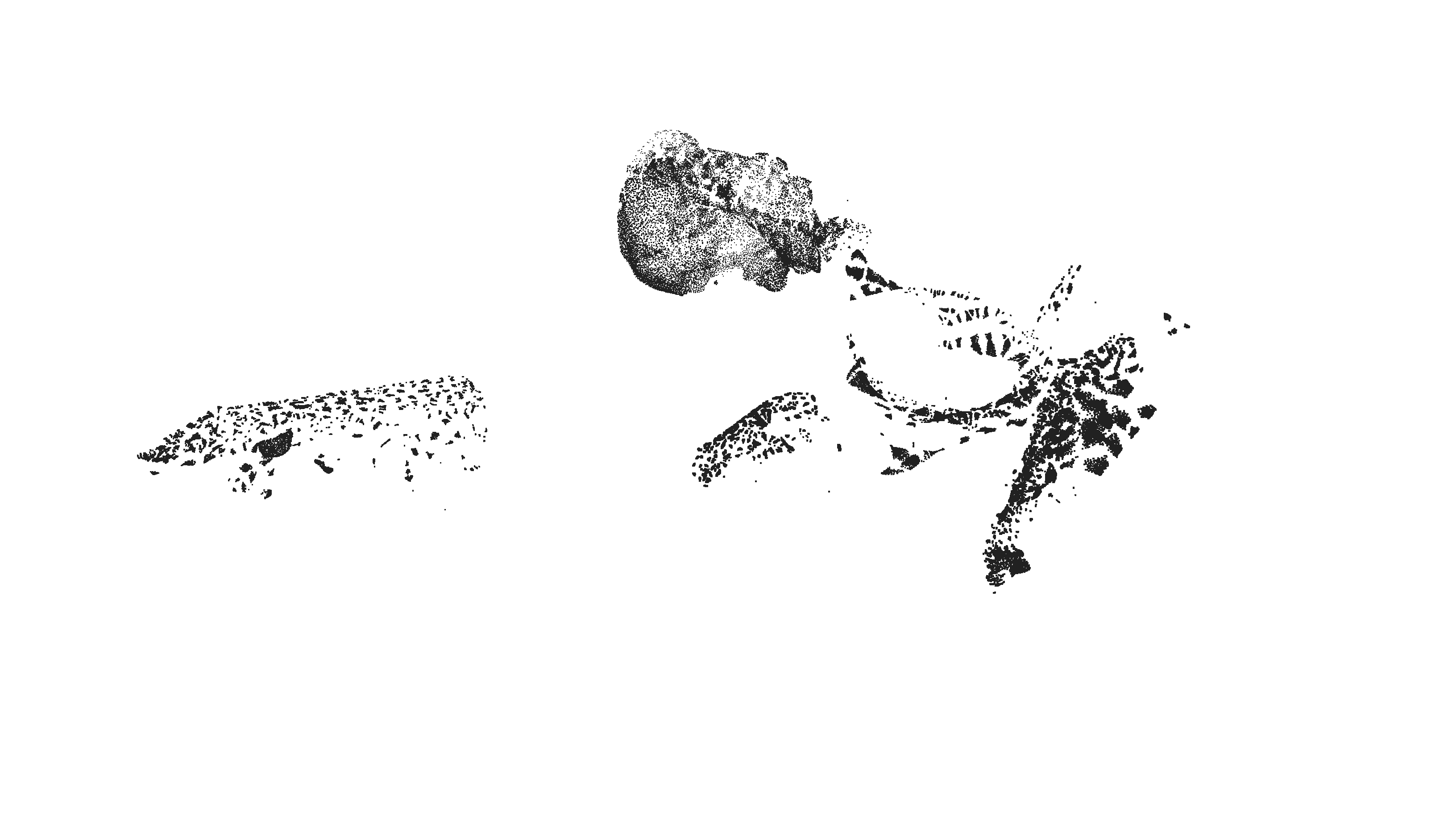}} &
        \adjustbox{valign=m}{\includegraphics[trim=20pt 10pt 20pt 10pt, clip, width=0.18\linewidth]{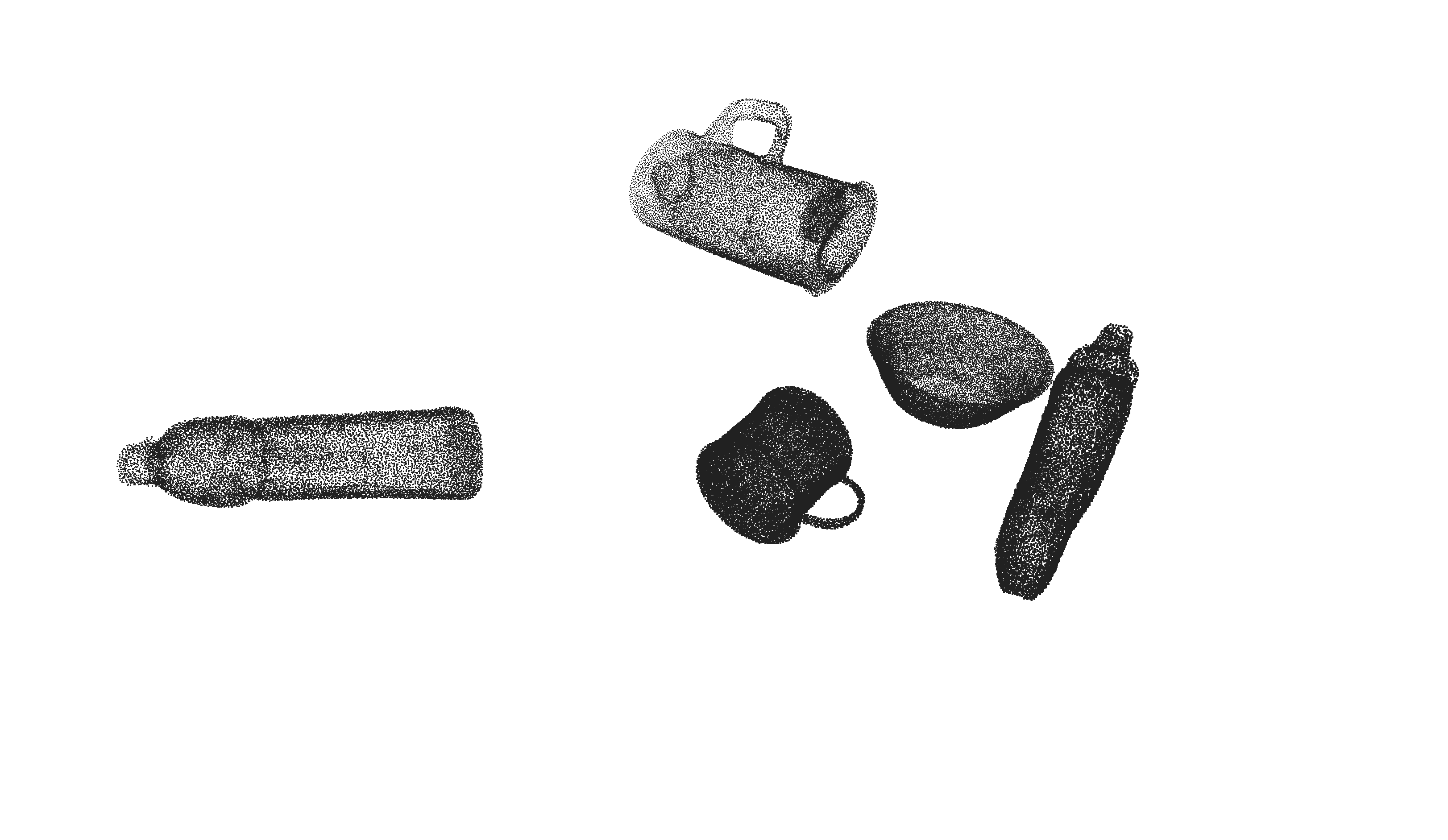}} \\

        \adjustbox{valign=m}{\includegraphics[trim=20pt 10pt 20pt 10pt, clip, width=0.18\linewidth]{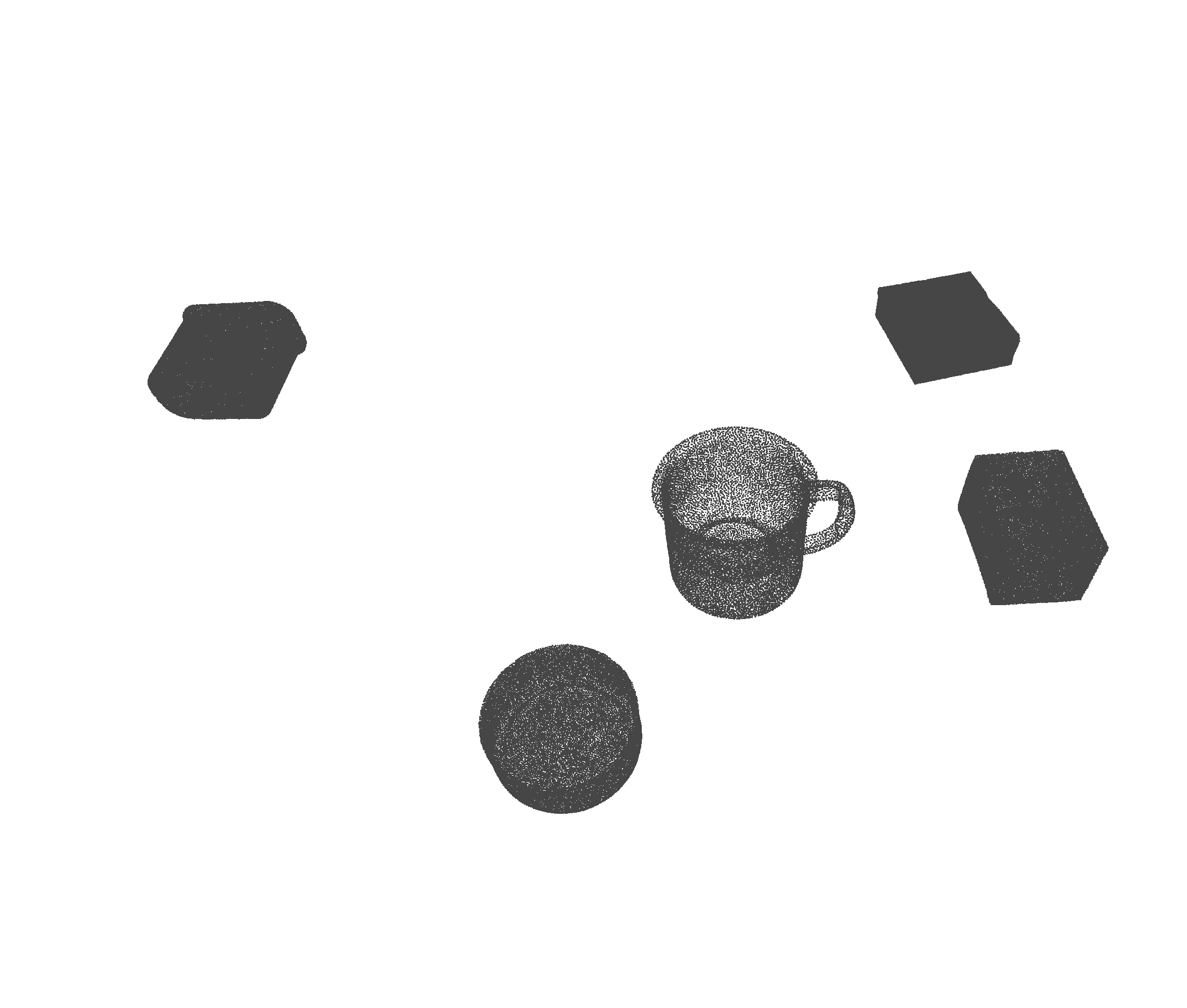}}
        & \adjustbox{valign=m}{\includegraphics[trim=20pt 10pt 20pt 10pt, clip, width=0.18\linewidth]{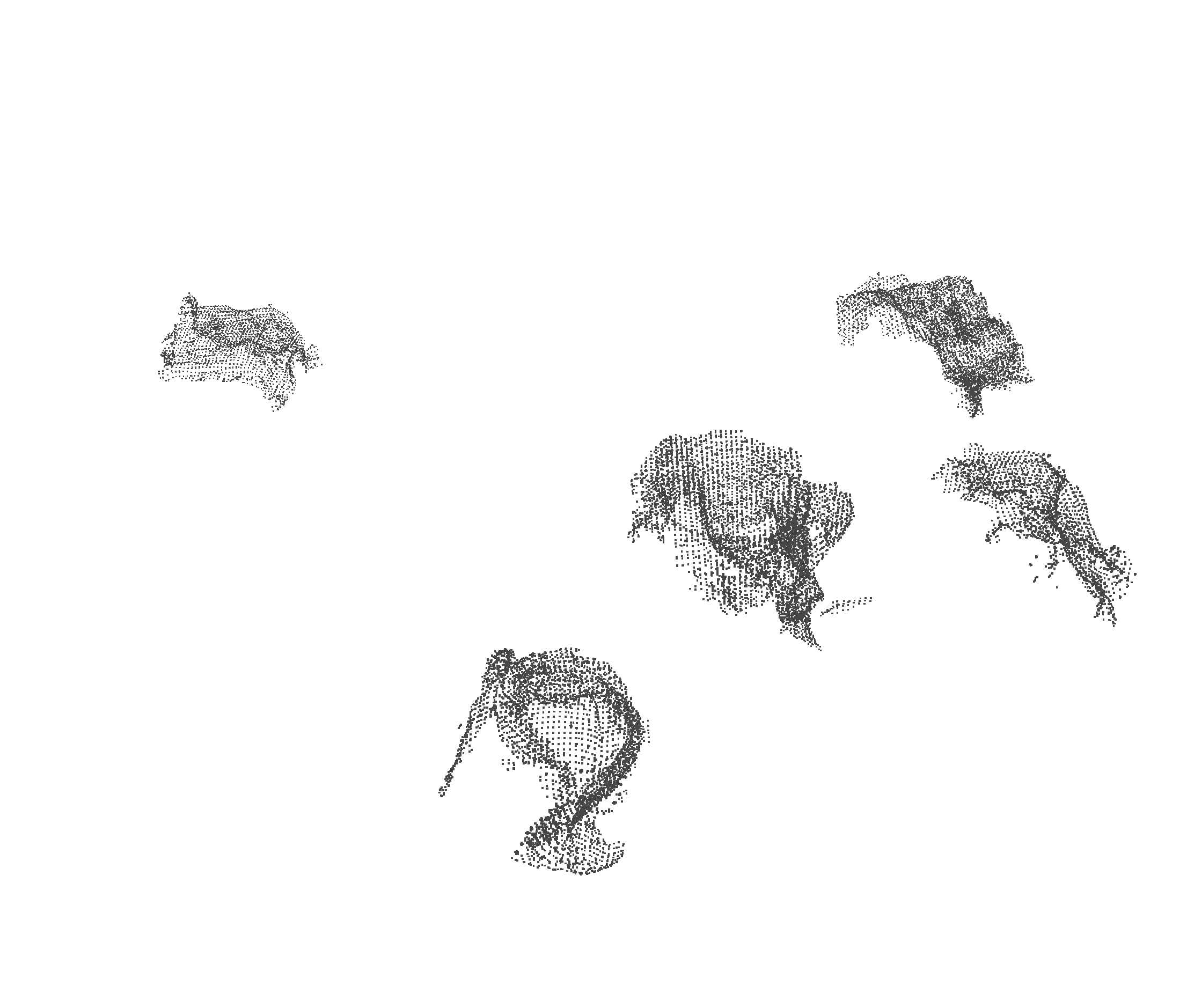}}
        & \adjustbox{valign=m}{\includegraphics[trim=20pt 10pt 20pt 10pt, clip, width=0.18\linewidth]{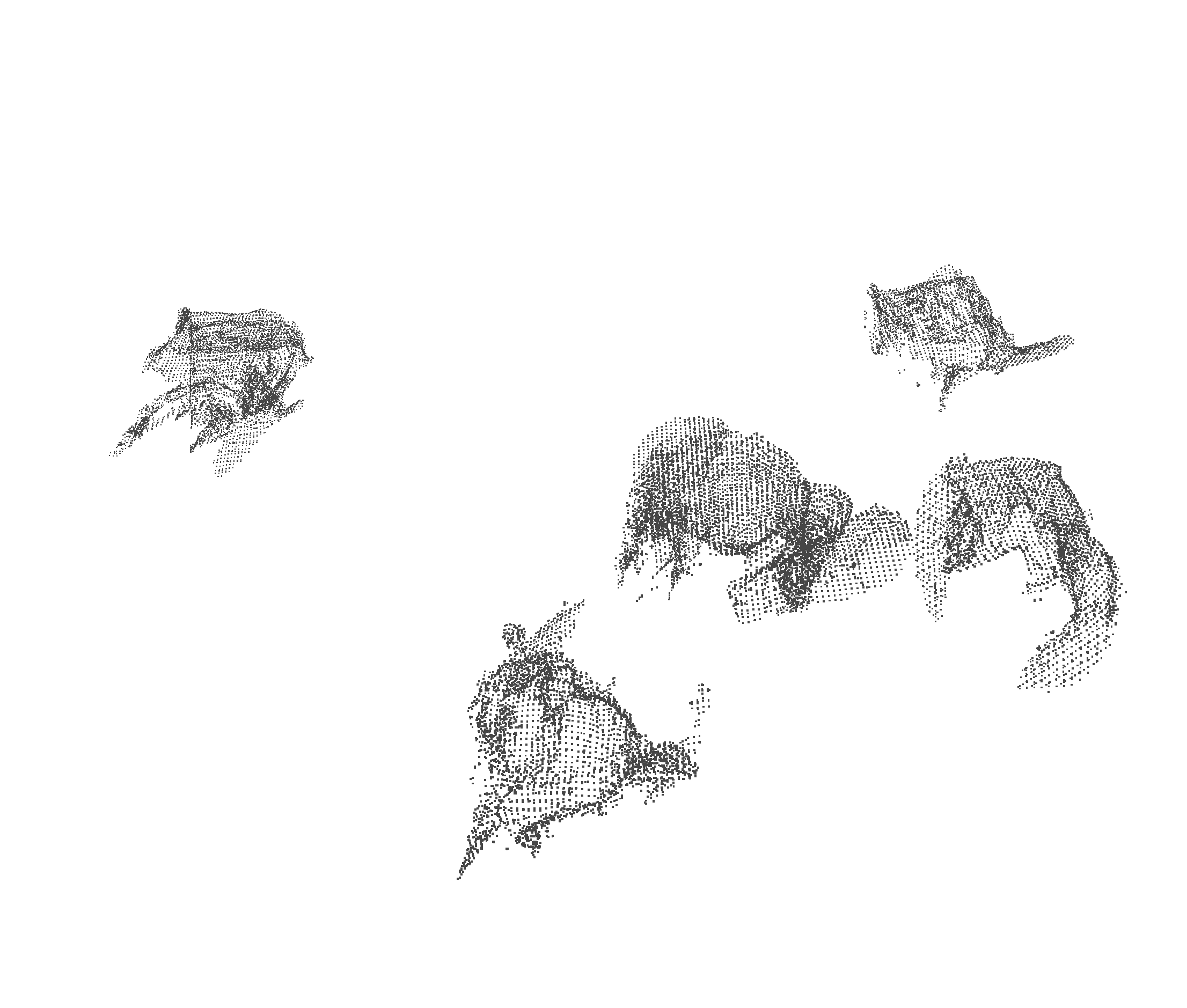}}
        & \adjustbox{valign=m}{\includegraphics[trim=20pt 10pt 20pt 10pt, clip, width=0.18\linewidth]{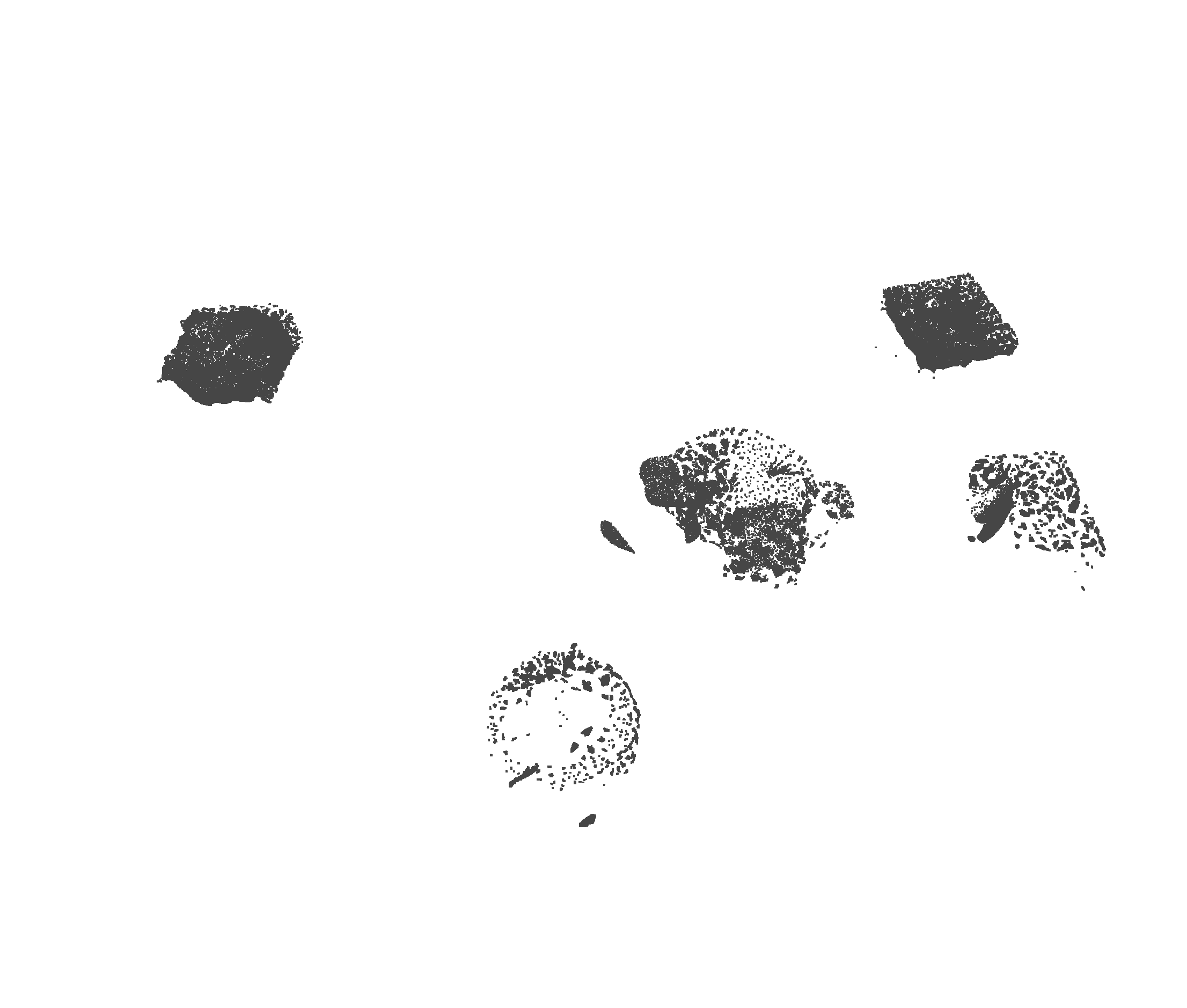}}
        & \adjustbox{valign=m}{\includegraphics[trim=20pt 10pt 20pt 10pt, clip, width=0.18\linewidth]{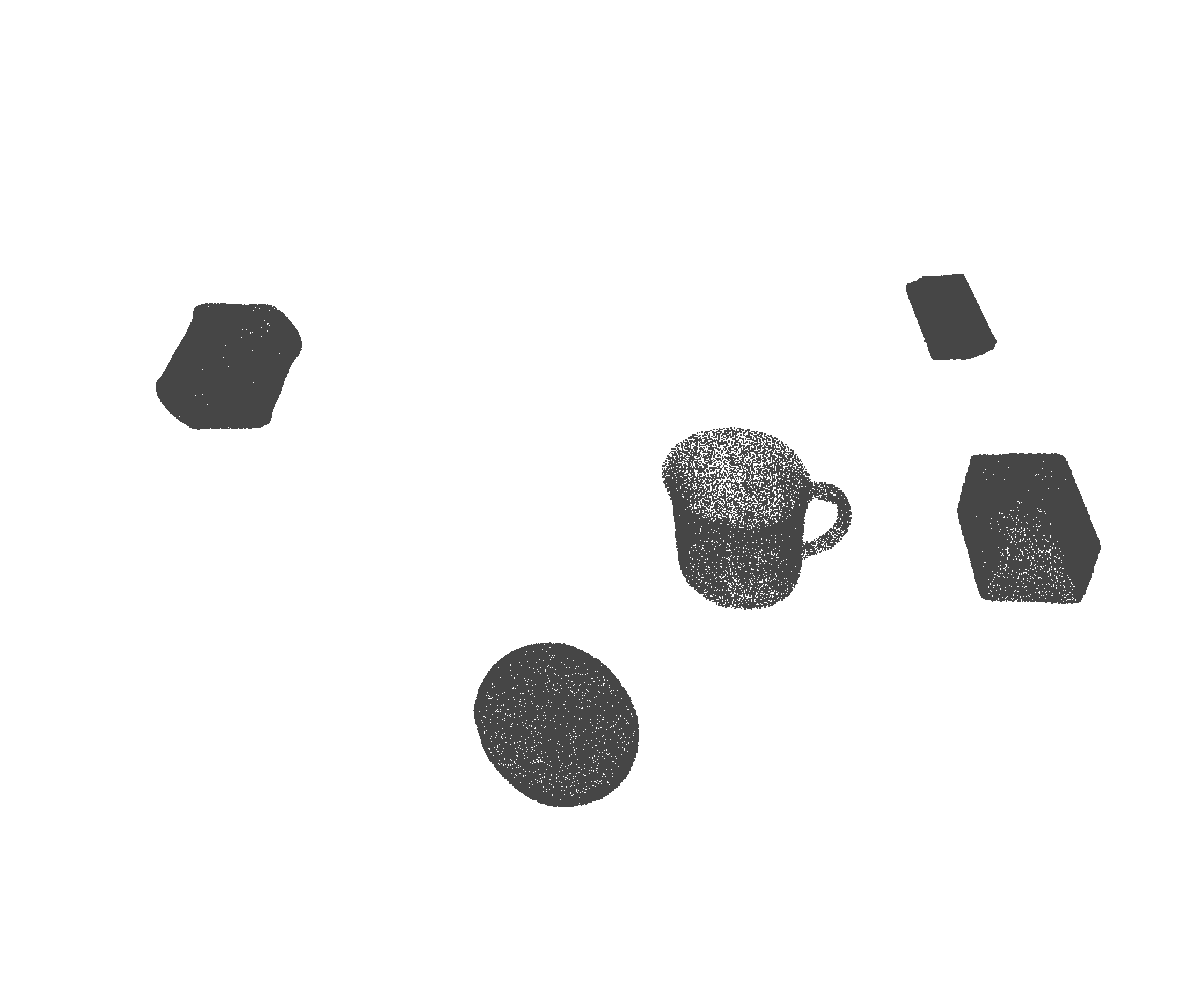}} \\

        \adjustbox{valign=m}{\includegraphics[trim=20pt 10pt 20pt 10pt, clip, width=0.18\linewidth]{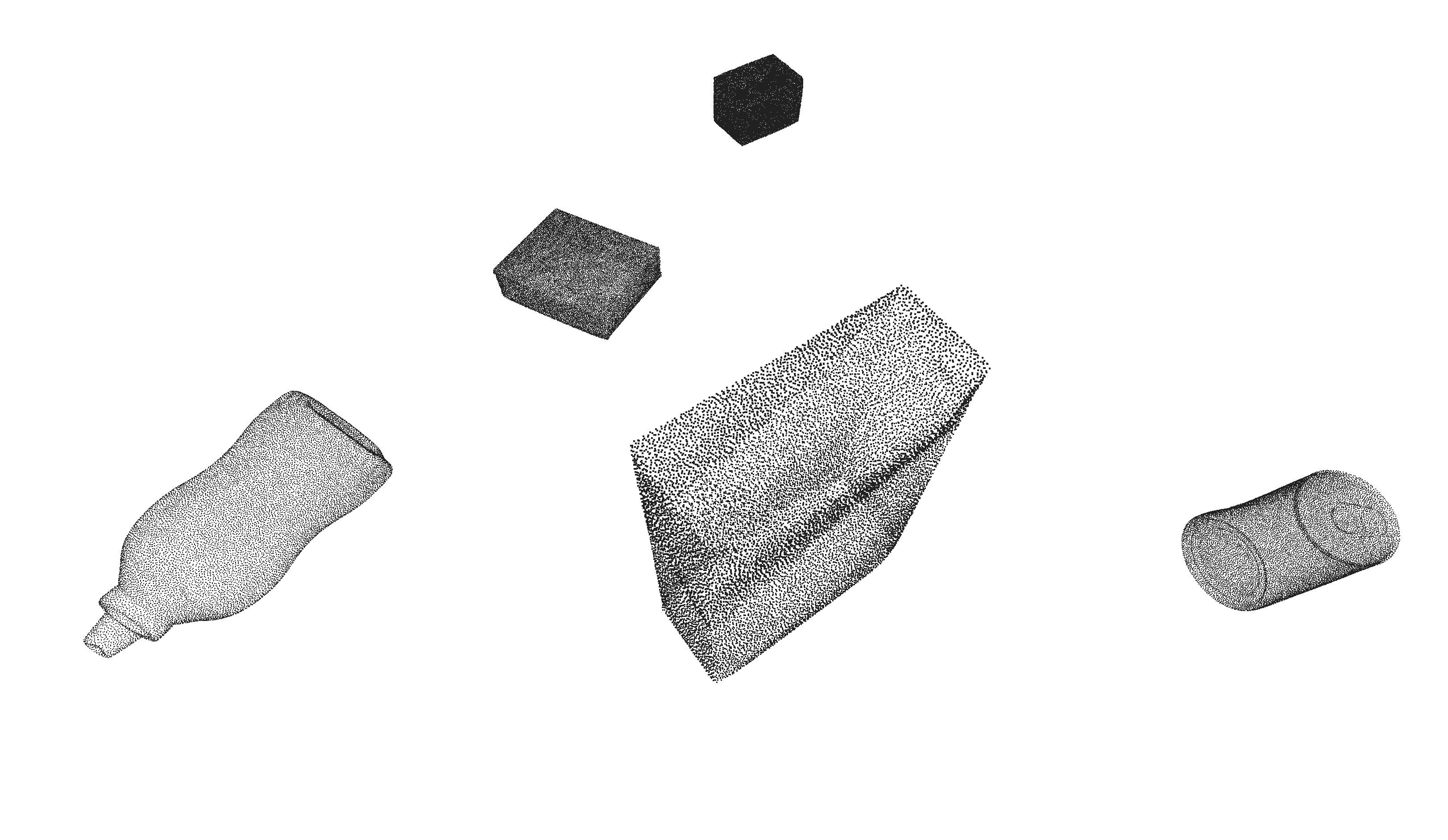}} &
        \adjustbox{valign=m}{\includegraphics[trim=20pt 10pt 20pt 10pt, clip, width=0.18\linewidth]{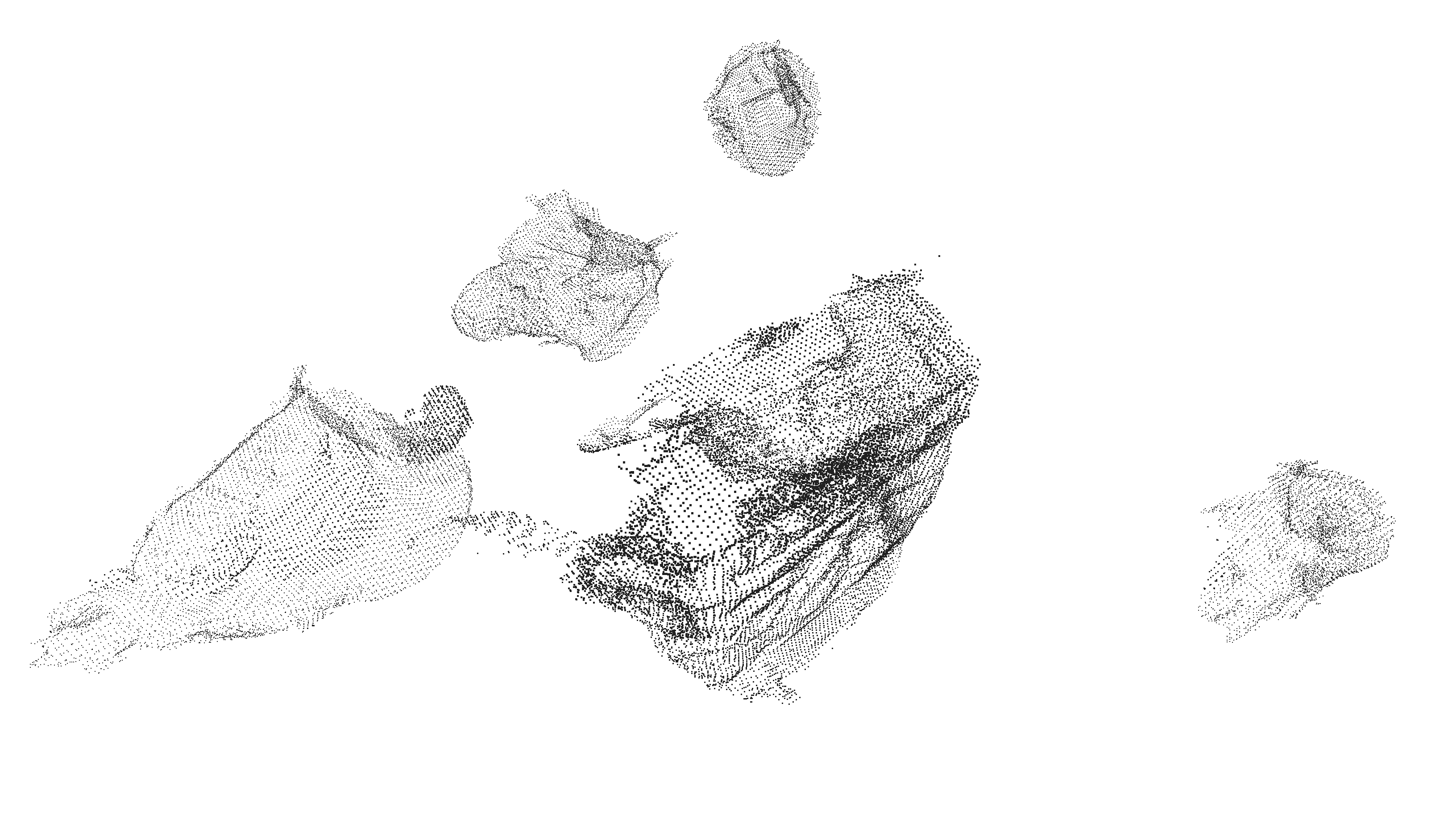}} &
        \adjustbox{valign=m}{\includegraphics[trim=20pt 10pt 20pt 10pt, clip, width=0.18\linewidth]{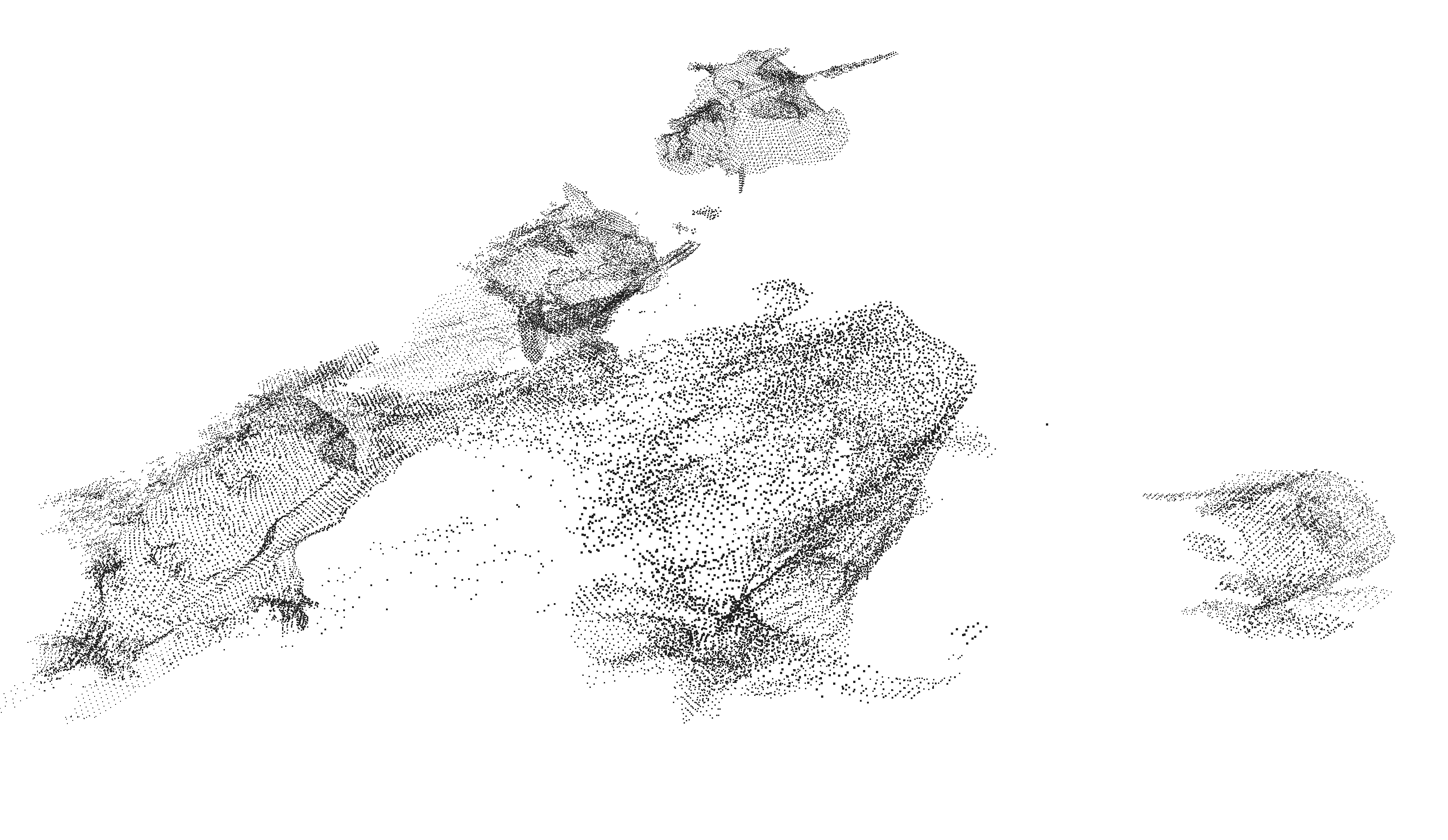}} &
        \adjustbox{valign=m}{\includegraphics[trim=20pt 10pt 20pt 10pt, clip, width=0.18\linewidth]{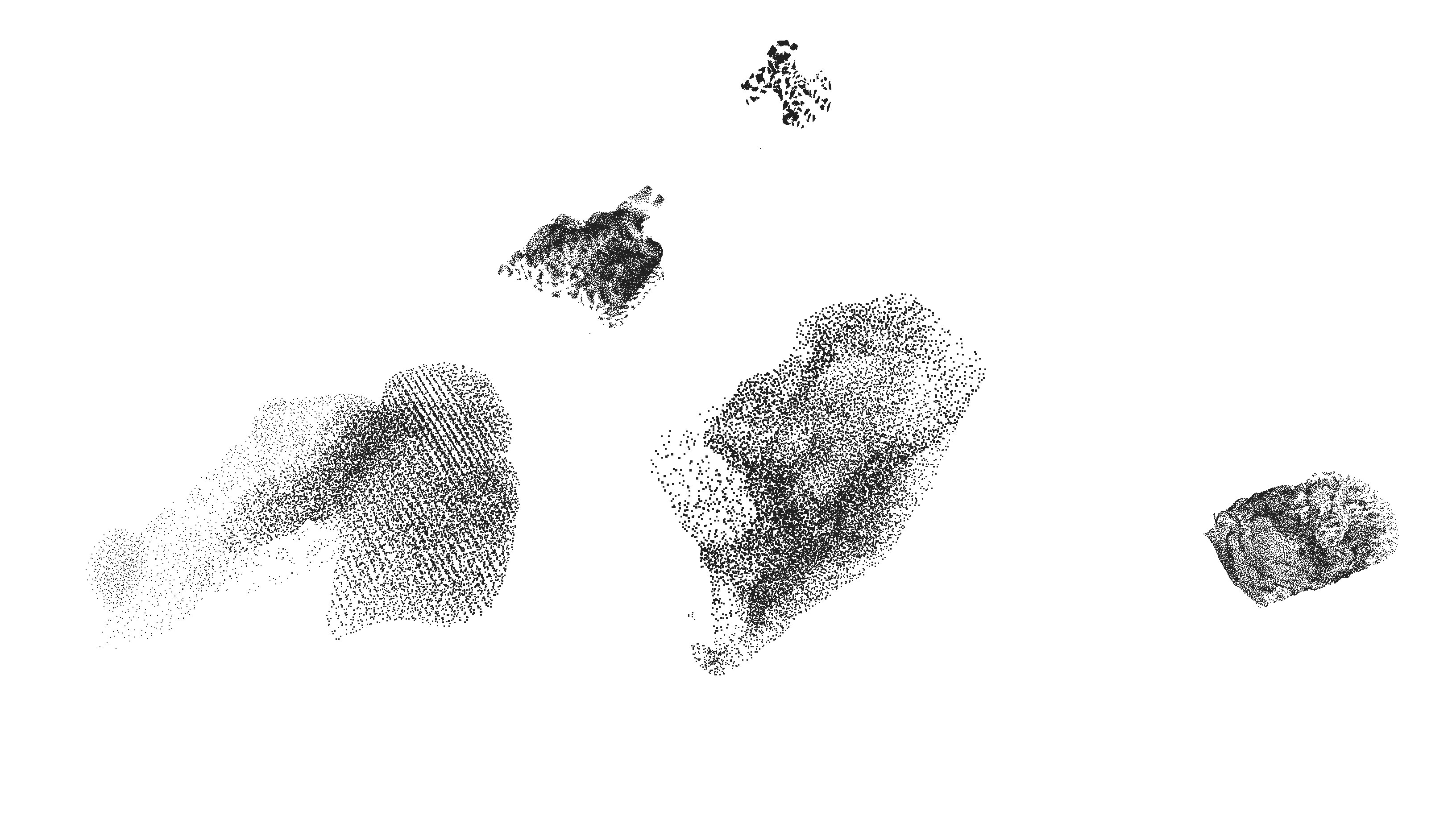}} &
        \adjustbox{valign=m}{\includegraphics[trim=20pt 10pt 20pt 10pt, clip, width=0.18\linewidth]{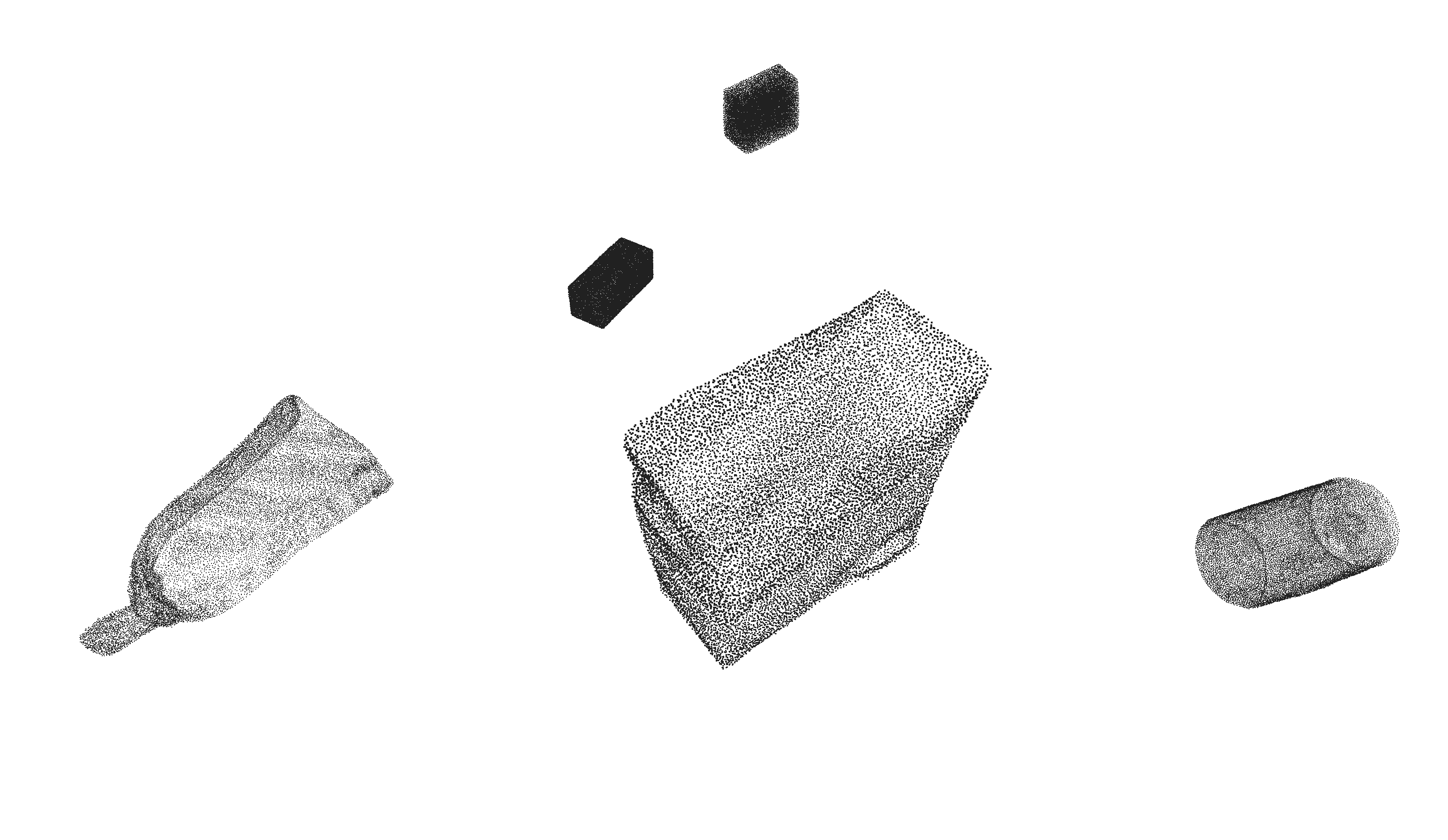}} \\
        

        GT & DNGaussian~\cite{dngaussian} & Genfusion~\cite{genfusion} & ComPC~\cite{ComPC} & Ours\\
    \end{tabular}
    \caption{Qualitative results of geometric comparisons. Our method outperforms all baselines in geometric integrity and visual effect.}
    \vspace{-2em}
    \label{fig:qualitative_compare_geo}
\end{figure*}

\vspace{-1em}
\paragraph{Appearance.} 
Quantitative comparisons under three training view configurations are summarized in Tab.~\ref{tab:quantitative-app}. \Ours consistently outperforms the baselines across most of metrics and settings. Remarkably, it achieves significant improvements in SSIM, LPIPS, and CLIPS, indicating that the refined object appearances closely match the real objects in unseen views in terms of both perceptual and semantic fidelity. \Ours also achieves significantly higher MUSIQ, showing that the native 3D generative model can plausibly hallucinate unseen regions of objects and produce realistic, artifact-free renderings. The favorable mIoU further suggests that our two-stage refinement successfully preserves object pose and silhouette, avoiding geometric drift of proxy objects. 
However, \ours does not always achieve the highest PSNR under missing views, since the generative prior can deviate from the true surface colors, causing slight color shifts. As PSNR mainly rewards pixel-wise color agreement rather than perceptual fidelity, baselines can obtain higher PSNR without delivering better visual quality (\eg, in Fig.~\ref{fig:qualitative_compare}, row 6, columns 2--6, PSNR values are 28.12, 28.20, 27.99, 28.10, and 28.06, with \ours not being the best in this case).
While PSNR is a traditionally used metric, it correlates imperfectly with perceptual quality for NVS~\cite{liang2024perceptual, jain2021putting, lin2023vision, du2023learning}.
Fig.~\ref{fig:qualitative_compare} provides qualitative comparisons between \ours and baselines across different scenes on unseen views.
In Fig.~\ref{fig:qualitative_compare}, baseline methods exhibit poor appearance details, with numerous artifacts and indistinguishable content, yielding blurry silhouettes. Meanwhile, \ours excels in preserving object appearance details, particularly in edge and texture retention, demonstrating higher realism and consistency. Extra experiments on seen views are available in the supplementary material.


\begin{table}[!ht]
    \caption{
        Quantitative comparisons of the geometry quality. The data listed are calculated on average across constructed scenarios.
    }
    \label{tab:quantitative-geo}
    \setlength{\tabcolsep}{3pt}
    \resizebox{\linewidth}{!}
    {
    \begin{tabular}{l|cccc|cccc}
        \toprule
        \multirow{2}{*}{Method} & \multicolumn{4}{c|}{Chamfer Distance [cm] $\downarrow$} & \multicolumn{4}{c}{Earth Mover Distance [cm]$\downarrow$} \\
        \addlinespace[1pt]
        \cline{2-9}
        \addlinespace[1pt]
        & $\nicefrac{2}{3}$ & $\nicefrac{4}{5}$ & $\nicefrac{6}{7}$ & Avg.& $\nicefrac{2}{3}$ & $\nicefrac{4}{5}$ & $\nicefrac{6}{7}$ & Avg. \\
        \midrule
        ComPC~\cite{ComPC}
        & \best 1.223 & \sbest 1.457 & \sbest 2.177 & \sbest 1.619 & \sbest 1.718 & \sbest 2.053 & \sbest 3.031 & \sbest 2.267 \\
        DNGaussian~\cite{dngaussian}
        & 3.183 & 3.738 & \tbest 3.699 & 3.540 & 3.700 & 4.122 & \tbest 3.851 & 3.891 \\
        GenFusion~\cite{genfusion}
        & \tbest 2.183 & \tbest 3.508 & 4.430 & 3.374 & \tbest 2.693 & \tbest 3.722 & 4.512 & 3.642 \\
        \textbf{Ours}
        & \best 1.223 & \best 1.261 & \best 1.564 & \best 1.349 & \best 1.395 & \best 1.469 & \best 1.818 & \best 1.561 \\
        \bottomrule
    \end{tabular}
    }
    \vspace{-2em}
\end{table}
\paragraph{Geometry.} 
We present quantitative results of \ours and baselines on the synthetic dataset in Tab.~\ref{tab:quantitative-geo}, showing that \ours outperforms all baselines in terms of CD and EMD metrics, indicating that \ours recovers a more plausible geometry of objects. Moreover, Fig.~\ref{fig:qualitative_compare_geo} provides a qualitative comparison of \ours under different reconstructed scenarios on an unseen view, where \ours exhibits less performance degradation despite viewpoint sparsity.

\subsection{Ablation Study}
\label{sec:ablation}

We ablate the effectiveness of each component, including Coarse Alignment (CA), Pose Alignment (PA), Shape Refinement (SR), and Appearance Refinement (AR), of our method, as shown in Tab.~\ref{tab:ablation_app_partial}.
Metrics on appearance are computed over three chosen scenes, 3DGS-CD-\textit{Bench}, 3DGS-CD-\textit{Desk}, and LERF-\textit{Donuts}. (a) and (g) are used to evaluate the effectiveness of AR, which approves the appearance consistency. (b), (c), and (g) are used to assess the effectiveness of SR. Especially, since imperfect 2D matches lead to wrong 3D–3D correspondences, the results of (b) suffer failures, where evaluation crashes caused by invalid solutions (\eg, a negative scale), confirming the numerical instability of \citet{chatrasingh2023generalized} mentioned in Sec.~\ref{sec:delicate_refine}. (d) and (g) are used to evaluate the effectiveness of PA. (f) and (g) are used to evaluate the effectiveness of CA in obtaining initial solutions, which increases the possibility of success of alignment. As for visual effects, the absence of regularization terms $\mathcal{L}_R,\mathcal{L}_S$ and orthogonal transformation term results in a distorted proxy object, as shown in Fig.~\ref{fig:ablation_demo_partial}.
\begin{table}[!htb]
  \vspace{-1em}
  \caption{
    Quantitative results of the ablation study on the rendering quality. ``$\triangle$'' indicates that SR is replaced with \citet{chatrasingh2023generalized}.
    Since Appearance Refinement only optimizes color, mIoU for (a) equals (g) and is omitted.
  }
  \label{tab:ablation_app_partial}
  \centering
  \resizebox{\linewidth}{!}{
  \begin{tabular}{cccc|l|cccc
      }
    \toprule
      CA& PA& SR& AR& & SSIM$\uparrow$& PSNR$\uparrow$ & LPIPS$\downarrow$&  mIoU$\uparrow$\\
    \midrule
    \checkmark& \checkmark& \checkmark& & a & 0.936\tbest&  21.445\tbest 
& 0.069\tbest & \diagbox{}{} \\
    \checkmark& \checkmark& $\triangle$& \checkmark& b & 0.525 
& 12.462  
& 0.476 & 0.362 \\
    \checkmark& \checkmark& & \checkmark& c & 0.940\sbest& 21.853\sbest 
& 0.066\sbest
      & 0.664\sbest \\
    \checkmark& & \checkmark& \checkmark& d & 0.797 
&16.892 
& 0.288
      & 0.289 \\
    \checkmark& & & \checkmark& e & 0.766 
&16.864 
& 0.305
      & 0.296 \\
    & \checkmark& \checkmark& \checkmark& f & 0.935 
& 21.051 
& 0.072
      & 0.618\tbest \\
    \checkmark& \checkmark& \checkmark& \checkmark& \textbf{g} & 0.943\best& 22.815\best & 0.063\best
      & 0.730\best
  \end{tabular}}
  \vspace{-2em}
\end{table}

\section{Conclusion}

To our knowledge, \ours is the first to leverage a native 3D generative model for object refinement within reconstructed scenes.
With the synthesized 3D proxy, registering it coarse-to-fine, and refining appearance under pose constraints, \ours improves appearance and geometry quality while preserving scene consistency, bridging the gap between scene-level reconstruction and object-level fidelity.

\newpage
{
    \small
    \bibliographystyle{ieeenat_fullname}
    \bibliography{reference}
}

\newpage

\appendix

\section*{Supplementary Material}

Supplementary material includes more implementation details in Sec.~\ref{app:implementation_details}, followed by experiment results in Sec.~\ref{app:additional_results}. Finally, we provide more dataset details in Sec.~\ref{sec:dataset_details}.

\section{Implementation Details}
\label{app:implementation_details}

We conducted our experiments primarily using Tesla V100-PCIE-32G GPUs. Inference time for each scene is proportional to the number of images present. Specifically, for a scene comprising approximately 100 images with a common resolution (\eg, 1080p), the complete refinement process of \ours requires about 10 minutes. A detailed analysis of time cost and VRAM usage is provided in Sec.~\ref{app:efficiency}.

\subsection{Object Segmentation}

For the object segmentation of a GS-based 3D scene, we initially identify an RGB image encompassing all relevant objects and employ Qwen \cite{bai2025qwen2} to generate initial object prompts. These prompts are subsequently refined by users through testing with Grounding DINO \cite{liu2023grounding} to obtain precise object masks in the selected image. Then, SAM2 \cite{ravi2024sam} is utilized to track and segment all objects throughout the entire RGB stream bidirectionally from the first frame, using the grounded masks as a reference.

\subsection{Proxy Objects Synthesis}

Starting from an initially selected view, we iteratively expand the input set by considering both the mask completeness of the segmented target object and view diversity to better support the generative model. As a preprocessing step, we discard the bottom 30\% of images where the target object occupies the smallest proportion of the image pixels. Subsequent selections are made from the remaining candidates. The shape completeness score $Q_{\text{shape}}$ for a given masked image is defined as:
\begin{align}
    Q_{\text{shape}} = \sum{M}/ \sum{\text{ConvexHull}(M  > 0)}
\end{align}
Here, $M > 0$ denotes the foreground region in the binary mask $M$, and $\text{ConvexHull}(M > 0)$ represents the convex hull of that region. A value of $Q_{\text{shape}}$ close to 1 indicates that the object shape is compact and complete, while lower values suggest fragmentation or missing regions in the mask. To assess viewpoint diversity, we define a viewpoint diversity score $Q_{\text{view}}$, which considers both positional and orientational differences among viewpoints:
\begin{align}
    Q_\text{view} &= 0.5 \cdot \tilde{d}_\text{pos} + 0.5 \cdot \tilde{d}_\text{rot}, \\
    \tilde{d}_\text{pos} &= \frac{d_\text{pos} - d_\text{min}}{d_\text{max} - d_\text{min} + \varepsilon}, \quad \tilde{d}_\text{rot} = d_\text{rot}
\end{align}
where $d_{\text{pos}}$ denotes the minimum Euclidean distance from the current view to the selected views, $d_{\text{min}}$ and $d_{\text{max}}$ represent the global minimum and maximum of all such distances respectively, and $\varepsilon$ is a small constant to prevent division by zero.  
Similarly, the term $d_{\text{rot}}$ indicates the maximum dot product between the camera's viewing direction (\ie, the unit vector along the $z$-axis) of the current view and those of the selected views. Since the dot product is in $[0,1]$, $d_{\text{rot}}$ is already normalized and requires no further scaling. 
The final image quality score is computed as a linear combination of the shape completeness and viewpoint diversity scores:
\begin{equation}
Q = \lambda_s Q_{\text{shape}} + \lambda_v Q_{\text{view}}.
\end{equation}
where the parameters $\lambda_s$ and $\lambda_v$ are weighting factors that balance the contribution of mask completeness and viewpoint diversity.
After selecting the complete set of input images, we perform center cropping around the target object and feed the cropped images into TRELLIS~\cite{trellis}.

\subsection{Coarse Alignment}

According to \citet{3dgs}, the covariance of a 3D Gaussian primitive can be represented as $\Sigma=RSS^\top R^\top$, where $R$ is a rotation matrix, and $S$ is a diagonal scaling matrix containing the scaling factors along the three principal axes. Obviously, $R$ and $S$ are the original parameters of the primitive. When applying a rotation $R'$ to the primitive, the parameters, position $\mu$ and rotation $R$, of the primitive are transformed as $ \mu \leftarrow R' \mu, R\leftarrow R' R$. Additionally, the SH is transformed according to \url{https://github.com/graphdeco-inria/gaussian-splatting/issues/176} from the official Gaussian Splatting \cite{3dgs} repository. Especially, when a scale $S'$ is applied to a 3D Gaussian primitive, the parameters are transformed as $ \mu \leftarrow S' \mu,S\leftarrow SS'$.

Since the point cloud formed by the positions of all Gaussian primitives in the degraded object $G^{\text{deg}}$ is incomplete and non-uniform, the Iterative Closest Point (ICP) \cite{besl1992method} method used in the initial alignment has a high probability of converging to a local optimal solution or failing. To address this limitation, we adopt a simple heuristic method to find the transformation for initial alignment. First, we randomly generate ${128}^2$ quaternions, each corresponding to a potential rotation matrix. Then, we traverse and select $128$ rotation matrices that form the largest angles with the already selected rotation matrices. These selected matrices serve as the initial rotations for the ICP process. Moreover, the initial translation is nearly ${\mathbf{0}}$ as a result of the alignment of the centroids between the proxy object and the degraded object. Next, we perform ICP optimization for each initial rotation matrix and finally select the transformation result with the best matching index as the transformation for initial alignment. We use the ICP implementation in Open3D \cite{zhou2018open3d}, setting the maximum correspondence points-pair distance to $0.16$ times the arithmetic mean on the dimension of the 3D bounding box of the degraded object $G^\text{deg}$ mentioned, and maximum number of iterations to $400$.

\subsection{Matching-Based Pose Adjustment}

As for the matching process of a single object, before inputting the masked target image $I^\text{deg}$ and the rendered image $I^\text{prx}$ into MASt3R \cite{leroy2024grounding} for object matching, we need to preprocess these two images. First, to reduce the subsequent computational load, we crop the images to focus on the area where the object mask or its projection is located. Next, we apply a padding of 200 pixels to the cropped images. This padding is intended to maintain a visually coherent appearance of the object in the image, which aids in the feature matching process of MASt3R.

For a pair of target image $I^\text{deg}$ and rendered image $I^\text{prx}$ in a dataset, we select the top $16$ matching points based on the confidence. However, when the number of images $|\mathcal{I}|$ in a scene is large, we further reduce the time overhead caused by matching by employing an equidistant sampling strategy with a step size of $|\mathcal{I}|/{15}$. Due to the low quality of the original images from certain viewpoints or occlusions of the target object, the final number of point pairs projected back into 3D space using depth maps and camera parameters will be close to $N_c$ (i.e., $\mathcal{P}^\text{prx}, \mathcal{P}^\text{deg}\in \mathbb{R}^{3\times N_c}$).

Notice that the problem addressed by Umeyama \cite{umeyama1991least} is a least-squares problem, motivating the use of RANSAC \cite{fischler1981random} to improve robustness. In our implementation, we have developed our own version of RANSAC, setting the maximum number of iterations to $2000$ and the inlier distance threshold to $0.01$ times the arithmetic mean on the dimension of the 3D bounding box of the degraded object $G^\text{deg}$.

\subsection{Scale-Undistorted Shape Refinement}

As for a rotation matrix $R=\exp{(\theta[u]_\times)}$, $u=[v_1,v_2,v_3]^\top$ is the rotation axis and $\theta$ is the rotation angle, where $||u||^2=(v_1^2+v_2^2+v_3^2)=1$.
$||[u]_\times||^2_F=\operatorname{Tr}\,([u]_\times^\top[u]_\times)$ equals the sum of the squares of all entries of $[u]_\times$,
\begin{align}
    ||[u]_\times||^2&=\sum_{i,j}{([u]_\times)_{i,j}^2}\nonumber\\
    &=2(v_1^2+v_2^2+v_3^2)=2
\end{align}
Therefore, $\mathcal{L}_R$ is the square of the rotation angle:
\begin{align}
    \mathcal{L}_R&=\tfrac{1}{2}||\log{R}||^2_F=\tfrac{1}{2}||\theta [u]_\times||^2\nonumber\\
    &=\theta^2\cdot\tfrac{1}{2}||[u]_\times||^2=\theta^2,
\end{align}
, limiting the angle of the rotation matrix. 
In the parameterization of the shape refinement problem, to make the scale parameter $S=\operatorname{diag}(s_1,s_2,s_3)$ more controllable, we set minimum and maximum values for the scale, denoted as $s_\text{min}$ and $s_\text{max}$, respectively. Specifically, for an original parameter $s_o \in \mathbb{R}$, we use the sigmoid function $f: x\mapsto \frac{1}{1+e^{-x}}$ to map it to the interval $(0,1)$. Then, we apply a simple linear transformation $s = s_\text{min} + (s_\text{max}-s_\text{min})f(s_o)$ to map it to the interval $(s_\text{min},s_\text{max})$.
This operation is considered as a part of the regularization term $\mathcal{L}_S$ for $S$.
We set $s_\text{min}=0.75$, $s_\text{max}=1.5$, $\lambda_S=2\times 10^{-5}$, and $\lambda_R=1\times 10^{-4}$ in our implementation.
To solve the proposed optimization objective, Adam \cite{kingma2014adam} optimizer is used with $3000$ iterations, a learning rate of $10^{-3}$, $\beta_1=0.9$, $\beta_2=0.999$, and $\epsilon=10^{-8}$. We utilize PyTorch for the optimizer and automatic differentiation.

\begin{algorithm}[!ht]
\caption{Gaussian Model Alignment}
\begin{algorithmic}[1]
\ENSURE{
\quad\\
$G^\text{deg}$: GS-based original degraded object \\
$G^\text{prx}$: GS-based proxy object after coarse alignment \\
$\mathcal{T}=\{T_i\}_{i=1}^{N_c}$: Camera extrinsic set \\
$K$: Camera intrinsic,
$\mathcal{I}=\{I_i\}_{i=1}^{N_c}$: Input images \\
$\mathcal{M} = \{M_{I_i}\}_{i=1}^{N_c}$: Binary masks for the target object \\
$T_{\text{max}}$: Max iterations \\
$\mathscr{T}$: Set of Shape Refinement iteration serial numbers
}
\REQUIRE{Aligned proxy object $G^\text{prx}$}

\FOR{$t \gets 1$ \textbf{to} $N_{\text{max}}$}
    \STATE $\mathcal{C}\gets \varnothing$
    \FOR{$n \gets 1$ \textbf{to} $N_c$}
        \STATE $I^\text{prx}_n,D^\text{prx}_n \gets \text{Render}(T_n, G^\text{prx})$ \\
        \STATE $\{(u^\text{prx}_i,u^\text{deg}_i)\}_{i=1}^{N_m} \gets \operatorname{Match}(I_n^\text{prx},M_{I_n}\odot I_n)$
        \STATE $\{(P^\text{prx}_i,P^\text{deg}_i)\}_{i=1}^{N_m}\gets \operatorname{UnProject}(\{(u^\text{prx}_i,u^\text{deg}_i)\}_{i=1}^{N_m},D^\text{prx}_n,D^\text{deg}_n,T_n,K)$
        \STATE $\{P^\text{prx}_i)\}_{i=1}^{N_m}\gets \operatorname{UnProject}(\{u^\text{prx}_i\}_{i=1}^{N_m},D^\text{prx}_n,T_n,K)$
        \STATE $\{P^\text{deg}_i)\}_{i=1}^{N_m}\gets \operatorname{UnProject}(\{u^\text{deg}_i\}_{i=1}^{N_m},D^\text{deg}_n,T_n,K)$
        \STATE $\mathcal{C}\gets \mathcal{C}\cup \{(P^\text{prx}_i,P^\text{deg}_i)\}_{i=1}^{N_m}$
    \ENDFOR
    \IF{$t\notin \mathscr{T}$}
        \STATE $(\hat{R},\hat{t},\hat{S})\gets\underset{R,t,s}{\operatorname{argmin}}\underset{(P^\text{prx},P^\text{deg})\in\mathcal{C}}{\sum}\left\|sRP^\text{prx}+t-P^\text{deg}\right\|^2$
        \STATE $\hat{R}'\gets I$
    \ELSE
        \STATE $(\hat{R},\hat{t},\hat{S}, \hat{R}')\gets\underset{R,t,S,R'}{\operatorname{argmin}}\Bigl\{\underset{(P^\text{prx},P^\text{deg})\in\mathcal{C}}{\sum}\left\|RR'^\top SR'P^\text{prx}+t-P^\text{deg}\right\|^2+\lambda_R\mathcal{L}_R+\lambda_S\mathcal{L}_S\Bigl\}$
    \ENDIF
    \STATE $G^\text{prx} \gets \operatorname{Rotate}(G^\text{prx},\hat{R}')$ \\
    \STATE $G^\text{prx} \gets \operatorname{Scale}(G^\text{prx},\hat{S})$ \\
    \STATE $G^\text{prx} \gets \operatorname{Rotate}(G^\text{prx},\hat{R}'^\top)$ \\
    \STATE $G^\text{prx} \gets \operatorname{Rotate}(G^\text{prx},\hat{R})$ \\
    \STATE $G^\text{prx} \gets \operatorname{Translate}(G^\text{prx},\hat{t})$ \\
    
\ENDFOR
\RETURN $G^\text{prx}$\\

\end{algorithmic}
\label{alg:alignment}
\end{algorithm}

\subsection{Iterative Optimization}

Note that the initial pose and scale alignment would be so rough that the results of 2D-2D correspondences and further 3D-3D correspondences may be inaccurate. Thus, the Correspondence Matching and Optimization are iteratively conducted to improve the matching and alignment results. Usually, there are $6$ iterations overall, the first $3$ iterations and the last are used for Pose Adjustment, and the remaining $2$ iterations are used for Shape Refinement.
Alg.~\ref{alg:alignment} summarizes the Iterative Optimization procedure for Pose Adjustment and Shape Refinement of the proxy object.

\subsection{Pose-Constrained Appearance Refinement}
To ensure that the surface texture of  $G^\text{prx}$ remains consistent with the training views under visible viewpoints, we perform a fine-tuning step. In order to preserve the previously established alignment of pose and shape, we adjust only the spherical harmonic coefficients of $G^\text{prx}$ without applying any adaptive density control, keeping the number and orientation of Gaussian points unchanged. The fine-tuning is conducted for a total of 600 iterations.

\begin{table}[!ht]{
    \caption{
        Additional quantitative comparisons of rendering quality were obtained using \ours alongside several baseline methods. The \colorbox{colorFst}{best}, the \colorbox{colorSnd}{second best}, and the \colorbox{colorTrd}{third best} are highlighted. Due to table linewidth constraints, partial method names and scene names have been abbreviated (\eg, ``DN.'' denotes DNGassian~\cite{dngaussian} and ``show\_r.'' represents the scene \textit{show\_rack}).
    }
    \vspace{-1em}
    \setlength{\tabcolsep}{1pt}
    \resizebox{1\linewidth}{!}{
    \begin{tabular}{l|
        ccccc|
        ccccc|
        ccccc}
        \toprule
         & \multicolumn{5}{c|}{SSIM$\uparrow$} & \multicolumn{5}{c|}{PSNR$\uparrow$} & \multicolumn{5}{c}{LPIPS$\downarrow$} \\
         \addlinespace[1pt]\cline{2-6}\cline{7-11}\cline{12-16}\addlinespace[1pt]
 & 3DGS & 2DGS & DN. & Gen. & Ours & 3DGS & 2DGS & DN. & Gen. & Ours & 3DGS & 2DGS & DN. & Gen. & Ours\\
        \midrule
 bonsai& \best 0.951 & \sbest 0.950 & \tbest 0.949 & 0.938 & 0.927 & \best 27.169 & \tbest 26.831 & \sbest 26.848 & 25.650 & 19.987 & \sbest 0.046 & \best 0.045 & \tbest 0.049 & 0.067 & 0.065 \\
 garden& 0.819 & \best 0.859 & \sbest 0.846 & 0.758 & \tbest 0.842 & \tbest 19.053 & \sbest 20.989 & \best 21.193 & 18.115 & 17.095 & 0.168 & \best 0.139 & \tbest 0.163 & 0.238 & \sbest 0.157 \\
 kitchen& 0.883 & \best 0.907 & \sbest 0.890 & 0.847 & \tbest 0.888 & \best 24.800 & \sbest 24.669 & \tbest 23.737 & 21.901 & 20.026 & \sbest 0.100 & \best 0.091 & \tbest 0.104 & 0.160 & 0.109 \\
 scene1& 0.751 & \tbest 0.774 & \sbest 0.852 & 0.738 & \best 0.868 & 9.648 & \tbest 10.507 & \sbest 14.389 & 8.617 & \best 16.259 & 0.237 & \tbest 0.219 & \sbest 0.140 & 0.251 & \best 0.136 \\
 scene2& \tbest 0.805 & \sbest 0.835 & 0.784 & 0.682 & \best 0.861 & \sbest 18.550 & \best 19.897 & \tbest 18.012 & 14.801 & 16.662 & \tbest 0.215 & \sbest 0.186 & 0.229 & 0.327 & \best 0.164 \\
 Bench& 0.905 & \sbest 0.955 & \sbest 0.955 & \tbest 0.944 & \best 0.969 & 18.316 & \tbest 23.227 & \sbest 23.311 & 22.264 & \best 24.776 & 0.099 & \sbest 0.052 & \tbest 0.056 & 0.071 & \best 0.029 \\
 Desk& \tbest 0.867 & \best 0.885 & \sbest 0.877 & 0.844 & \best 0.885 & \tbest 17.498 & \best 18.762 & \sbest 17.641 & 16.350 & 17.225 & 0.165 & \sbest 0.142 & \tbest 0.155 & 0.188 & \best 0.131 \\
 donuts& \tbest 0.907 & \sbest 0.936 & 0.900 & 0.861 & \best 0.973 & \tbest 17.332 & \sbest 19.430 & 16.281 & 15.149 & \best 26.443 & \tbest 0.112 & \sbest 0.078 & 0.114 & 0.172 & \best 0.030 \\
 figurines& 0.933 & \sbest 0.940 & \tbest 0.939 & 0.929 & \best 0.957 & 20.466 & \sbest 21.395 & \tbest 20.965 & 19.519 & \best 22.640 & 0.079 & \sbest 0.069 & \tbest 0.072 & 0.088 & \best 0.049 \\
 shoe\_r.& 0.940 & \sbest 0.956 & \tbest 0.948 & 0.943 & \best 0.962 & 22.556 & \best 25.248 & \tbest 23.951 & 22.870 & \sbest 24.430 & 0.068 & \sbest 0.049 & \tbest 0.059 & 0.070 & \best 0.040 \\
 teatime& 0.928 & \tbest 0.949 & \sbest 0.950 & 0.926 & \best 0.968 & 17.828 & \sbest 20.313 & \tbest 20.014 & 18.127 & \best 25.539 & 0.084 & \sbest 0.060 & \tbest 0.061 & 0.087 & \best 0.043 \\
 Avg.& 0.881 & \sbest 0.904 & \tbest 0.899 & 0.855 & \best 0.918 & 19.383 & \best 21.024 & \tbest 20.577 & 18.488 & \sbest 21.007 & 0.125 & \sbest 0.103 & \tbest 0.109 & 0.156 & \best 0.087 \\
        \midrule
         & \multicolumn{5}{c|}{MUSIQ$\uparrow$} & \multicolumn{5}{c|}{CLIPS$\uparrow$} & \multicolumn{5}{c}{mIoU$\uparrow$}\\
         \addlinespace[1pt]\cline{2-6}\cline{7-11}\cline{12-16}\addlinespace[1pt]
 & 3DGS & 2DGS & DN. & Gen. & Ours & 3DGS & 2DGS & DN. & Gen. & Ours & 3DGS & 2DGS & DN. & Gen. & Ours\\
        \midrule
 bonsai& \tbest 48.711 & \best 51.993 & 44.882 & 20.048 & \sbest 49.205 & \best 0.921 & \sbest 0.920 & \tbest 0.905 & 0.875 & 0.882 & \sbest 0.769 & \best 0.798 & \tbest 0.740 & 0.652 & 0.394 \\
 garden& \tbest 55.415 & \sbest 61.179 & 54.888 & 42.411 & \best 63.410 & \tbest 0.845 & \best 0.896 & \tbest 0.845 & 0.797 & \sbest 0.892 & 0.611 & \sbest 0.702 & \tbest 0.667 & 0.524 & \best 0.731 \\
 kitchen& \tbest 48.983 & \sbest 51.518 & 46.280 & 33.304 & \best 54.907 & \tbest 0.920 & \sbest 0.922 & 0.917 & 0.833 & \best 0.947 & 0.553 & \sbest 0.743 & \tbest 0.615 & 0.550 & \best 0.838 \\
 scene1& \tbest 55.485 & \sbest 57.513 & 33.900 & 50.191 & \best 59.004 & \tbest 0.765 & \sbest 0.769 & 0.623 & 0.754 & \best 0.863 & \tbest 0.481 & \sbest 0.527 & 0.001 & 0.458 & \best 0.636 \\
 scene2& \tbest 42.301 & \sbest 48.475 & 39.973 & 33.004 & \best 61.706 & \tbest 0.799 & \sbest 0.825 & 0.773 & 0.735 & \best 0.887 & \tbest 0.594 & \sbest 0.693 & 0.584 & 0.470 & \best 0.800 \\
 Bench& 38.889 & \sbest 44.997 & \tbest 41.130 & 30.470 & \best 58.641 & 0.866 & \sbest 0.909 & \tbest 0.890 & 0.868 & \best 0.956 & 0.405 & \sbest 0.722 & \tbest 0.697 & 0.625 & \best 0.819 \\
 Desk& 47.242 & \sbest 52.417 & \tbest 47.392 & 41.457 & \best 65.700 & 0.821 & \tbest 0.833 & \sbest 0.835 & 0.795 & \best 0.873 & 0.703 & \best 0.765 & \sbest 0.744 & 0.654 & \tbest 0.724 \\
 donuts& \tbest 35.328 & \sbest 40.032 & 34.682 & 25.904 & \best 53.335 & 0.829 & \sbest 0.861 & \tbest 0.836 & 0.824 & \best 0.922 & \tbest 0.264 & \sbest 0.395 & 0.259 & 0.184 & \best 0.648 \\
 figurines& 34.804 & \sbest 40.740 & \tbest 35.066 & 20.399 & \best 48.797 & 0.871 & \sbest 0.880 & \tbest 0.878 & 0.848 & \best 0.899 & \tbest 0.465 & \best 0.510 & \sbest 0.495 & 0.409 & 0.448 \\
 shoe\_r.& 33.428 & \sbest 40.262 & \tbest 34.970 & 23.596 & \best 50.221 & 0.879 & \best 0.908 & \tbest 0.890 & 0.879 & \sbest 0.901 & 0.399 & \sbest 0.444 & \best 0.472 & 0.322 & \tbest 0.408 \\
 teatime& 27.961 & \sbest 35.051 & \tbest 30.117 & 20.654 & \best 50.358 & 0.823 & \tbest 0.837 & \sbest 0.841 & 0.827 & \best 0.893 & 0.290 & \tbest 0.408 & \best 0.434 & 0.255 & \sbest 0.426 \\
 Avg.& \tbest 42.595 & \sbest 47.653 & 40.298 & 31.040 & \best 55.935 & \tbest 0.849 & \sbest 0.869 & 0.839 & 0.821 & \best 0.901 & 0.503 & \sbest 0.610 & \tbest 0.519 & 0.464 & \best 0.625 \\
        \bottomrule
    \end{tabular}}
    \label{tab:quantitative-app-full}}
    \vspace{-1em}
\end{table}

\section{Additional Results}
\label{app:additional_results}
\subsection{Quantitative Comparison Results}

Tab.~\ref{tab:quantitative-app-full} presents the quantitative evaluation of our method alongside baseline approaches on the comprehensive dataset described in Sec.~\ref{sec:dataset_details_app}, where all data represent averages computed over the three difficulty levels discussed in the paper. Our results demonstrate that our method outperforms all baselines, including 3DGS~\cite{3dgs}, 2DGS~\cite{2dgs}, DNGaussian~\cite{dngaussian}, and GenFusion~\cite{genfusion}, across nearly all metrics.

\begin{table}[!hbt]{
        \caption{
         More Quantitative results of the ablation study on the rendering quality.  
          Metrics on appearance are computed over three scenes (3DGS-CD-\textit{Bench}, 3DGS-CD-\textit{Desk}, LERF-\textit{Donuts}).  
          \colorbox{colorFst}{best}, \colorbox{colorSnd}{second best}, \colorbox{colorTrd}{third best} and  
          \colorbox{colorLow}{false case} are highlighted.  
          Since the Appearance Refinement branch only optimizes color, the mIoU of (a) and (g) are the same and therefore omitted.
        }
        \label{tab:ablation_app}
        \vspace{-1em}
        \setlength{\tabcolsep}{1pt}
        \resizebox{\linewidth}{!}{%
        \begin{tabular}{l
            |cccc
            |cccc
            |cccc
            |cccc}
          \toprule

    \multirow{2}{*}{\rotatebox[origin=c]{90}{SSIM$\uparrow$}}
      & \multicolumn{4}{c|}{Bench}
      & \multicolumn{4}{c|}{Desk}
      & \multicolumn{4}{c|}{Donuts}
      & \multicolumn{4}{c}{Avg.}\\
    \addlinespace[1pt]\cline{2-17}\addlinespace[1pt]
      & $\nicefrac{2}{3}$ & $\nicefrac{4}{5}$ & $\nicefrac{6}{7}$ & Avg.
      & $\nicefrac{2}{3}$ & $\nicefrac{4}{5}$ & $\nicefrac{6}{7}$ & Avg.
      & $\nicefrac{2}{3}$ & $\nicefrac{4}{5}$ & $\nicefrac{6}{7}$ & Avg.
      & $\nicefrac{2}{3}$ & $\nicefrac{4}{5}$ & $\nicefrac{6}{7}$ & Avg.\\
    \midrule
    a & 0.961 & 0.962 & 0.963 & 0.962 & 0.889\sbest & 0.881\sbest & 0.858 & 0.876\tbest & 0.974\sbest & 0.968\tbest & 0.972\sbest & 0.971\sbest & 0.942\sbest & 0.937\tbest & 0.931 & 0.936\tbest \\
    b & 0.963\tbest & 0.968\best & 0.970\best & 0.967\sbest & 0.000\false & 0.848 & 0.000\false & 0.283 & 0.973 & 0.000\false & 0.000\false & 0.324 & 0.645 & 0.605 & 0.323 & 0.525 \\
    c & 0.964\sbest & 0.964\tbest & 0.963 & 0.964\tbest & 0.886\tbest & 0.884\best & 0.885\best & 0.885\best & 0.974\tbest & 0.969\sbest & 0.971 & 0.971\tbest & 0.941\tbest & 0.939\sbest & 0.940\sbest & 0.940\sbest \\
    d & 0.920 & 0.700 & 0.309 & 0.643 & 0.770 & 0.774 & 0.622 & 0.722 & 0.951 & 0.935 & 0.915 & 0.934 & 0.881 & 0.803 & 0.615 & 0.766 \\
    e & 0.908 & 0.698 & 0.309 & 0.638 & 0.772 & 0.775 & 0.634 & 0.727 & 0.950 & 0.935 & 0.913 & 0.933 & 0.877 & 0.803 & 0.619 & 0.766 \\
    f & 0.957 & 0.963 & 0.964\tbest & 0.961 & 0.883 & 0.875 & 0.867\tbest & 0.875 & 0.972 & 0.964 & 0.972\tbest & 0.969 & 0.937 & 0.934 & 0.934\tbest & 0.935 \\
    \textbf{g}& 0.970\best & 0.968\sbest & 0.969\sbest & 0.969\best & 0.895\best & 0.881\tbest & 0.880\sbest & 0.885\sbest & 0.975\best & 0.971\best & 0.973\best & 0.973\best & 0.947\best & 0.940\best & 0.941\best & 0.943\best \\

          \midrule
          \multirow{2}{*}{\rotatebox[origin=c]{90}{PSNR$\uparrow$}}
            & \multicolumn{4}{c|}{Bench}
            & \multicolumn{4}{c|}{Desk}
            & \multicolumn{4}{c|}{Donuts}
            & \multicolumn{4}{c}{Avg.}\\
          \addlinespace[1pt]\cline{2-17}\addlinespace[1pt]
            & $\nicefrac{2}{3}$ & $\nicefrac{4}{5}$ & $\nicefrac{6}{7}$ & Avg.
            & $\nicefrac{2}{3}$ & $\nicefrac{4}{5}$ & $\nicefrac{6}{7}$ & Avg.
            & $\nicefrac{2}{3}$ & $\nicefrac{4}{5}$ & $\nicefrac{6}{7}$ & Avg.
            & $\nicefrac{2}{3}$ & $\nicefrac{4}{5}$ & $\nicefrac{6}{7}$ & Avg.\\
          \midrule
          a 
& 22.568 & 22.967\tbest & 23.072\tbest & 22.869\tbest
            & 17.133\sbest & 16.132 & 14.724 & 15.997
            & 26.235 & 24.976\tbest & 25.197 & 25.469
            & 21.979\tbest & 21.358\tbest & 20.998 & 21.445\tbest \\
          b 
& 23.484\sbest & 24.130\sbest & 24.639\sbest & 24.084\sbest
            & 0.000\false & 13.549 & 0.000\false & 4.516
            & 26.356\tbest & 0.000\false & 0.000\false & 8.785
            & 16.613 & 12.560 & 8.213 & 12.462 \\
          c 
& 23.316\tbest & 22.540 & 22.349 & 22.735
            & 16.794\tbest & 16.778\sbest & 17.540\best & 17.038\sbest
            & 26.479\sbest & 25.268\sbest & 25.612\tbest & 25.786\sbest
            & 22.196\sbest & 21.529\sbest & 21.833\sbest & 21.853\sbest \\
          d
& 21.185 & 16.310 & 13.420 & 16.972
            & 14.498 & 14.744 & 9.440 & 12.894
            & 23.471 & 21.486 & 17.223 & 20.726
            & 19.718 & 17.513 & 13.361 & 16.864 \\
          e
& 18.954 & 15.935 & 13.437 & 16.109
            & 14.079 & 14.349 & 9.410 & 12.613
            & 22.853 & 21.066 & 16.893 & 20.271
            & 18.629 & 17.117 & 13.247 & 16.331 \\
          f
& 20.222 & 21.747 & 22.281 & 21.416
            & 16.676 & 16.518\tbest & 15.461\tbest & 16.218\tbest
            & 26.179 & 24.336 & 26.039\sbest & 25.518\tbest
            & 21.026 & 20.867 & 21.260\tbest & 21.051 \\
          \textbf{g}& 25.122\best & 24.306\best & 24.901\best & 24.776\best
            & 17.635\best & 17.106\best & 16.935\sbest & 17.225\best
            & 27.032\best & 25.681\best & 26.615\best & 26.443\best
            & 23.263\best & 22.364\best & 22.817\best & 22.815\best \\

    \midrule
    \multirow{2}{*}{\rotatebox[origin=c]{90}{LPIPS$\downarrow$}}
      & \multicolumn{4}{c|}{Bench}
      & \multicolumn{4}{c|}{Desk}
      & \multicolumn{4}{c|}{Donuts}
      & \multicolumn{4}{c}{Avg.}\\
    \addlinespace[1pt]\cline{2-17}\addlinespace[1pt]
      & $\nicefrac{2}{3}$ & $\nicefrac{4}{5}$ & $\nicefrac{6}{7}$ & Avg.
      & $\nicefrac{2}{3}$ & $\nicefrac{4}{5}$ & $\nicefrac{6}{7}$ & Avg.
      & $\nicefrac{2}{3}$ & $\nicefrac{4}{5}$ & $\nicefrac{6}{7}$ & Avg.
      & $\nicefrac{2}{3}$ & $\nicefrac{4}{5}$ & $\nicefrac{6}{7}$ & Avg.\\
    \midrule
    a & 0.035\sbest & 0.035\tbest & 0.034\tbest & 0.035\tbest & 0.133\tbest & 0.134\tbest & 0.154 & 0.141\tbest & 0.028\sbest & 0.038\tbest & 0.032\sbest & 0.032\sbest & 0.065\sbest & 0.069\tbest & 0.074 & 0.069\tbest \\
    b & 0.039 & 0.031\sbest & 0.029\best & 0.033\sbest & 1.000\false & 0.154 & 1.000\false & 0.718 & 0.030 & 1.000\false & 1.000\false & 0.677 & 0.356 & 0.395 & 0.676 & 0.476 \\
    c & 0.036\tbest & 0.035 & 0.035 & 0.036 & 0.130\sbest & 0.130\best & 0.134\tbest & 0.131\sbest & 0.029\tbest & 0.034\sbest & 0.032 & 0.032\tbest & 0.065\tbest & 0.067\sbest & 0.067\sbest & 0.066\sbest \\
    d & 0.076 & 0.322 & 0.760 & 0.386 & 0.424 & 0.294 & 0.635 & 0.451 & 0.055 & 0.078 & 0.104 & 0.079 & 0.185 & 0.231 & 0.500 & 0.305 \\
    e & 0.085 & 0.330 & 0.756 & 0.390 & 0.435 & 0.295 & 0.635 & 0.455 & 0.056 & 0.080 & 0.105 & 0.080 & 0.192 & 0.235 & 0.499 & 0.309 \\
    f & 0.042 & 0.037 & 0.035 & 0.038 & 0.140 & 0.145 & 0.145\best & 0.144 & 0.031 & 0.040 & 0.031\best & 0.034 & 0.071 & 0.074 & 0.070\tbest & 0.072 \\
    \textbf{g}& 0.029\best & 0.030\best & 0.030\sbest & 0.030\best & 0.121\best & 0.134\sbest & 0.134\sbest & 0.130\best & 0.026\best & 0.032\best & 0.032\sbest & 0.030\best & 0.059\best & 0.066\best & 0.066\best & 0.063\best \\

          \midrule

          \multirow{2}{*}{\rotatebox[origin=c]{90}{mIoU$\uparrow$}}
            & \multicolumn{4}{c|}{Bench}
            & \multicolumn{4}{c|}{Desk}
            & \multicolumn{4}{c|}{Donuts}
            & \multicolumn{4}{c}{Avg.}\\
          \addlinespace[1pt]\cline{2-17}\addlinespace[1pt]
            & $\nicefrac{2}{3}$ & $\nicefrac{4}{5}$ & $\nicefrac{6}{7}$ & Avg.
            & $\nicefrac{2}{3}$ & $\nicefrac{4}{5}$ & $\nicefrac{6}{7}$ & Avg.
            & $\nicefrac{2}{3}$ & $\nicefrac{4}{5}$ & $\nicefrac{6}{7}$ & Avg.
            & $\nicefrac{2}{3}$ & $\nicefrac{4}{5}$ & $\nicefrac{6}{7}$ & Avg.\\
          \midrule
          a 
& \diagbox{}{} & \diagbox{}{} & \diagbox{}{} & \diagbox{}{}
            & \diagbox{}{} & \diagbox{}{} & \diagbox{}{} & \diagbox{}{}
            & \diagbox{}{} & \diagbox{}{} & \diagbox{}{} & \diagbox{}{}
            & \diagbox{}{} & \diagbox{}{} & \diagbox{}{} & \diagbox{}{} \\
          b 
& 0.773\tbest & 0.747\sbest & 0.804\sbest & 0.775\sbest
            & 0.000\false & 0.299 & 0.000\false & 0.100
            & 0.636\tbest & 0.000\false & 0.000\false & 0.212
            & 0.470 & 0.349 & 0.268 & 0.362 \\
          c 
& 0.803\sbest & 0.681\tbest & 0.674 & 0.719\tbest
            & 0.675\tbest & 0.663\sbest & 0.735\best & 0.691\sbest
            & 0.656\sbest & 0.508\sbest & 0.582\tbest & 0.582\sbest
            & 0.711\sbest & 0.617\sbest & 0.664\sbest & 0.664\sbest \\
          d
& 0.432 & 0.128 & 0.048 & 0.203
            & 0.343 & 0.463 & 0.227 & 0.345
            & 0.473 & 0.335 & 0.217 & 0.342
            & 0.416 & 0.309 & 0.164 & 0.296 \\
          e
& 0.432 & 0.128 & 0.048 & 0.203
            & 0.343 & 0.463 & 0.227 & 0.345
            & 0.473 & 0.335 & 0.217 & 0.342
            & 0.416 & 0.309 & 0.164 & 0.296 \\
          f
& 0.494 & 0.662 & 0.718\tbest & 0.625
            & 0.747\sbest & 0.630\tbest & 0.599\tbest & 0.659\tbest
            & 0.615 & 0.485\tbest & 0.608\sbest & 0.569\tbest
            & 0.619\tbest & 0.592\tbest & 0.642\tbest & 0.618\tbest \\
          \textbf{g}& 0.862\best & 0.781\best & 0.813\best & 0.819\best
            & 0.754\best & 0.698\best & 0.720\sbest & 0.724\best
            & 0.712\best & 0.606\best & 0.627\best & 0.648\best
            & 0.776\best & 0.695\best & 0.720\best & 0.730\best \\
          \bottomrule
        \end{tabular}}
        }
    \vspace{-1em}
        
\end{table}

\newcommand{\renderwidthvis}{0.14}
\begin{figure*}[!htb]
    \vspace{-1em}
    \centering
    \addtolength{\tabcolsep}{-6.5pt}
    
    \footnotesize{
        \setlength{\tabcolsep}{1pt} 
        \begin{tabular}{cccccc}

            \includegraphics[width=\renderwidthvis\textwidth]{figures/qrj/DSC08018.jpg} &
            \includegraphics[width=\renderwidthvis\textwidth]{figures/qrj/DSC08018_3dgs.jpg} &
            \includegraphics[width=\renderwidthvis\textwidth]{figures/qrj/DSC08018_2dgs.jpg} &
            \includegraphics[width=\renderwidthvis\textwidth]{figures/qrj/DSC08018_dn.jpg} &
            \includegraphics[width=\renderwidthvis\textwidth]{figures/qrj/DSC08018_genfusion.jpg} &
            \includegraphics[width=\renderwidthvis\textwidth]{figures/qrj/DSC08018_ours.jpg} \\

            \includegraphics[width=\renderwidthvis\textwidth]{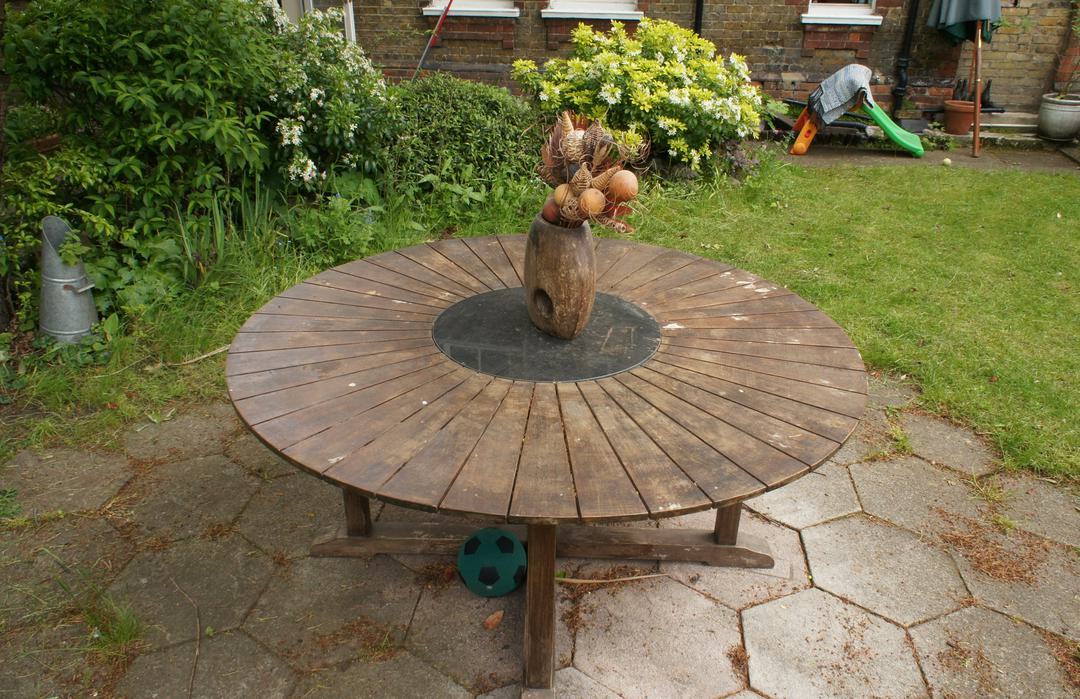} &
            \includegraphics[width=\renderwidthvis\textwidth]{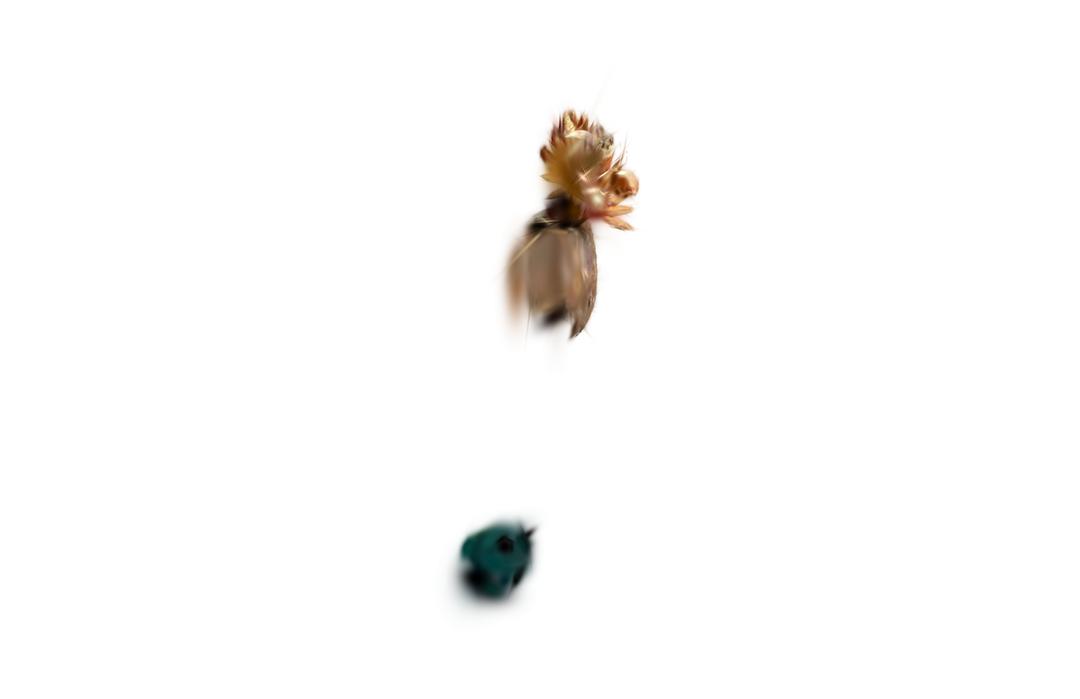} &
            \includegraphics[width=\renderwidthvis\textwidth]{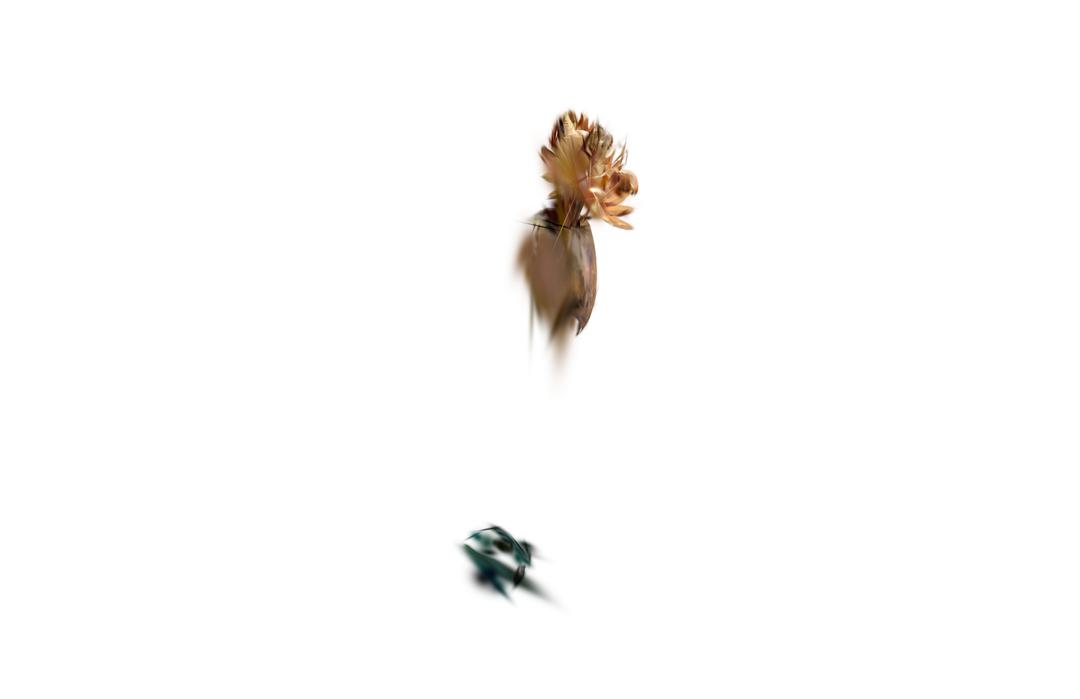} &
            \includegraphics[width=\renderwidthvis\textwidth]{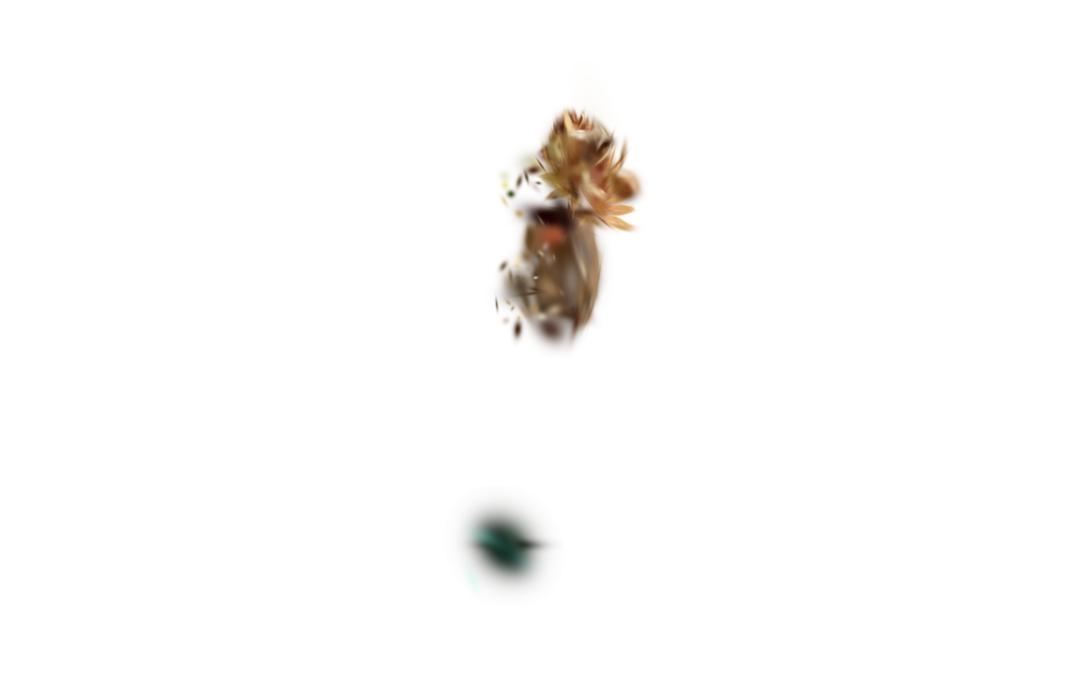} &
            \includegraphics[width=\renderwidthvis\textwidth]{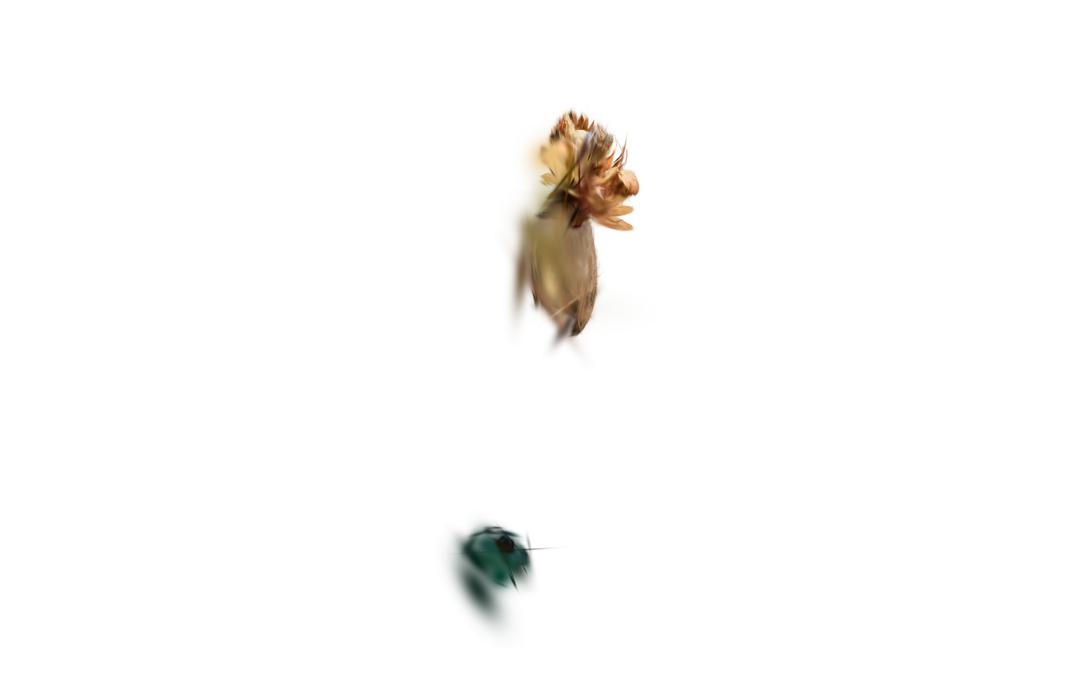} &
            \includegraphics[width=\renderwidthvis\textwidth]{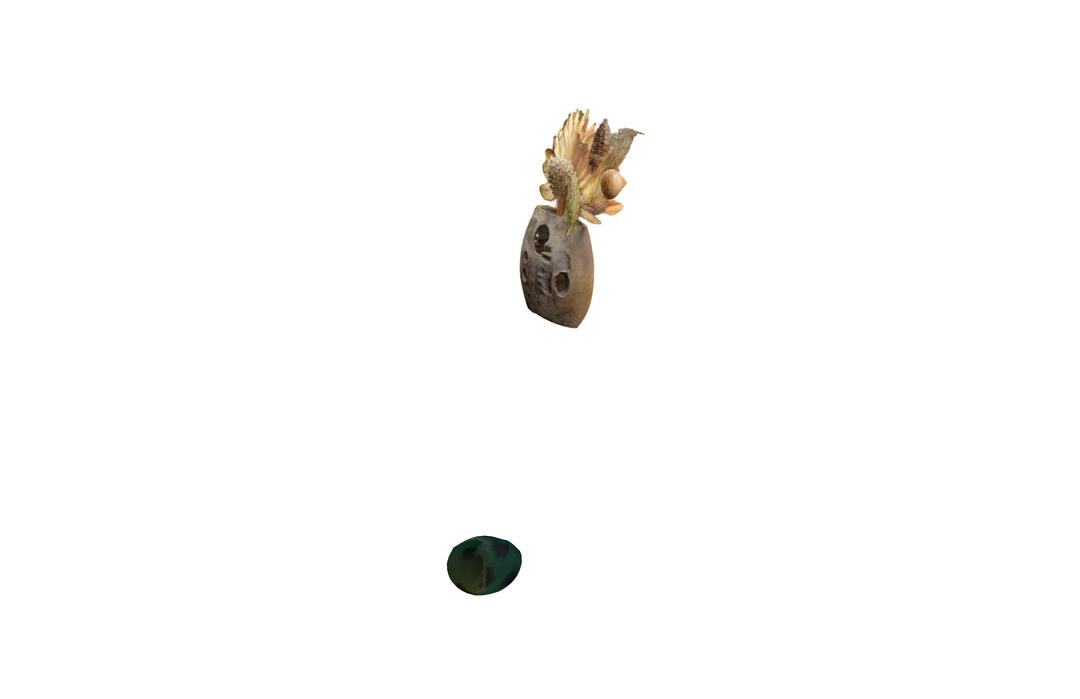} \\

            \includegraphics[width=\renderwidthvis\textwidth]{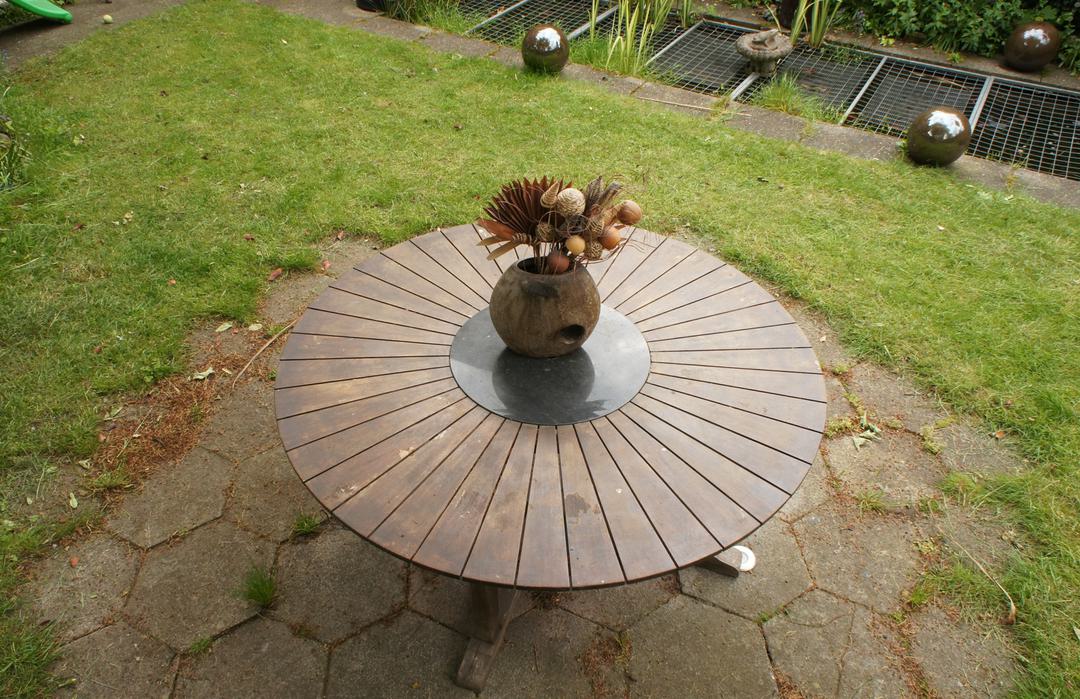} &
            \includegraphics[width=\renderwidthvis\textwidth]{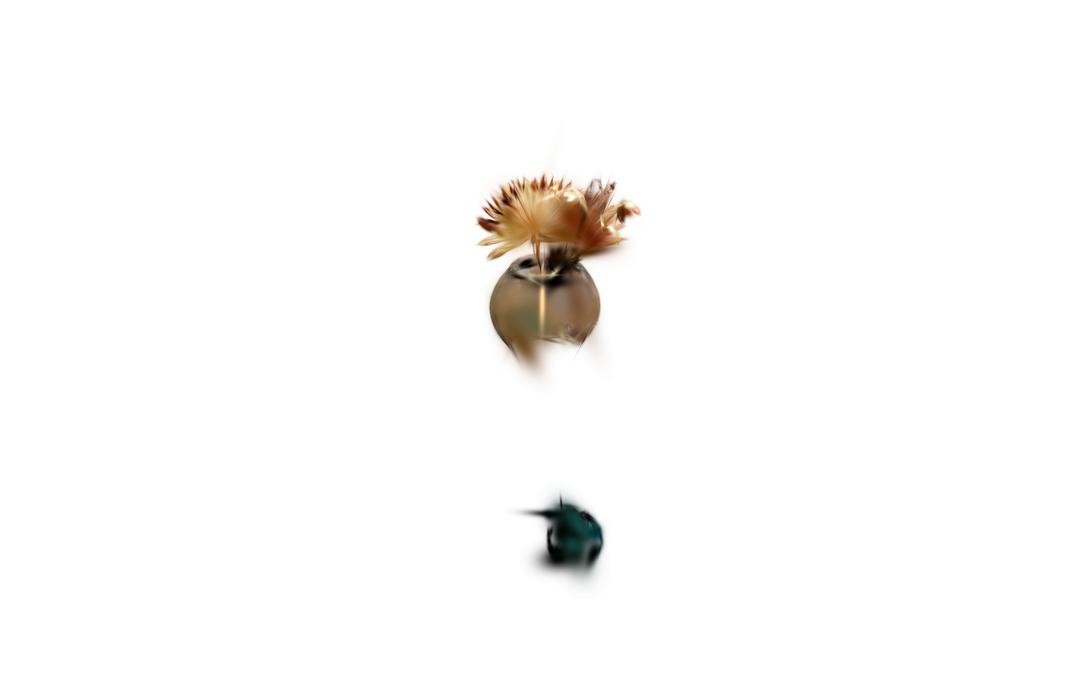} &
            \includegraphics[width=\renderwidthvis\textwidth]{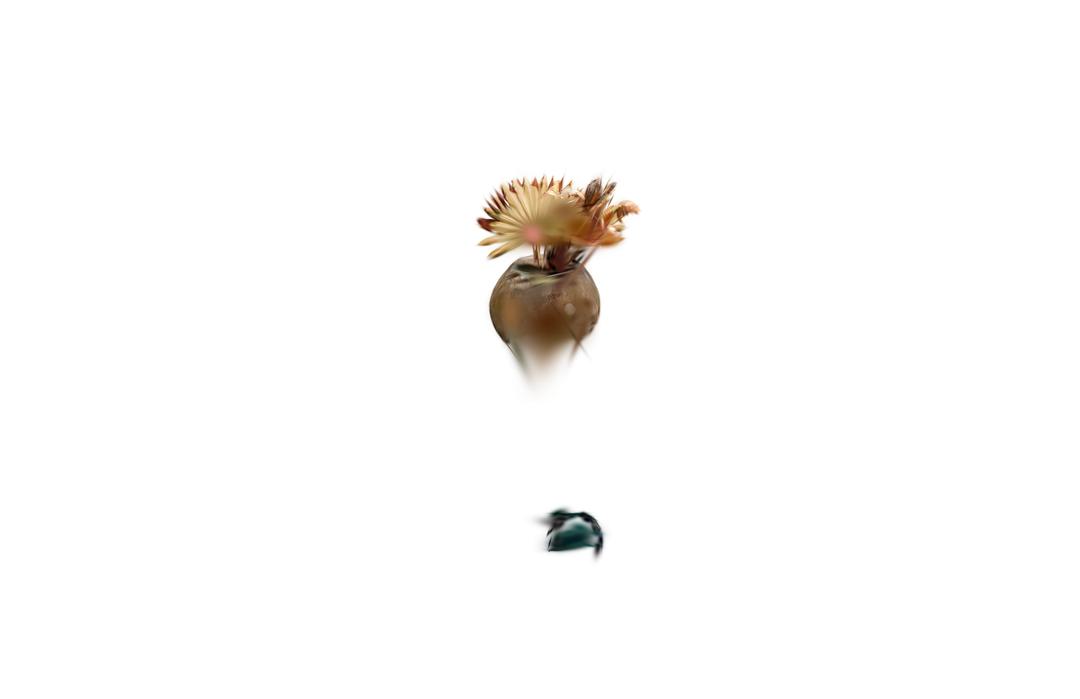} &
            \includegraphics[width=\renderwidthvis\textwidth]{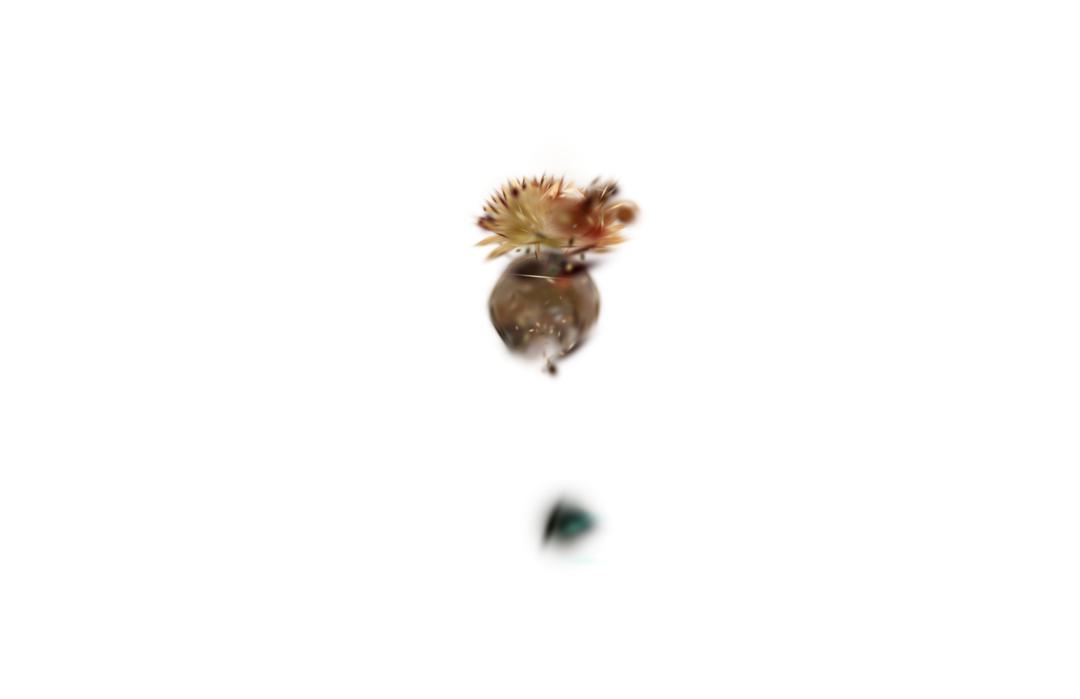} &
            \includegraphics[width=\renderwidthvis\textwidth]{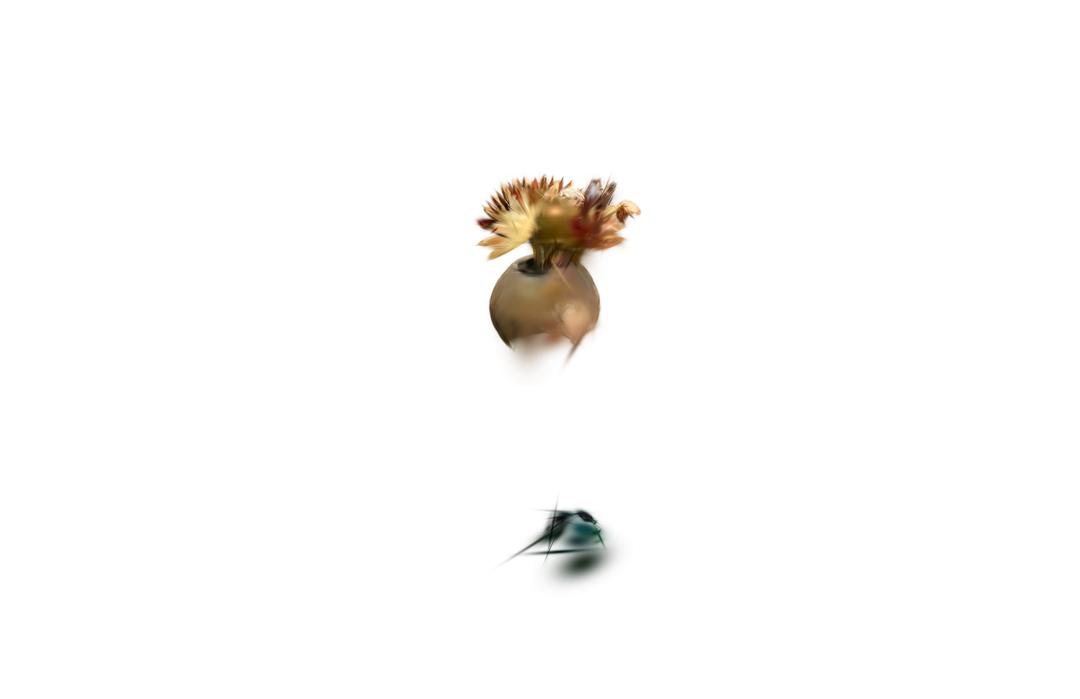} &
            \includegraphics[width=\renderwidthvis\textwidth]{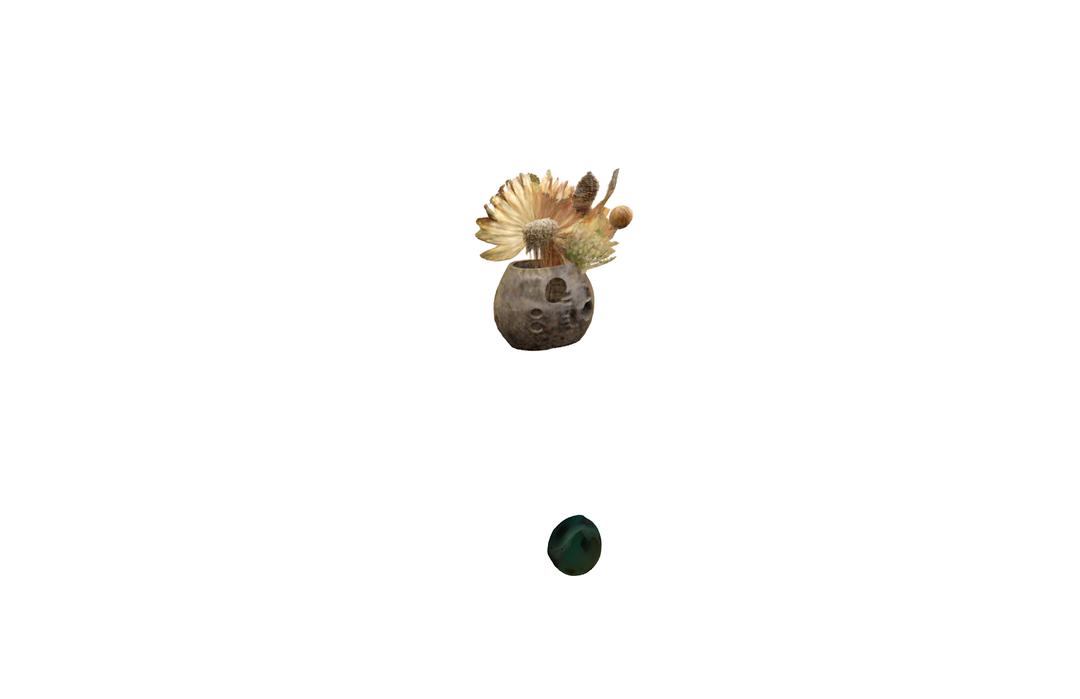} \\
            \includegraphics[width=\renderwidthvis\textwidth]{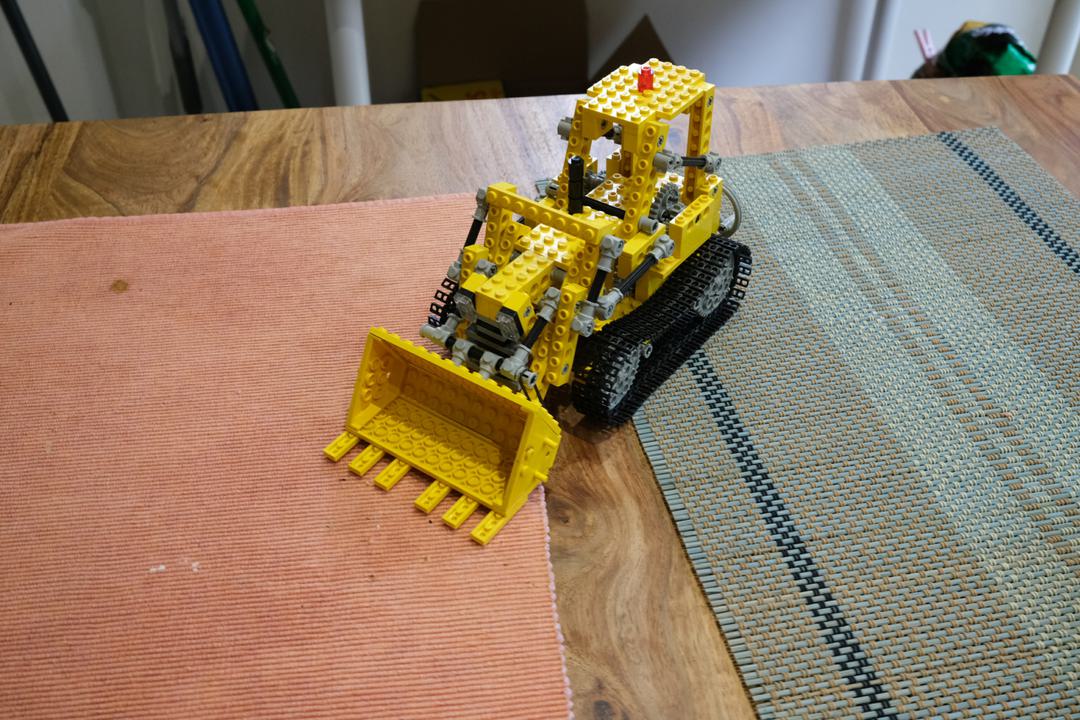} &
            \includegraphics[width=\renderwidthvis\textwidth]{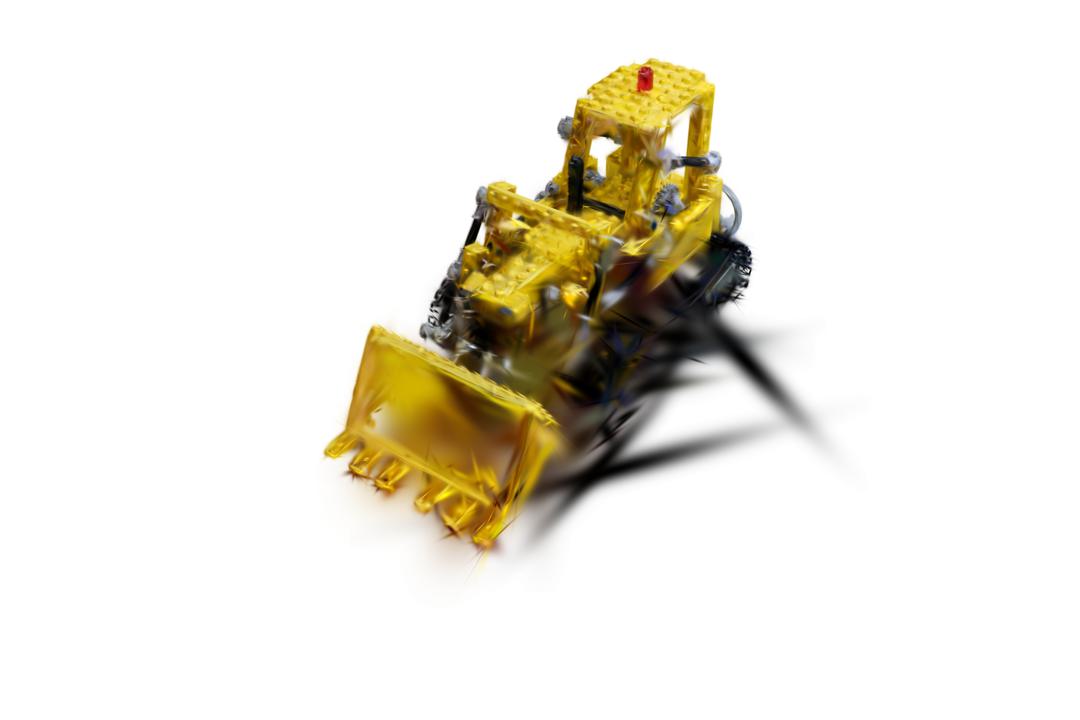} &
            \includegraphics[width=\renderwidthvis\textwidth]{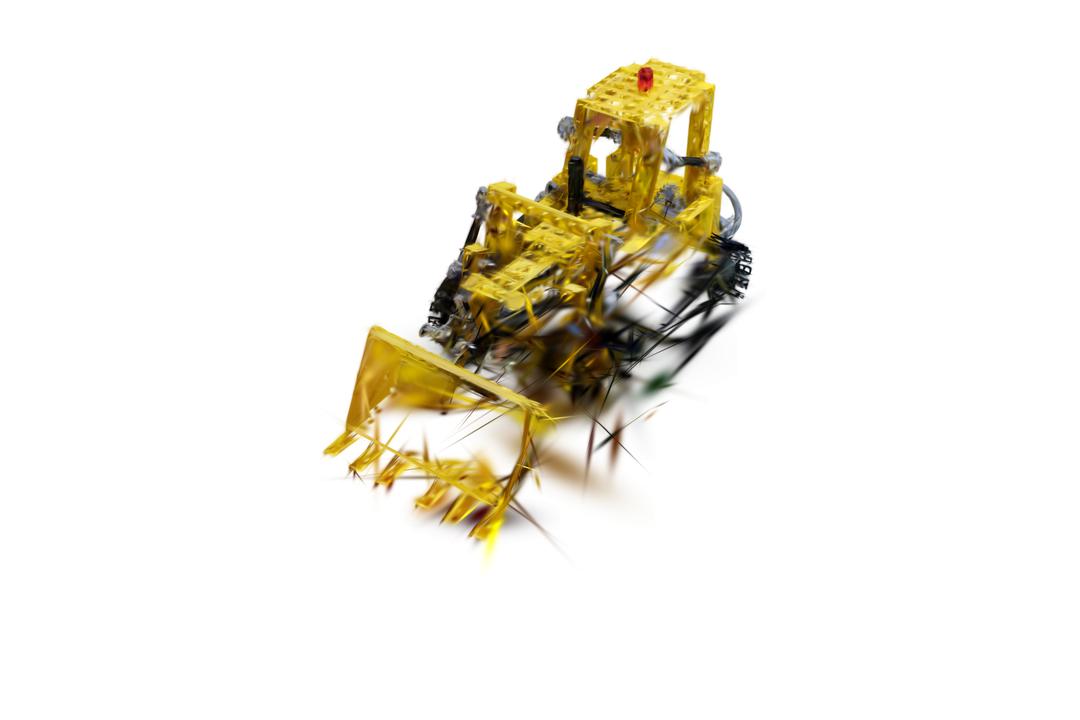} &
            \includegraphics[width=\renderwidthvis\textwidth]{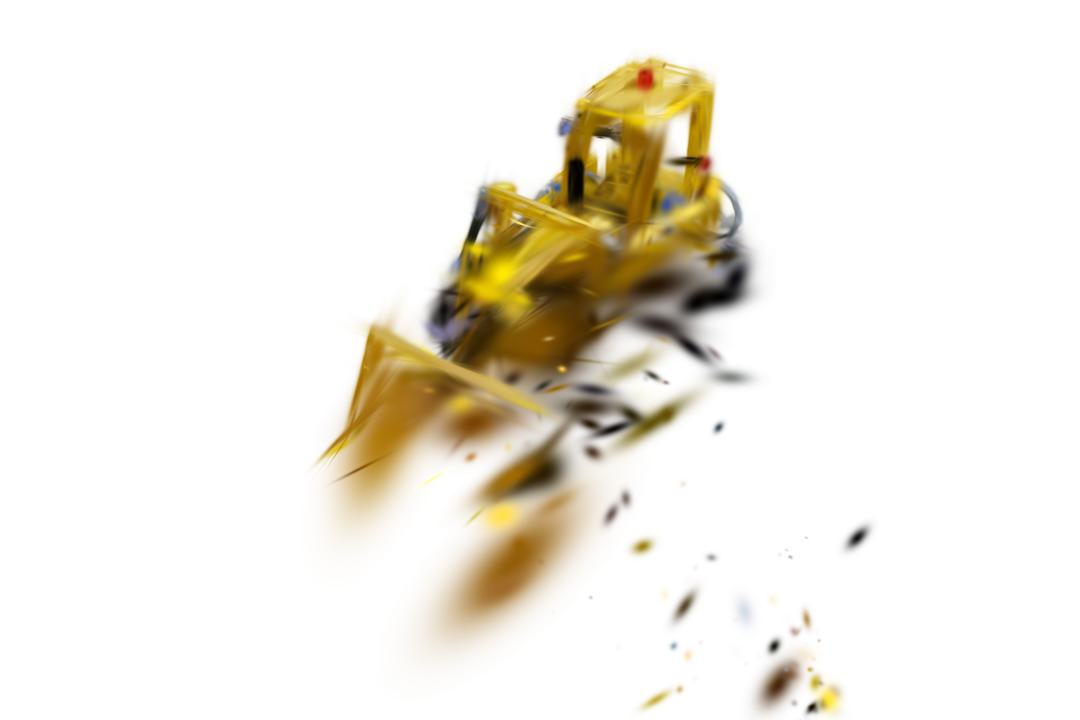} &
            \includegraphics[width=\renderwidthvis\textwidth]{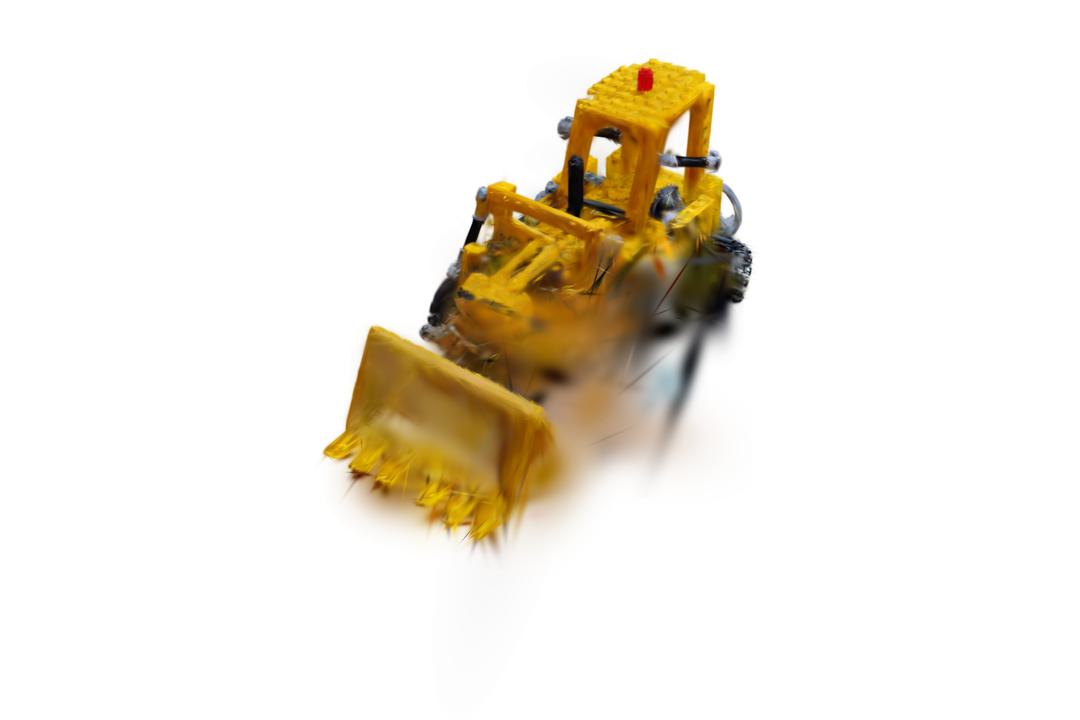} &
            \includegraphics[width=\renderwidthvis\textwidth]{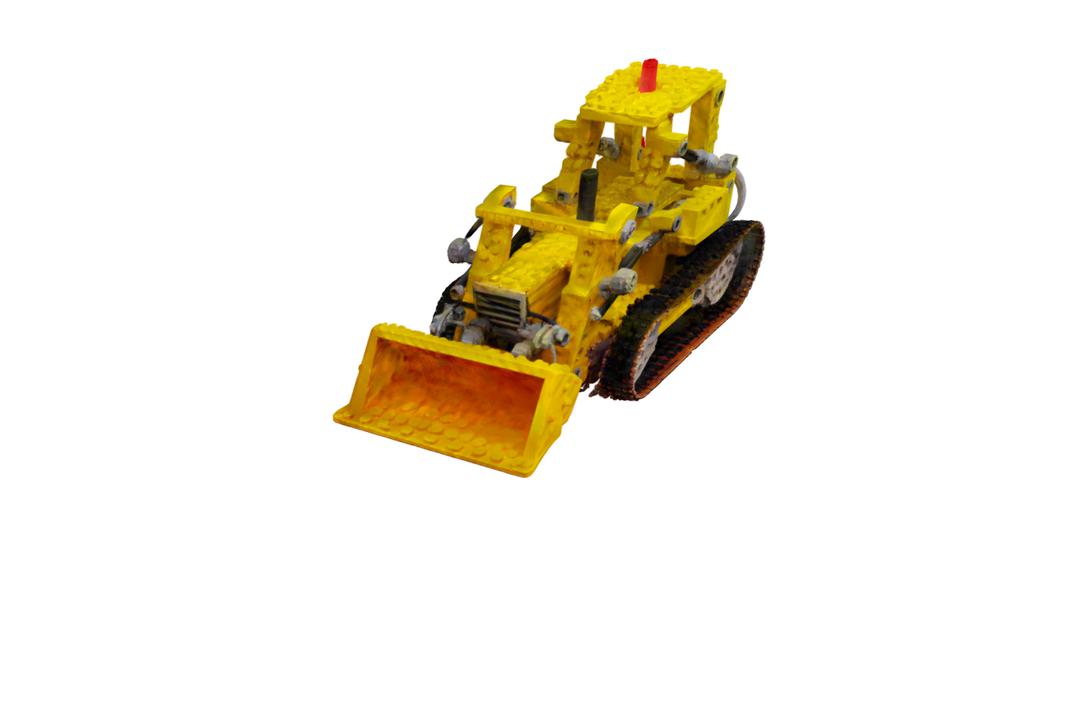} \\

            \includegraphics[width=\renderwidthvis\textwidth]{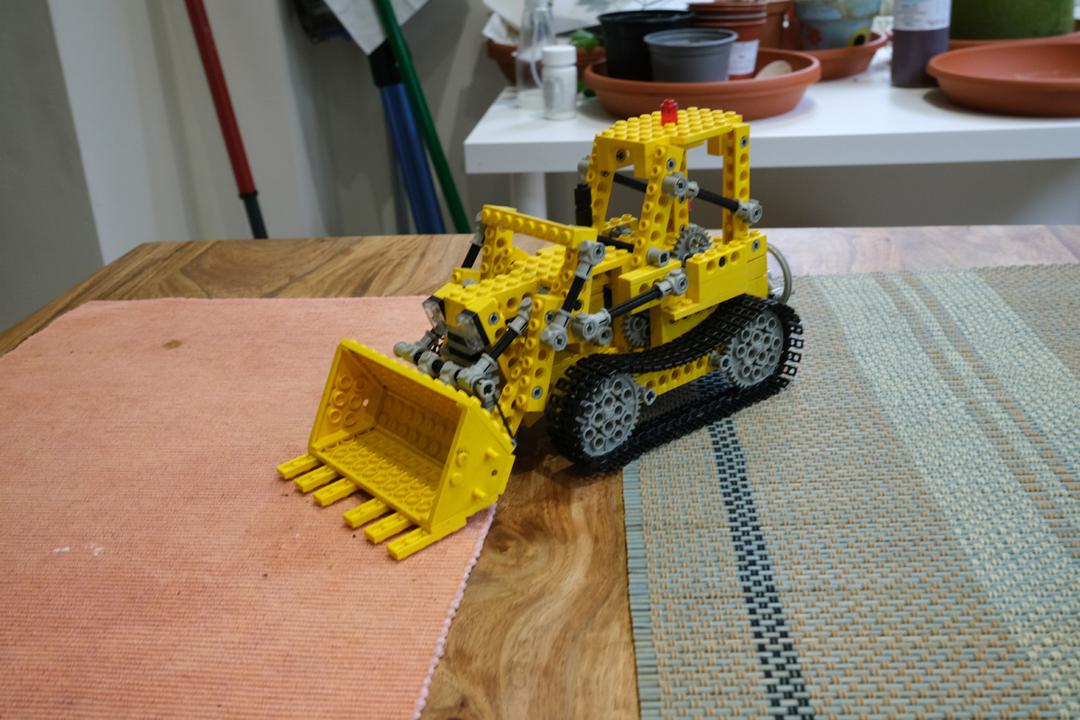} &
            \includegraphics[width=\renderwidthvis\textwidth]{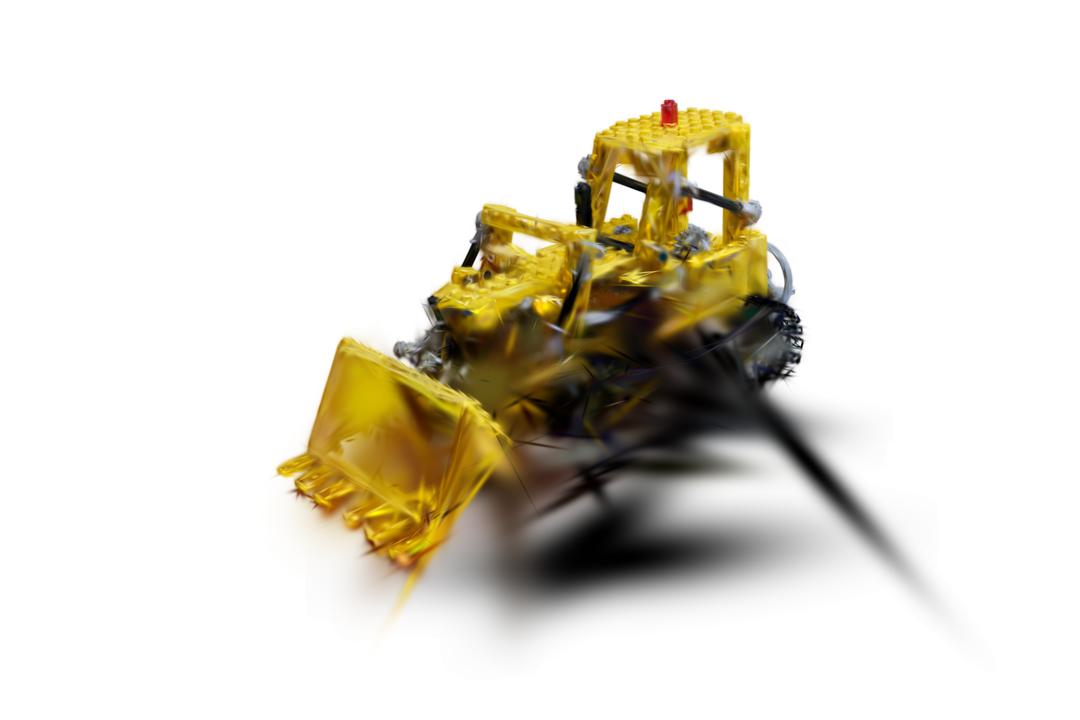} &
            \includegraphics[width=\renderwidthvis\textwidth]{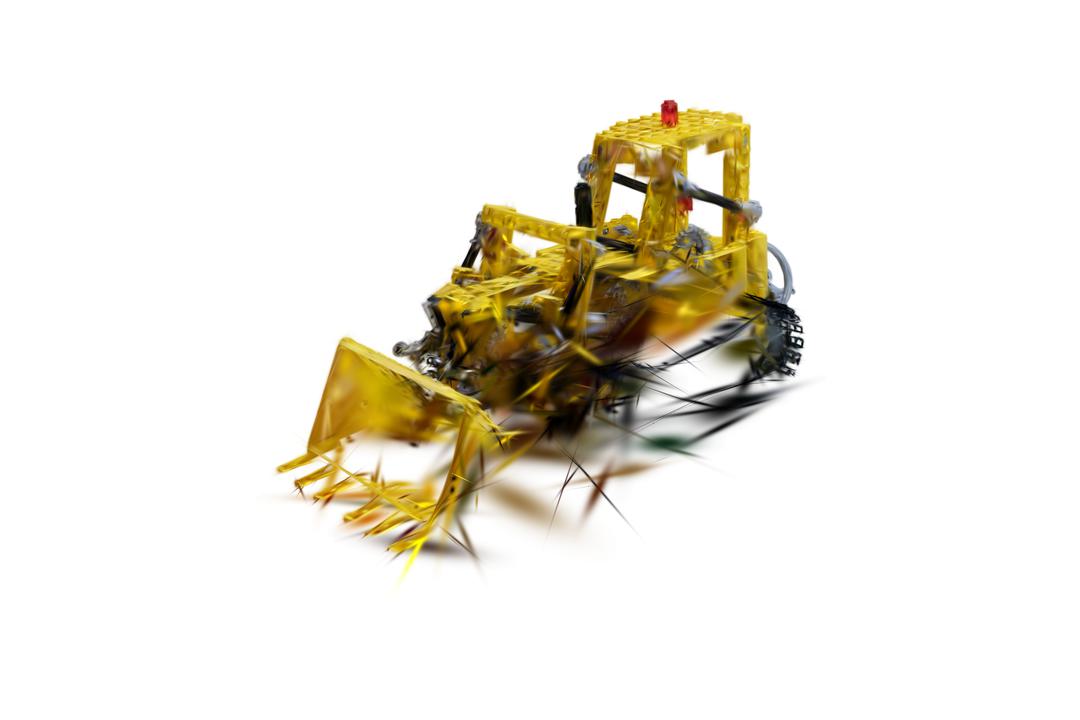} &
            \includegraphics[width=\renderwidthvis\textwidth]{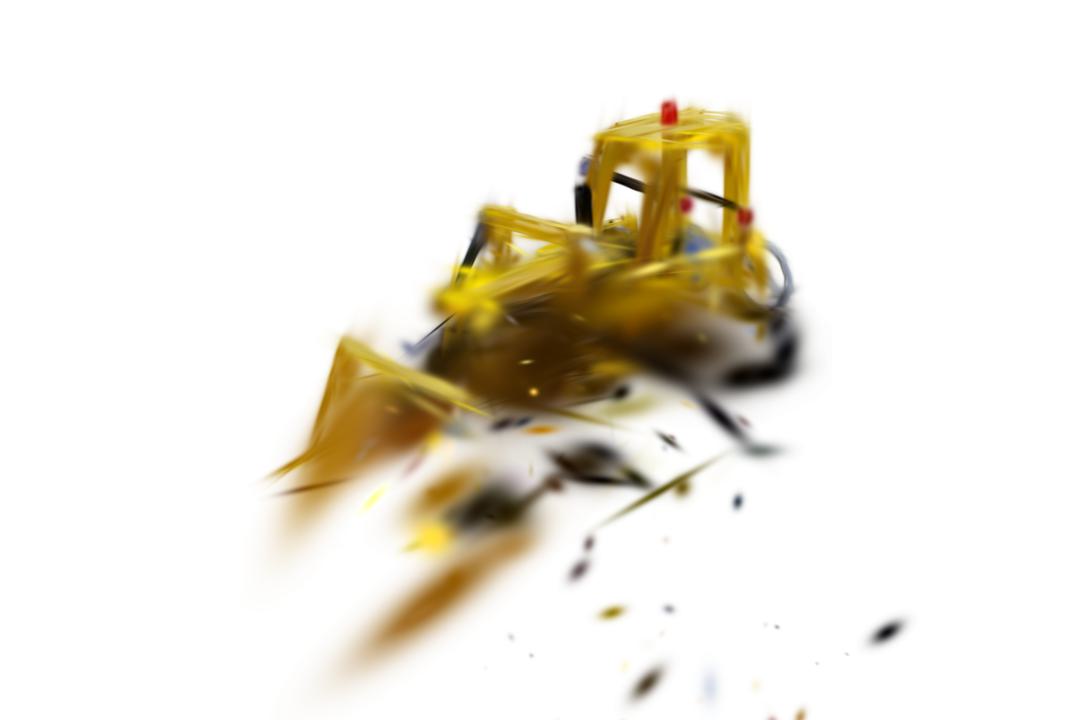} &
            \includegraphics[width=\renderwidthvis\textwidth]{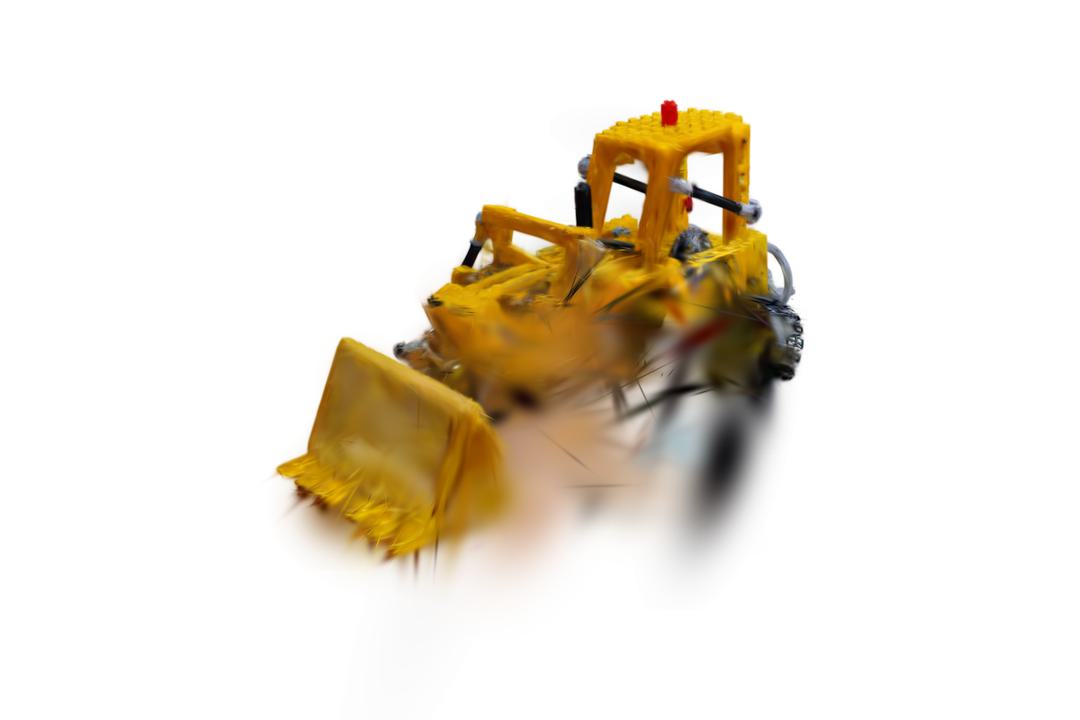} &
            \includegraphics[width=\renderwidthvis\textwidth]{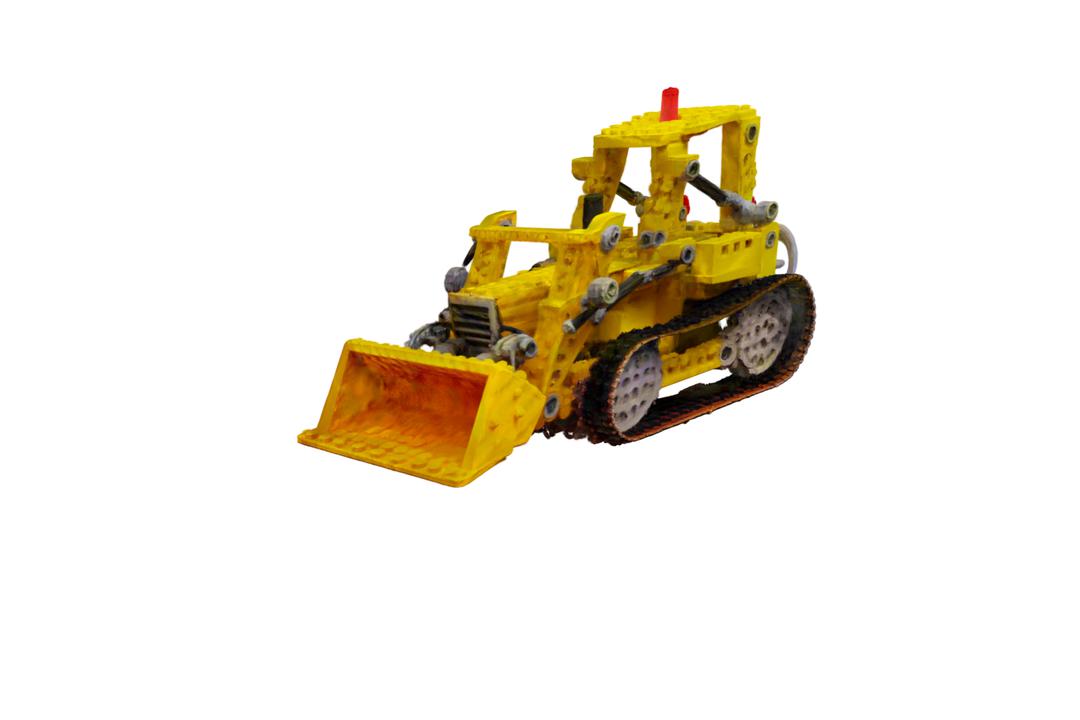} \\

            \includegraphics[width=\renderwidthvis\textwidth]{figures/qrj/DSCF0860.jpg} &
            \includegraphics[width=\renderwidthvis\textwidth]{figures/qrj/DSCF0860_3dgs.jpg} &
            \includegraphics[width=\renderwidthvis\textwidth]{figures/qrj/DSCF0860_2dgs.jpg} &
            \includegraphics[width=\renderwidthvis\textwidth]{figures/qrj/DSCF0860_dn.jpg} &
            \includegraphics[width=\renderwidthvis\textwidth]{figures/qrj/DSCF0860_genfusion.jpg} &
            \includegraphics[width=\renderwidthvis\textwidth]{figures/qrj/DSCF0860_ours.jpg} \\
            \includegraphics[width=\renderwidthvis\textwidth]{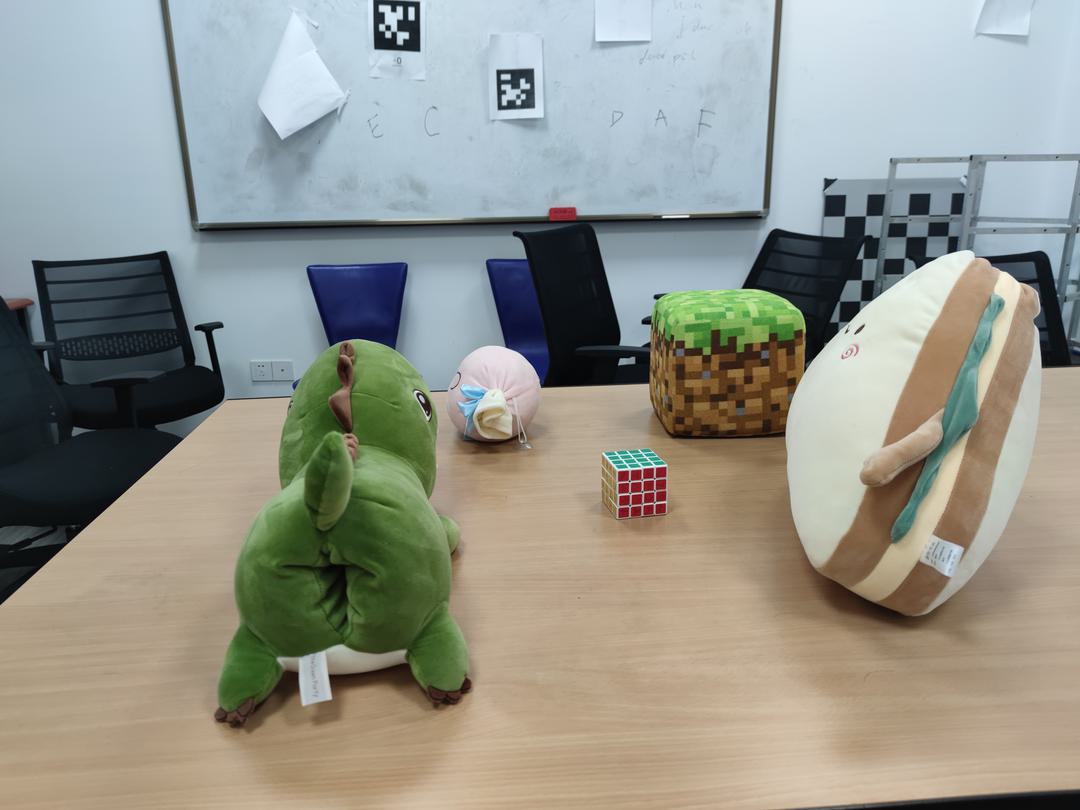} &
            \includegraphics[width=\renderwidthvis\textwidth]{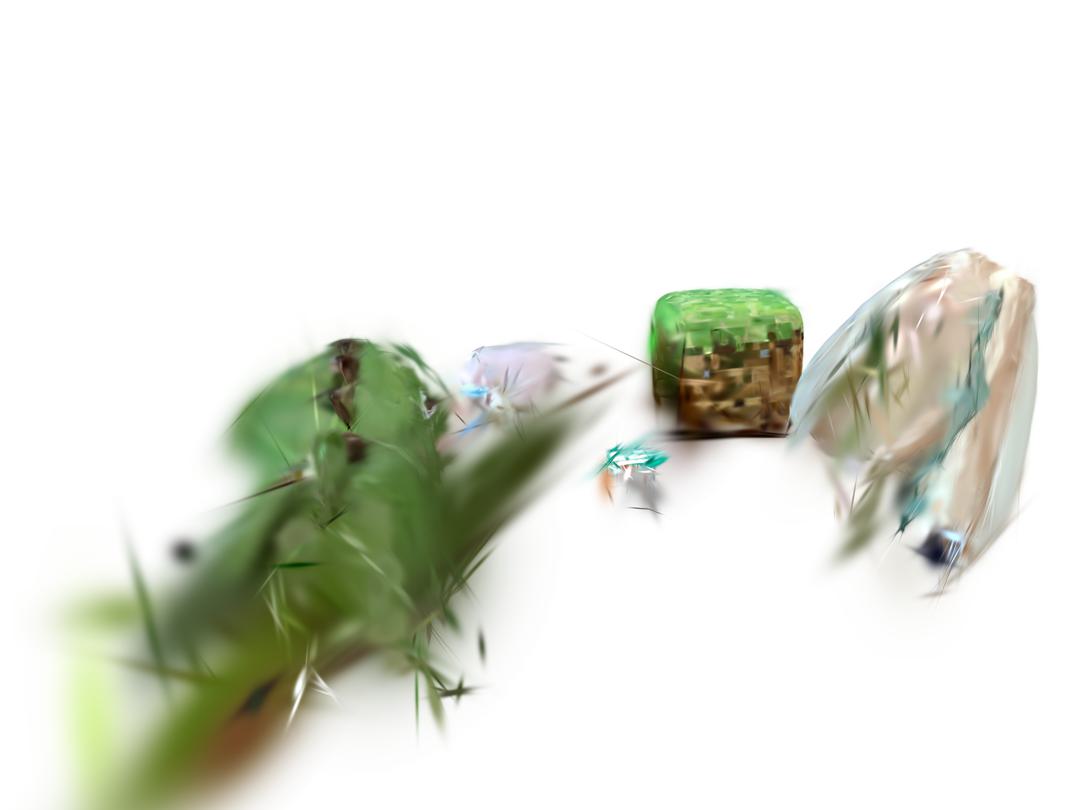} &
            \includegraphics[width=\renderwidthvis\textwidth]{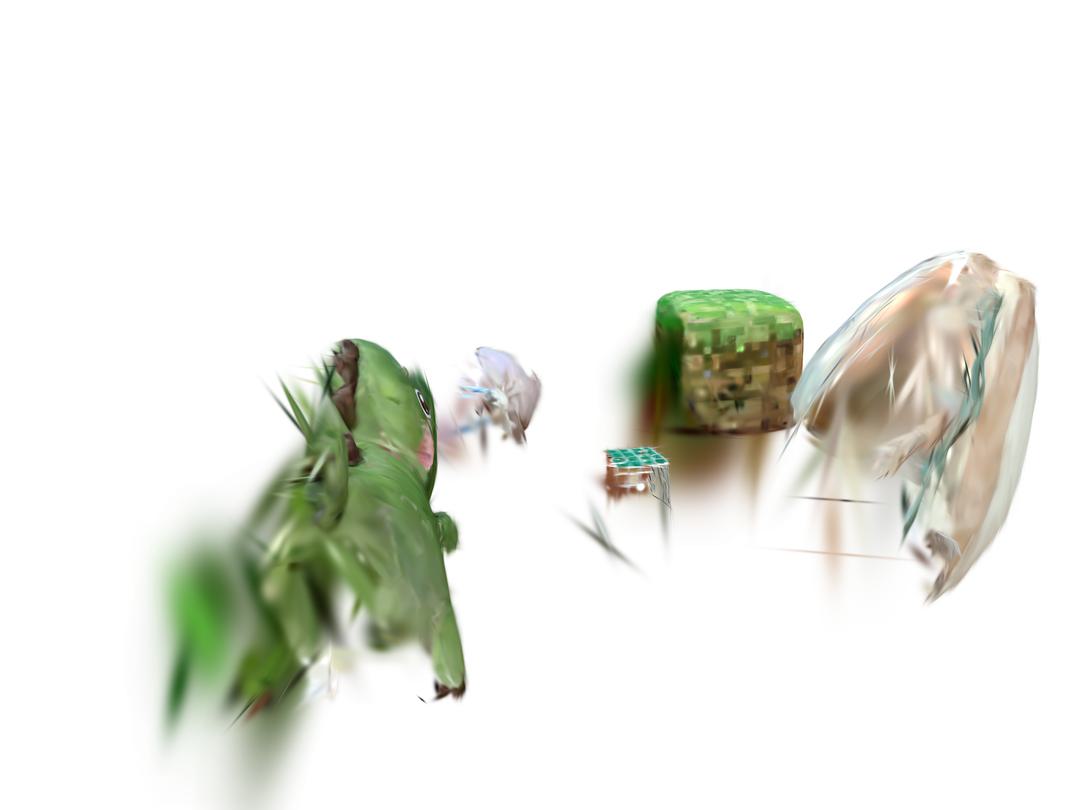} &
            \includegraphics[width=\renderwidthvis\textwidth]{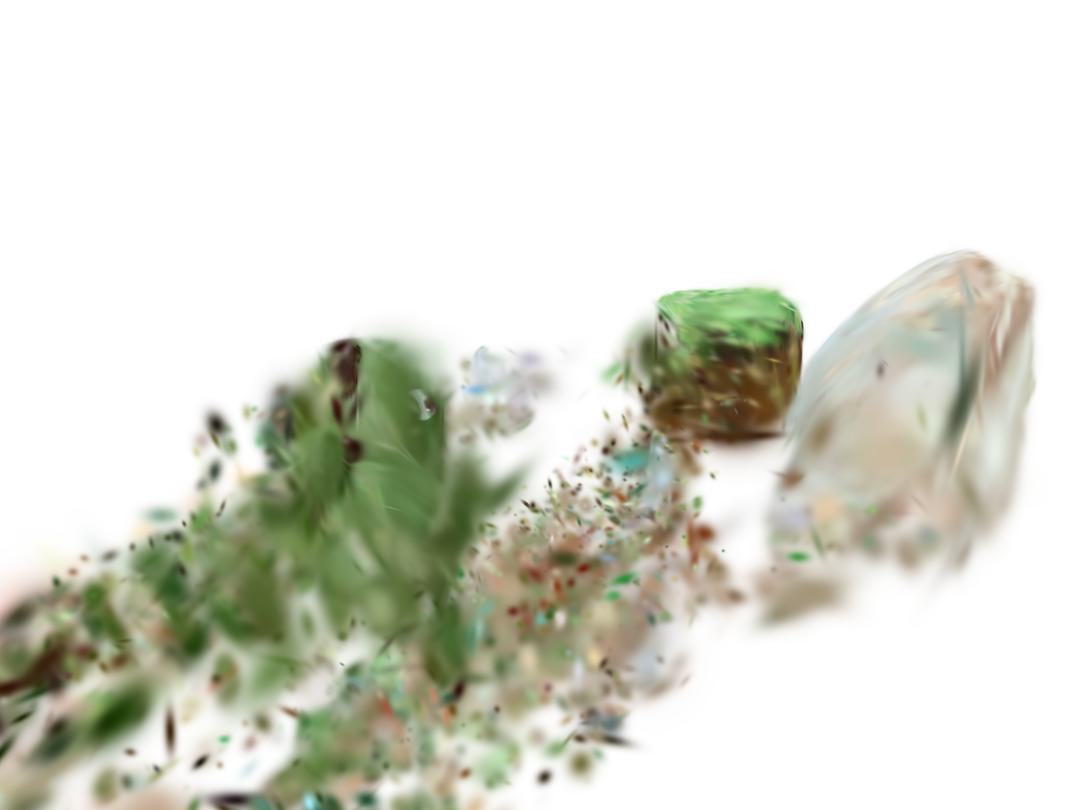} &
            \includegraphics[width=\renderwidthvis\textwidth]{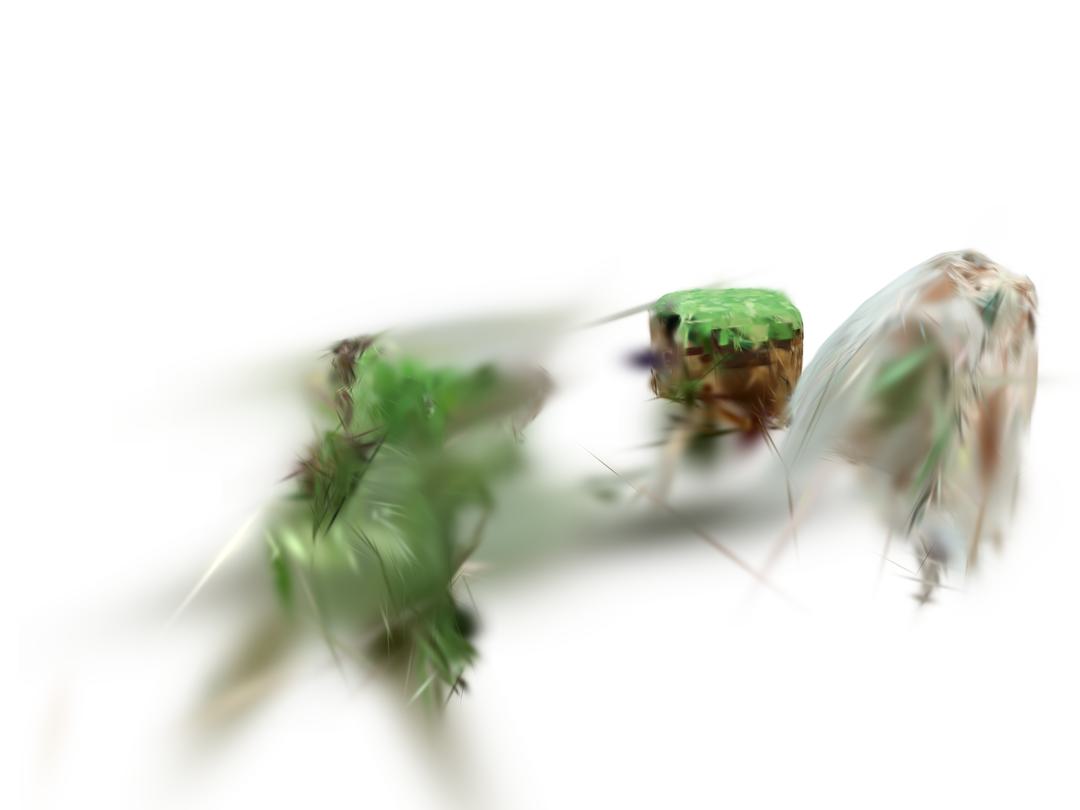} &
            \includegraphics[width=\renderwidthvis\textwidth]{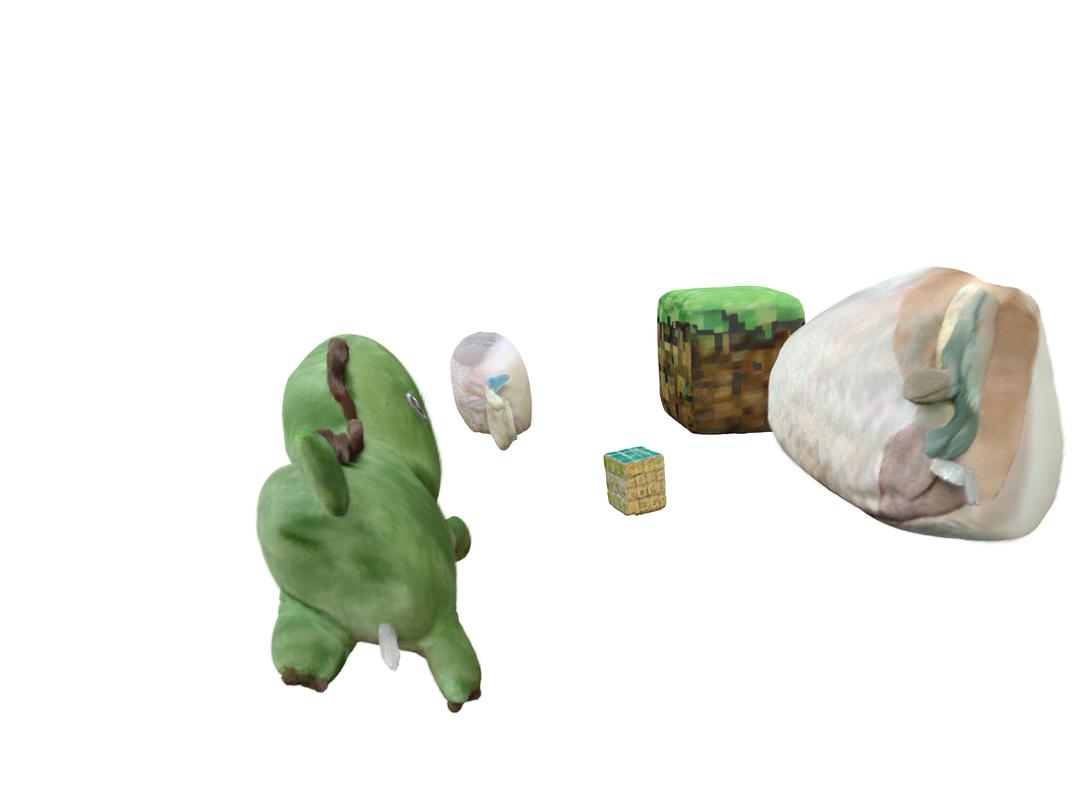} \\

            \includegraphics[width=\renderwidthvis\textwidth]{figures/qrj/0056.jpg} &
            \includegraphics[width=\renderwidthvis\textwidth]{figures/qrj/0056_3dgs.jpg} &
            \includegraphics[width=\renderwidthvis\textwidth]{figures/qrj/0056_2dgs.jpg} &
            \includegraphics[width=\renderwidthvis\textwidth]{figures/qrj/0056_dn.jpg} &
            \includegraphics[width=\renderwidthvis\textwidth]{figures/qrj/0056_genfusion.jpg} &
            \includegraphics[width=\renderwidthvis\textwidth]{figures/qrj/0056_ours.jpg} \\

            \includegraphics[width=\renderwidthvis\textwidth]{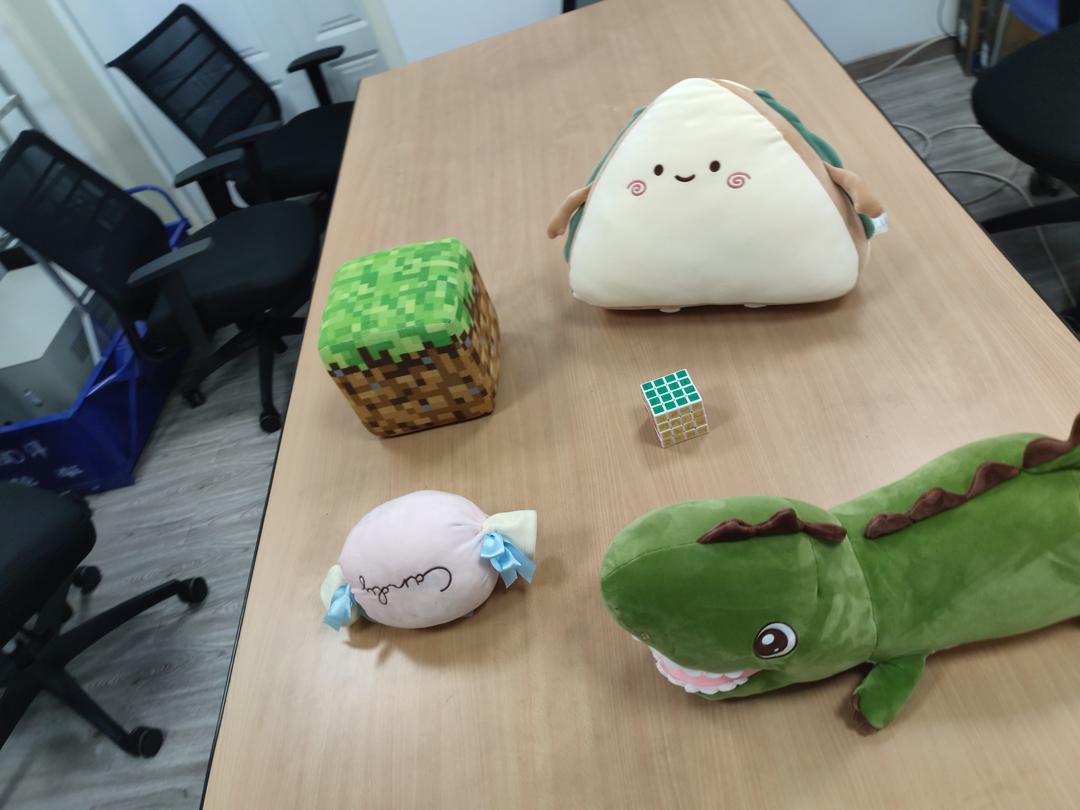} &
            \includegraphics[width=\renderwidthvis\textwidth]{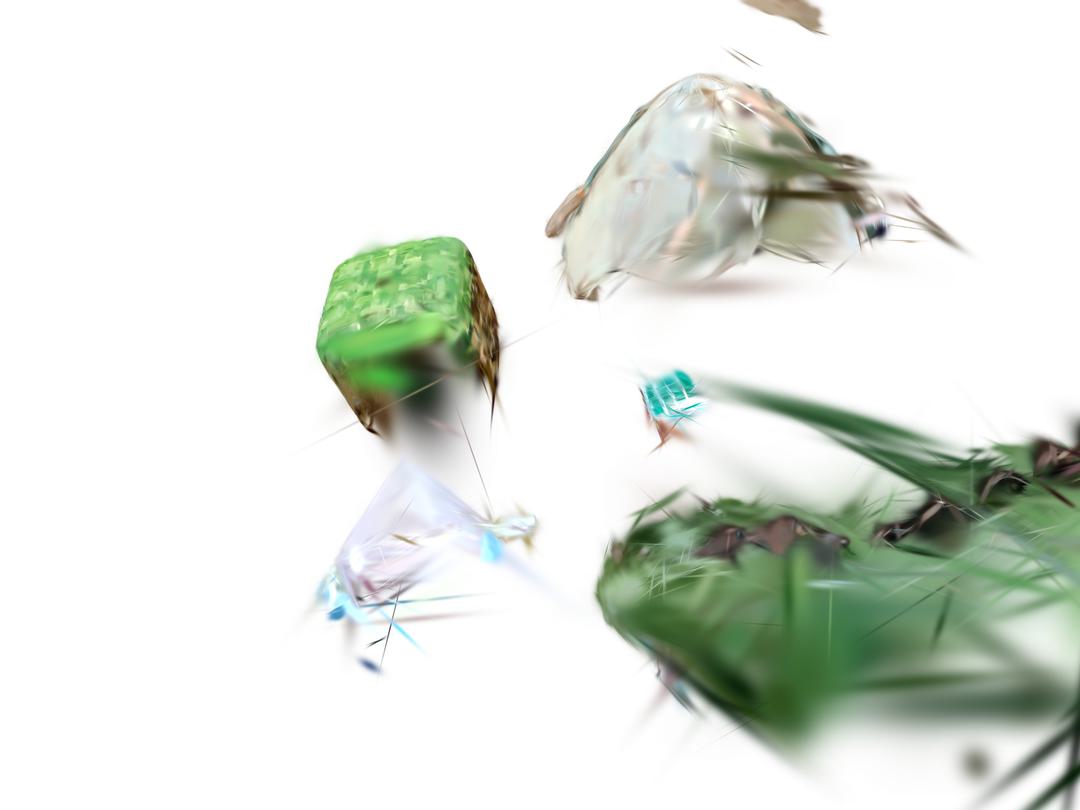} &
            \includegraphics[width=\renderwidthvis\textwidth]{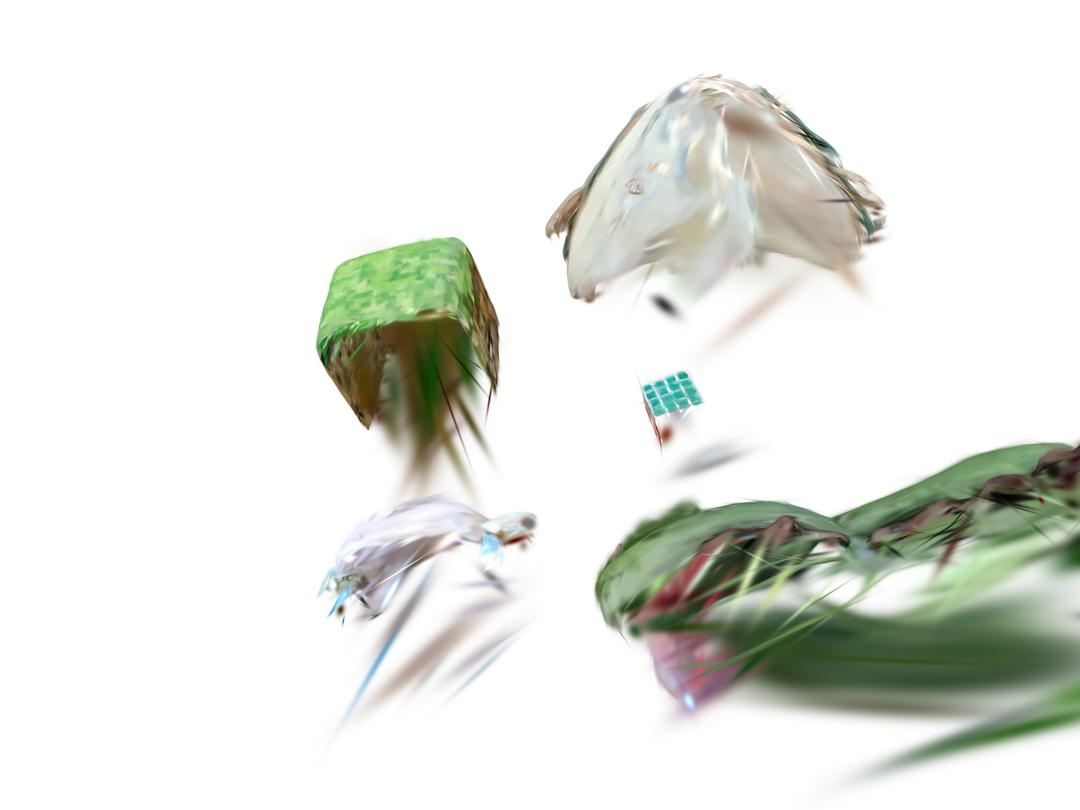} &
            \includegraphics[width=\renderwidthvis\textwidth]{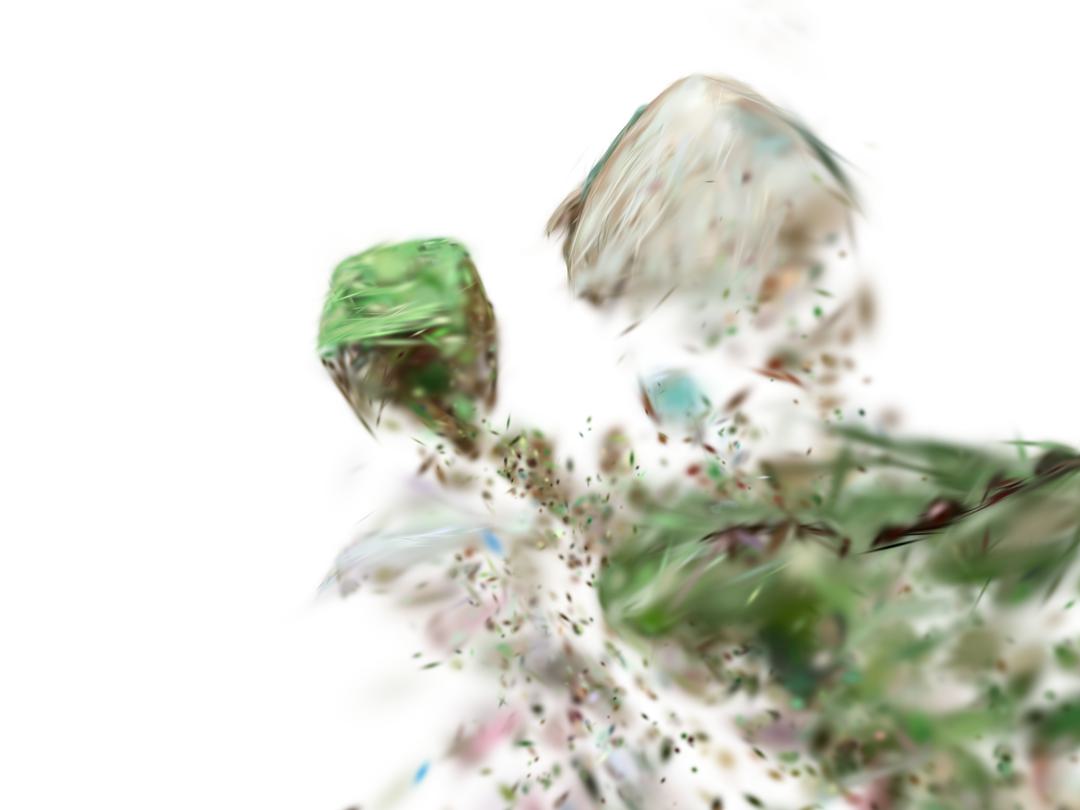} &
            \includegraphics[width=\renderwidthvis\textwidth]{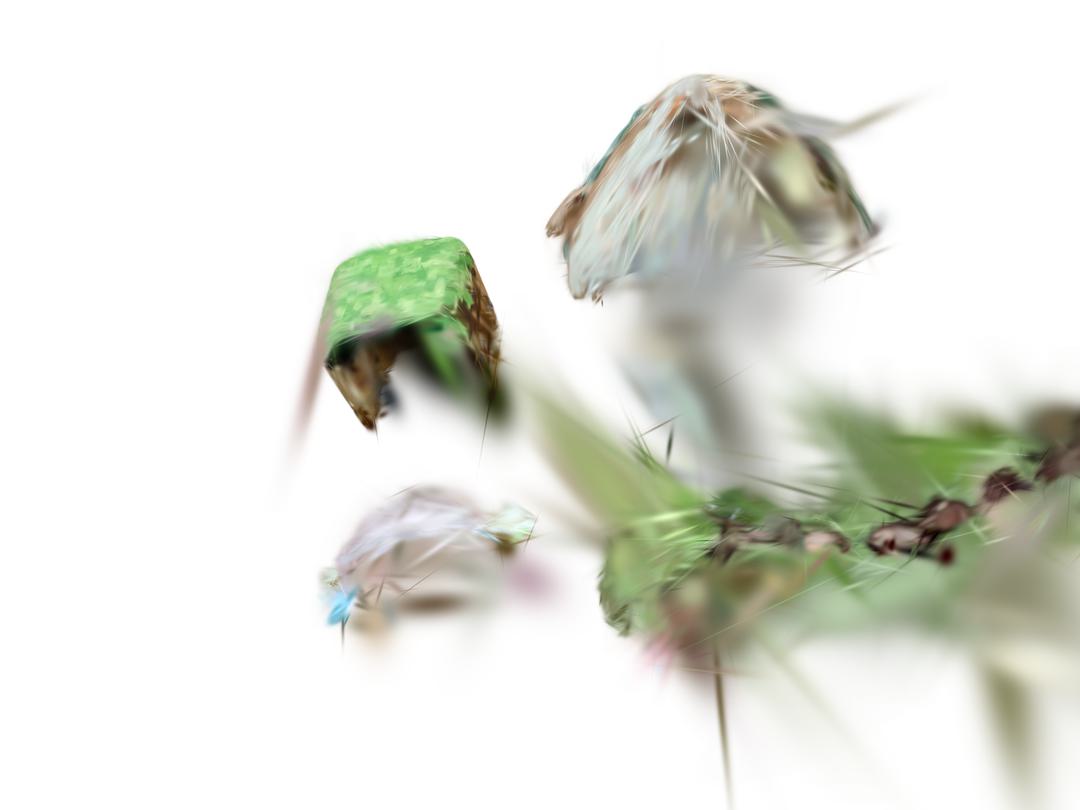} &
            \includegraphics[width=\renderwidthvis\textwidth]{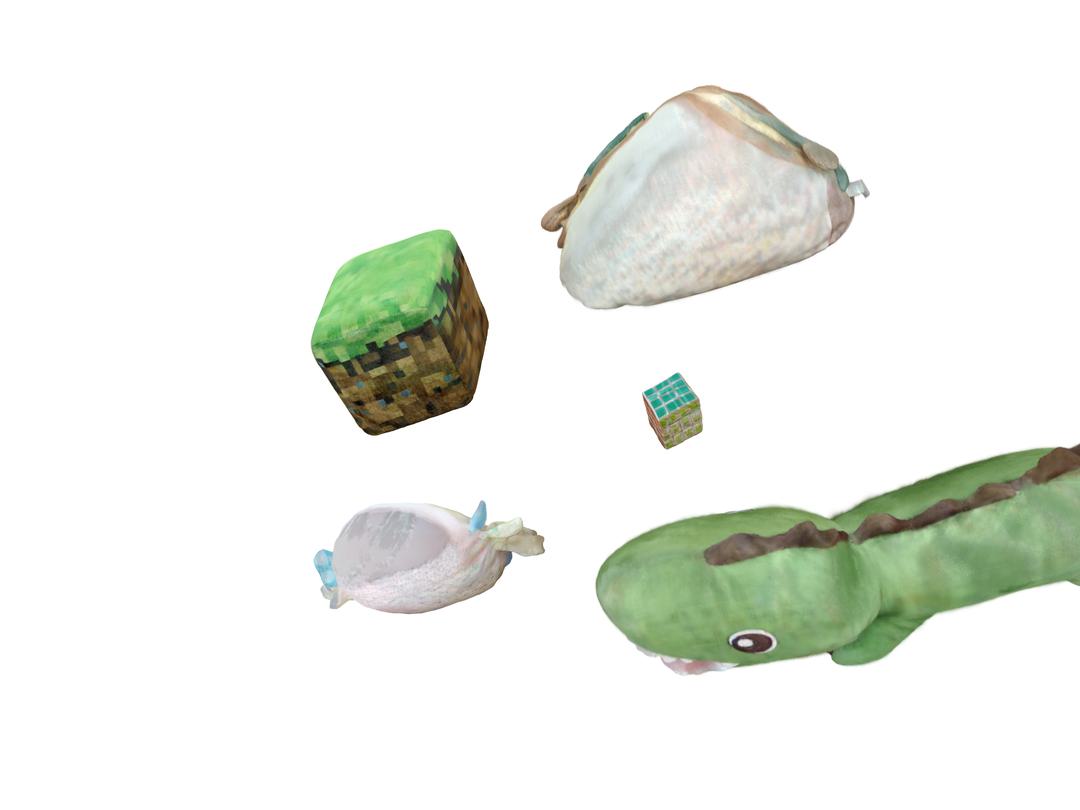} \\
            \includegraphics[width=\renderwidthvis\textwidth]{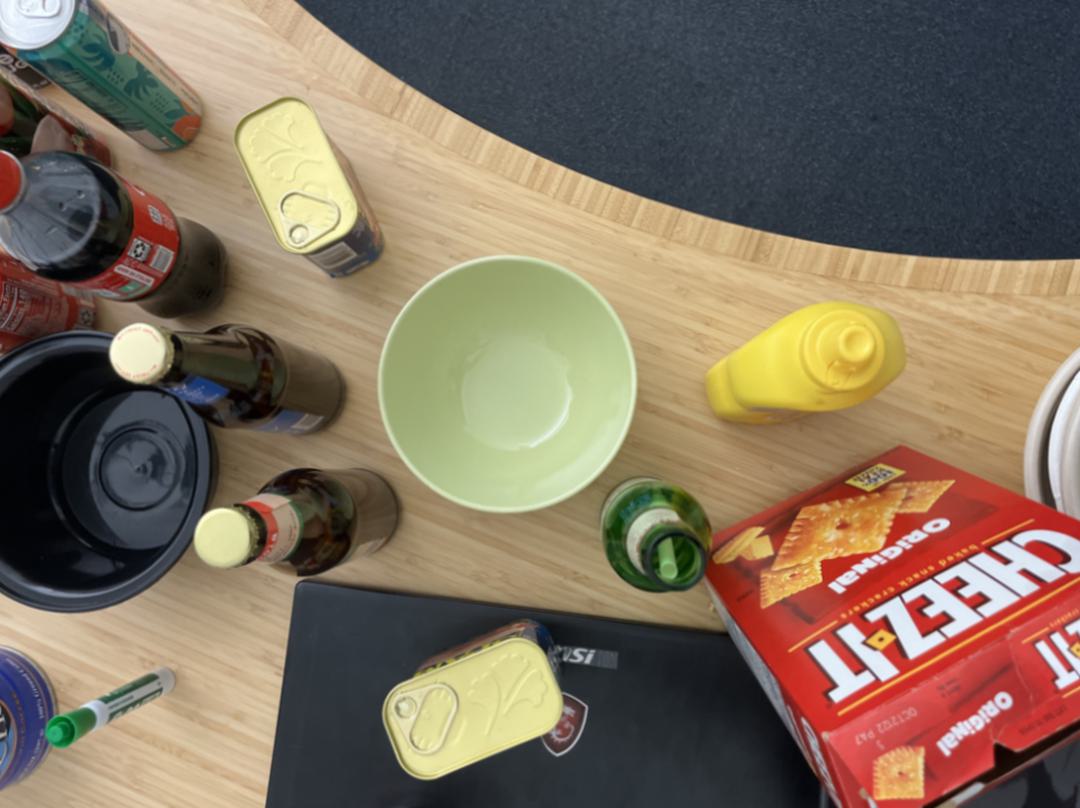} &
            \includegraphics[width=\renderwidthvis\textwidth]{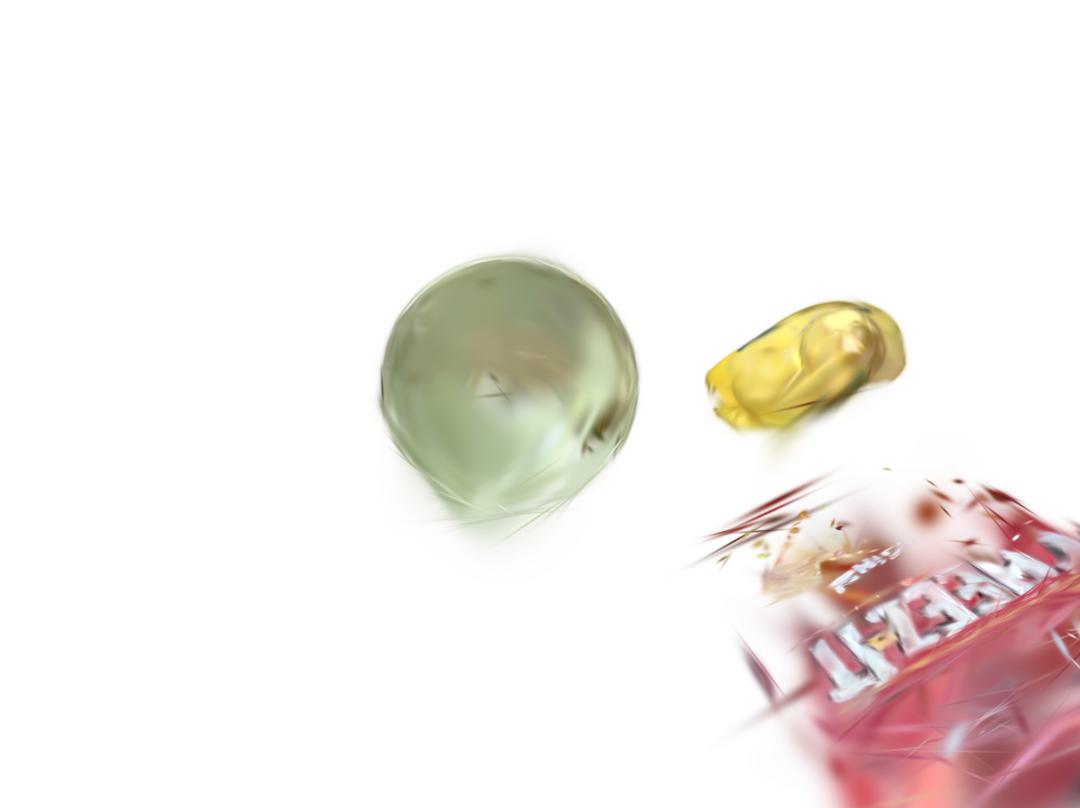} &
            \includegraphics[width=\renderwidthvis\textwidth]{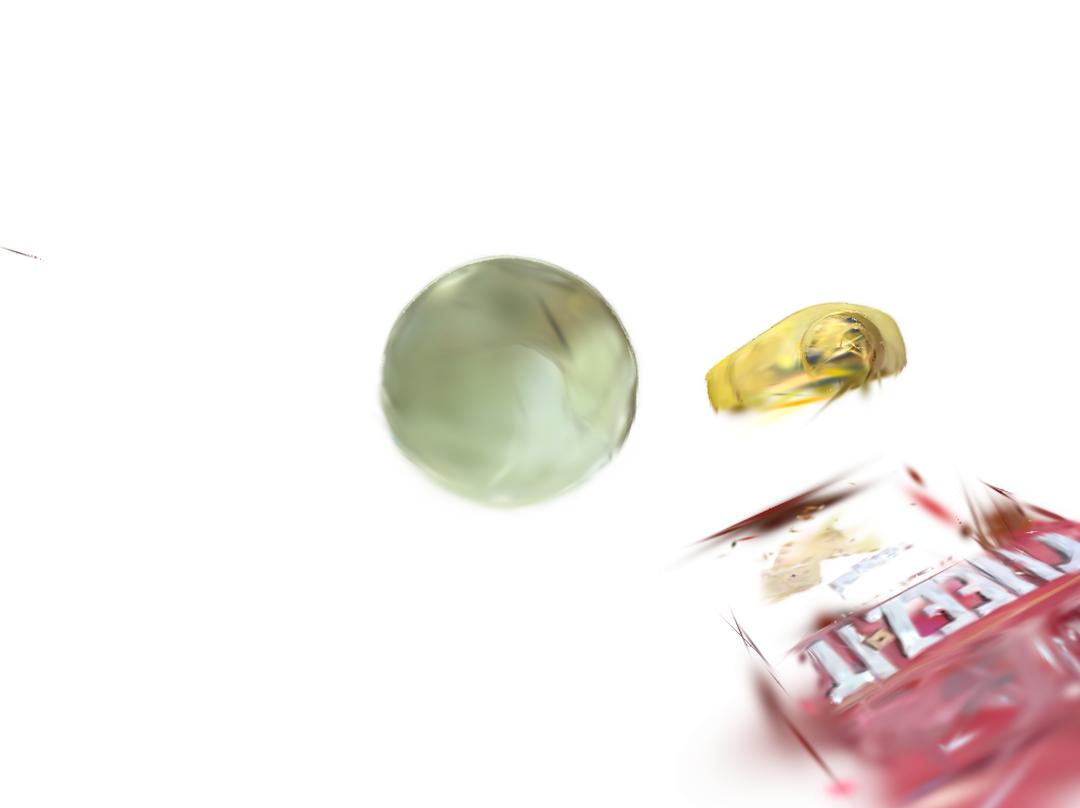} &
            \includegraphics[width=\renderwidthvis\textwidth]{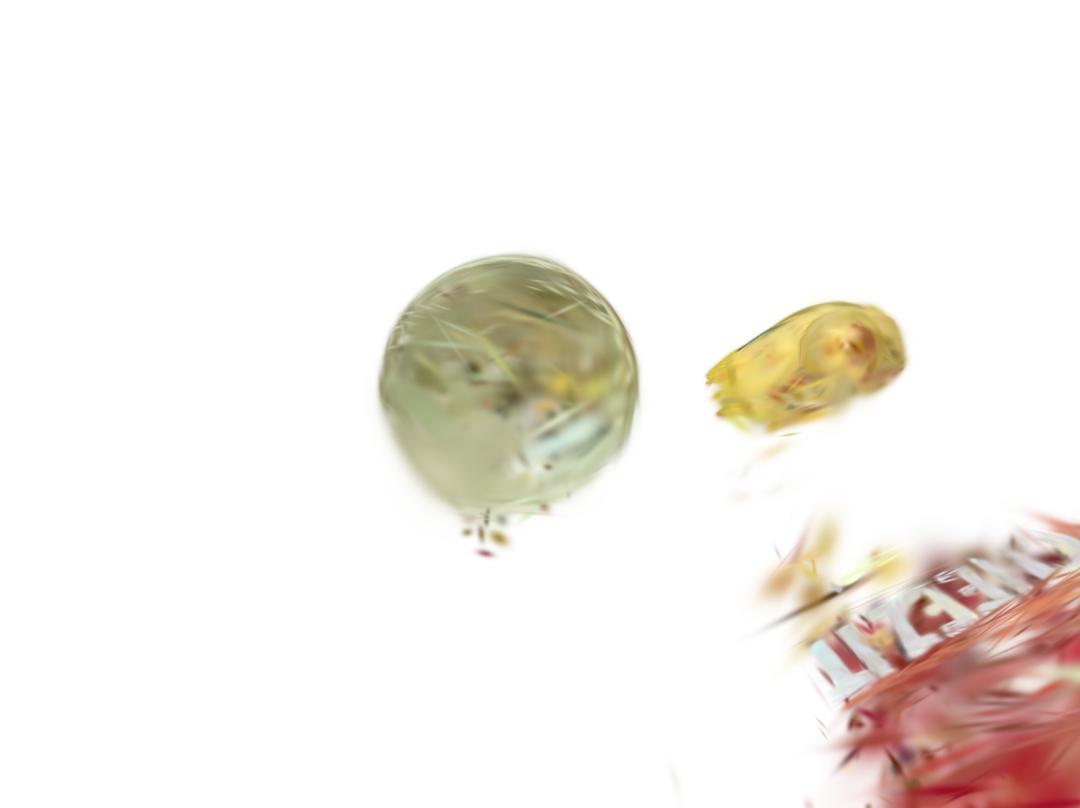} &
            \includegraphics[width=\renderwidthvis\textwidth]{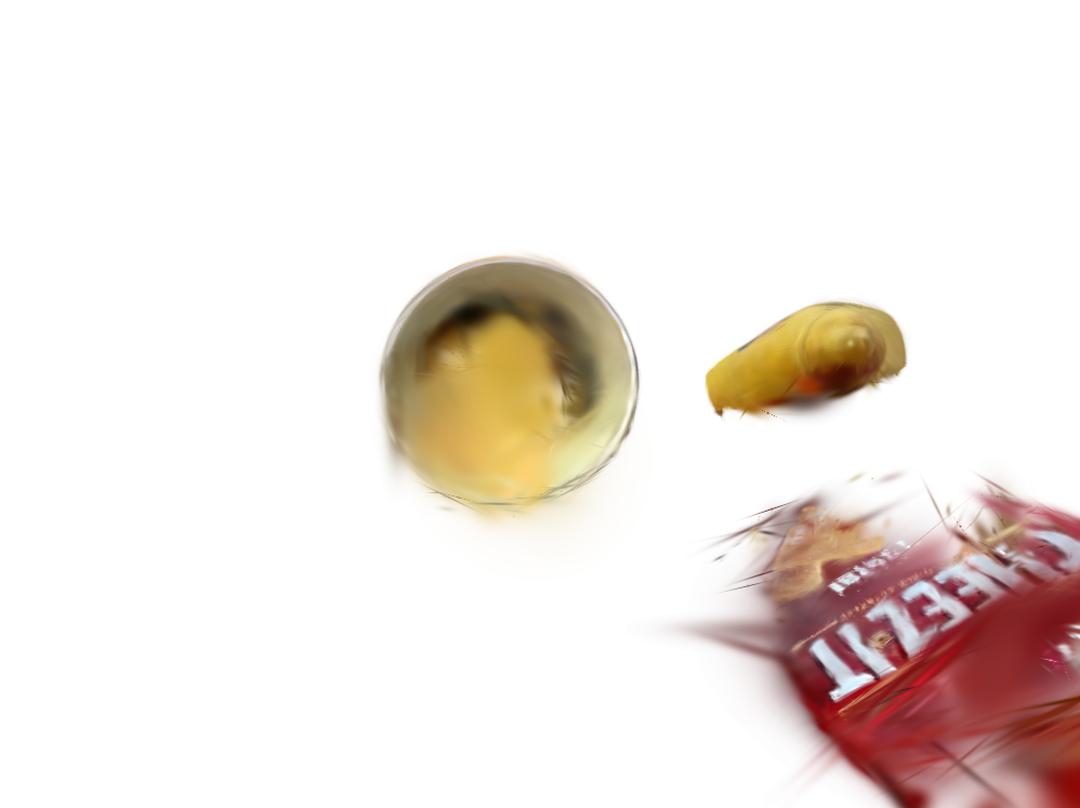} &
            \includegraphics[width=\renderwidthvis\textwidth]{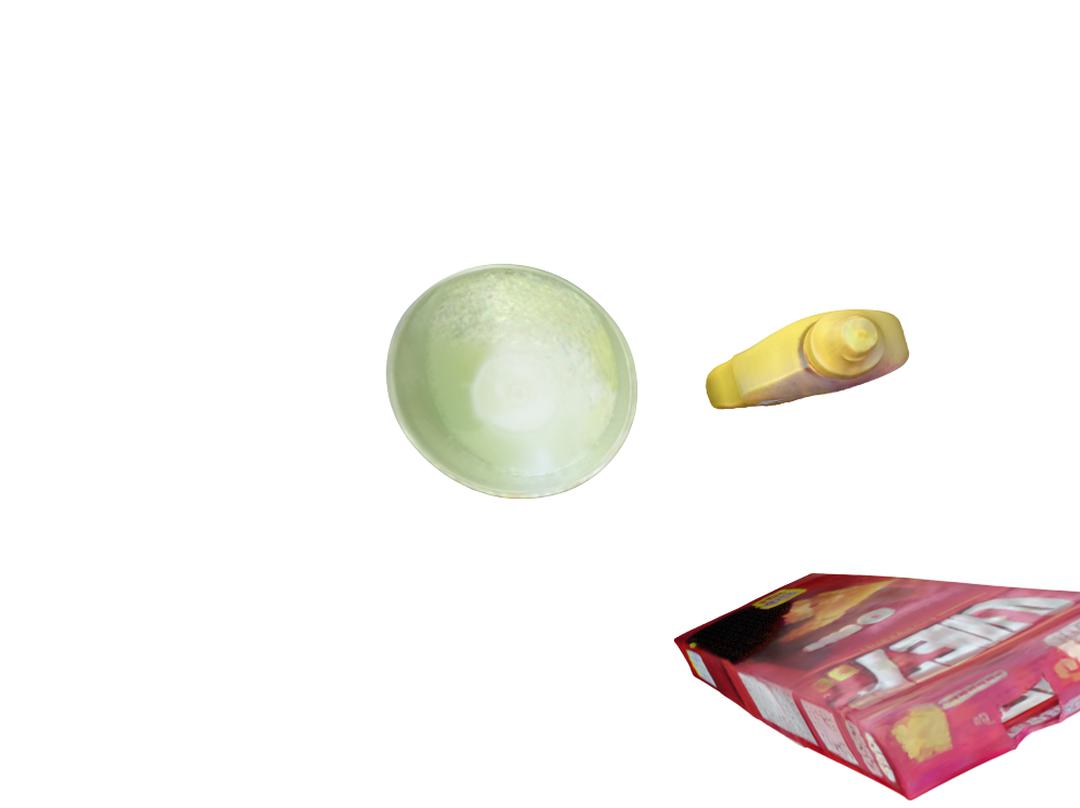} \\

            \includegraphics[width=\renderwidthvis\textwidth]{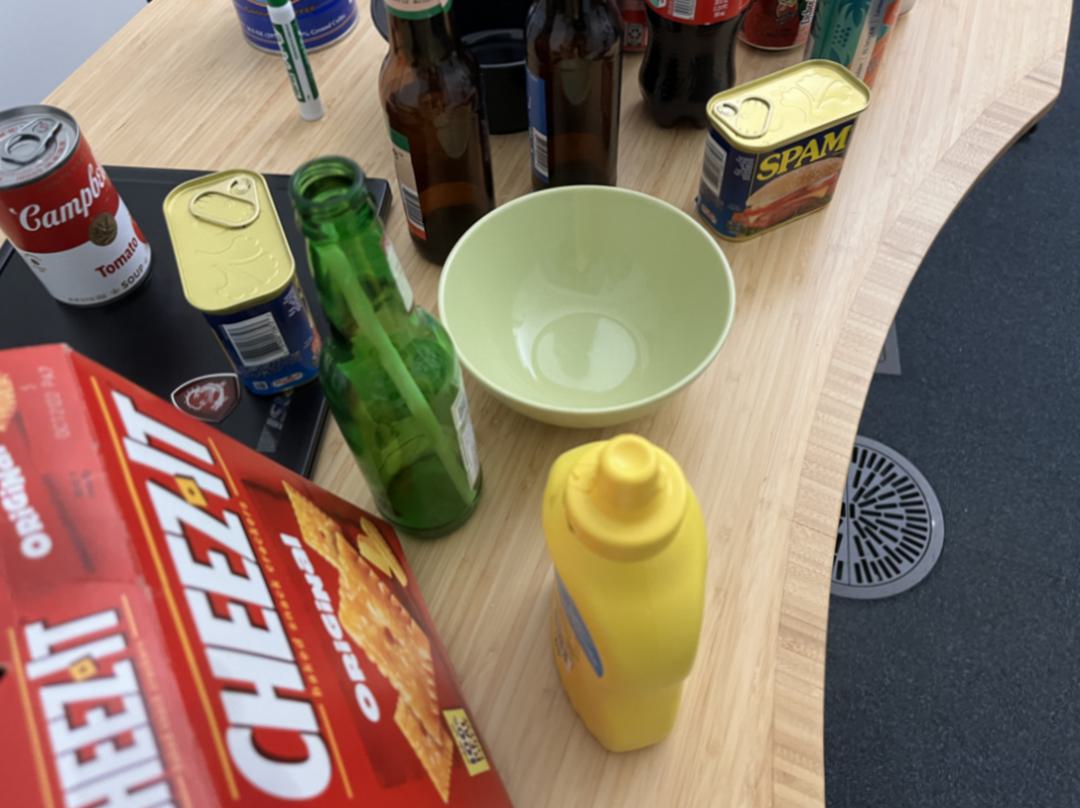} &
            \includegraphics[width=\renderwidthvis\textwidth]{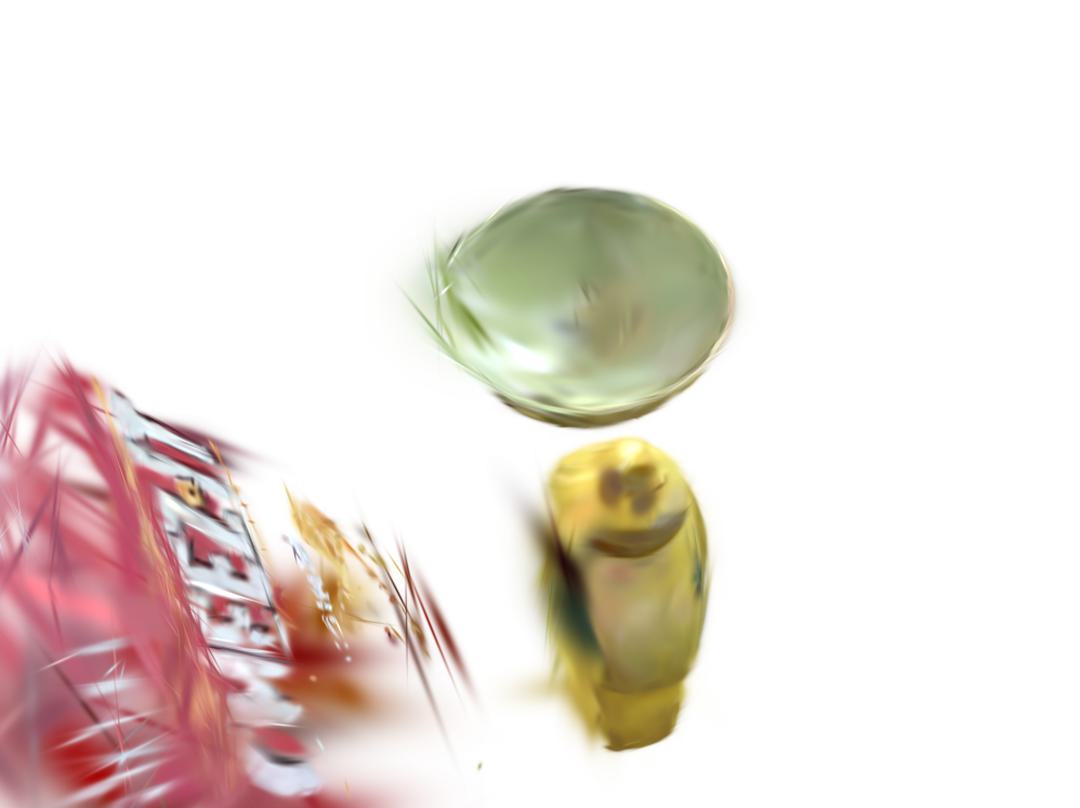} &
            \includegraphics[width=\renderwidthvis\textwidth]{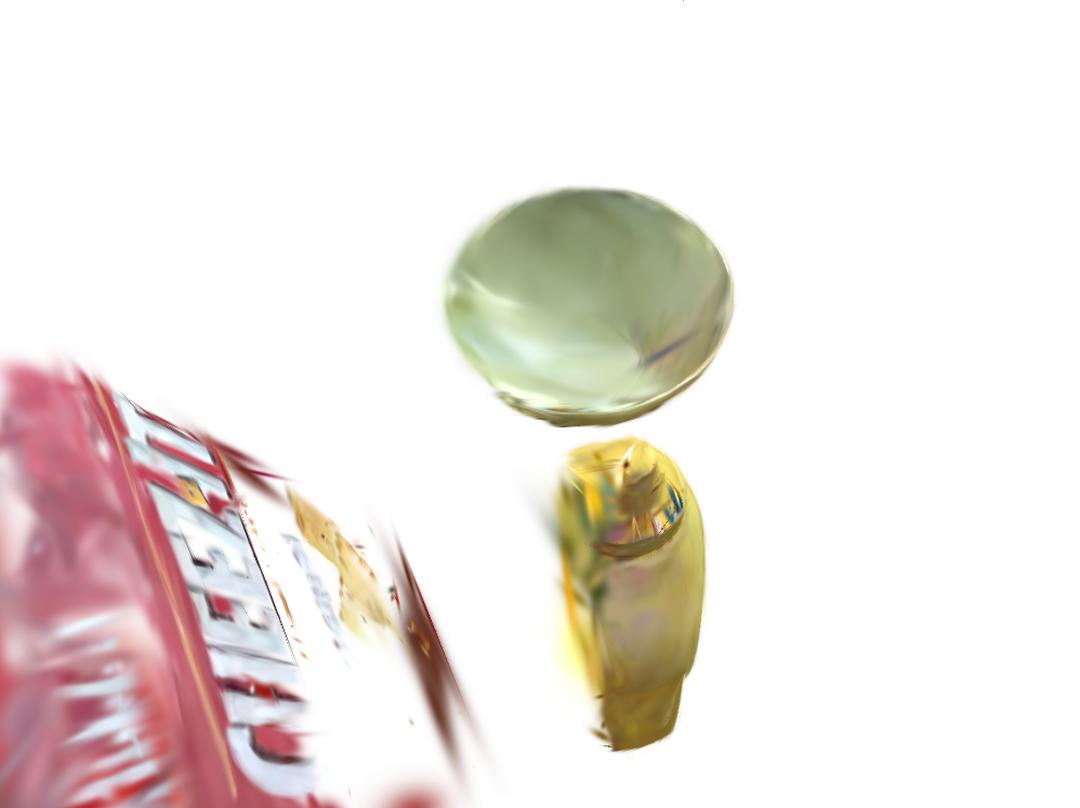} &
            \includegraphics[width=\renderwidthvis\textwidth]{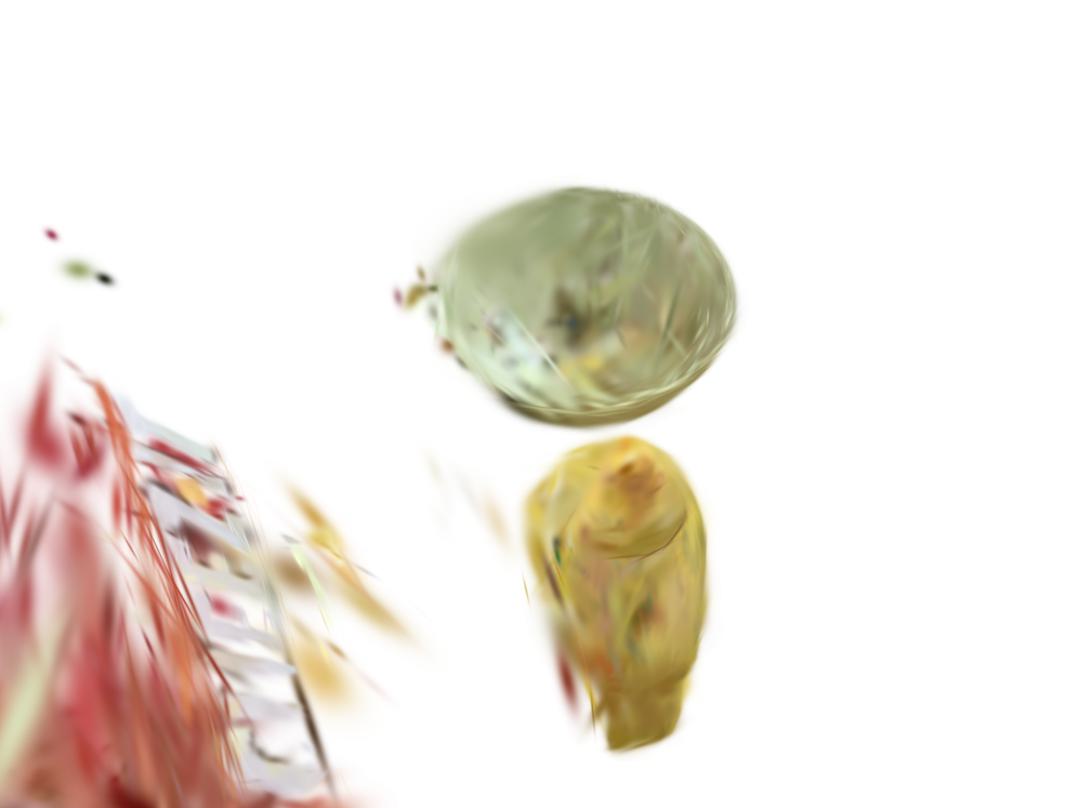} &
            \includegraphics[width=\renderwidthvis\textwidth]{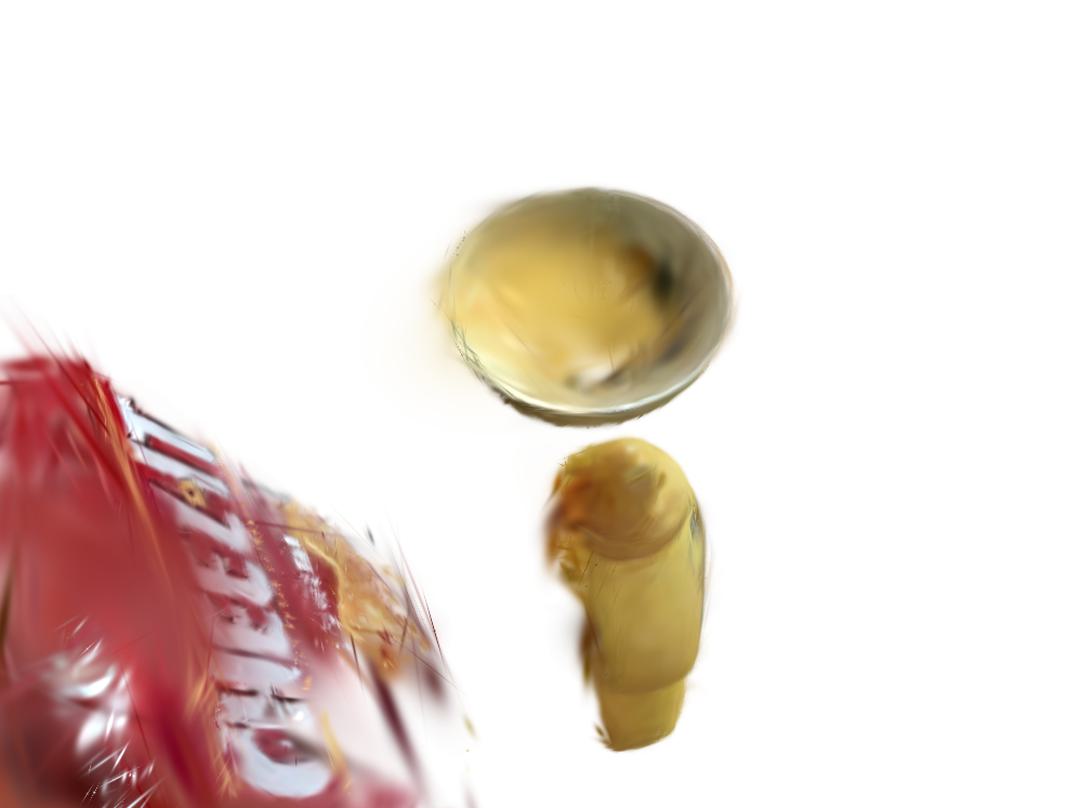} &
            \includegraphics[width=\renderwidthvis\textwidth]{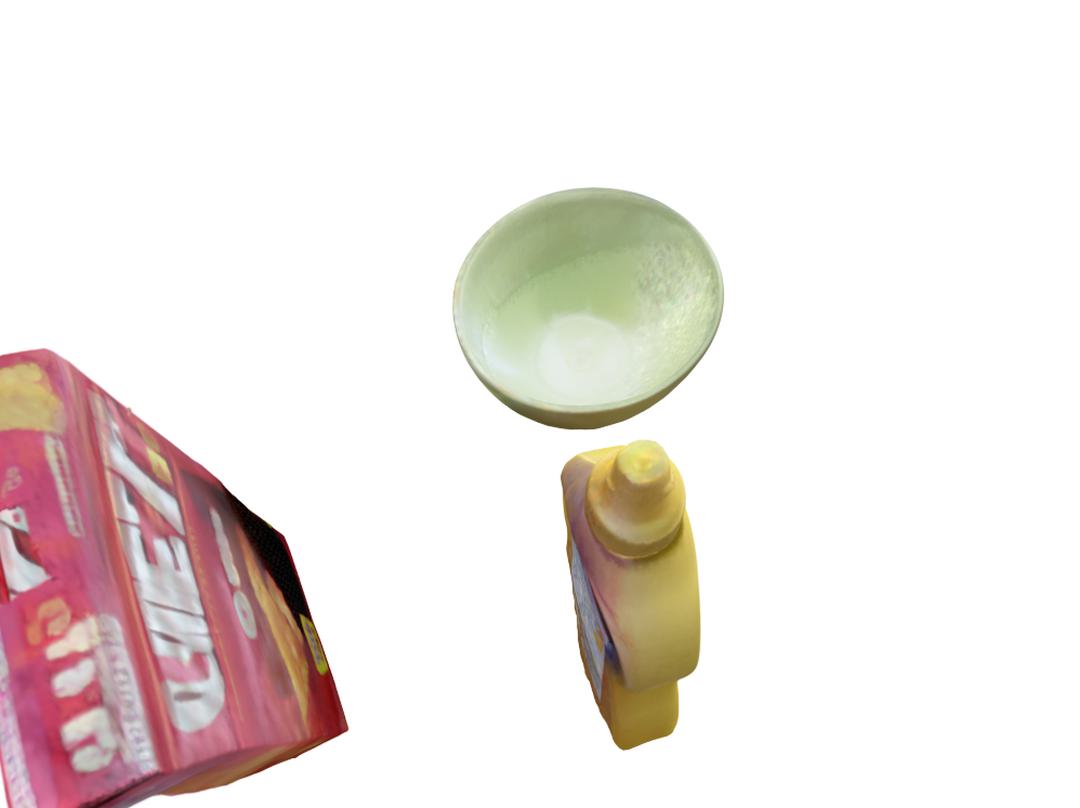} \\

            \includegraphics[width=\renderwidthvis\textwidth]{figures/qrj/frame_00162.jpg} &
            \includegraphics[width=\renderwidthvis\textwidth]{figures/qrj/frame_00162_3dgs.jpg} &
            \includegraphics[width=\renderwidthvis\textwidth]{figures/qrj/frame_00162_2dgs.jpg} &
            \includegraphics[width=\renderwidthvis\textwidth]{figures/qrj/frame_00162_dn.jpg} &
            \includegraphics[width=\renderwidthvis\textwidth]{figures/qrj/frame_00162_genfusion.jpg} &
            \includegraphics[width=\renderwidthvis\textwidth]{figures/qrj/frame_00162_ours.jpg} \\

            GT & 3DGS~\cite{3dgs} & 2DGS~\cite{2dgs} & DNGaussian~\cite{dngaussian} & GenFusion~\cite{genfusion} & Ours \\
        \end{tabular}
    }
    \vspace{-1em}
    \caption{Qualitative results of rendered images of target objects produced by \ours and several baseline methods. From top to bottom, images are selected from Mip360-\textit{garden} \texttt{DSC08018 DSC08038 DSC08130}, Mip360-\textit{kitchen} \texttt{DSCF0740 DSCF0759 DSCF0860}, ToyDesk-\textit{scene2} \texttt{0052 0056 0145} and 3DGS-CD-\textit{Desk} \texttt{frame\_00020 frame\_00090 frame\_00162}.}
    \label{fig:qualitative_full}
    \vspace{-2em}
\end{figure*}

\newcommand{\renderwidthgeo}{0.19}

\begin{figure*}[!htb]
    \centering
    \addtolength{\tabcolsep}{-6.5pt}
    \footnotesize{
        \setlength{\tabcolsep}{1pt}
        \begin{tabular}{@{}ccccc@{}}
            \includegraphics[width=\renderwidthgeo\textwidth]{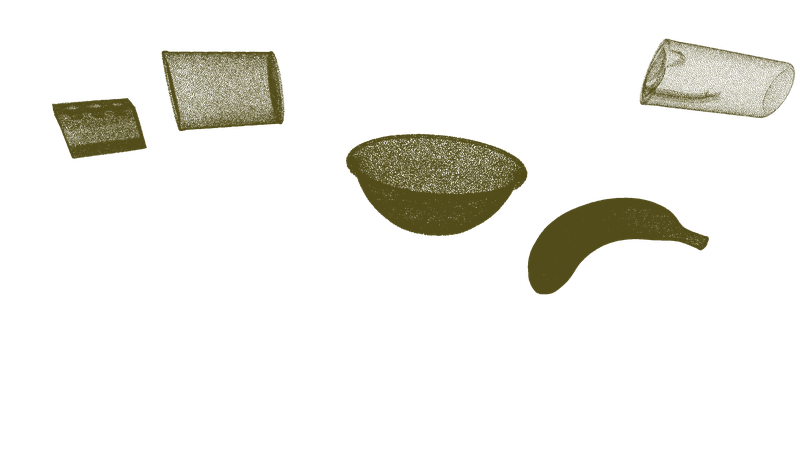} &
            \includegraphics[width=\renderwidthgeo\textwidth]{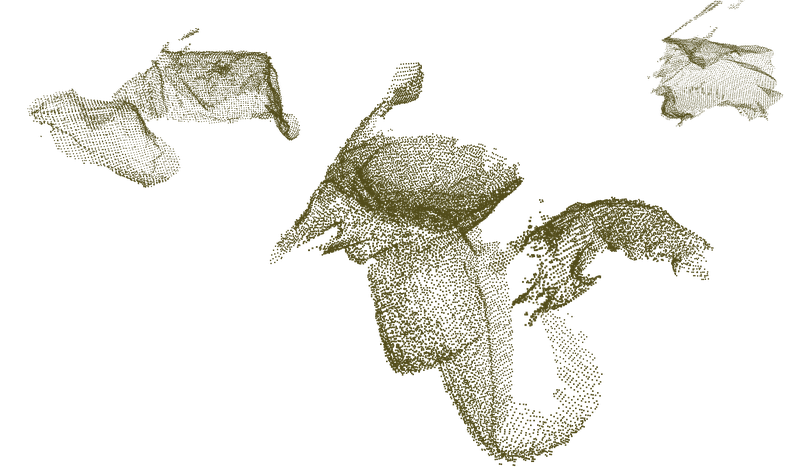} &
            \includegraphics[width=\renderwidthgeo\textwidth]{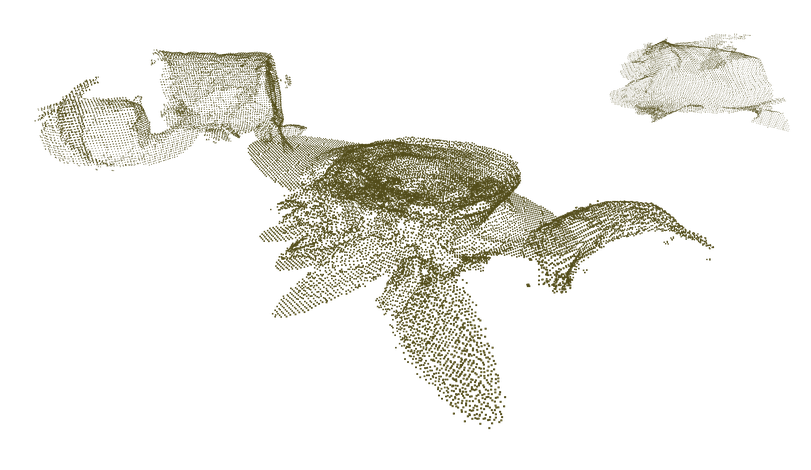} &
            \includegraphics[width=\renderwidthgeo\textwidth]{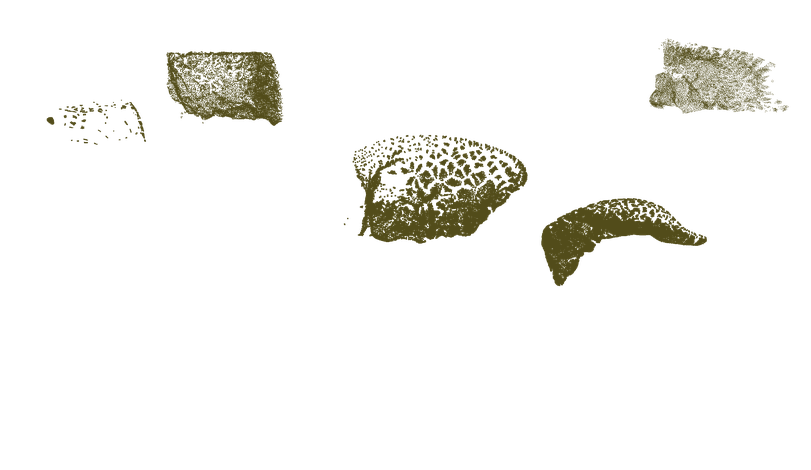} &
            \includegraphics[width=\renderwidthgeo\textwidth]{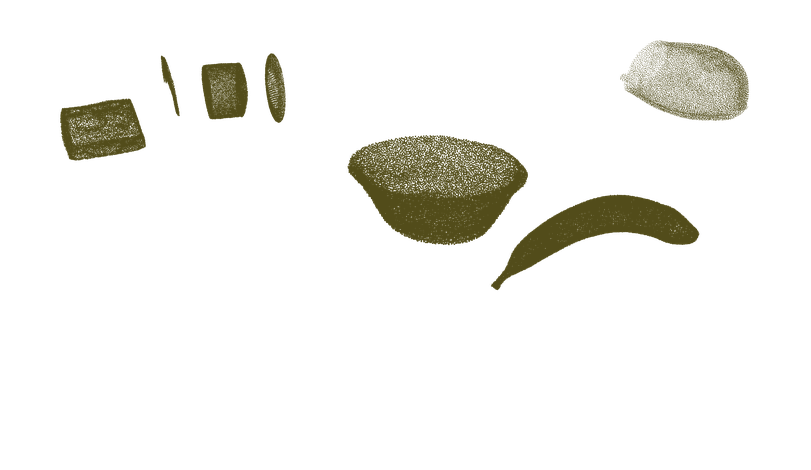} \\

            \includegraphics[width=\renderwidthgeo\textwidth]{figures/supply_geo_resized/scene_1_gt.ply00.png} &
            \includegraphics[width=\renderwidthgeo\textwidth]{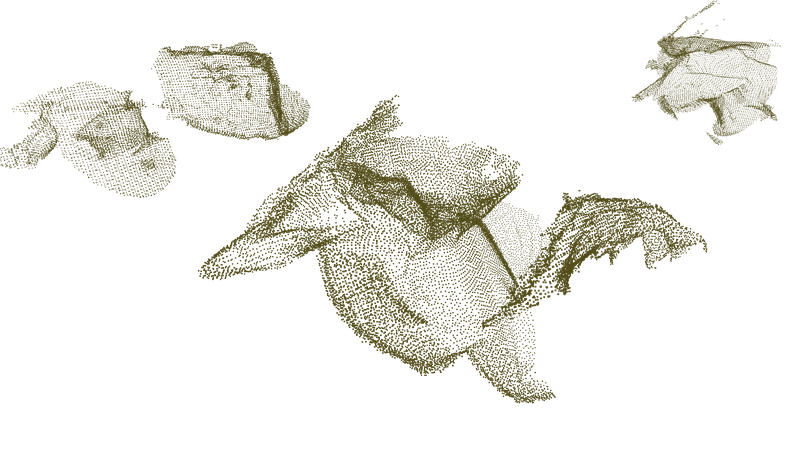} &
            \includegraphics[width=\renderwidthgeo\textwidth]{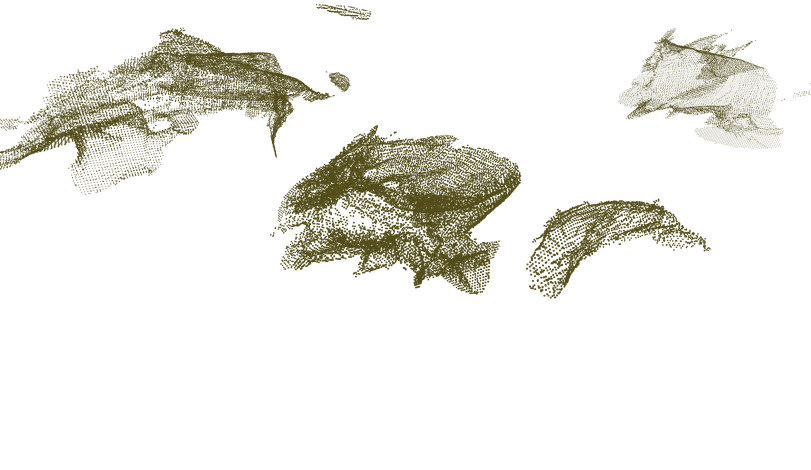} &
            \includegraphics[width=\renderwidthgeo\textwidth]{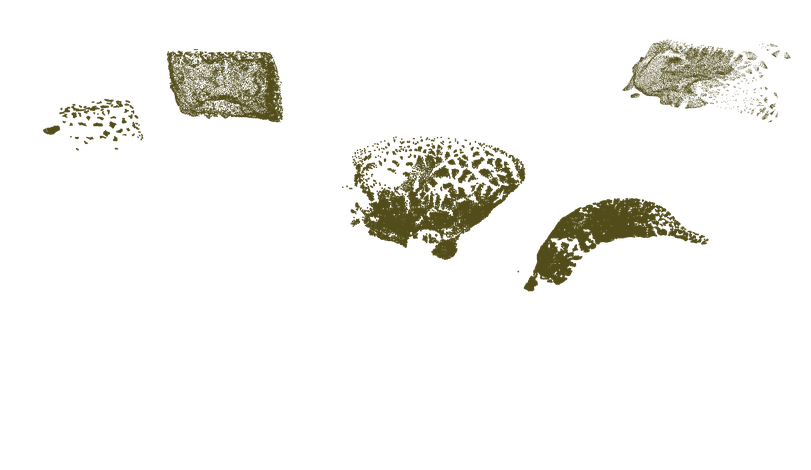} &
            \includegraphics[width=\renderwidthgeo\textwidth]{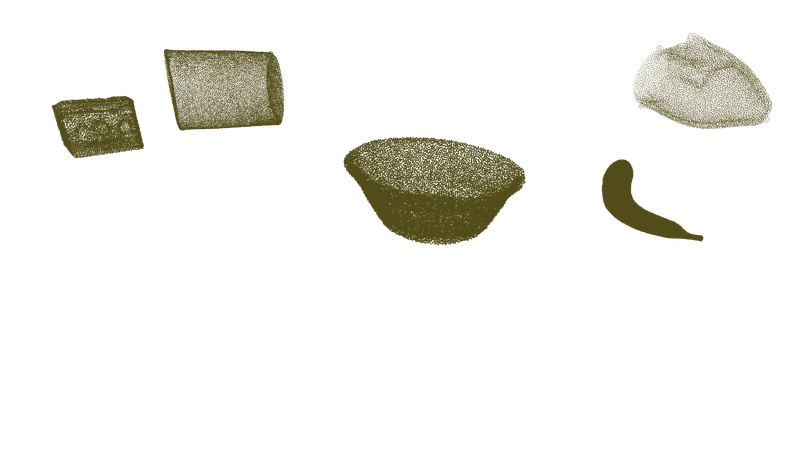} \\

            \includegraphics[width=\renderwidthgeo\textwidth]{figures/supply_geo_resized/scene_1_gt.ply00.png} &
            \includegraphics[width=\renderwidthgeo\textwidth]{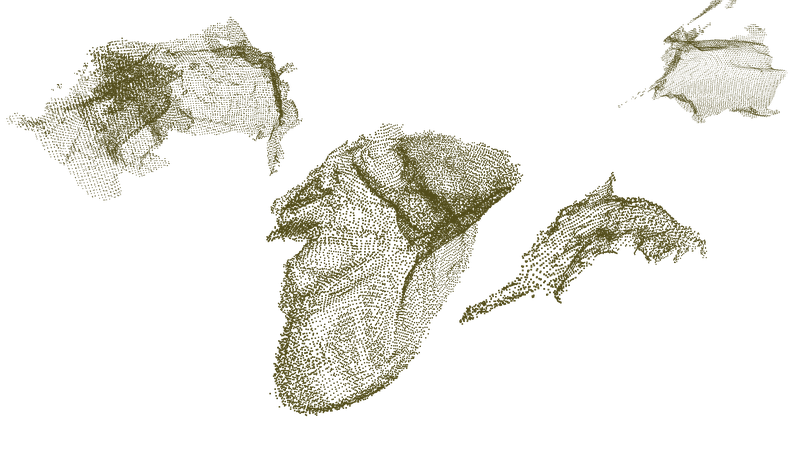} &
            \includegraphics[width=\renderwidthgeo\textwidth]{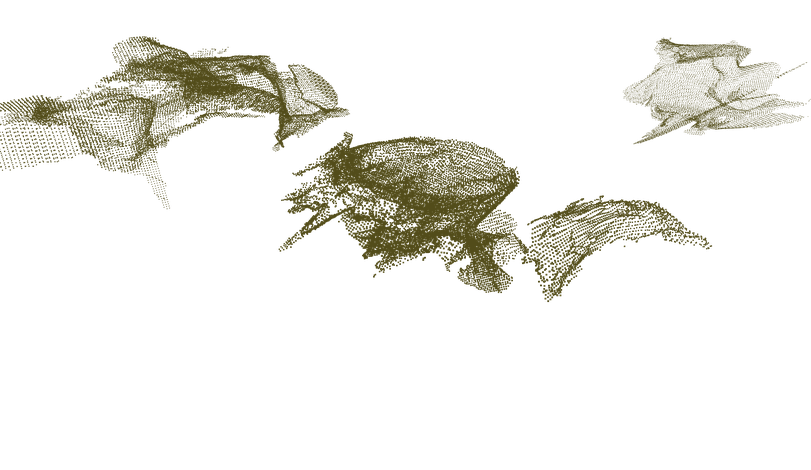} &
            \includegraphics[width=\renderwidthgeo\textwidth]{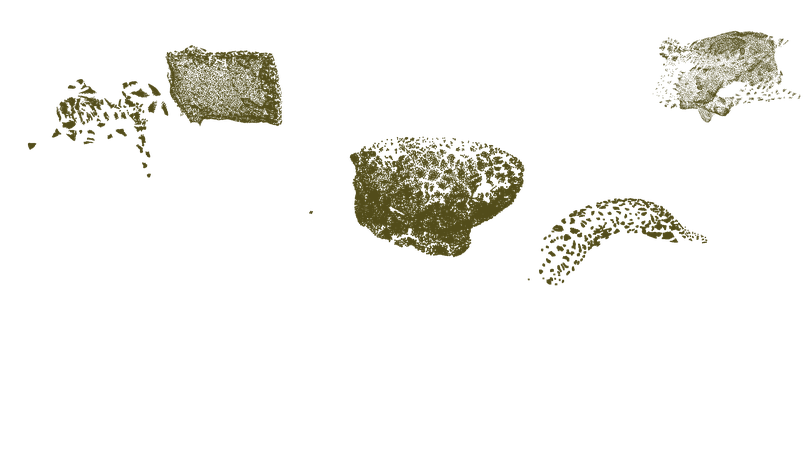} &
            \includegraphics[width=\renderwidthgeo\textwidth]{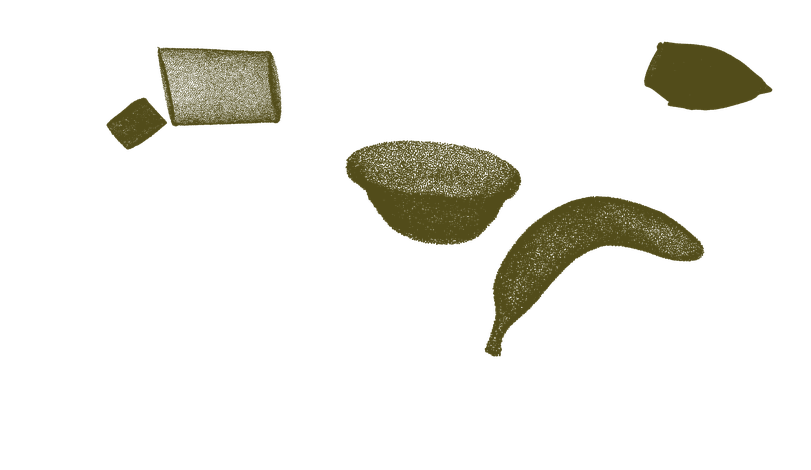} \\
            GT & DNGaussian~\cite{dngaussian} & Genfusion~\cite{genfusion} & ComPC~\cite{ComPC} & Ours \\
        \end{tabular}
    }
    \vspace{-5pt}
    \caption{Qualitative results of point cloud data for Scene 1 in our synthetic dataset. From top to bottom: easy, medium, and hard settings. All point clouds are downsampled to 16384 points to keep similar visual effect.}
    \label{fig:qualitative_compare_geo_scene1}
\end{figure*}

\begin{figure*}[!htb]
    \centering
    \addtolength{\tabcolsep}{-6.5pt}
    \footnotesize{
        \setlength{\tabcolsep}{1pt}
        \begin{tabular}{@{}ccccc@{}}
            \includegraphics[width=\renderwidthgeo\textwidth]{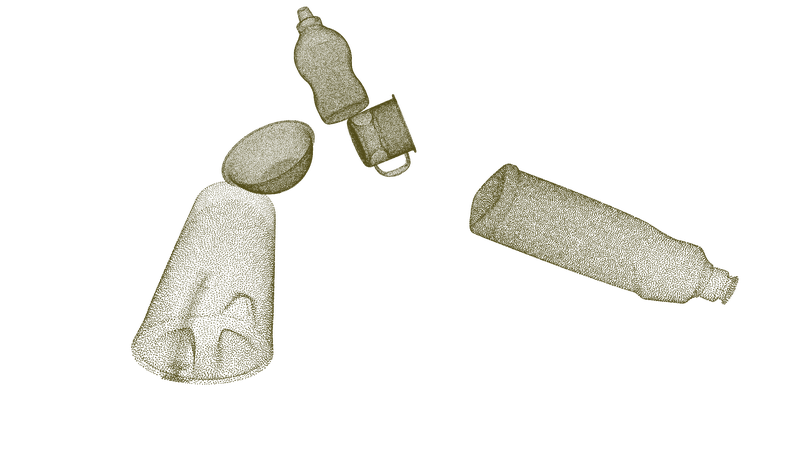} &
            \includegraphics[width=\renderwidthgeo\textwidth]{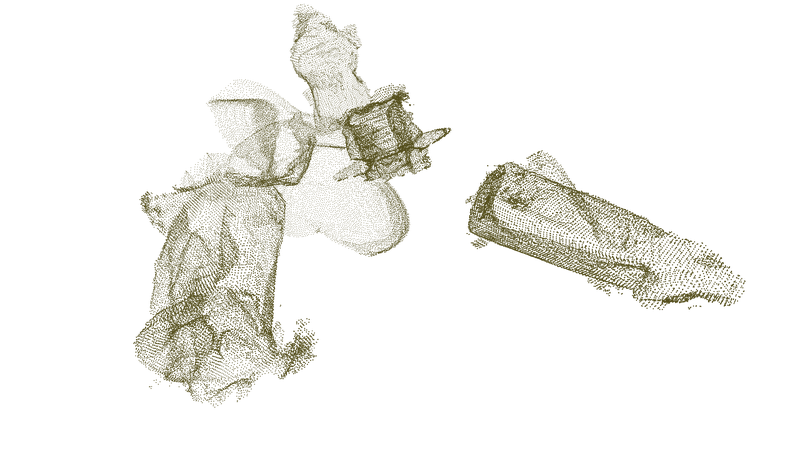} &
            \includegraphics[width=\renderwidthgeo\textwidth]{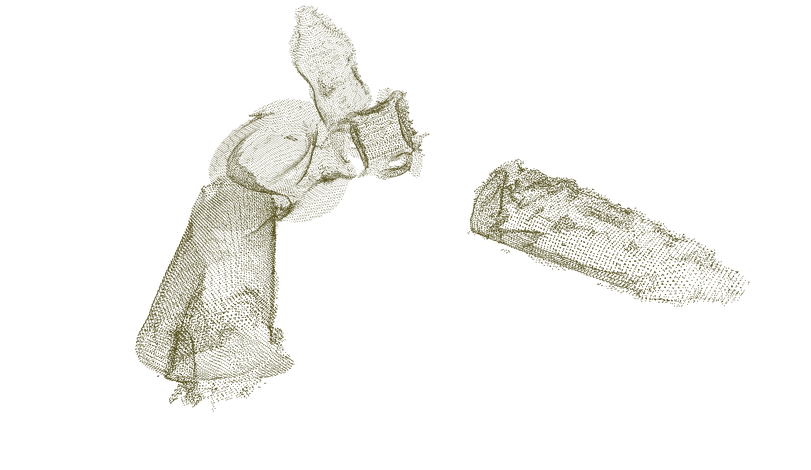} &
            \includegraphics[width=\renderwidthgeo\textwidth]{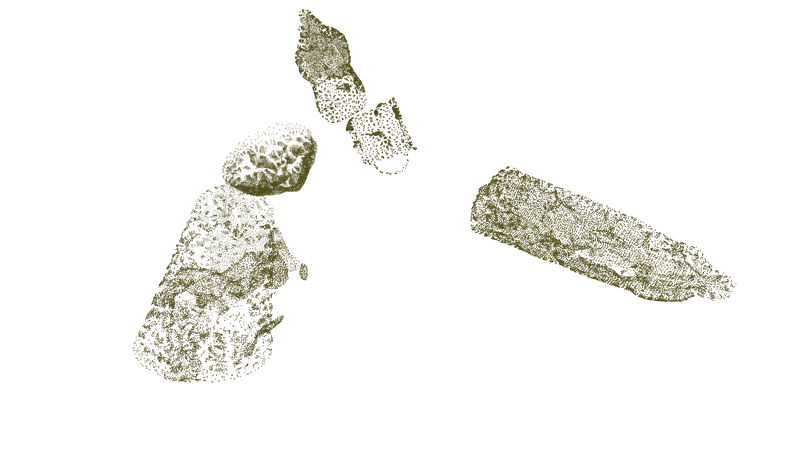} &
            \includegraphics[width=\renderwidthgeo\textwidth]{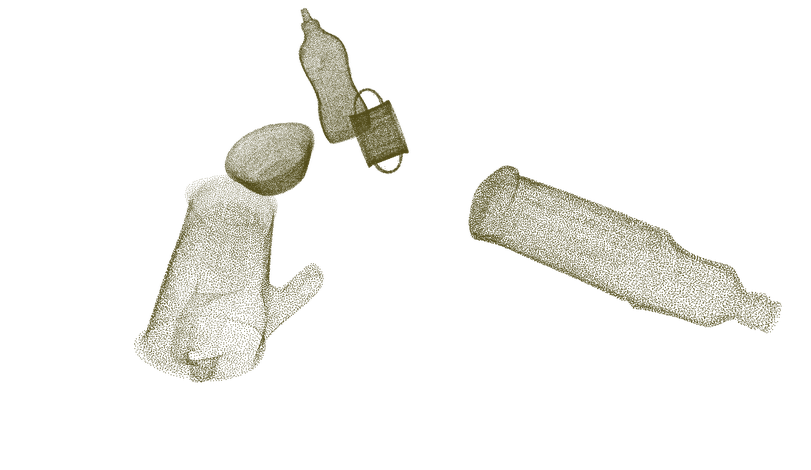} \\

            \includegraphics[width=\renderwidthgeo\textwidth]{figures/supply_geo_resized/scene_2_gt.ply00.png} &
            \includegraphics[width=\renderwidthgeo\textwidth]{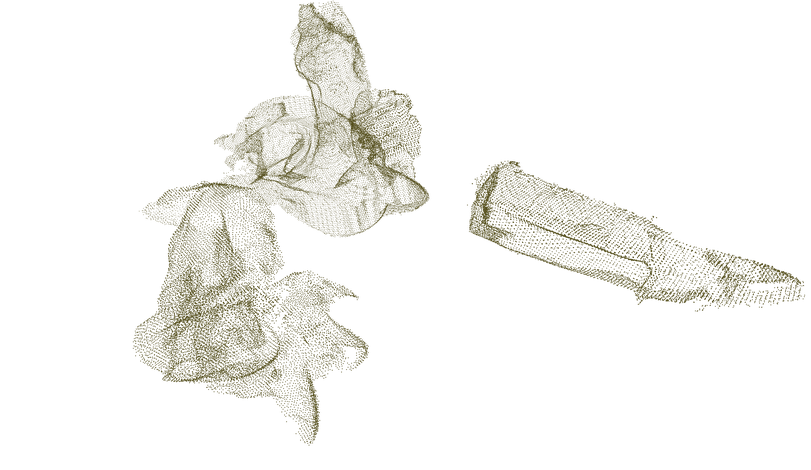} &
            \includegraphics[width=\renderwidthgeo\textwidth]{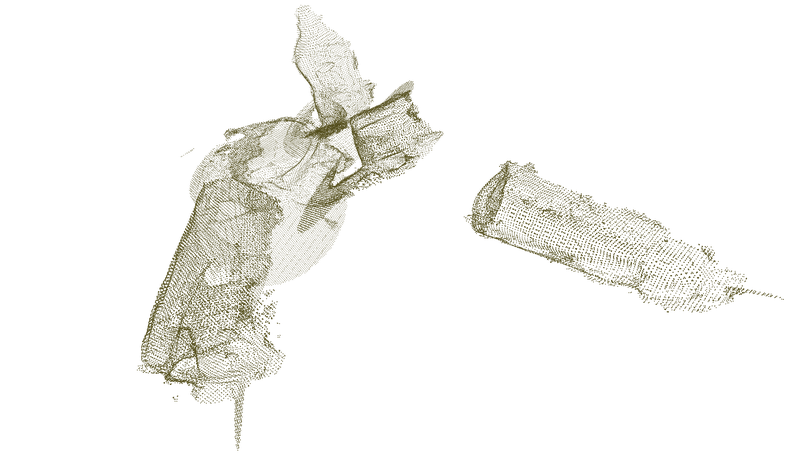} &
            \includegraphics[width=\renderwidthgeo\textwidth]{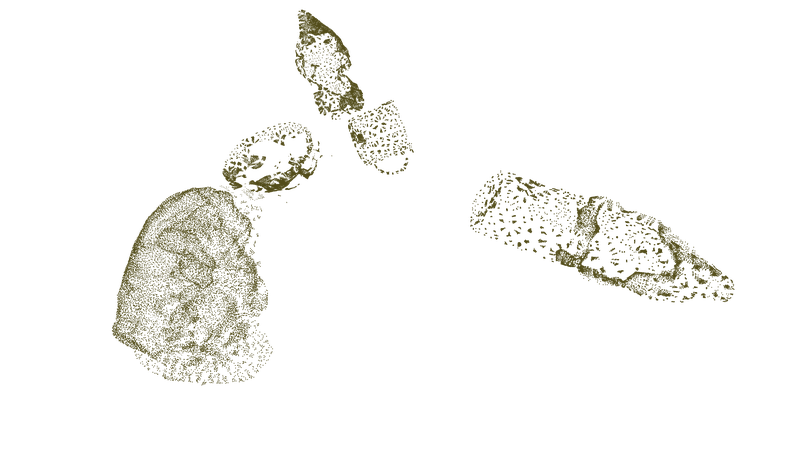} &
            \includegraphics[width=\renderwidthgeo\textwidth]{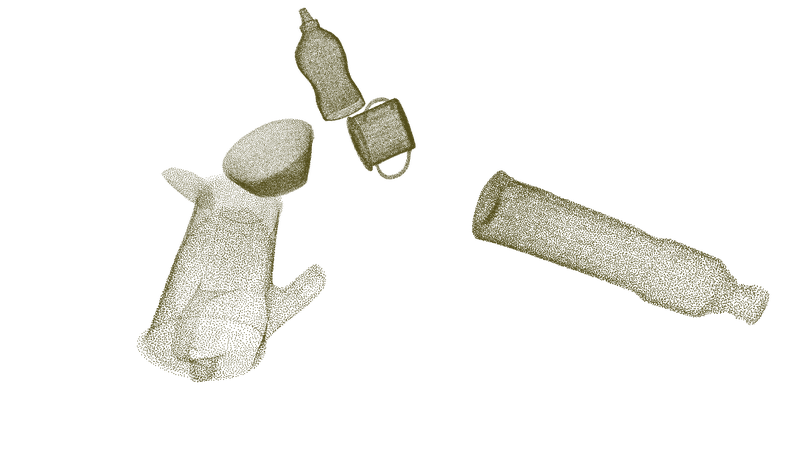} \\

            \includegraphics[width=\renderwidthgeo\textwidth]{figures/supply_geo_resized/scene_2_gt.ply00.png} &
            \includegraphics[width=\renderwidthgeo\textwidth]{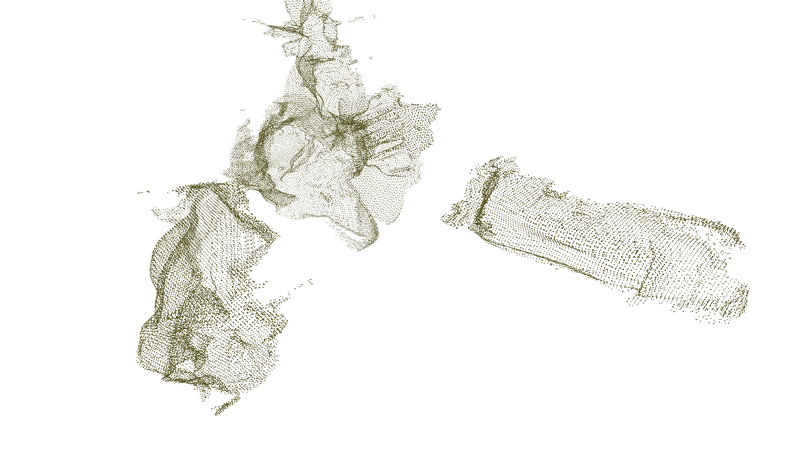} &
            \includegraphics[width=\renderwidthgeo\textwidth]{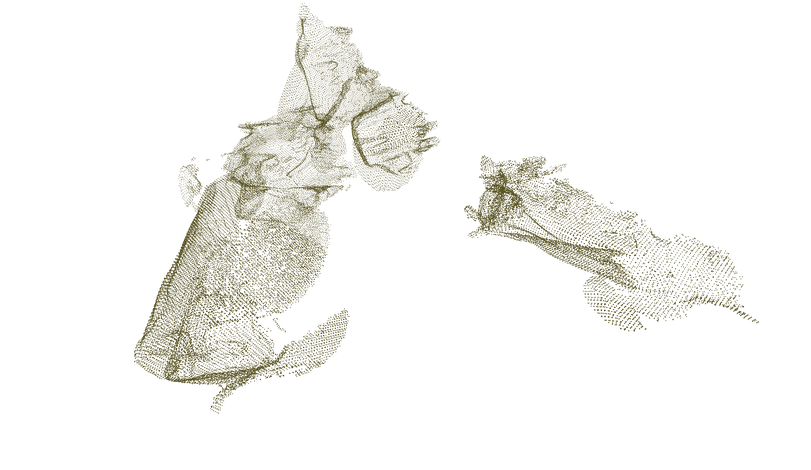} &
            \includegraphics[width=\renderwidthgeo\textwidth]{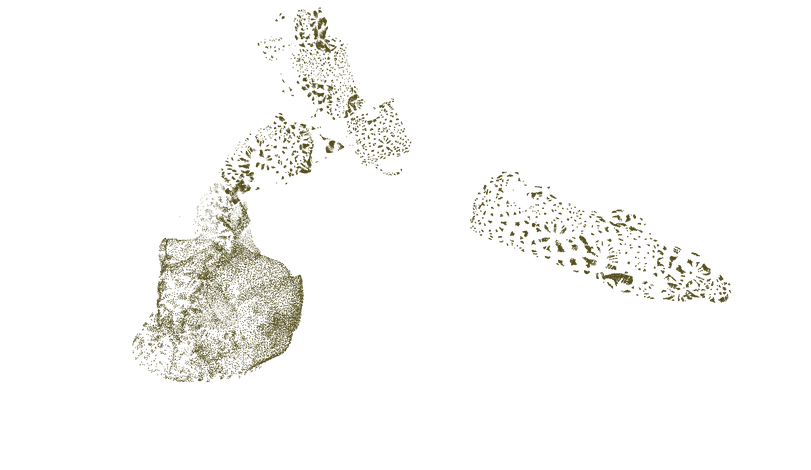} &
            \includegraphics[width=\renderwidthgeo\textwidth]{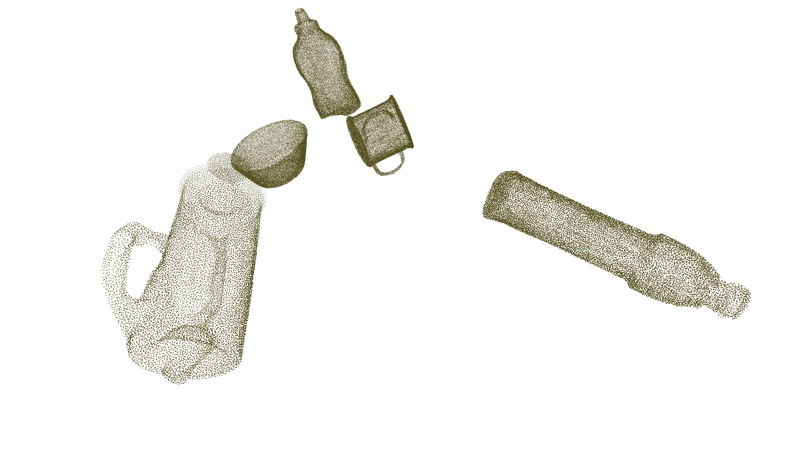} \\
            GT & DNGaussian~\cite{dngaussian} & Genfusion~\cite{genfusion} & ComPC~\cite{ComPC} & Ours \\
        \end{tabular}
    }
    \vspace{-5pt}
    \caption{Qualitative results of point cloud data for Scene 2 in our synthetic dataset.}
    \label{fig:qualitative_compare_geo_scene2}
\end{figure*}

\begin{figure*}[!htb]
    \centering
    \addtolength{\tabcolsep}{-6.5pt}
    \footnotesize{
        \setlength{\tabcolsep}{1pt}
        \begin{tabular}{@{}ccccc@{}}
            \includegraphics[width=\renderwidthgeo\textwidth]{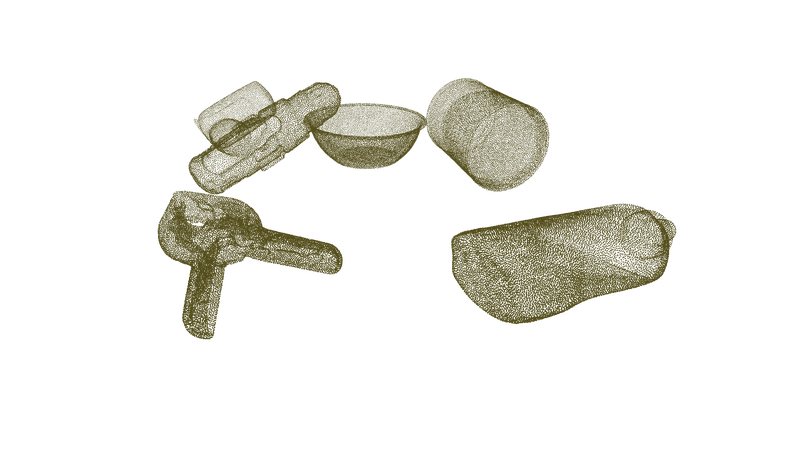} &
            \includegraphics[width=\renderwidthgeo\textwidth]{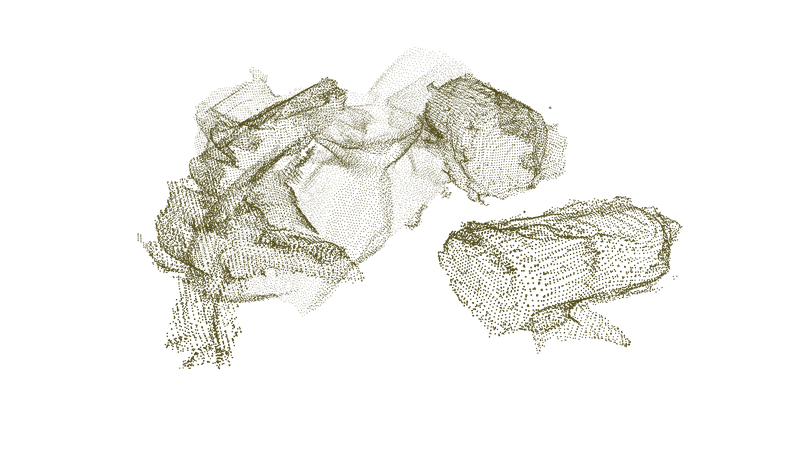} &
            \includegraphics[width=\renderwidthgeo\textwidth]{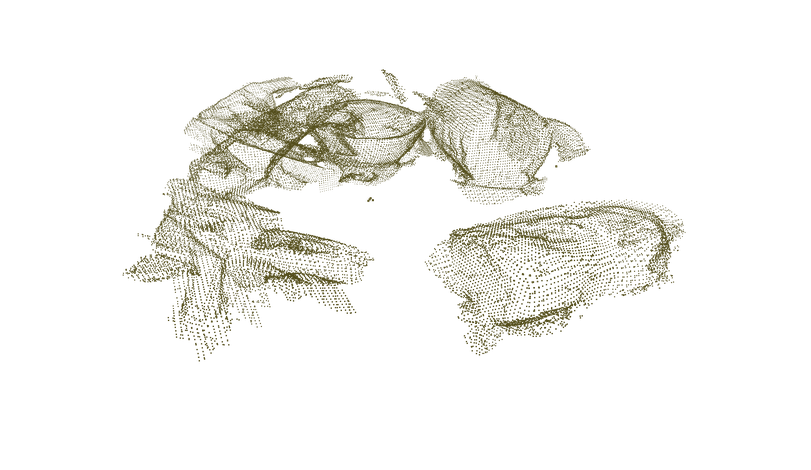} &
            \includegraphics[width=\renderwidthgeo\textwidth]{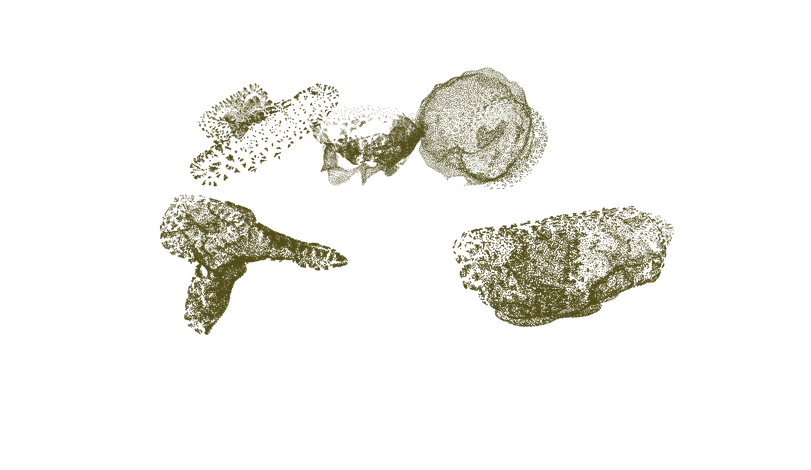} &
            \includegraphics[width=\renderwidthgeo\textwidth]{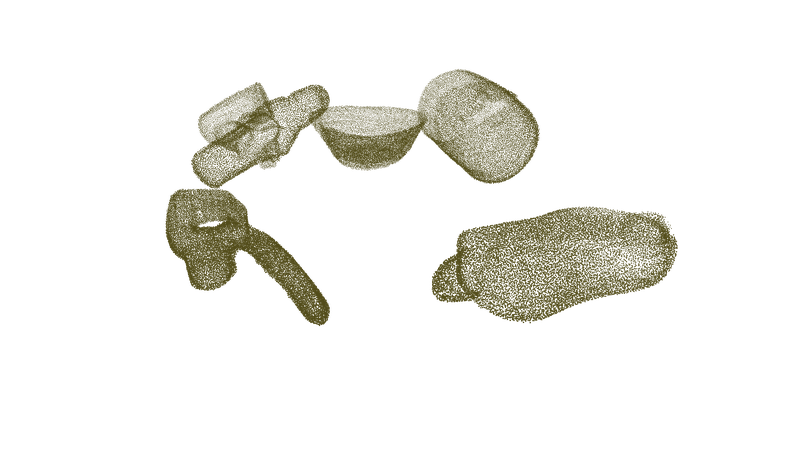} \\

            \includegraphics[width=\renderwidthgeo\textwidth]{figures/supply_geo_resized/scene_3_gt.ply00.png} &
            \includegraphics[width=\renderwidthgeo\textwidth]{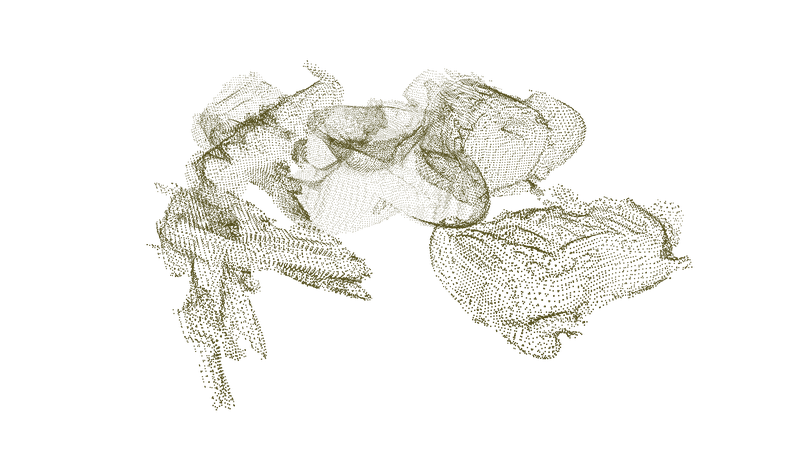} &
            \includegraphics[width=\renderwidthgeo\textwidth]{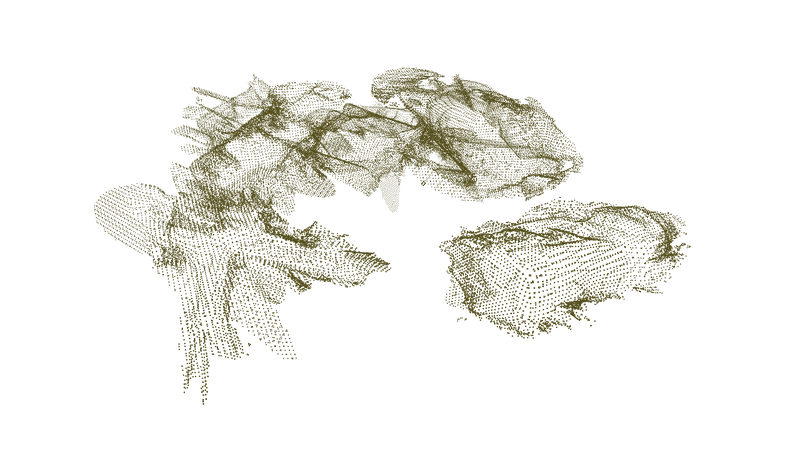} &
            \includegraphics[width=\renderwidthgeo\textwidth]{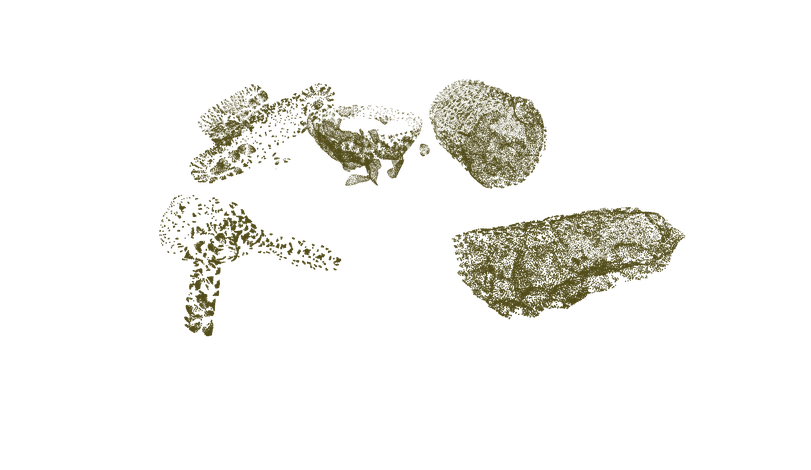} &
            \includegraphics[width=\renderwidthgeo\textwidth]{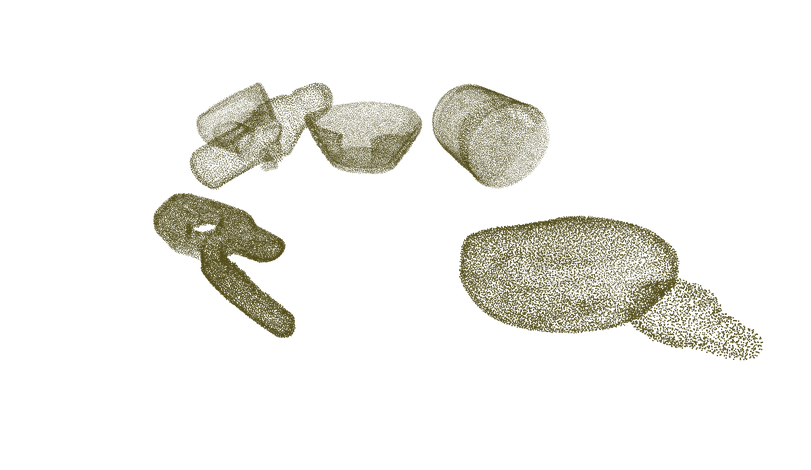} \\

            \includegraphics[width=\renderwidthgeo\textwidth]{figures/supply_geo_resized/scene_3_gt.ply00.png} &
            \includegraphics[width=\renderwidthgeo\textwidth]{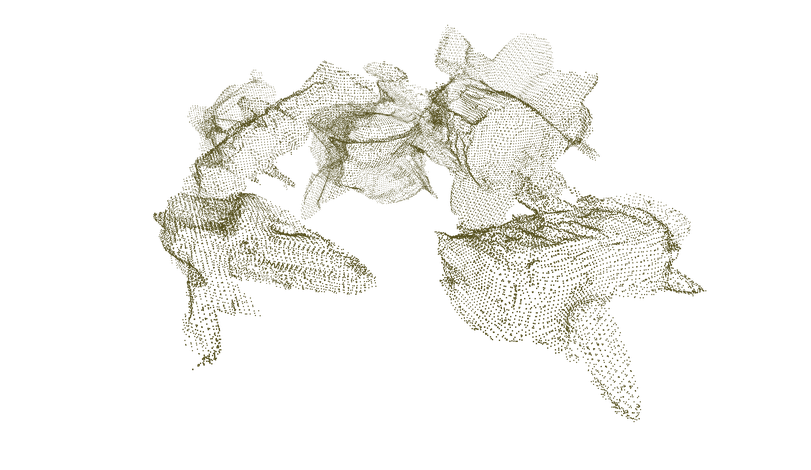} &
            \includegraphics[width=\renderwidthgeo\textwidth]{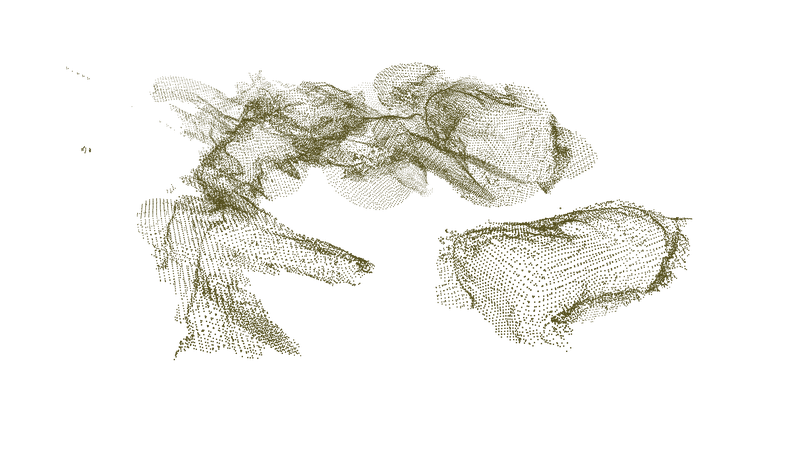} &
            \includegraphics[width=\renderwidthgeo\textwidth]{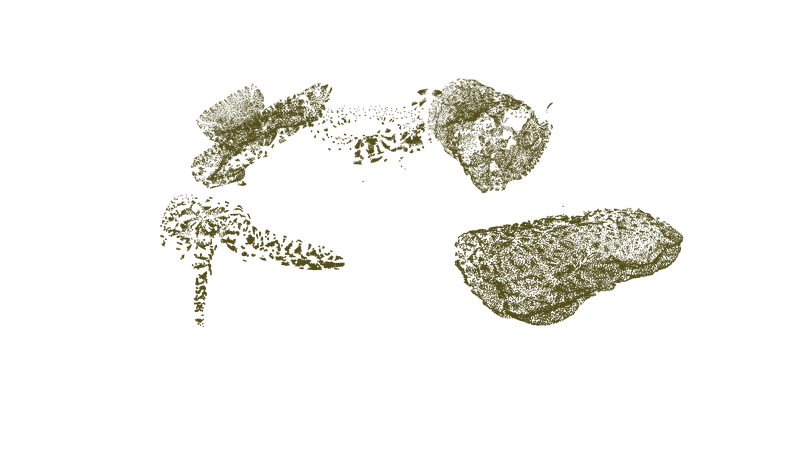} &
            \includegraphics[width=\renderwidthgeo\textwidth]{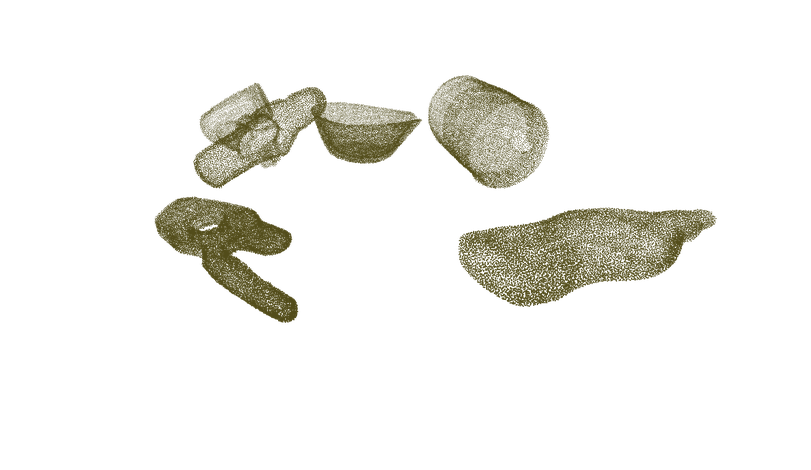} \\
            GT & DNGaussian~\cite{dngaussian} & Genfusion~\cite{genfusion} & ComPC~\cite{ComPC} & Ours \\
        \end{tabular}
    }
    \vspace{-5pt}
    \caption{Qualitative results of point cloud data for Scene 3 in our synthetic dataset.}
    \label{fig:qualitative_compare_geo_scene3}
\end{figure*}

\begin{figure*}[!htb]
    \centering
    \addtolength{\tabcolsep}{-6.5pt}
    \footnotesize{
        \setlength{\tabcolsep}{1pt}
        \begin{tabular}{@{}ccccc@{}}
            \includegraphics[width=\renderwidthgeo\textwidth]{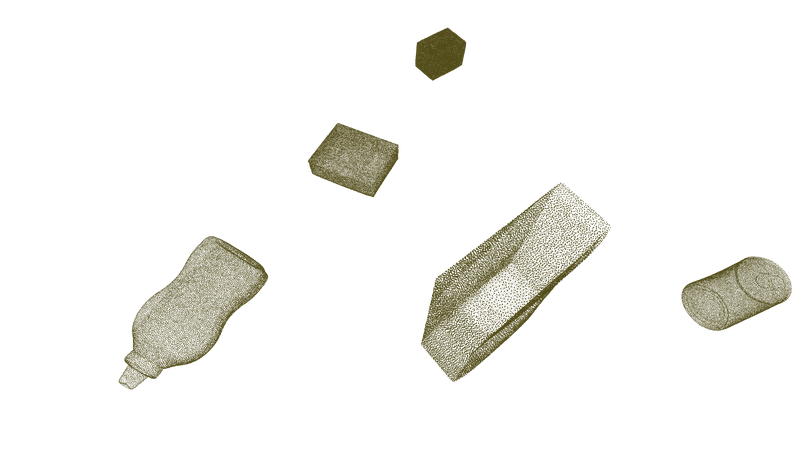} &
            \includegraphics[width=\renderwidthgeo\textwidth]{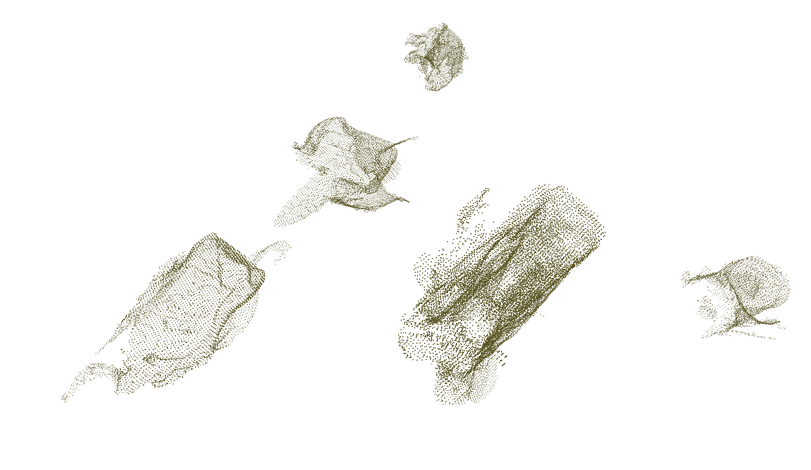} &
            \includegraphics[width=\renderwidthgeo\textwidth]{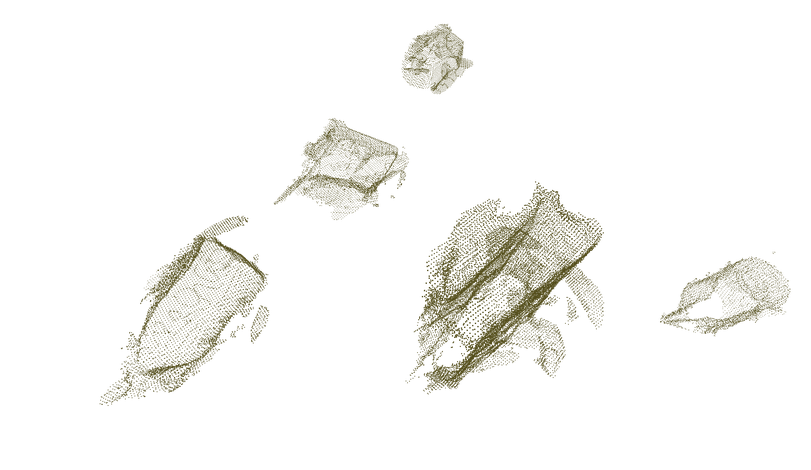} &
            \includegraphics[width=\renderwidthgeo\textwidth]{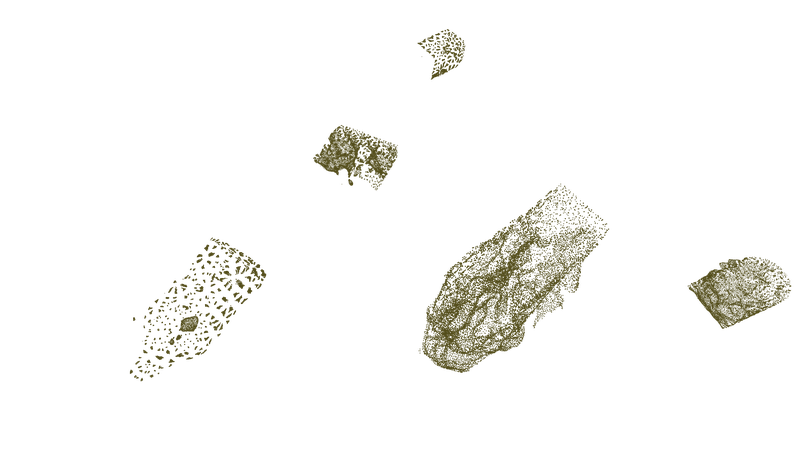} &
            \includegraphics[width=\renderwidthgeo\textwidth]{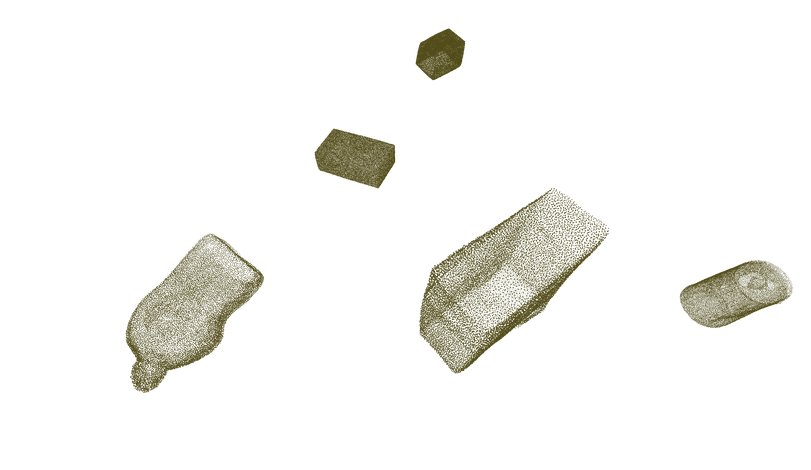} \\

            \includegraphics[width=\renderwidthgeo\textwidth]{figures/supply_geo_resized/scene_4_gt.ply00.png} &
            \includegraphics[width=\renderwidthgeo\textwidth]{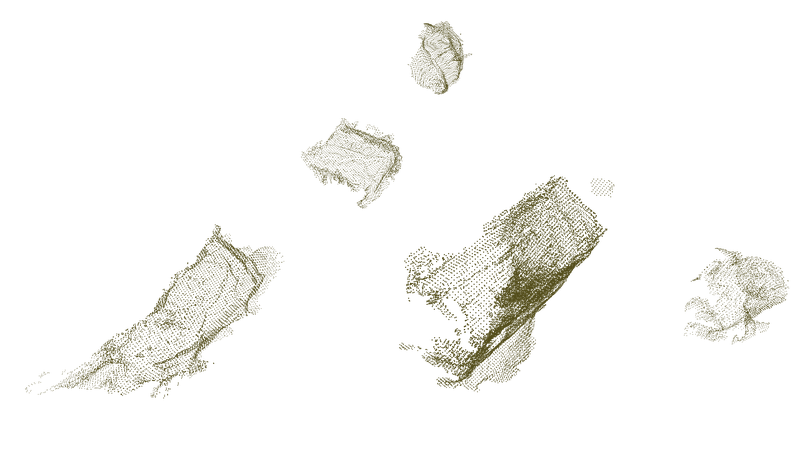} &
            \includegraphics[width=\renderwidthgeo\textwidth]{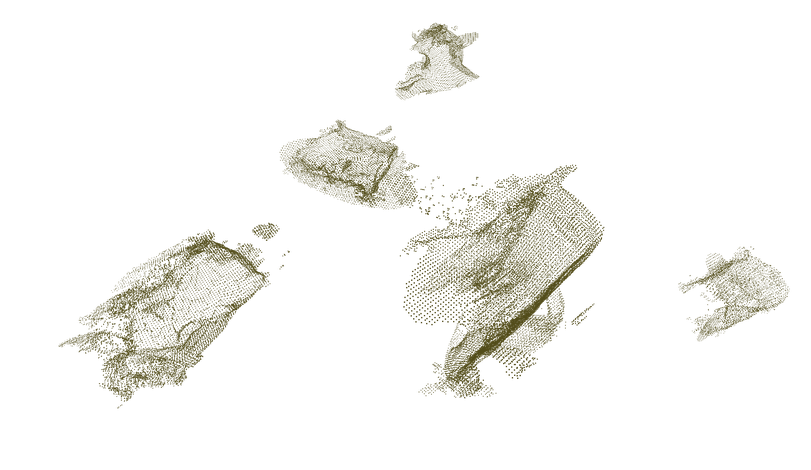} &
            \includegraphics[width=\renderwidthgeo\textwidth]{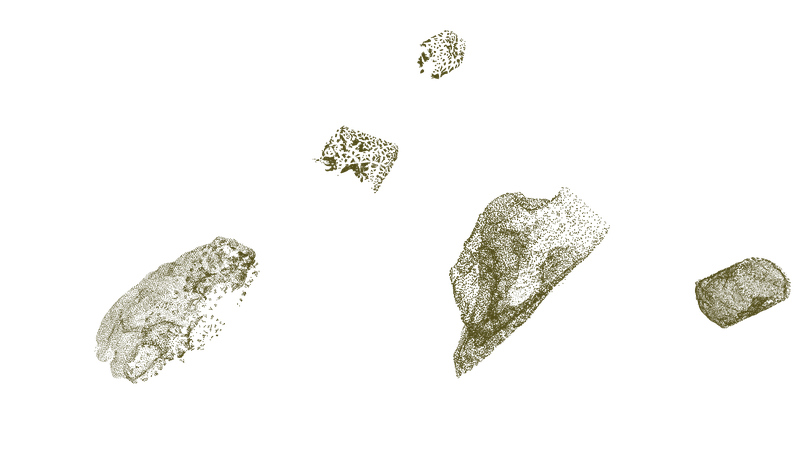} &
            \includegraphics[width=\renderwidthgeo\textwidth]{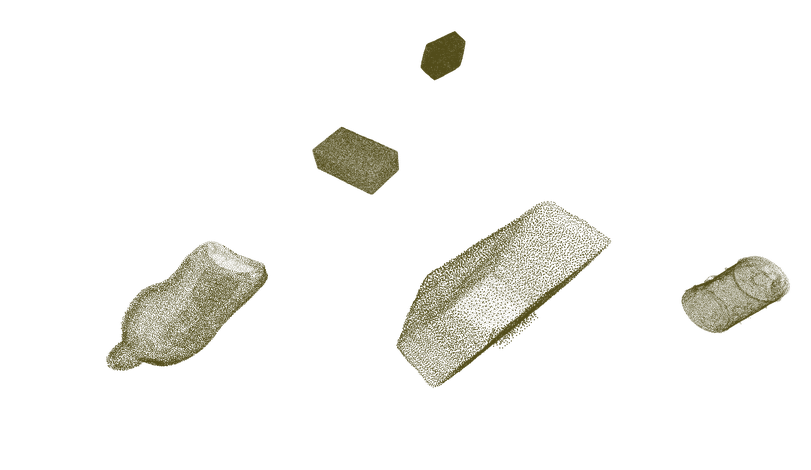} \\

            \includegraphics[width=\renderwidthgeo\textwidth]{figures/supply_geo_resized/scene_4_gt.ply00.png} &
            \includegraphics[width=\renderwidthgeo\textwidth]{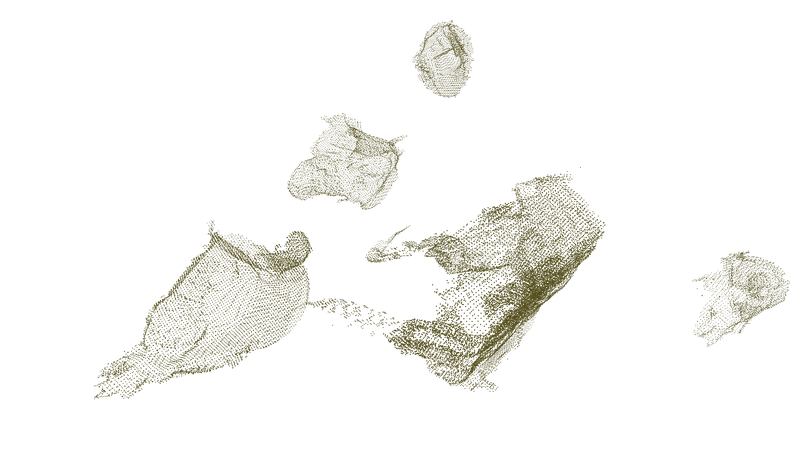} &
            \includegraphics[width=\renderwidthgeo\textwidth]{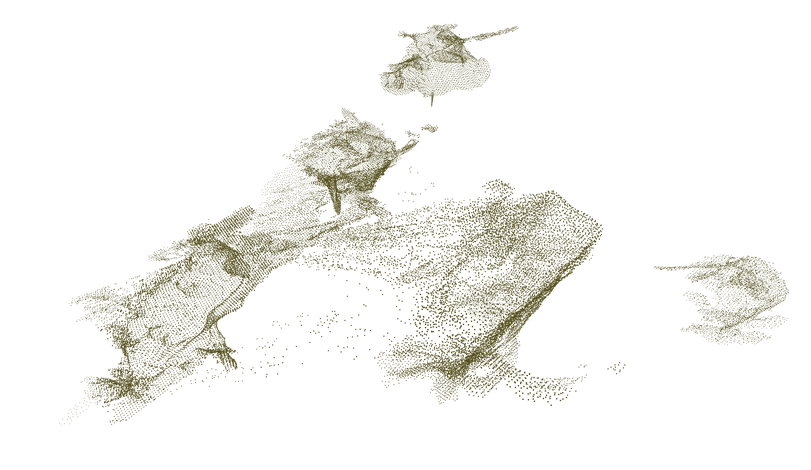} &
            \includegraphics[width=\renderwidthgeo\textwidth]{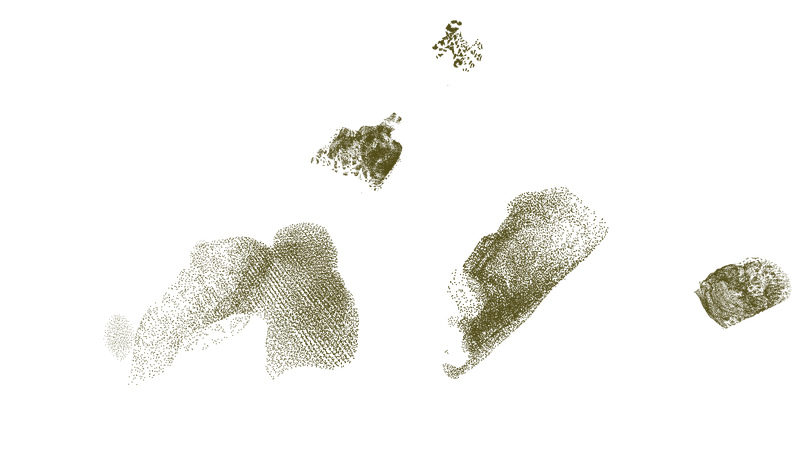} &
            \includegraphics[width=\renderwidthgeo\textwidth]{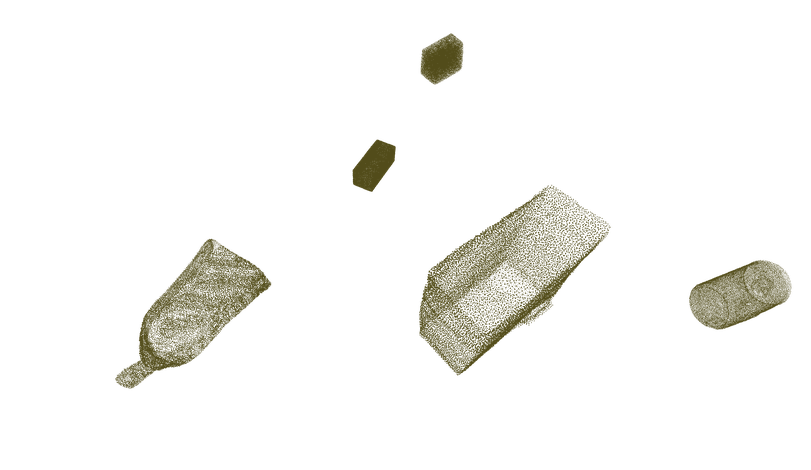} \\
            GT & DNGaussian~\cite{dngaussian} & Genfusion~\cite{genfusion} & ComPC~\cite{ComPC} & Ours \\
        \end{tabular}
    }
    \vspace{-5pt}
    \caption{Qualitative results of point cloud data for Scene 4 in our synthetic dataset.}
    \label{fig:qualitative_compare_geo_scene4}
\end{figure*}

\subsection{Qualitative Comparison Results}

Fig.~\ref{fig:qualitative_full} illustrates additional qualitative results obtained from various methods applied to the same scene under medium difficulty (\ie, with the removal of $\nicefrac{4}{5}$ of the images). Although our method does not guarantee pixel-level correspondence with the ground truth, it consistently maintains a high level of visual quality and overall coherence.
Figs.~\ref{fig:qualitative_compare_geo_scene1},~\ref{fig:qualitative_compare_geo_scene2},~\ref{fig:qualitative_compare_geo_scene3}, and~\ref{fig:qualitative_compare_geo_scene4} provide further qualitative comparisons between our approach and other baseline methods, demonstrating that our approach offers significant advantages in terms of geometric quality.

\subsection{Quantitative Ablation Results}
Tab.~\ref{tab:ablation_app} and Tab.~\ref{tab:ablation_geo} report the outcomes of the ablation experiments described in the main paper, evaluated across different scenes using appearance and geometry metrics, respectively. Besides, the failure cases associated with the method proposed by \citet{chatrasingh2023generalized} are explicitly presented.

\begin{table}[!ht]
    \caption{
        More Quantitative results of the ablation study on the geometry quality. 
        Metrics on appearance are computed over 3 randomly selected scenes from the constructed scenarios. Performance is evaluated using Chamfer Distance (CD) [cm] and Earth Mover Distance (EMD) [cm]. Results under the worst-performing ablation condition of the same scenario and difficulty level are included in the average calculation when false cases are present.
        The value 5.000 is chosen for the \colorbox{colorLow}{false case} since it is the upper bound value among the remaining values.
        Ablation (a), (e) are not included in the table since the remaining ones are sufficient to prove the effectiveness of the modules of \ours.
    }
    \label{tab:ablation_geo}
    \vspace{-1em}
    
    \setlength{\tabcolsep}{1pt}
    \renewcommand{\arraystretch}{1}
    \resizebox{\linewidth}{!}{%
    \begin{tabular}{l
              |cccc
              |cccc
              |cccc
              |cccc}
            \toprule
    
        \multirow{2}{*}{\rotatebox[origin=c]{90}{CD$\downarrow$}}
          & \multicolumn{4}{c|}{Scene 1}
          & \multicolumn{4}{c|}{Scene 2}
          & \multicolumn{4}{c|}{Scene 3}
          & \multicolumn{4}{c}{Avg.}\\
        \addlinespace[1pt]\cline{2-17}\addlinespace[1pt]
          & $\nicefrac{2}{3}$ & $\nicefrac{4}{5}$ & $\nicefrac{6}{7}$ & Avg.
          & $\nicefrac{2}{3}$ & $\nicefrac{4}{5}$ & $\nicefrac{6}{7}$ & Avg.
          & $\nicefrac{2}{3}$ & $\nicefrac{4}{5}$ & $\nicefrac{6}{7}$ & Avg.
          & $\nicefrac{2}{3}$ & $\nicefrac{4}{5}$ & $\nicefrac{6}{7}$ & Avg.\\
        \midrule
        b  & 5.000\false & 1.201\best  & 4.511       & 3.571 & 1.965      & 1.954      & 2.547      & 2.155   & 1.302\tbest & 1.362\sbest & 1.881\tbest & 1.515 \sbest & 1.733 & 1.506\sbest & 2.980 & 2.073\\
        c  & 1.869\tbest & 1.999       & 1.174\best  & 1.681\sbest & 1.597      & 1.720\tbest & 2.305\sbest & 1.874\sbest   & 1.318      & 1.863      & 1.425\sbest & 1.535 \tbest & 1.594\tbest & 1.860 & 1.634\sbest & 1.696\sbest \\
        d  & 1.819\sbest & 1.284\sbest & 1.903\tbest & 1.669\best & 1.576\tbest & 1.956      & 2.543\tbest & 2.025   & 1.422      & 1.362\tbest & 1.881      & 1.555 & 1.605 & 1.534\tbest & 2.109\tbest & 1.749\tbest \\
        f  & 1.558\best  & 2.940       & 4.520       & 3.006 & 1.186\best & 1.234\sbest & 3.306       & 1.909\tbest   & 1.250\sbest & 3.304      & 2.778      & 2.444 & 1.331\best &	2.492 &	3.534 & 2.452 \\
        \textbf{g} & 1.932       & 1.802\tbest & 1.684\sbest & 1.806\tbest & 1.192\sbest & 1.162\best & 1.684\best & 1.346\best & 1.176\best & 1.269\best & 1.355\best & 1.267 \best & 1.433\sbest & 1.411\best & 1.574\best & 1.472\best \\
              \midrule
              \multirow{2}{*}{\rotatebox[origin=c]{90}{\small{EMD$\downarrow$}}}
                & \multicolumn{4}{c|}{Scene 1}
                & \multicolumn{4}{c|}{Scene 2}
                & \multicolumn{4}{c|}{Scene 3}
                & \multicolumn{4}{c}{Avg.}\\
              \addlinespace[1pt]\cline{2-17}\addlinespace[1pt]
                & $\nicefrac{2}{3}$ & $\nicefrac{4}{5}$ & $\nicefrac{6}{7}$ & Avg.
                & $\nicefrac{2}{3}$ & $\nicefrac{4}{5}$ & $\nicefrac{6}{7}$ & Avg.
                & $\nicefrac{2}{3}$ & $\nicefrac{4}{5}$ & $\nicefrac{6}{7}$ & Avg.
                & $\nicefrac{2}{3}$ & $\nicefrac{4}{5}$ & $\nicefrac{6}{7}$ & Avg.\\
              \midrule
              b & 5.000\false & 1.406\best & 4.442 & 3.616 & 2.223 & 2.581 & 3.224\tbest & 2.676 & 1.337 & 1.825\tbest & 2.515 & 1.892 \tbest & 2.052 & 1.937\sbest & 3.393 & 2.461 \\
              c & 2.595 & 1.932 & 1.717\best & 2.081\tbest & 2.070 & 1.954\tbest & 3.271 & 2.432\tbest & 1.296\best & 2.033 & 1.623\best & 1.651 \sbest & 1.987\tbest	& 1.973	& 2.203\sbest & 2.054\sbest\\
              d & 2.474\tbest & 1.485\sbest & 2.057\sbest & 2.005\sbest & 1.878\sbest & 2.585 & 3.219\sbest & 2.561 & 1.664 & 1.822\sbest & 2.509\tbest & 1.998 & 2.005 & 1.964\tbest	& 2.595\tbest & 2.188\tbest\\
              f & 1.791\best & 2.933 & 4.462 & 3.062 & 1.614\best & 1.599\best & 3.812 & 2.342 \sbest & 1.307\tbest & 3.518 & 3.104 & 2.643 & 1.570\best & 2.683 & 3.792 & 2.682\\
  \textbf{g} & 1.840\sbest & 1.902\tbest & 2.021\sbest & 1.921\best & 1.880\tbest & 1.729\sbest & 1.803\best & 1.804 \best & 1.299\sbest & 1.498\best & 1.671\sbest & 1.489 \best & 1.673\sbest	& 1.709\best & 1.831\best & 1.738\best \\
              \bottomrule
            \end{tabular}}
    \vspace{-1em}
    \end{table}

\subsection{Qualitative Ablation Results}

We conducted additional ablation experiments to evaluate the effectiveness of various components in our Shape Refinement solution. These components include (i) initial alignment using only ICP, (ii) w/o the rotation term $R'$, (iii) w/o the regularization terms $\mathcal{L}_R$ and $\mathcal{L}_S$, (iv) w/o $R'$, $\mathcal{L}_R$, and $\mathcal{L}_S$, and (v) the full model. Fig.~\ref{fig:ablation_qualitative_supply} presents the qualitative results of these ablation experiments. Configuration (i) demonstrates that the proxy object maintains its shape following generation, whereas (ii) and (iii) result in the shape distortion after alignment. In (iv), additional deformation and unrealistic object dimensions are observed. These results indicate that the full model consistently produces correctly proportioned objects and achieves distortion-free alignment.
\newcommand{\figwidthb}{0.191}
\begin{figure*}[!htb]
    \centering
    \addtolength{\tabcolsep}{-6.5pt}
    \footnotesize{
        \setlength{\tabcolsep}{1pt} 
        \begin{tabular}{ccccc}

            \includegraphics[width=\figwidthb\textwidth]{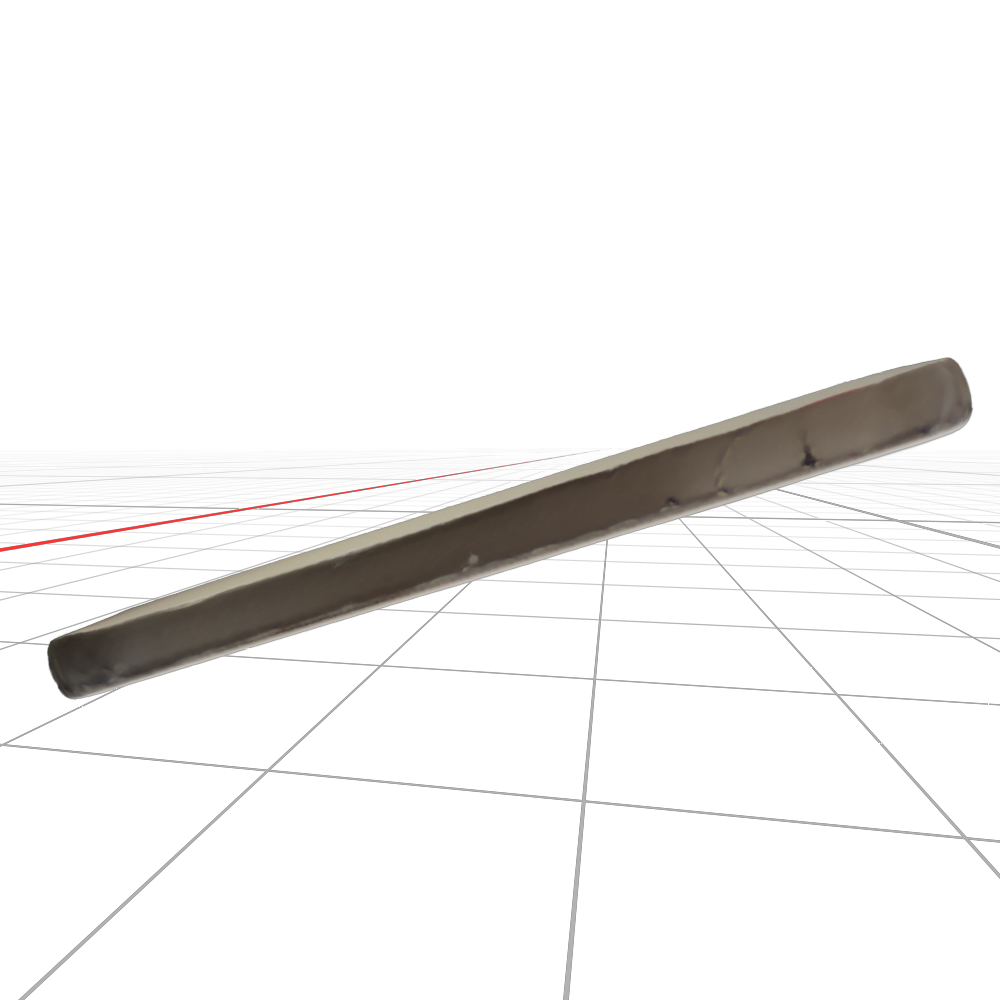} &
            \includegraphics[width=\figwidthb\textwidth]{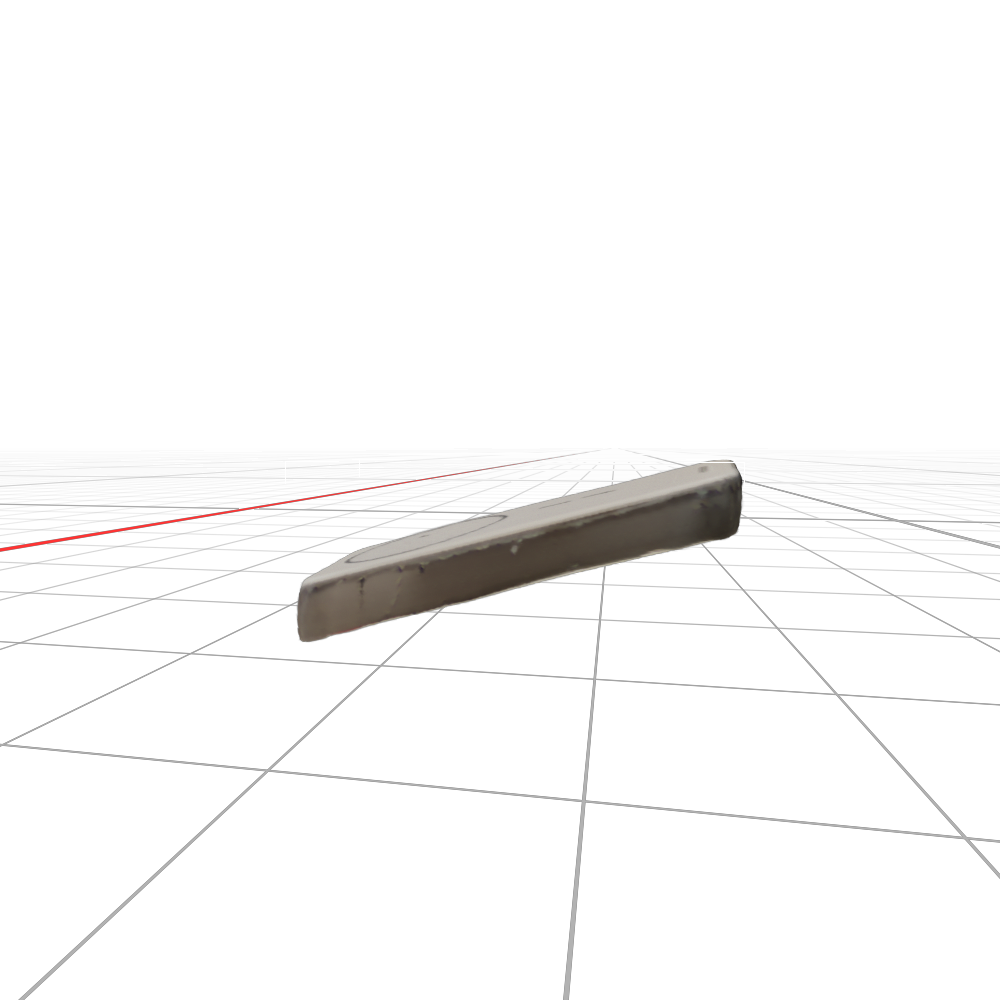} &
            \includegraphics[width=\figwidthb\textwidth]{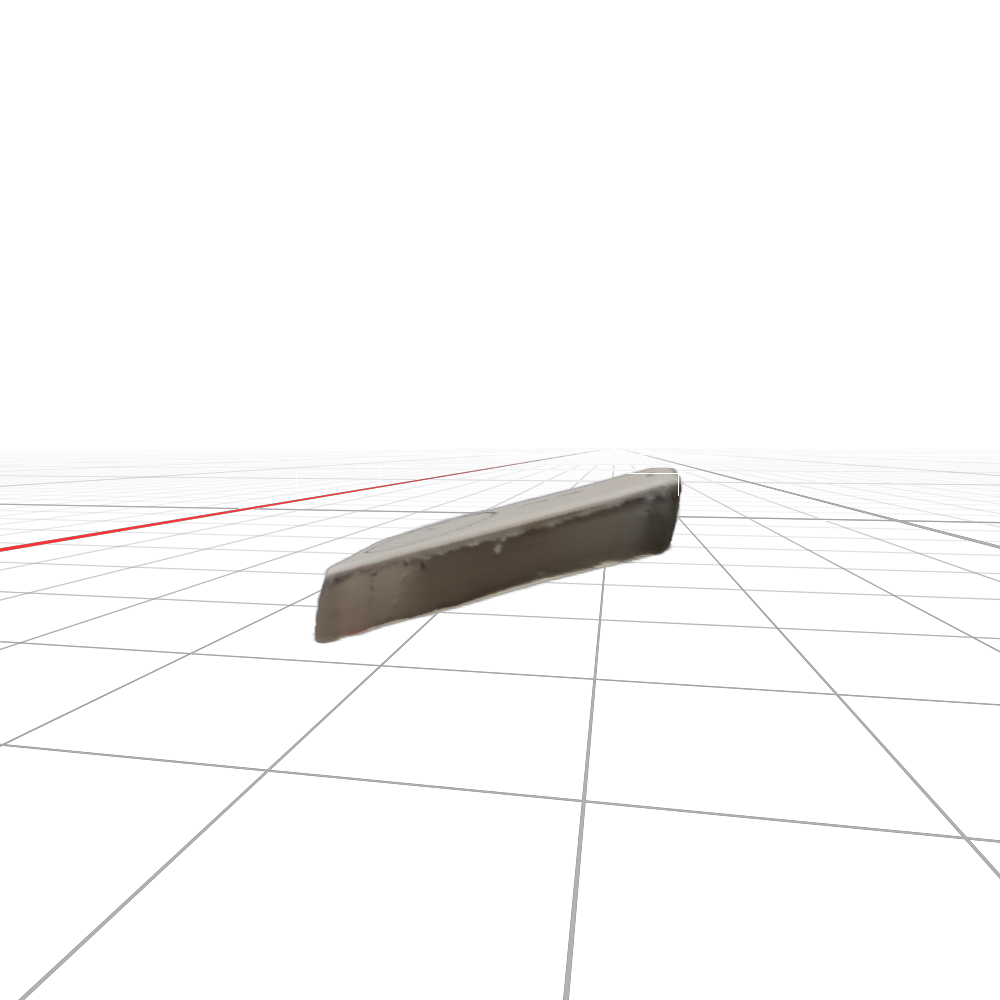} &
            \includegraphics[width=\figwidthb\textwidth]{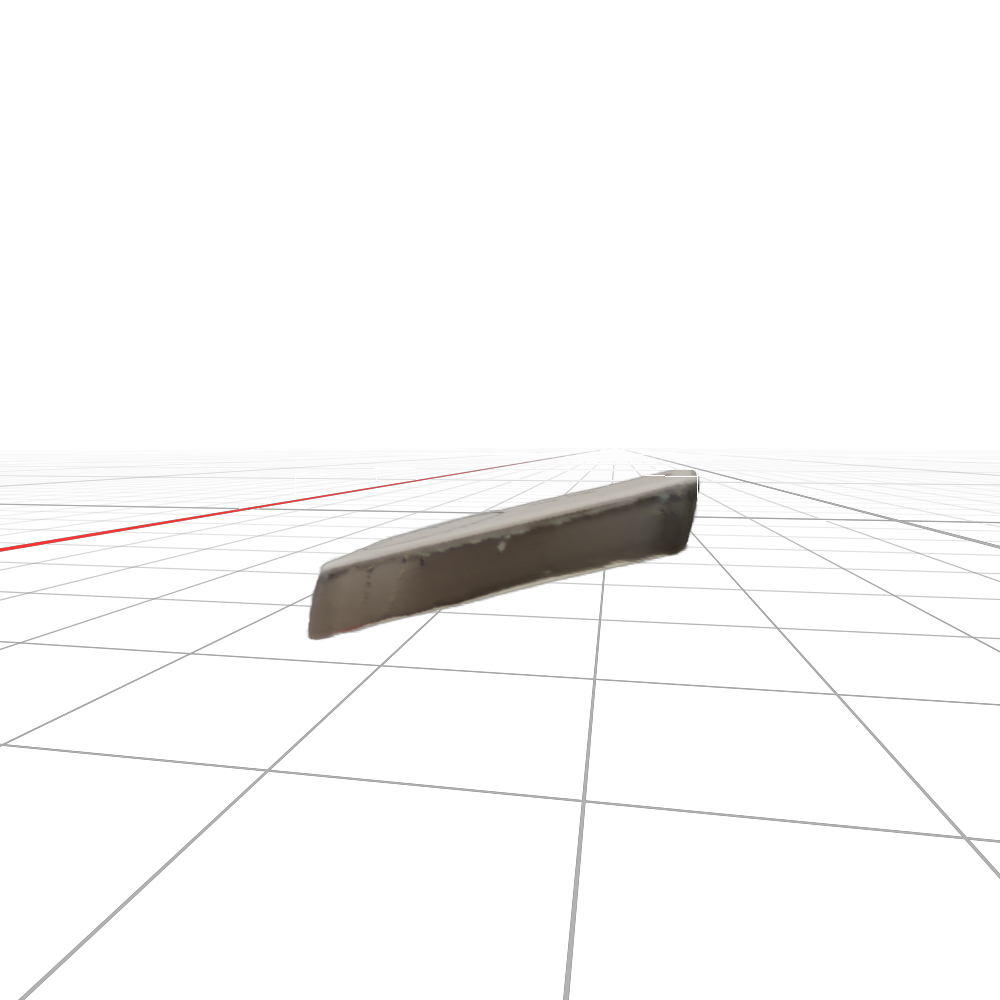} &
            \includegraphics[width=\figwidthb\textwidth]{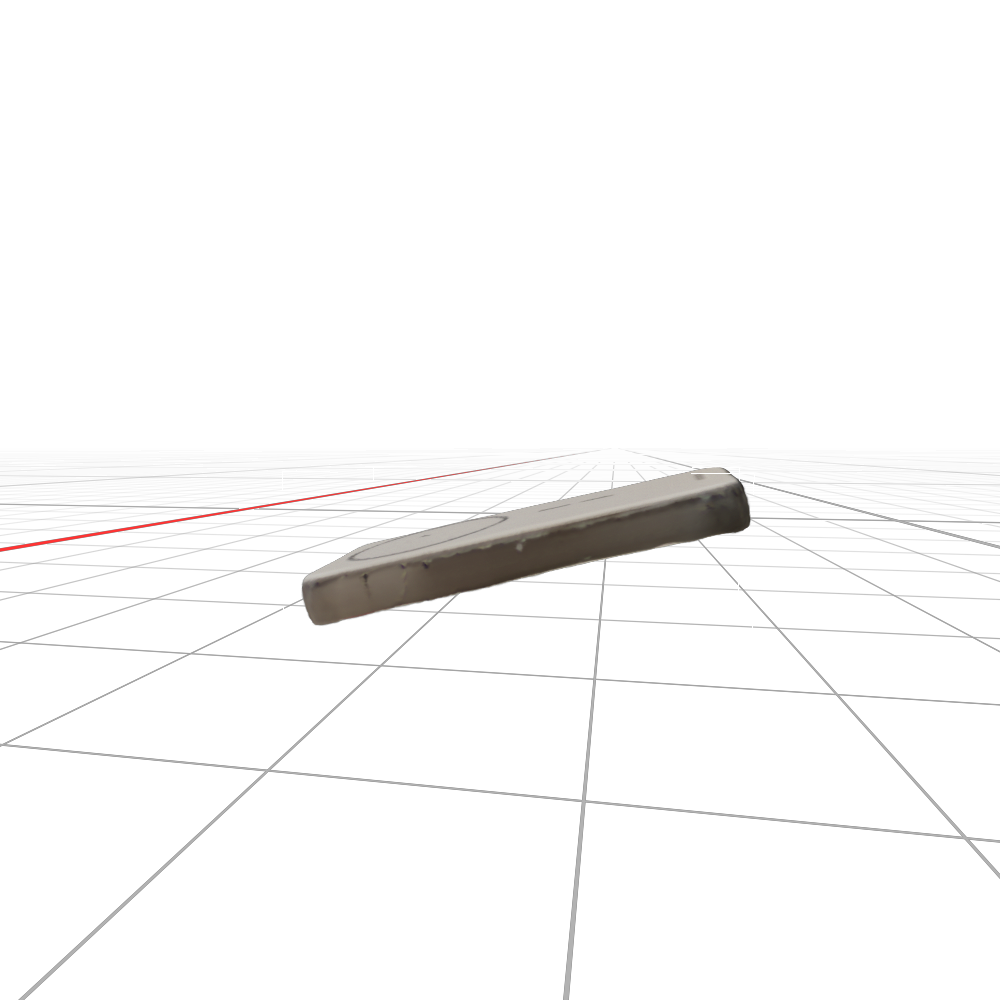} \\

            \includegraphics[width=\figwidthb\textwidth]{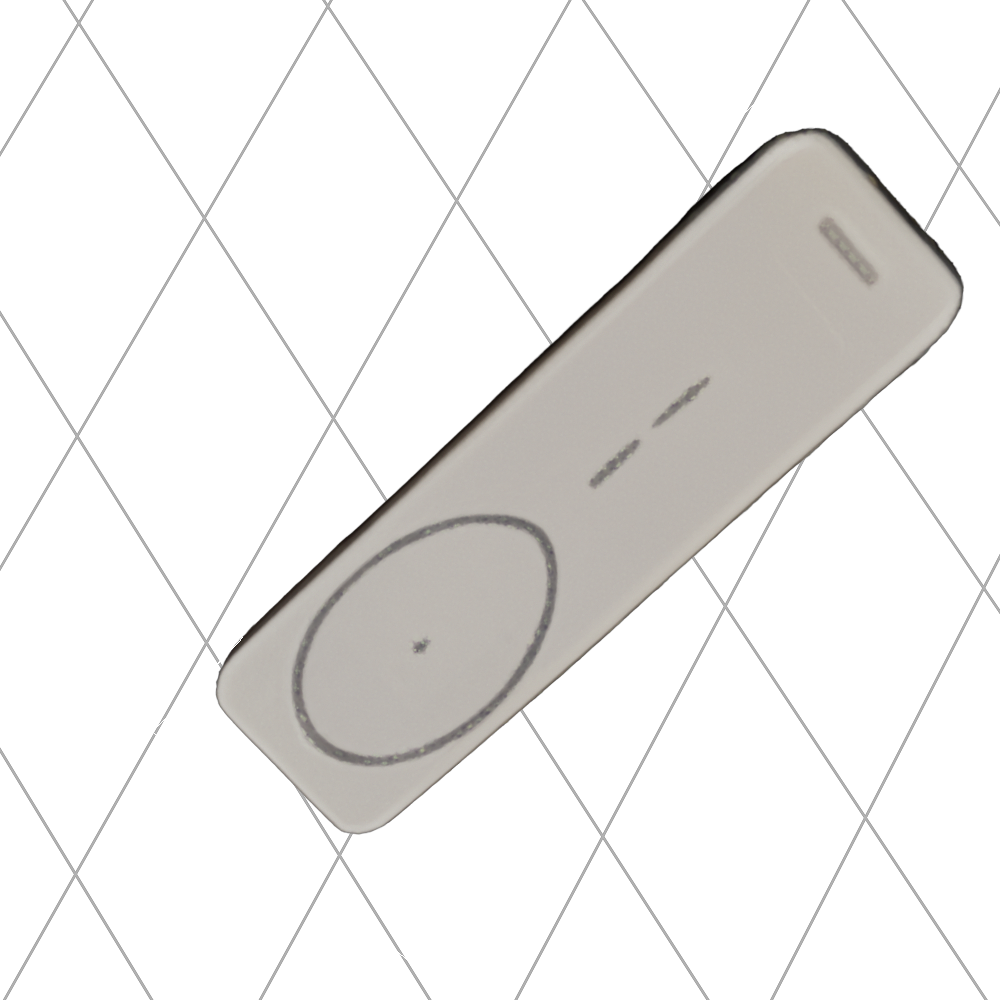} &
            \includegraphics[width=\figwidthb\textwidth]{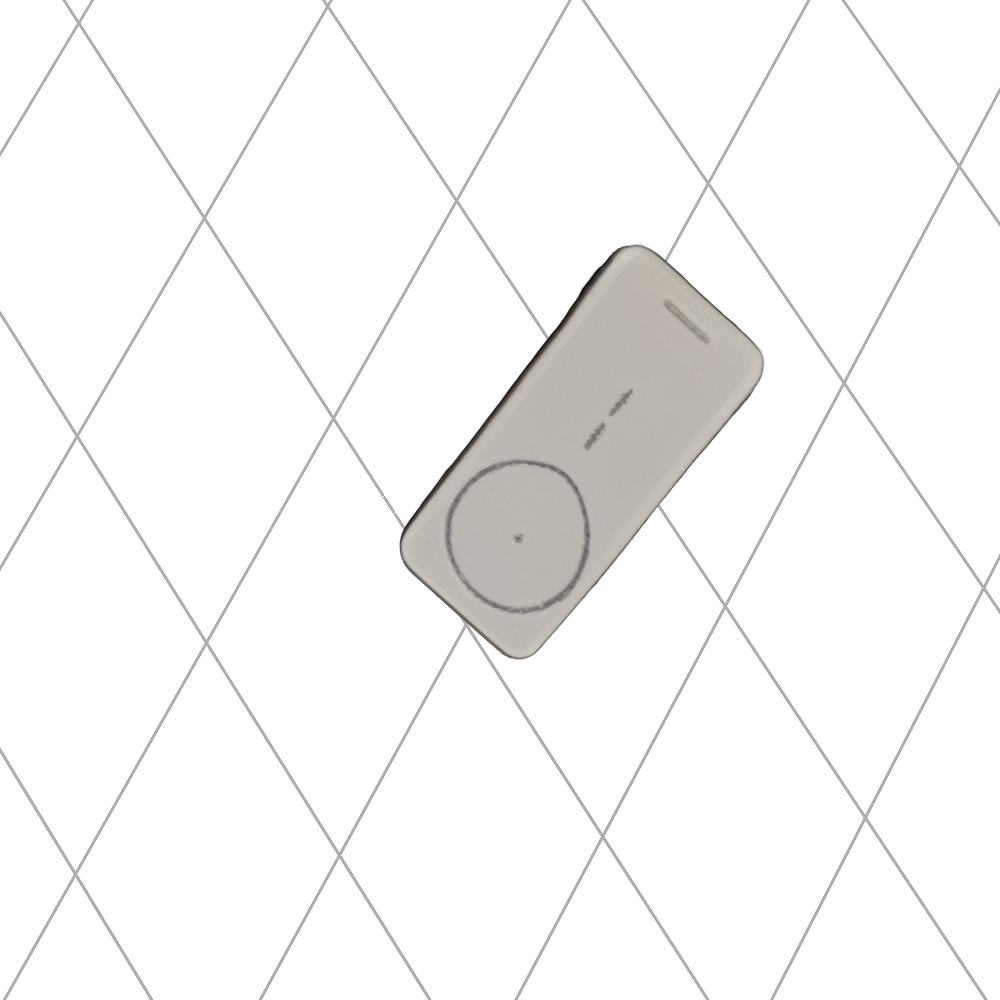} &
            \includegraphics[width=\figwidthb\textwidth]{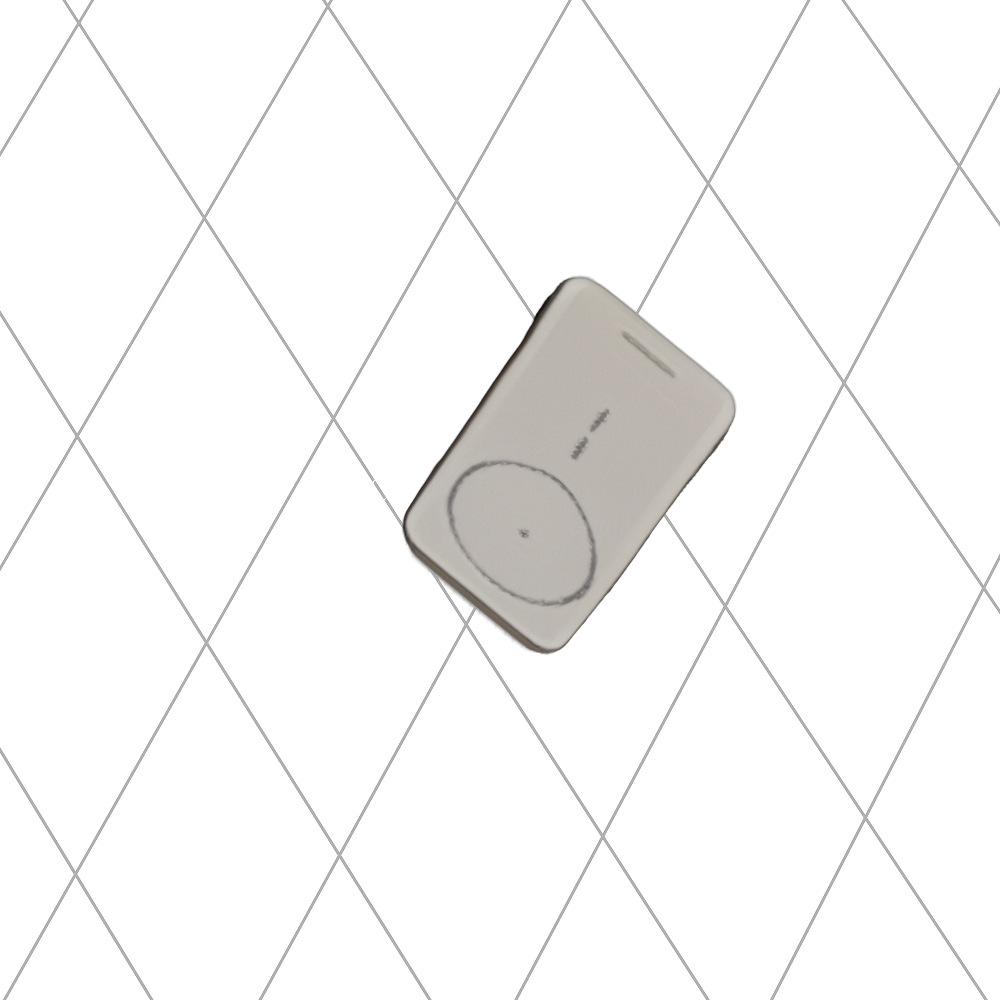} &
            \includegraphics[width=\figwidthb\textwidth]{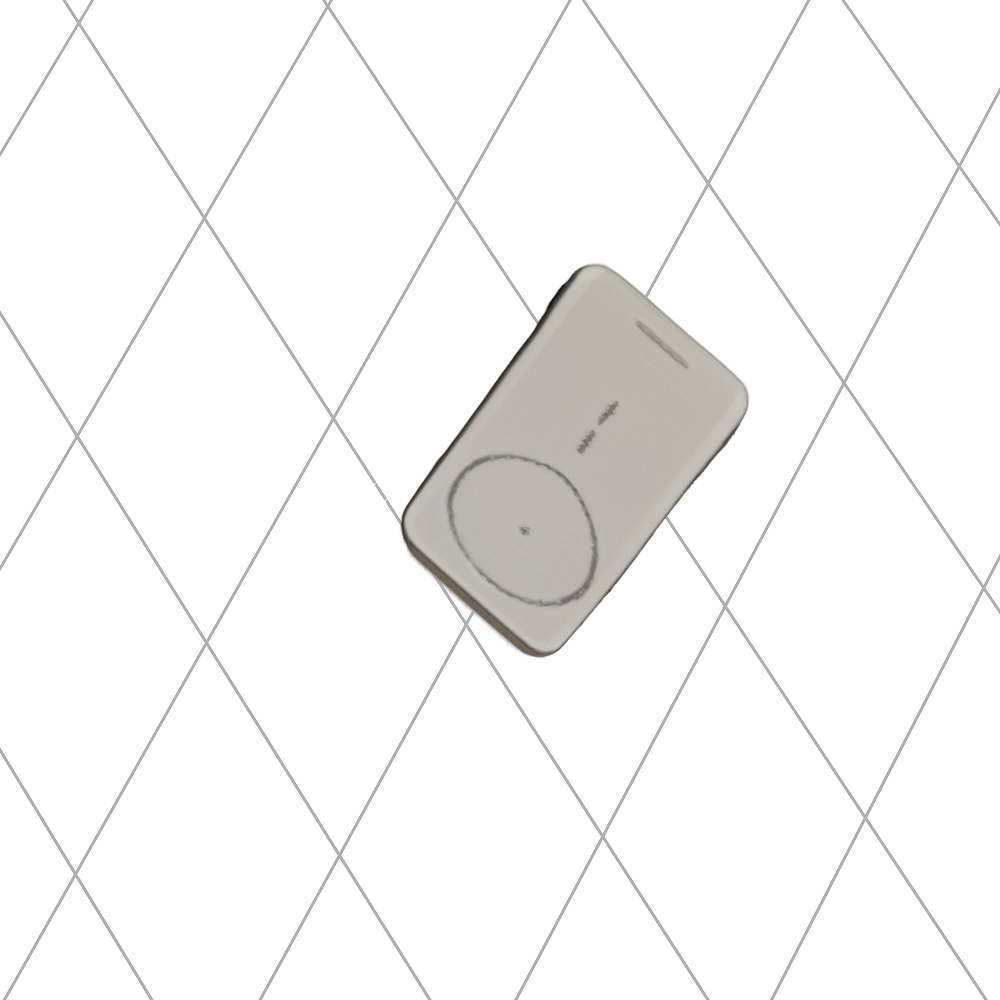} &
            \includegraphics[width=\figwidthb\textwidth]{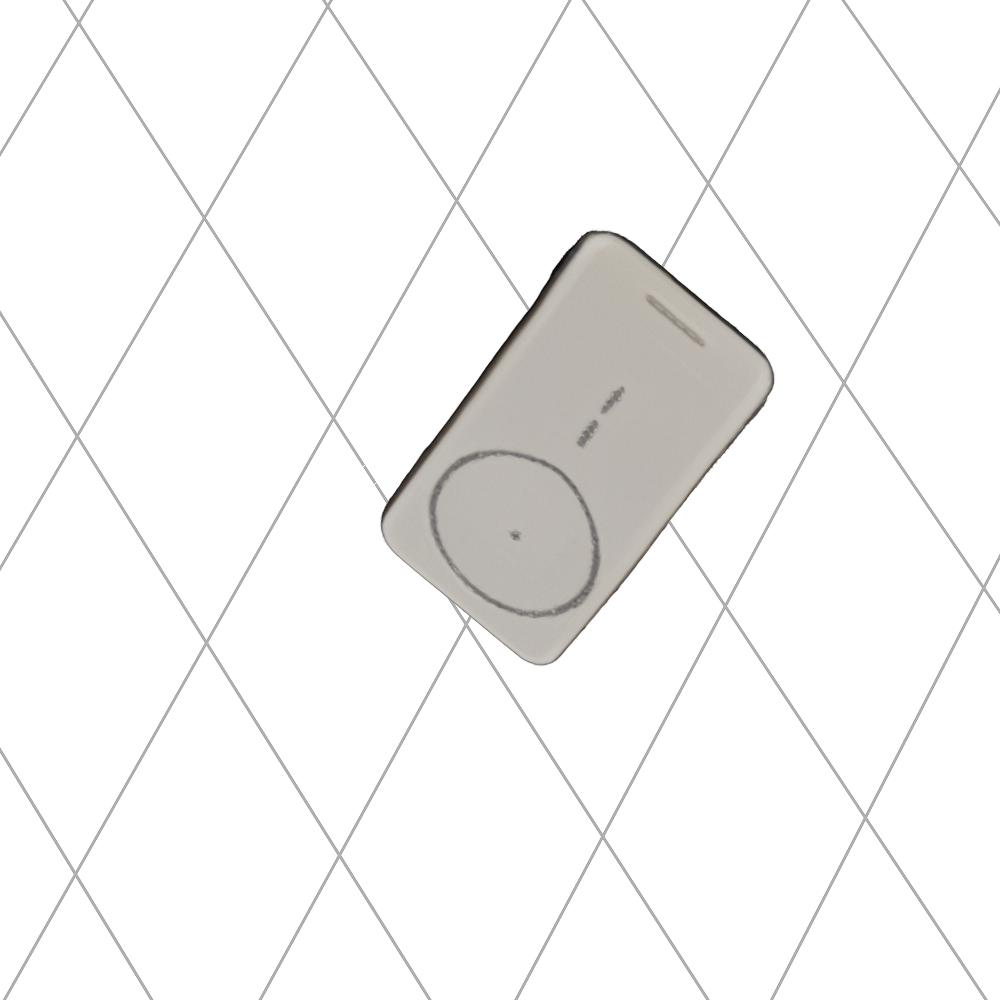} \\

            \includegraphics[width=\figwidthb\textwidth]{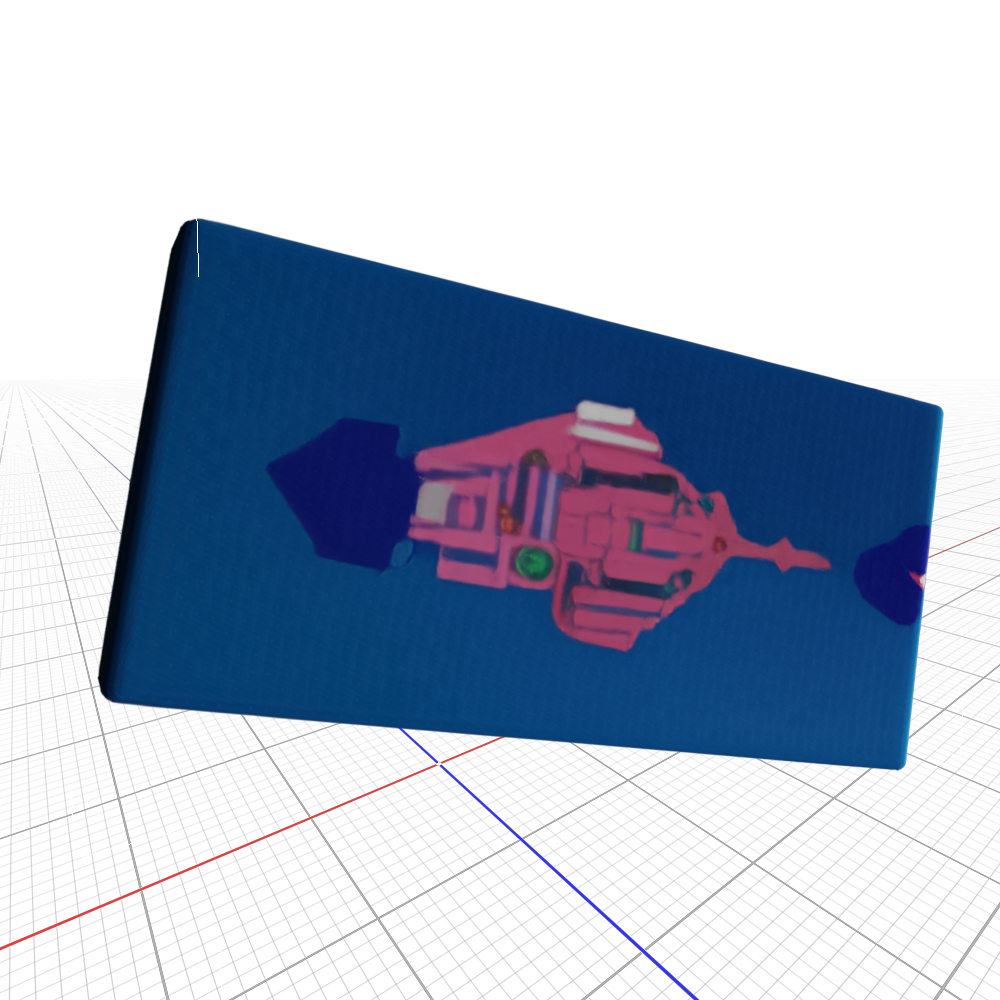} &
            \includegraphics[width=\figwidthb\textwidth]{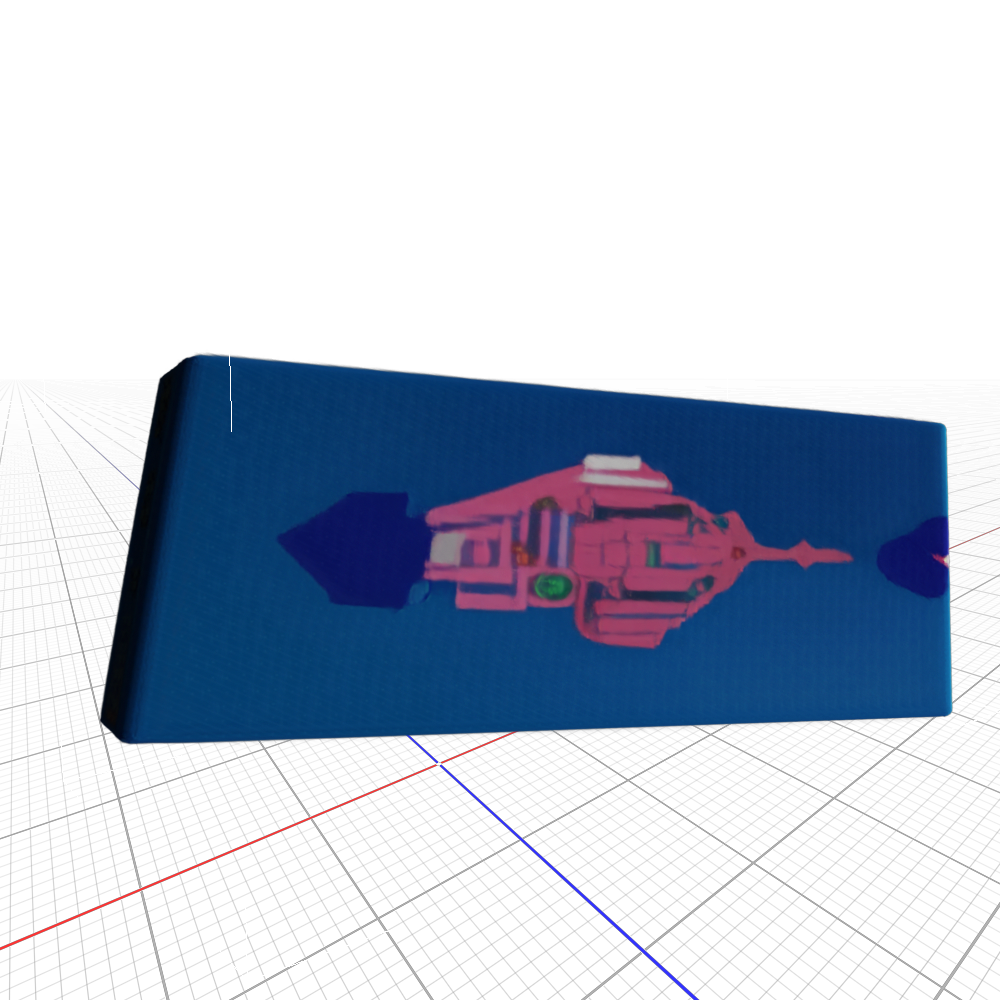} &
            \includegraphics[width=\figwidthb\textwidth]{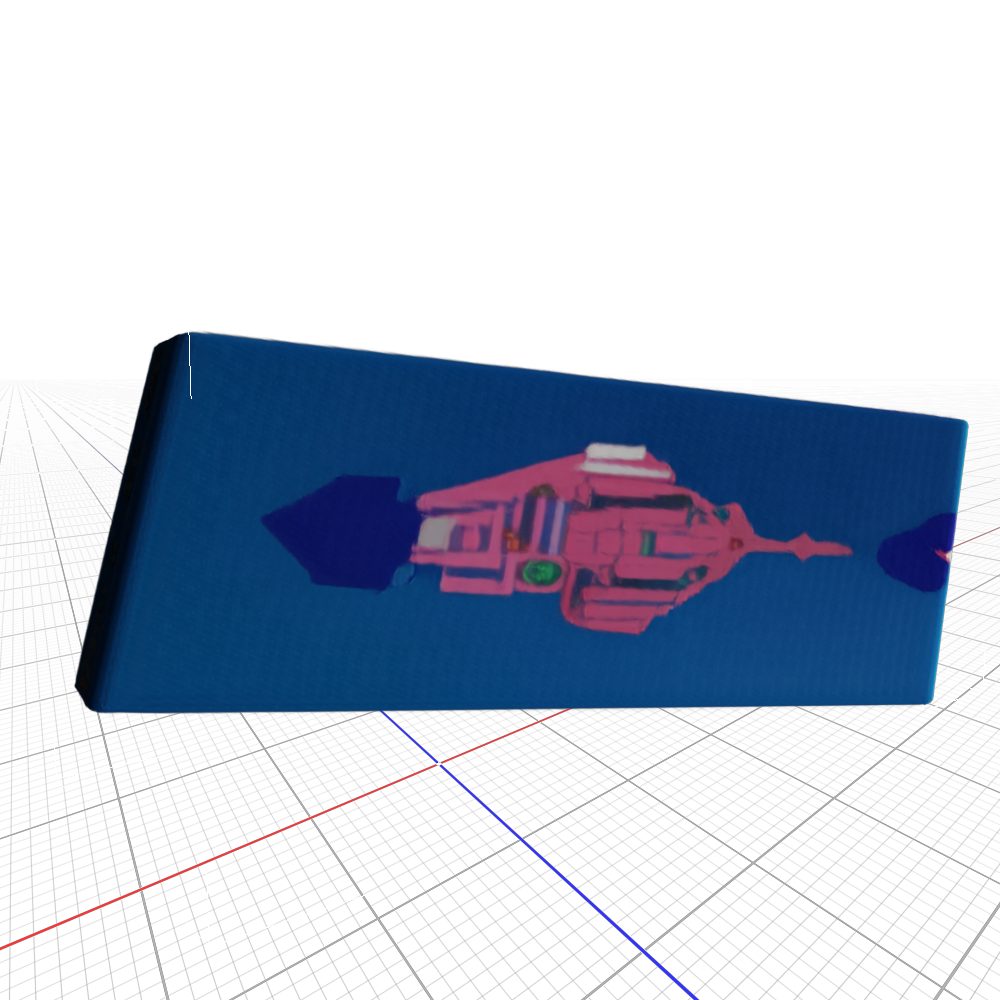} &
            \includegraphics[width=\figwidthb\textwidth]{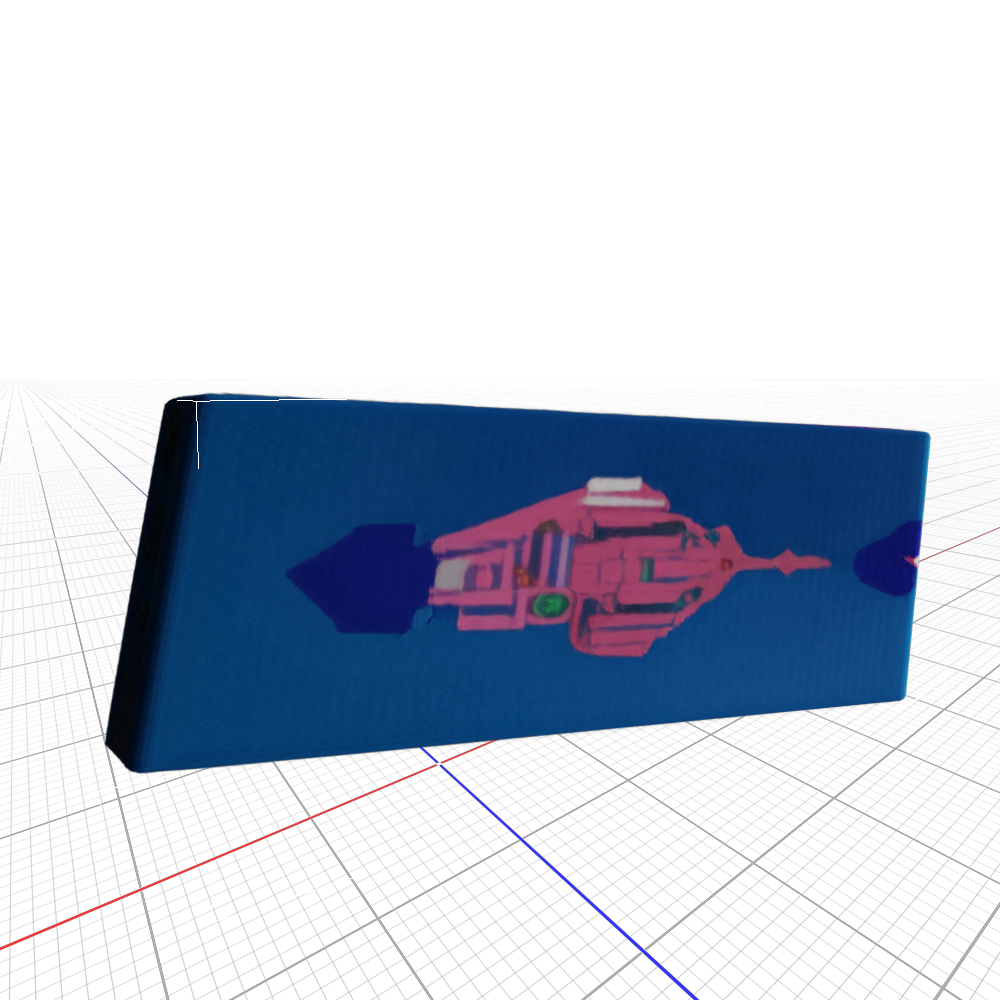} &
            \includegraphics[width=\figwidthb\textwidth]{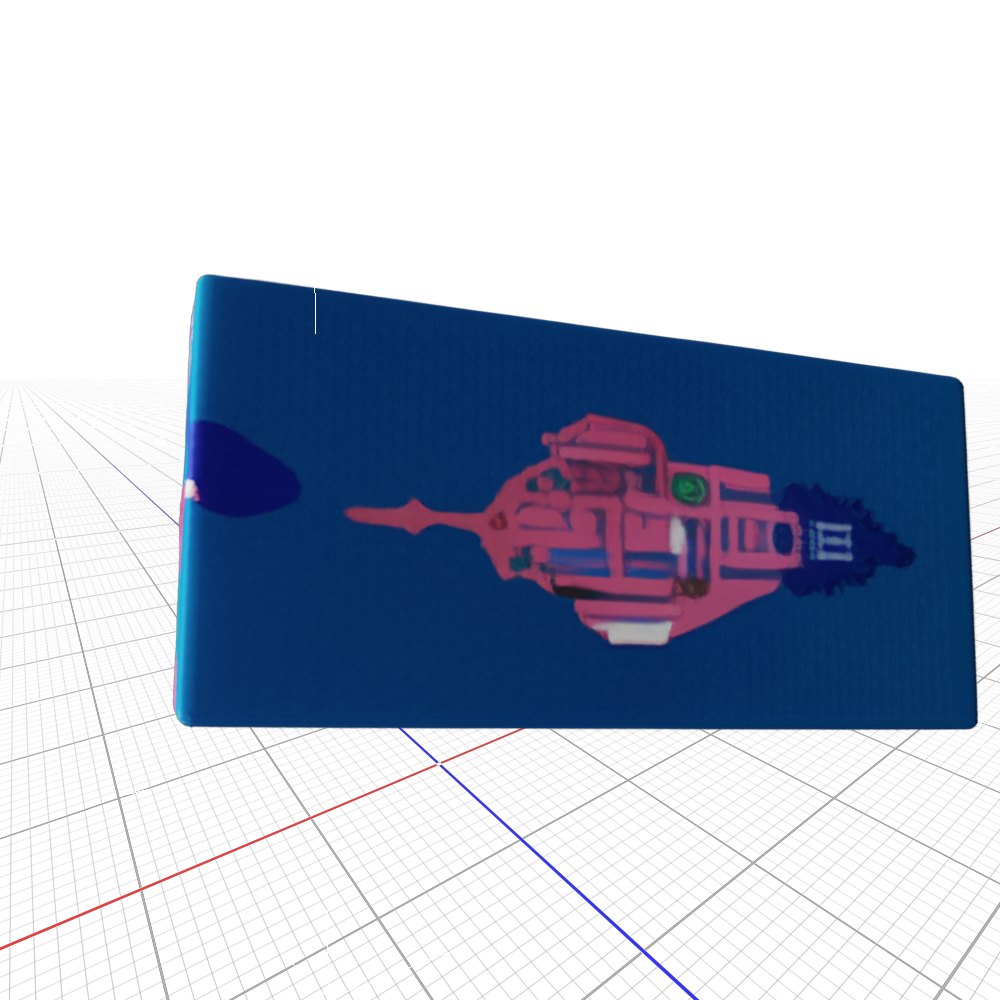} \\

            \includegraphics[width=\figwidthb\textwidth]{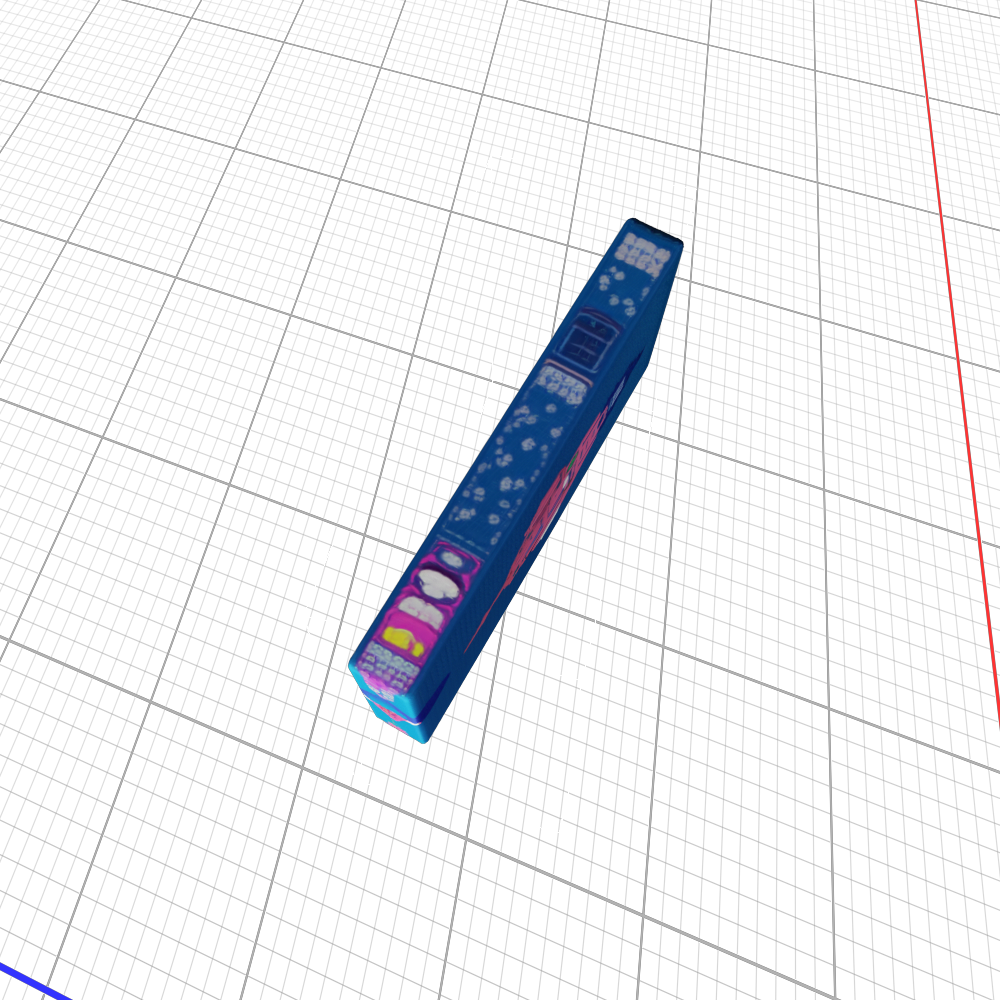} &
            \includegraphics[width=\figwidthb\textwidth]{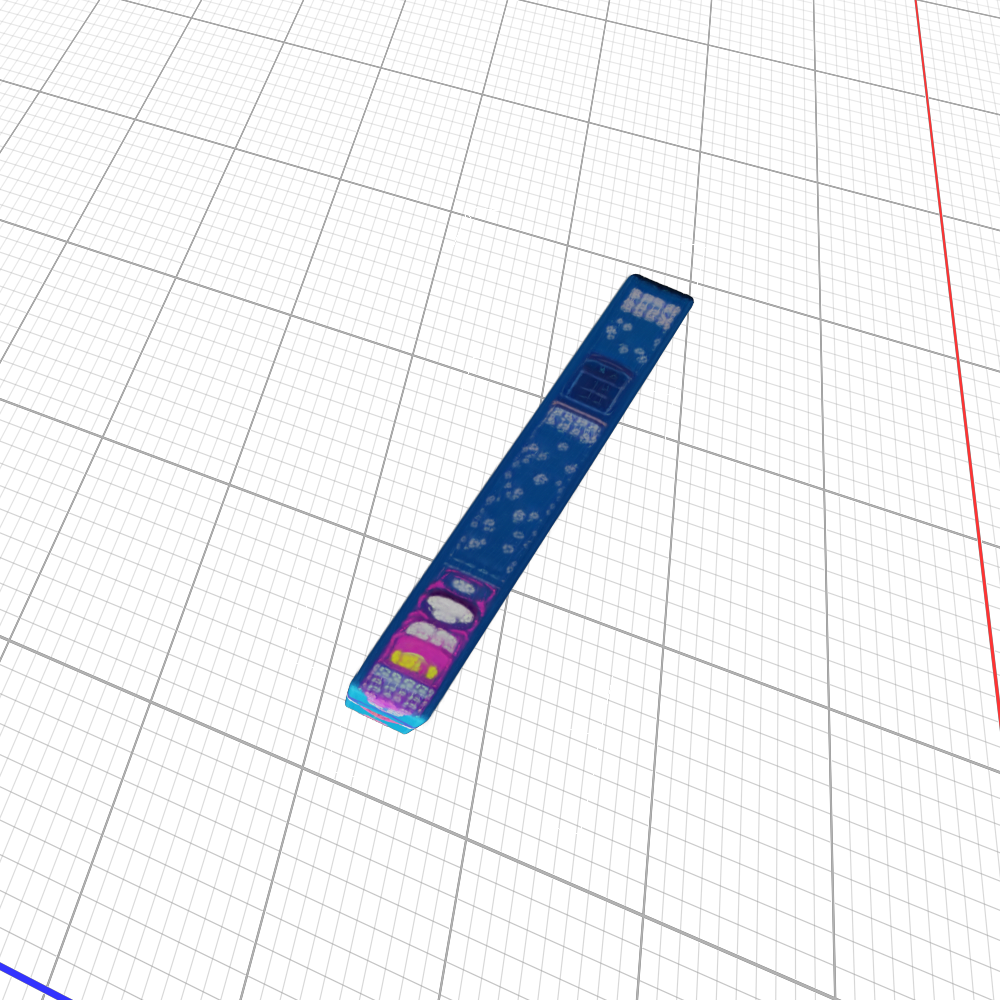} &
            \includegraphics[width=\figwidthb\textwidth]{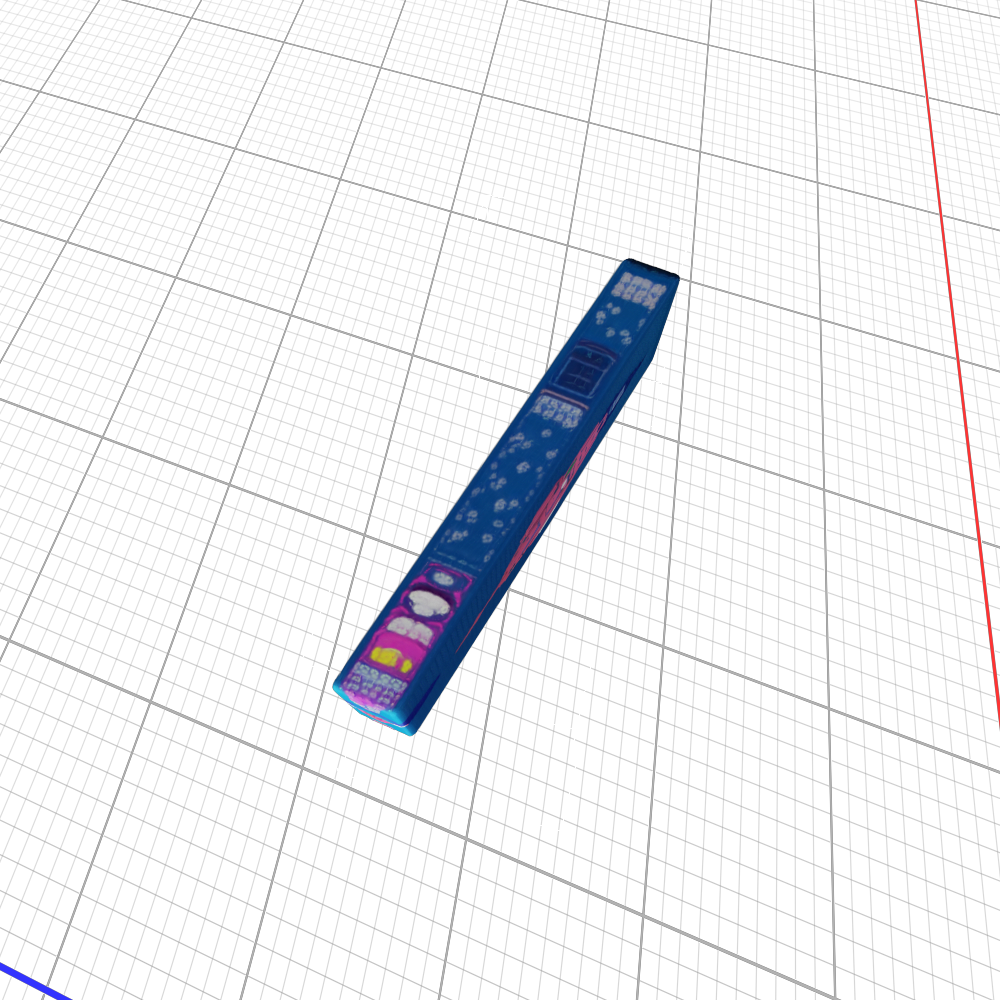} &
            \includegraphics[width=\figwidthb\textwidth]{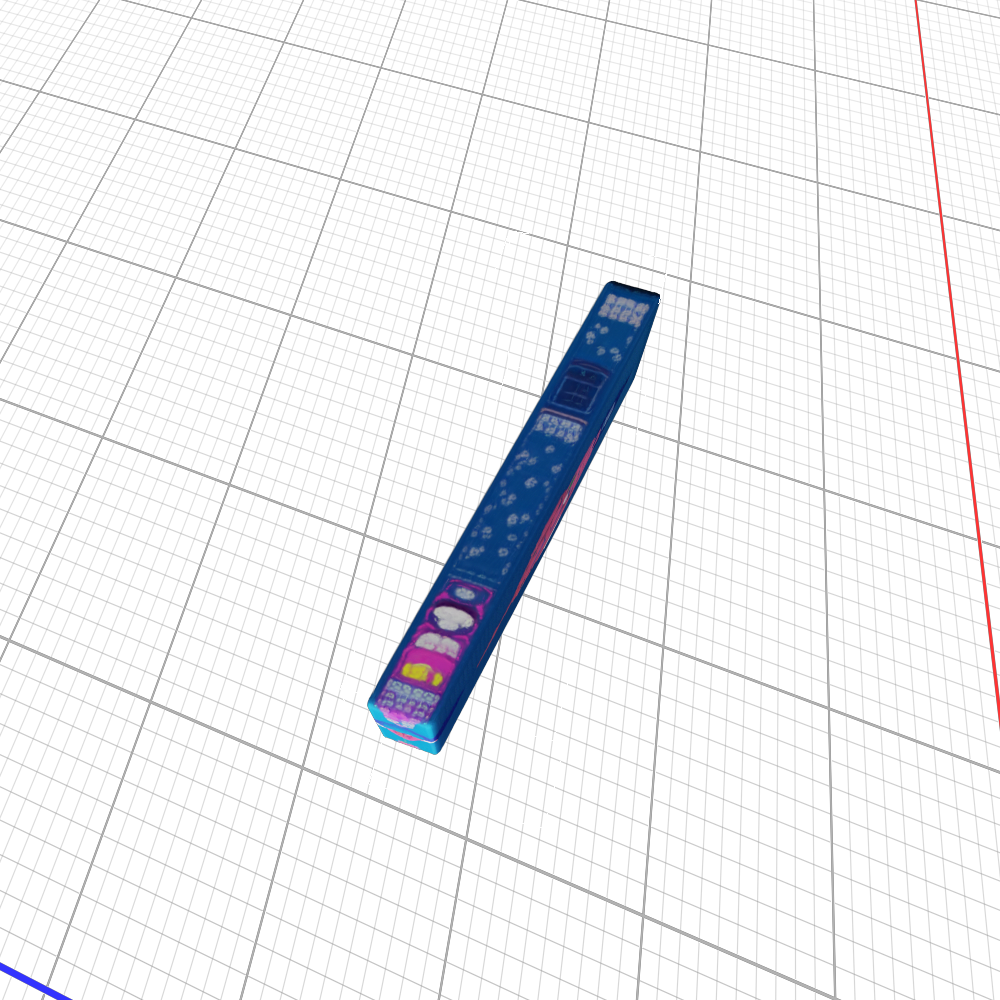} &
            \includegraphics[width=\figwidthb\textwidth]{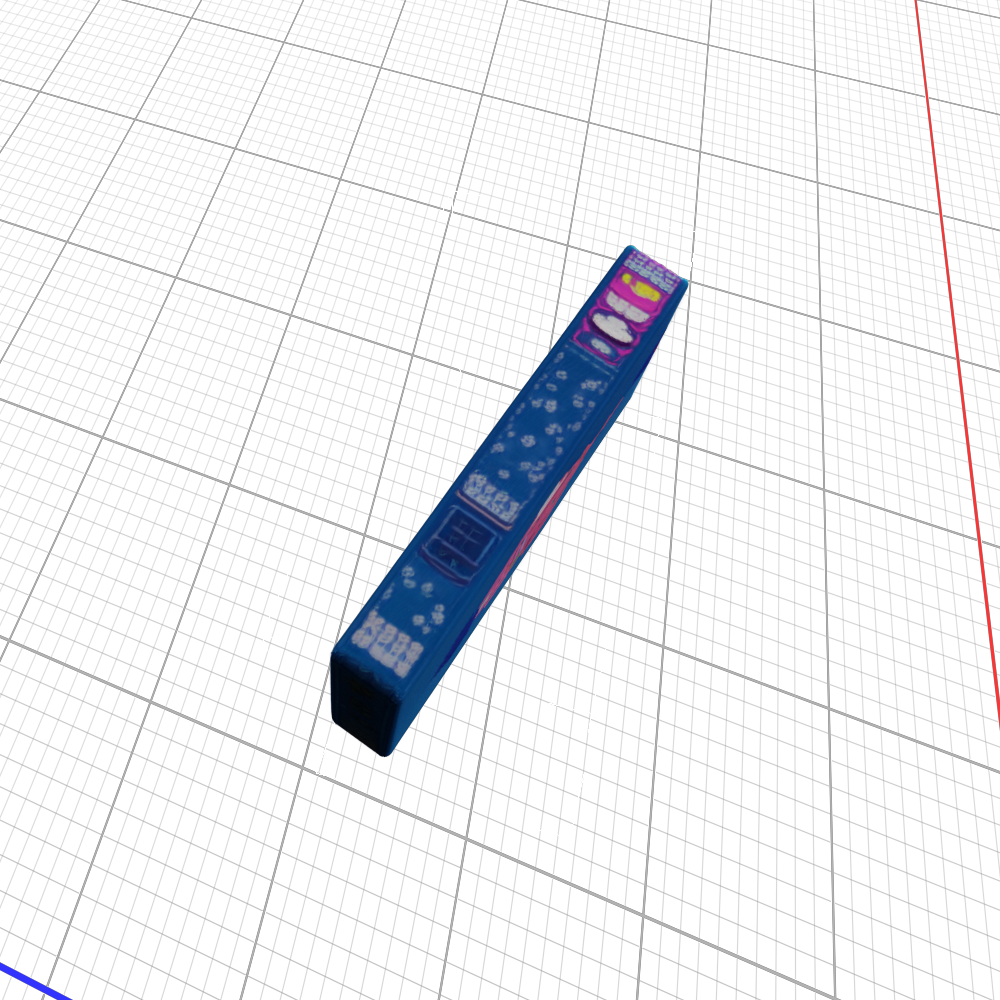} \\

            \includegraphics[width=\figwidthb\textwidth]{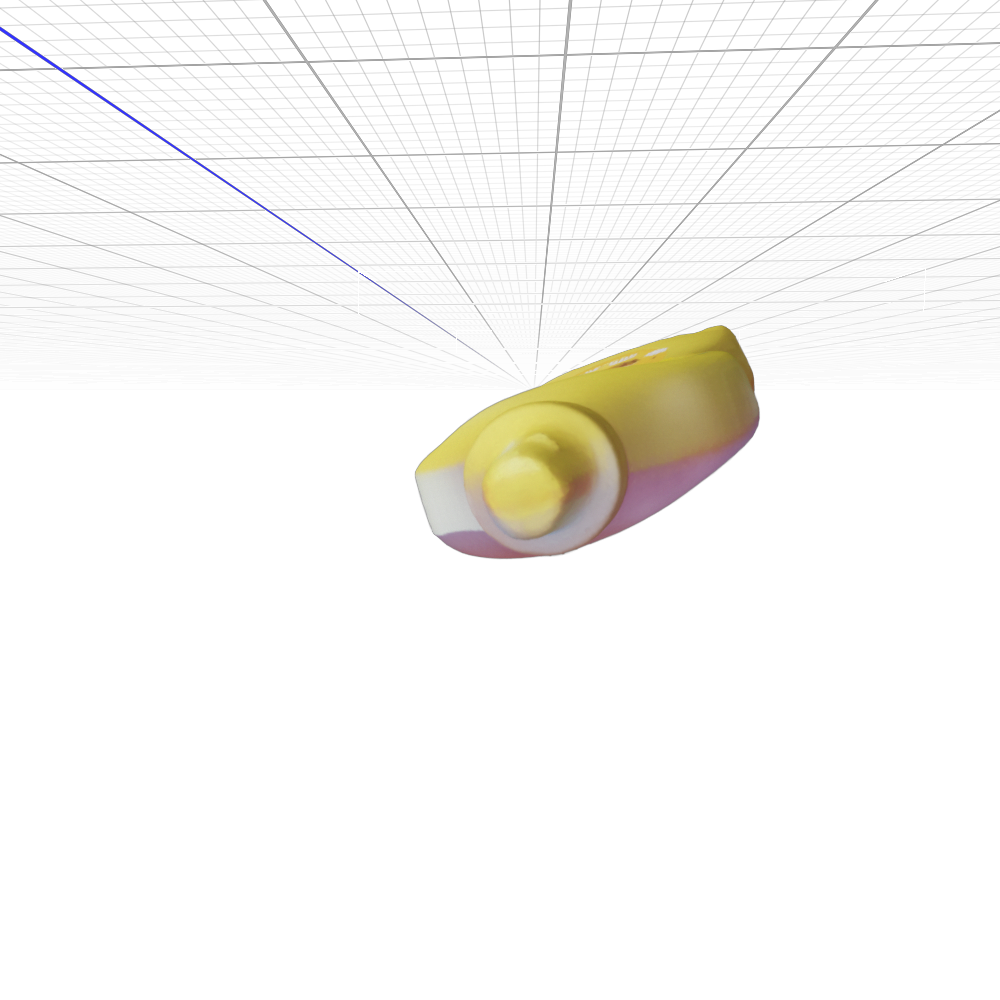} &
            \includegraphics[width=\figwidthb\textwidth]{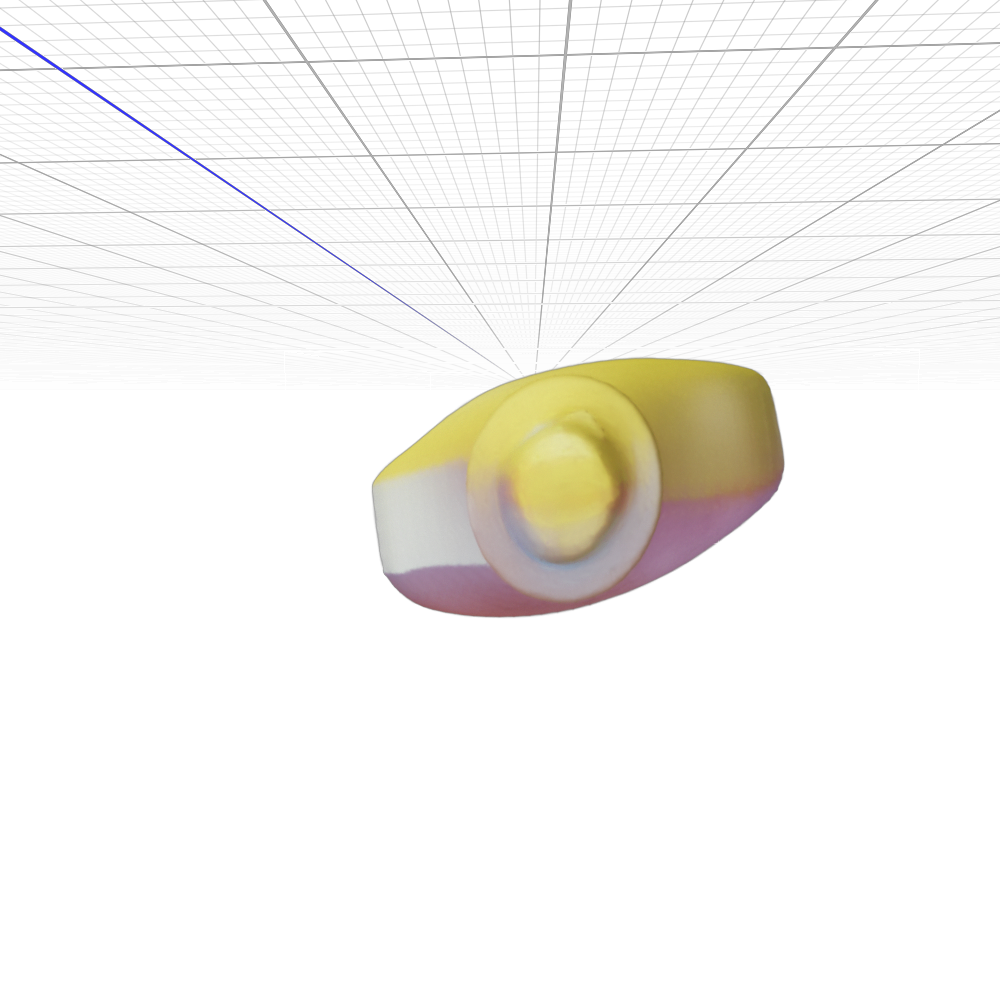} &
            \includegraphics[width=\figwidthb\textwidth]{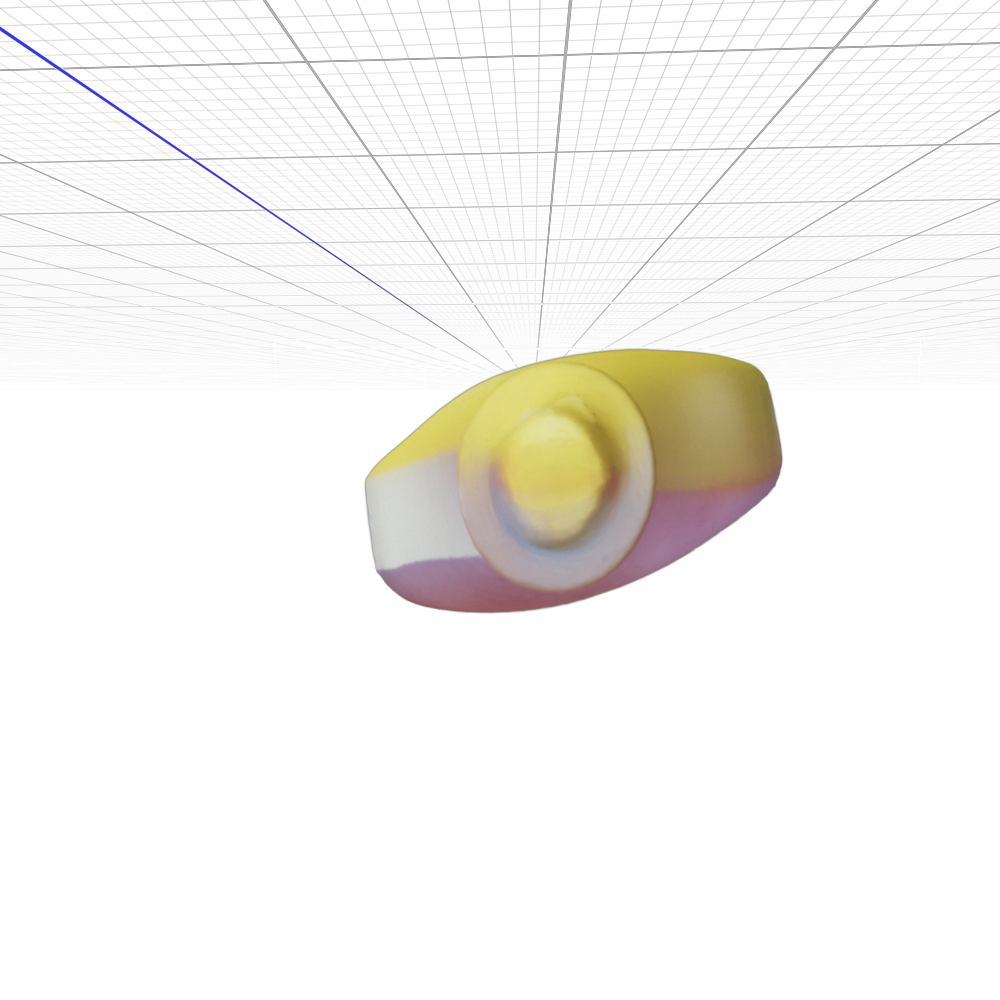} &
            \includegraphics[width=\figwidthb\textwidth]{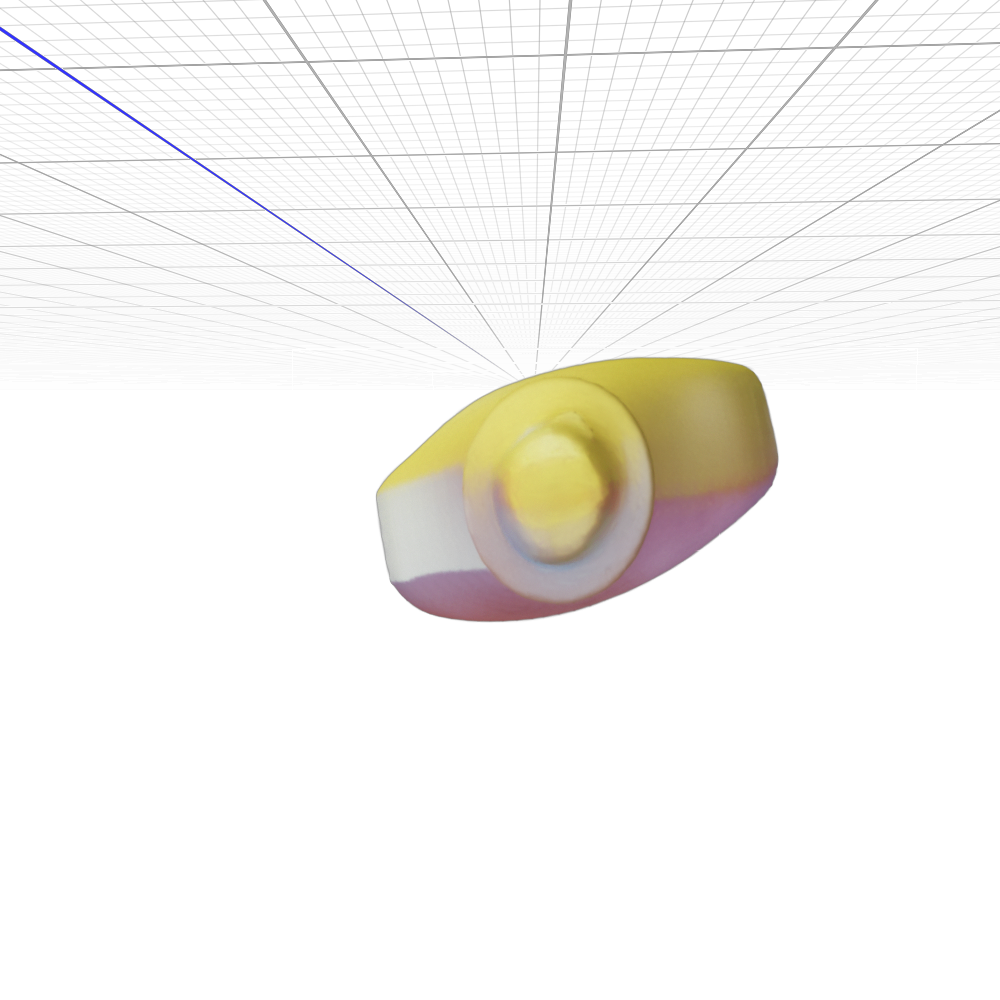} &
            \includegraphics[width=\figwidthb\textwidth]{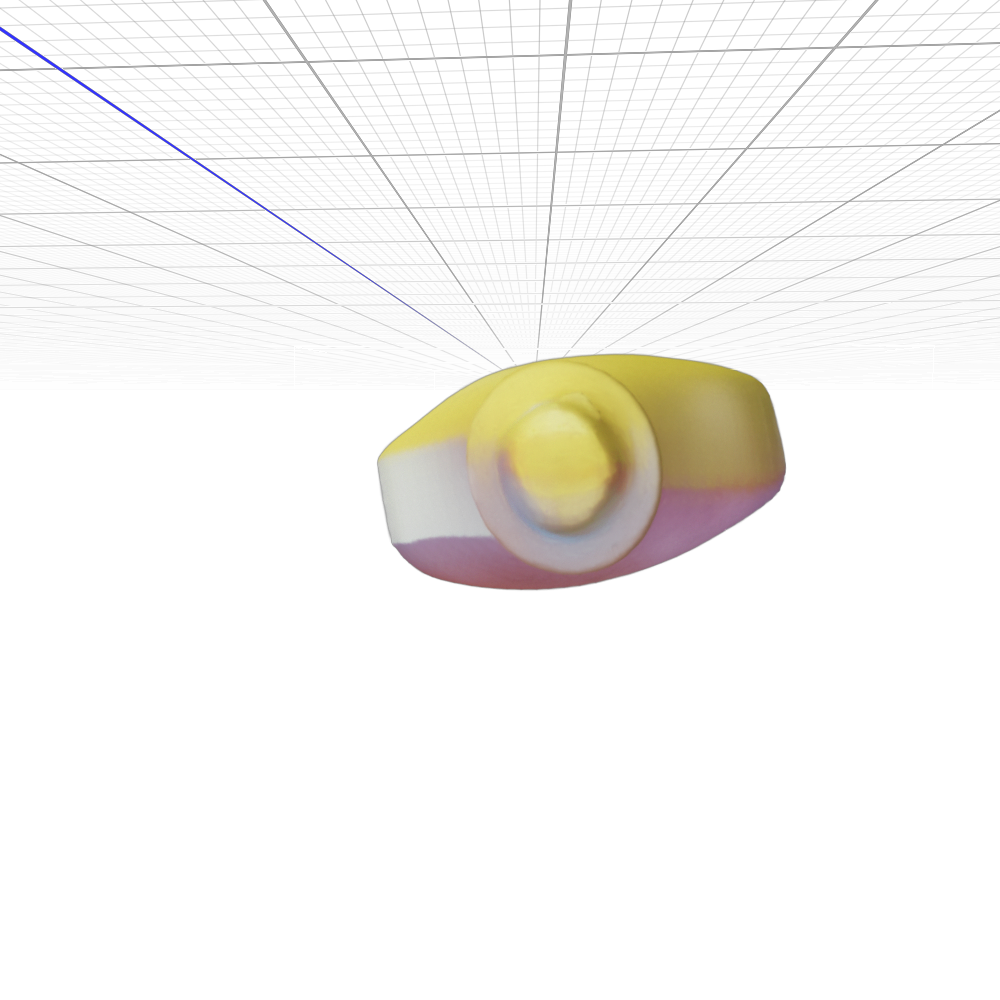} \\

            \includegraphics[width=\figwidthb\textwidth]{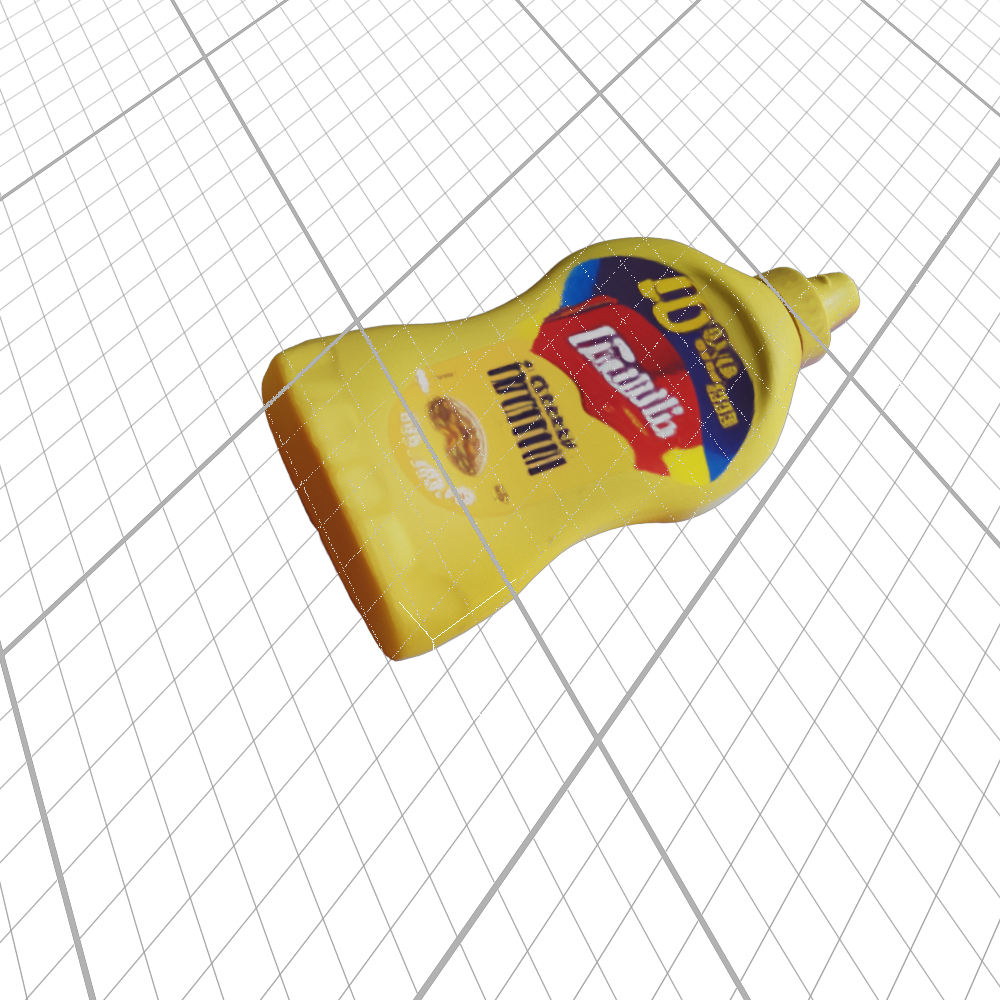} &
            \includegraphics[width=\figwidthb\textwidth]{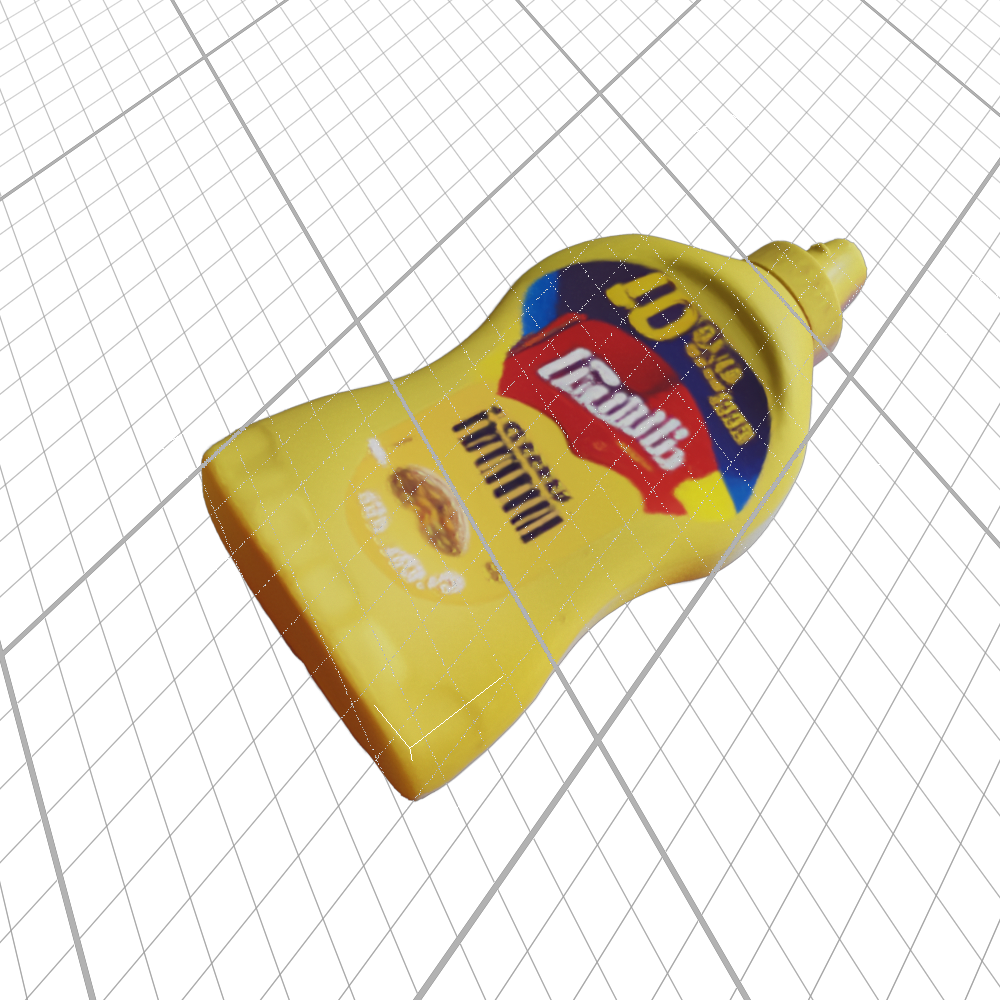} &
            \includegraphics[width=\figwidthb\textwidth]{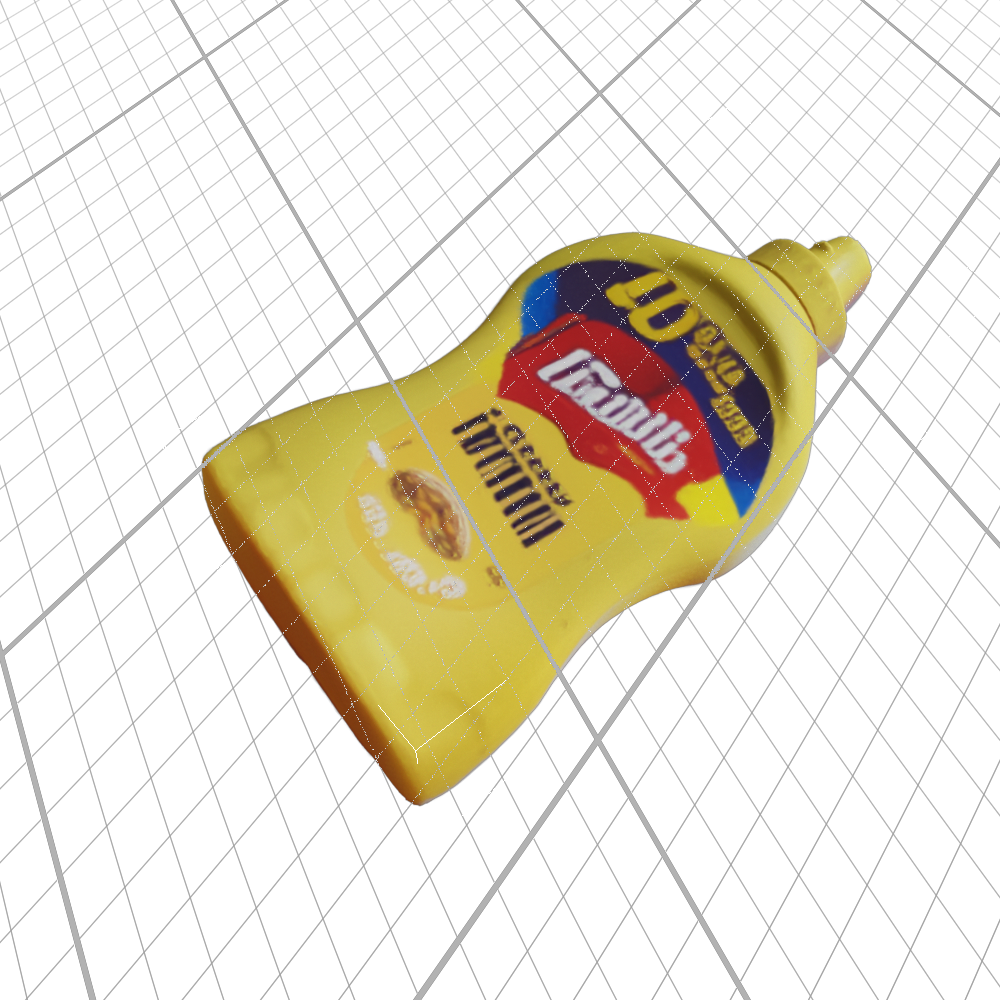} &
            \includegraphics[width=\figwidthb\textwidth]{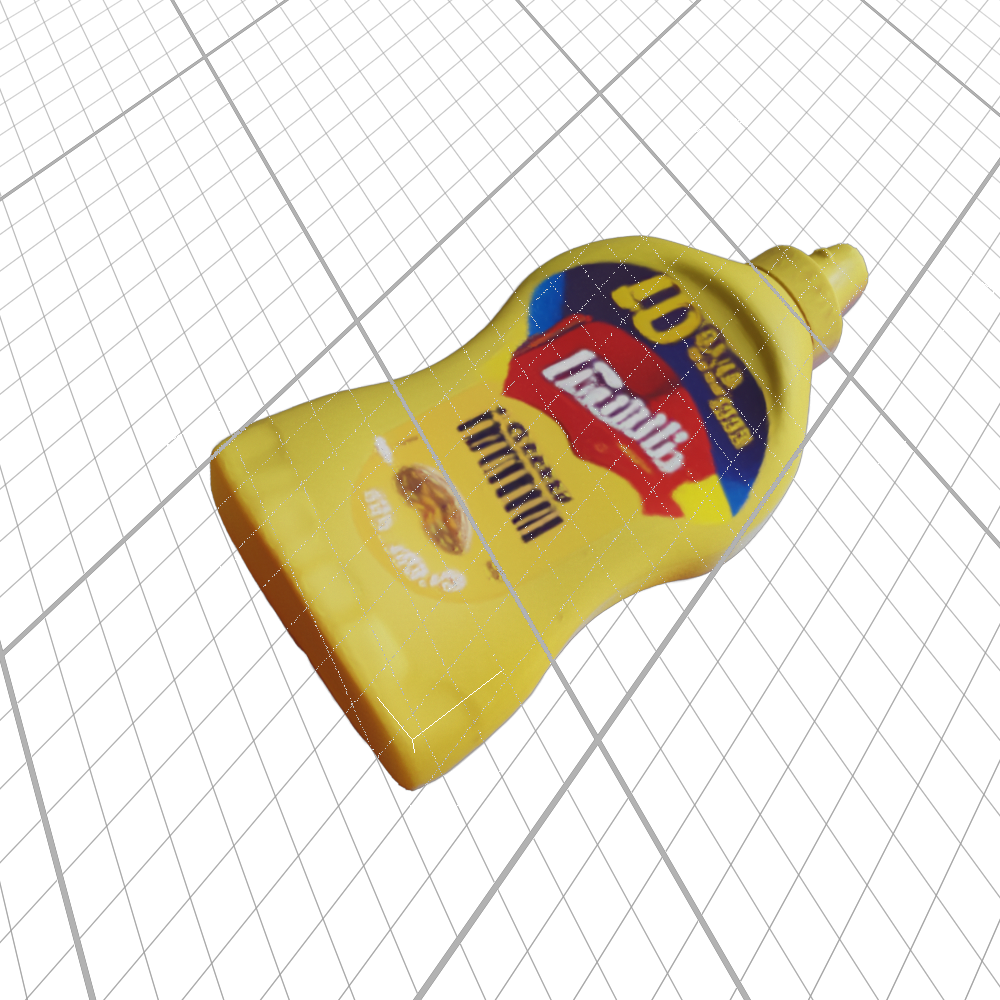} &
            \includegraphics[width=\figwidthb\textwidth]{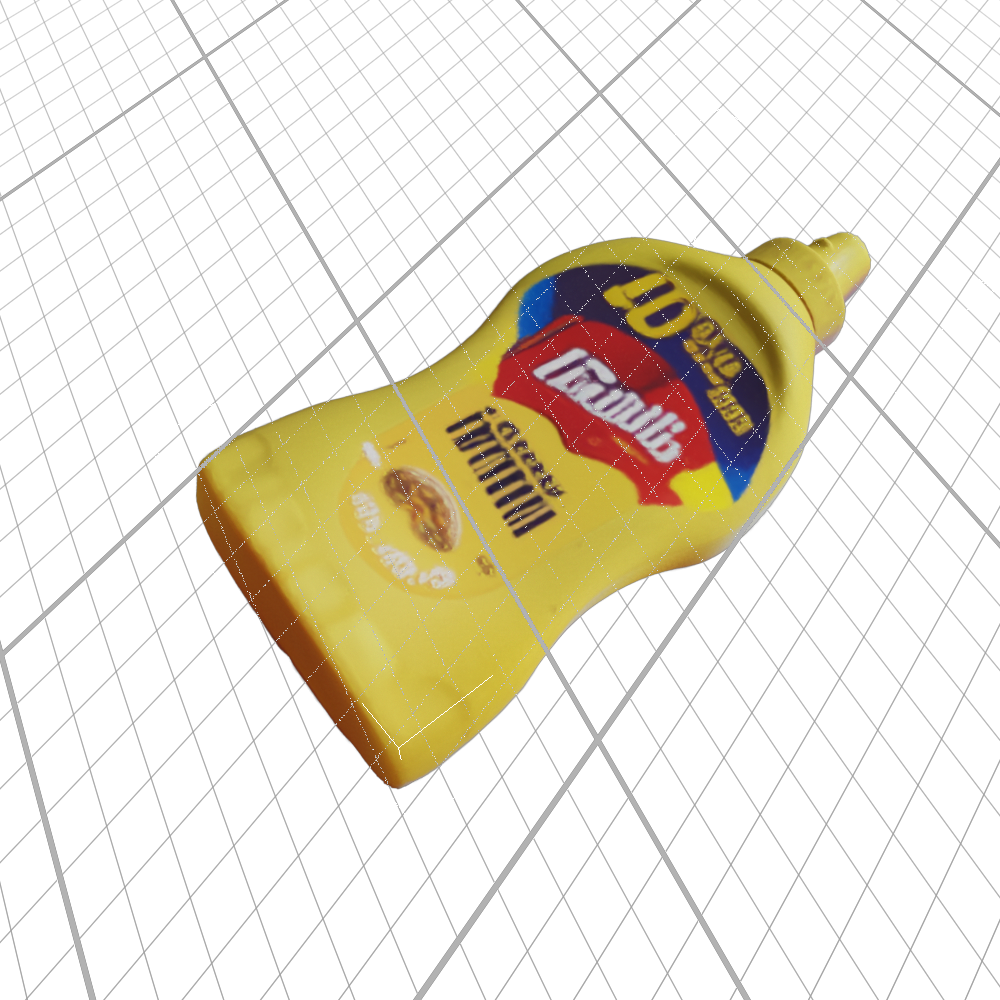} \\
            i & ii & iii & iv & v \\
        \end{tabular}
    }
    \vspace{-5pt}
    \caption{Qualitative ablations on Scale-Undistorted Shape Refinement for a real-world object ``wireless charger'', ``blue box'' from 3DGS-CD-\textit{Bench}, and ``yellow bottle'' from 3DGS-CD-\textit{Desk}.}
    \label{fig:ablation_qualitative_supply}
\end{figure*}

\subsection{Performance on Visible Viewpoints}

We conducted an additional experiment that evaluates the PSNR, SSIM, and LPIPS specifically on visible views, comparing our method against 3DGS~\cite{3dgs}. As shown in Tab.~R5, \ours achieves comparable performance in visible views due to the effective Appearance Refinement (AR) module. As some cases in the dataset contain complex geometry, such as the ``lego bulldozer'' with so complex detailed structures including ``buckets'', ``tracks'', ``connecting rod structures'', and ``cabs'' in Mip360-\textit{kitchen} scene that TRELLIS~\cite{trellis} could not fully capture, \ours may not always achieve the higher PSNR, resulting in a lower average PSNR. Moreover, mathematically, the same $\Delta\text{PSNR}$ could be hard to perceive when PSNR is high enough. Since our method achieves a similar performance on appearance to the baseline in regions that are actually observed during training, we believe that our method still faithfully reconstructs the geometry and appearance of known regions in most cases.
\begin{table}[!htb]
  \caption{
    Quantitative comparison of appearance on visible views against 3DGS~\cite{3dgs} among different settings. Our method achieves comparable performance to 3DGS on seen views
  }
  \label{tab:seen_view}
  \vspace{-1em}
  \centering
  \resizebox{0.6\linewidth}{!}{
  \begin{tabular}{c|c|cccc}
    \toprule
    & & \nicefrac{1}{3} & \nicefrac{1}{5} & \nicefrac{1}{7} & Avg. \\
    \midrule
    \multirow{2}{*}{{PSNR$\uparrow$}} 
    & 3DGS & 33.938 & 33.907 & 34.834 & 34.226 \\
    & Ours & 33.114 & 33.401 & 33.630 & 33.381 \\
    \midrule
    \multirow{2}{*}{{SSIM$\uparrow$}} 
    & 3DGS & 0.9877 & 0.9883 & 0.9896 & 0.9885 \\
    & Ours & 0.9833 & 0.9855 & 0.9855 & 0.9848 \\
    \midrule
    \multirow{2}{*}{{LPIPS$\downarrow$}} 
    & 3DGS & 0.0197 & 0.0185 & 0.0170 & 0.0184 \\
    & Ours & 0.9833 & 0.9855 & 0.9855 & 0.9848 \\
    \bottomrule
  \end{tabular}
  }
  \vspace{-1em}
\end{table}

\newcommand{\graspheight}{0.5in}
\begin{figure*}[!ht]
\vspace{-1em}
\footnotesize
\centering
\setlength{\tabcolsep}{3pt}
\begin{tabular}{cccc}
\includegraphics[height=\graspheight]{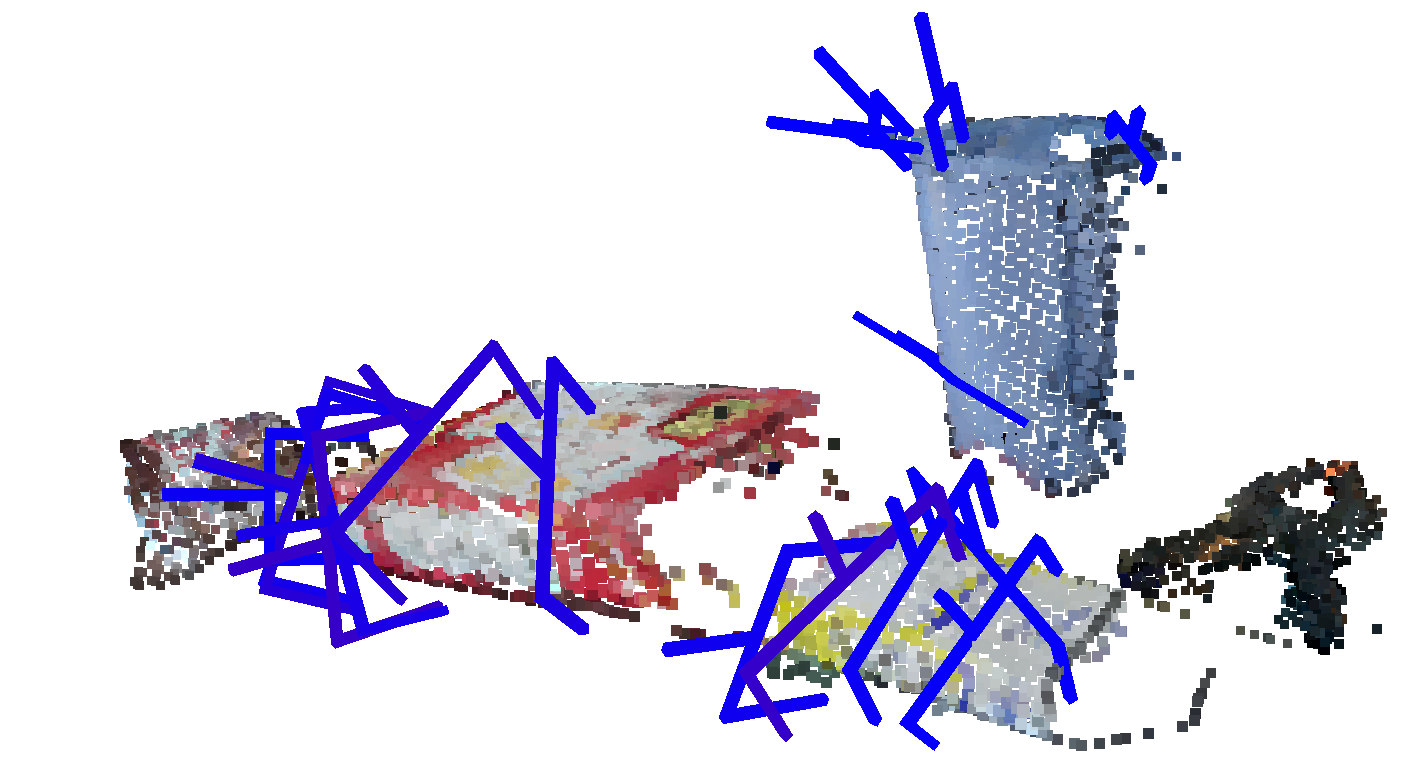} &
\includegraphics[height=\graspheight]{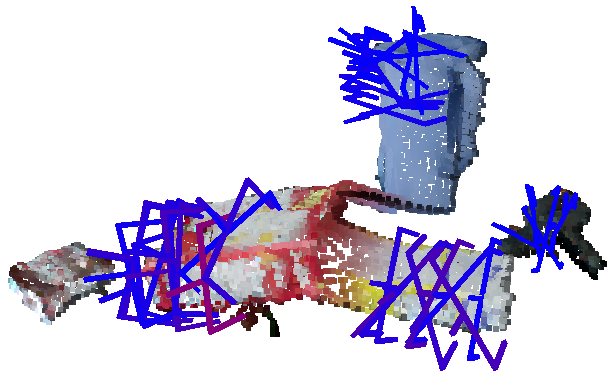} & 
\includegraphics[height=\graspheight]{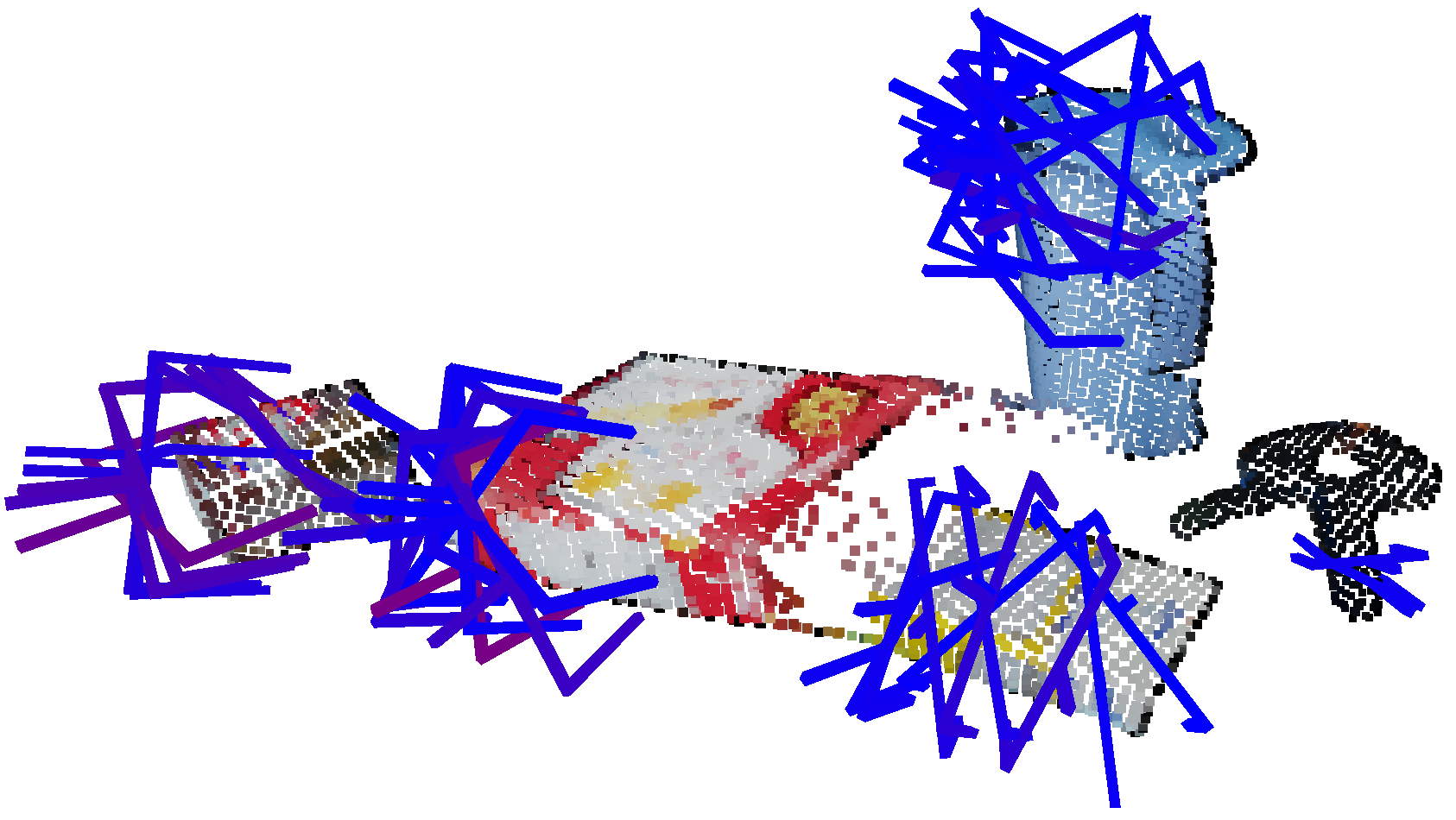} &
\includegraphics[height=\graspheight]{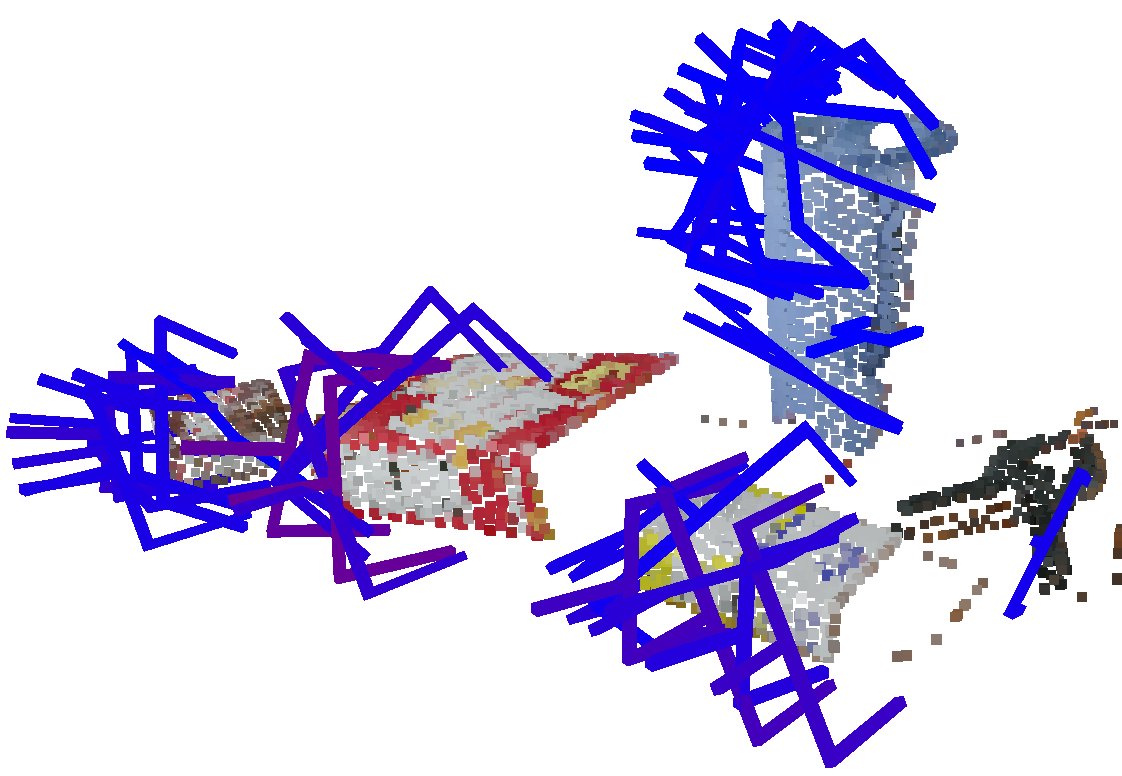}  \\
\includegraphics[height=\graspheight]{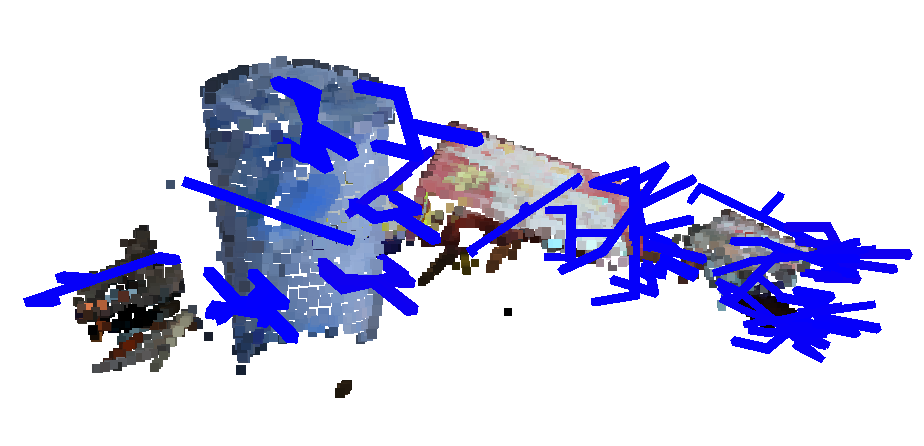} &
\includegraphics[height=\graspheight]{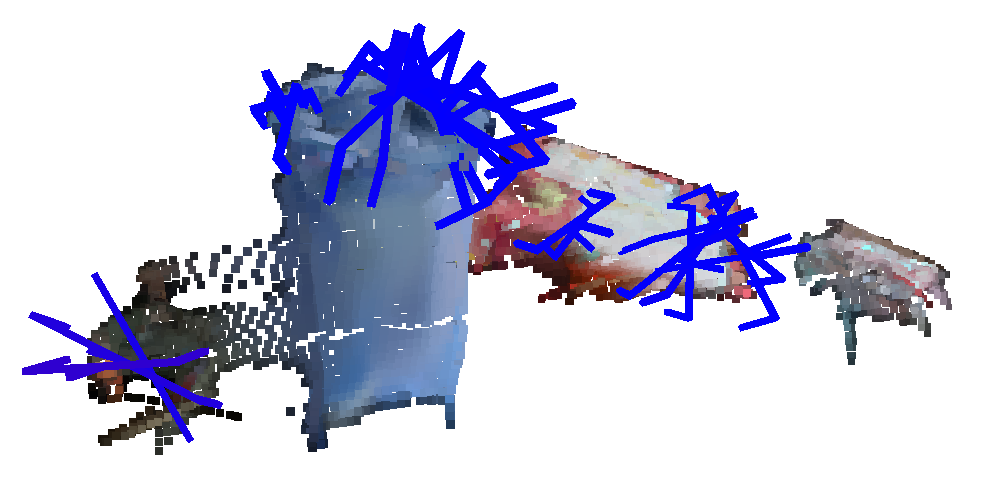} & 
\includegraphics[height=\graspheight]{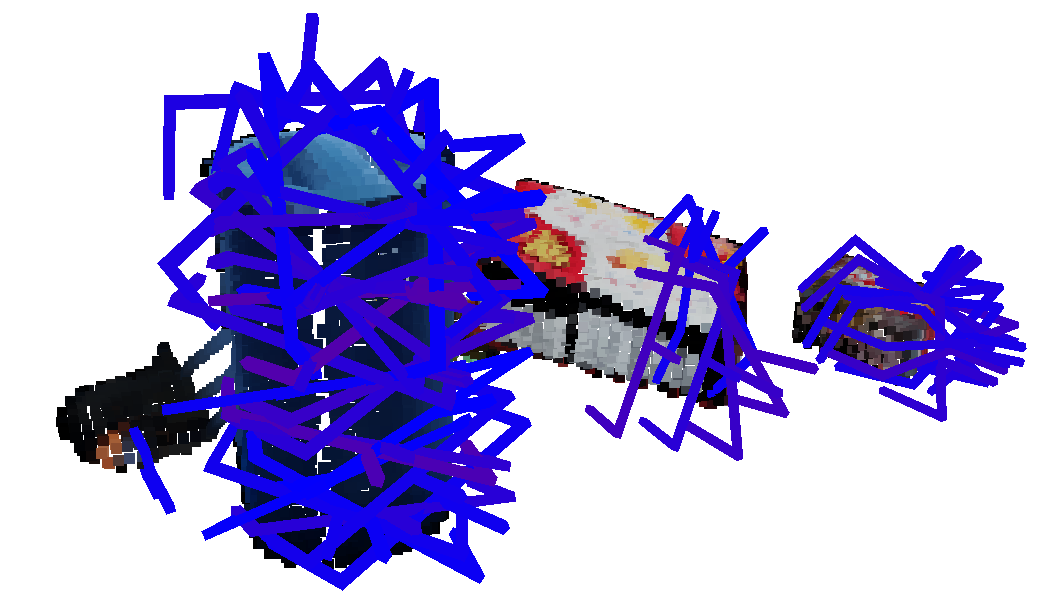} &
\includegraphics[height=\graspheight]{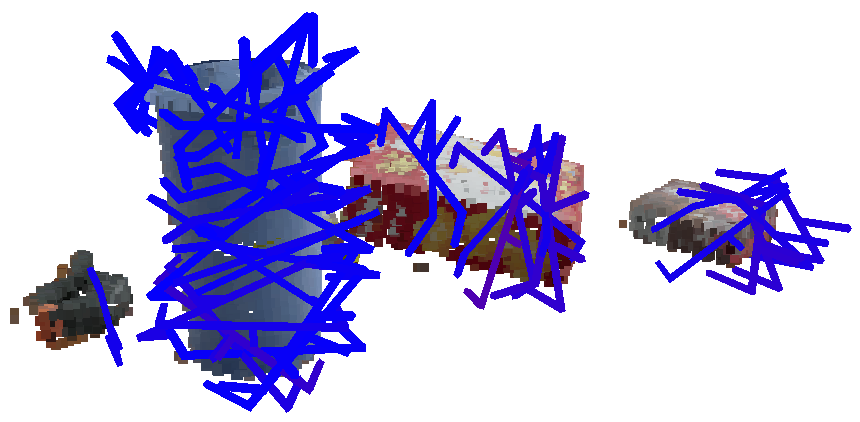}  \\
\includegraphics[height=\graspheight]{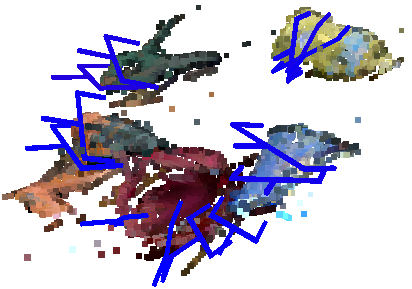} &
\includegraphics[height=\graspheight]{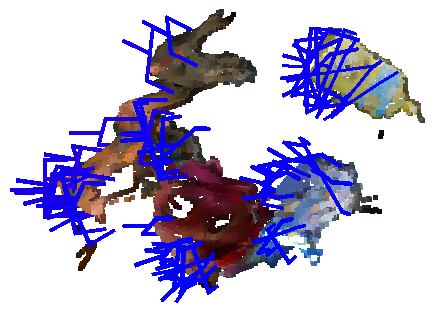} & 
\includegraphics[height=\graspheight]{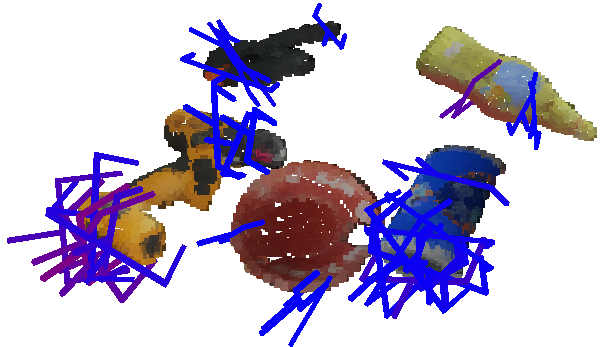} &
\includegraphics[height=\graspheight]{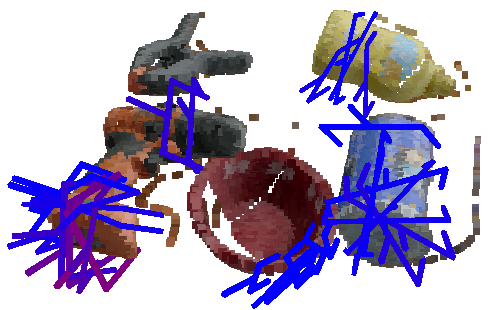}  \\
\includegraphics[height=\graspheight]{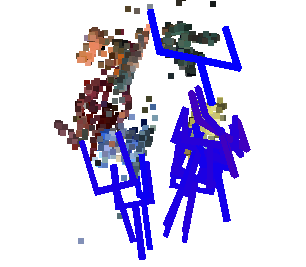} &
\includegraphics[height=\graspheight]{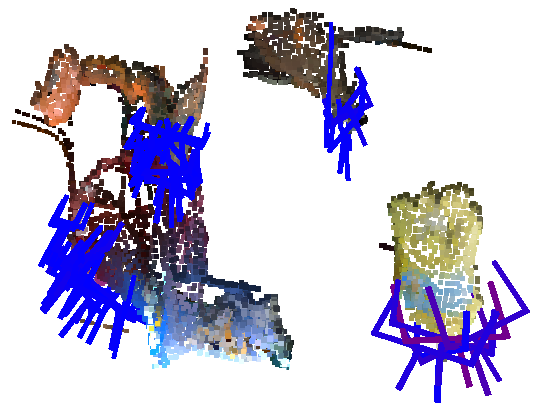} & 
\includegraphics[height=\graspheight]{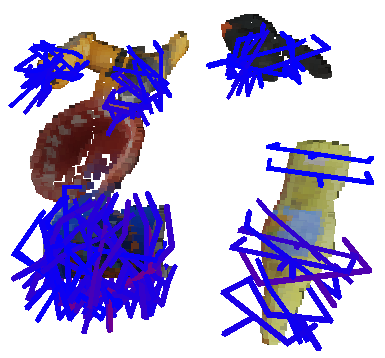} &
\includegraphics[height=\graspheight]{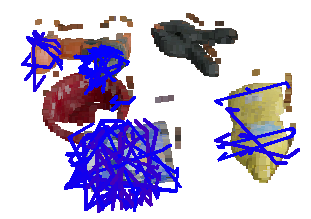}  \\
\includegraphics[height=\graspheight]{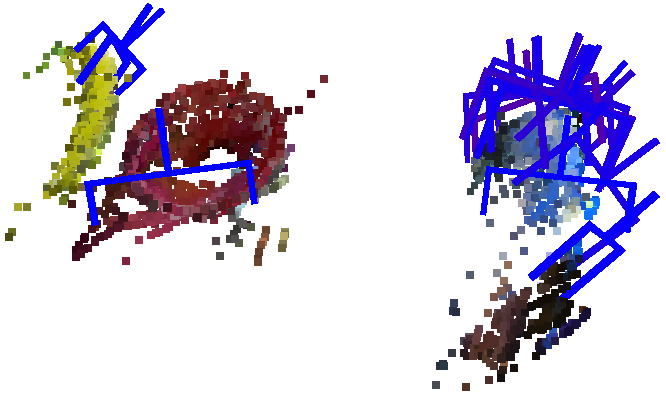} &
\includegraphics[height=\graspheight]{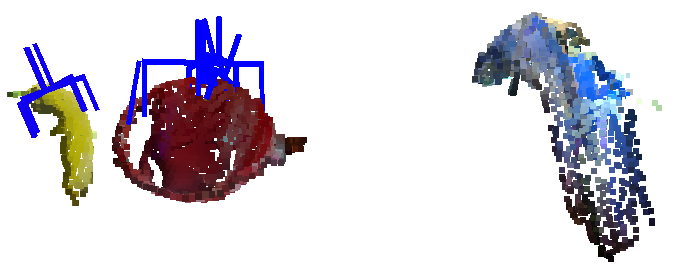} & 
\includegraphics[height=\graspheight]{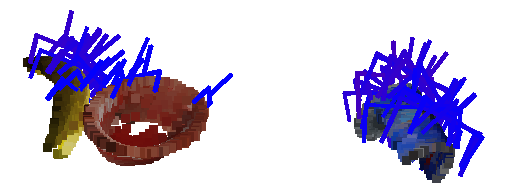} &
\includegraphics[height=\graspheight]{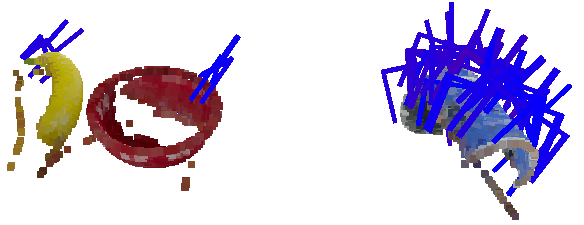}  \\
2DGS &  3DGS &  Ours & GT Depth\\
\end{tabular}
\caption{More AnyGrasp \cite{anygrasp} results on the rendering depth of test view.}
\label{fig:grasp_all}
\vspace{-1em}
\end{figure*}

\subsection{Robot Grasping}  
We further validate the application of our method in the robotic grasping task. In real-world scenarios, missing observations from specific viewpoints often lead to degraded reconstruction quality and reduced grasping performance under those views. As shown in Fig.~\ref{fig:grasp_all}, compared to the fragmented objects from baseline, rendered RGB-D of objects refined by \ours enable the grasping model~\cite{anygrasp} to generate denser and more plausible grasp candidates. Results highlight our method's sim-to-real potential application with inferred high-quality grasps on limited views.

\subsection{Efficiency Analysis}
\label{app:efficiency}
We report wall-clock time and peak GPU memory for each stage of \ours on a single Tesla V100 GPU with 32 GB on the high-resolution (3115x2078) kitchen scene with a single target object. 
The results are summarized in Tab.~\ref{tab:efficiency}, which provides a breakdown of VRAM usage and time cost for each component, including 3DGS Scene Initialization (Scene Init.), 3D Segmentation (3D Seg.), Object Synthesis (OS), Coarse Alignment (CA), Pose Adjustment (PA), Scale Refinement (SR), and Appearance Refinement (AR). There are 4 iterations in PA and 2 iterations in SR, which could be adjusted as needed on custom scenes. The results indicate that the peak VRAM consumption in our pipeline consistently arises during the OS stage.

\begin{table}[!htb]
  \caption{
    Time cost and peak VRAM usage statistics on Mip360-\textit{kitchen} among different difficulty settings, which are shown in ``Time [min]$\downarrow$ / VRAM [MB]$\downarrow$'' columns respectively. 
    Since CA uses none of the VRAM, and PA obviously uses less VRAM than SR, we do not report the VRAM usage of CA and PA.
  }
  \label{tab:efficiency}
  \vspace{-1em}
  \centering
  \resizebox{\linewidth}{!}{
  \begin{tabular}{c|ccccccc}
    \toprule
     & 3DGS Init. & 3D Seg. & OS & CA & PA & SR & AR \\
    \midrule
    \nicefrac{1}{3} & 8.55 / 1344.2 & 0.58 / 827.7 & 1.62 / 11048.2 & 1.65 / - & 3.47 / - & 2.93 / 6061.3 & 1.57 / 2120.7 \\
    \nicefrac{1}{5} & 7.40 / 1310.3 & 0.37 / 823.5 & 1.60 / 11203.6 & 1.73 / - & 3.58 / - & 2.88 / 5726.9 & 1.55 / 2149.7 \\
    \nicefrac{1}{7} & 7.22 / 1302.5 & 0.20 / 813.0 & 1.08 / 11098.4 & 1.77 / - & 3.83 / - & 2.47 / 5556.8 & 1.25 / 2130.0 \\
    \bottomrule
  \end{tabular}
  }
  \vspace{-1em}
\end{table}

Additionally, we compare both the computation time and the memory usage between our method and a baseline, GenFusion~\cite{genfusion}, in Tab.~\ref{tab:efficiency_compare}. This comparison highlights the practical efficiency of our approach and provides a more comprehensive understanding of its performance characteristics. During our experiments, we tried to maintain the default settings of the baseline method and fixed the number of Gaussian training iterations to 10k across methods to ensure a fair comparison. 
It is worth noting that GenFusion resizes all images used in the diffusion model to 960x512, which is consistent with its default settings. Besides, we noticed that our method's time cost on 3DGS-CD-\textit{Bench} is slightly higher than Mip360-\textit{kitchen} since there are 3 target objects in the \textit{Bench} scene, which triple (3x) the time cost of CA, PA, and SR in our method compared to the single object in the same scene. In the high-resolution \textit{kitchen} scene, our method achieves significantly lower VRAM usage and competitive computation time cost compared to GenFusion.

\begin{table}[!htb]
  \caption{
    Quantitative comparison of time cost and peak VRAM usage between our method and GenFusion~\cite{genfusion}, on 3DGS-CD-\textit{Bench} (1006x753) and Mip360-\textit{kitchen} (3115x2078). Consistently, our method achieves a lower VRAM usage and a better computation time cost. The better score is highlighted in \textbf{bold}.
  }
  \label{tab:efficiency_compare}
  \vspace{-1em}
  \centering
  \resizebox{\linewidth}{!}{
  \begin{tabular}{c|c|cccccc}
    \toprule
    &&  \textit{Bench}-\nicefrac{1}{3} & \textit{Bench}-\nicefrac{1}{5} & \textit{Bench}-\nicefrac{1}{7} & \textit{kitchen}-\nicefrac{1}{3} & \textit{kitchen}-\nicefrac{1}{5} & \textit{kitchen}-\nicefrac{1}{7} \\
    \midrule
    \#images & \diagbox{}{} & 52 & 32 & 26 & 93 & 59 & 45 \\
    \midrule
    \multirow{2}{*}{{Time [min]$\downarrow$}} 
    & Gen. & 51.98 & 49.45 & 48.70 & 82.68 & 78.18 & 75.70 \\
    & Ours & \textbf{22.45} & \textbf{20.67} & \textbf{22.27} & \textbf{20.45} & \textbf{19.20} & \textbf{18.03} \\
    \midrule
    \multirow{2}{*}{{VRAM [GB]$\downarrow$}} 
    & Gen. & 20.70 & 20.60 & 20.55 & 20.58 & 20.56 & 20.56 \\
    & Ours & \textbf{9.76} & \textbf{9.59} & \textbf{9.79}  & \textbf{10.79} & \textbf{10.94} & \textbf{10.84} \\
    \bottomrule
  \end{tabular}
  }
  \vspace{-1em}
\end{table}
\section{Dataset Details}
\label{sec:dataset_details}

\subsection{Appearance}
\label{sec:dataset_details_app}

\begin{figure}[!htb]
    \vspace{-1em}
    \centering
    \footnotesize{
        \setlength{\tabcolsep}{1pt} 
        \begin{tabular}{cc}
            \includegraphics[width=0.49\linewidth]{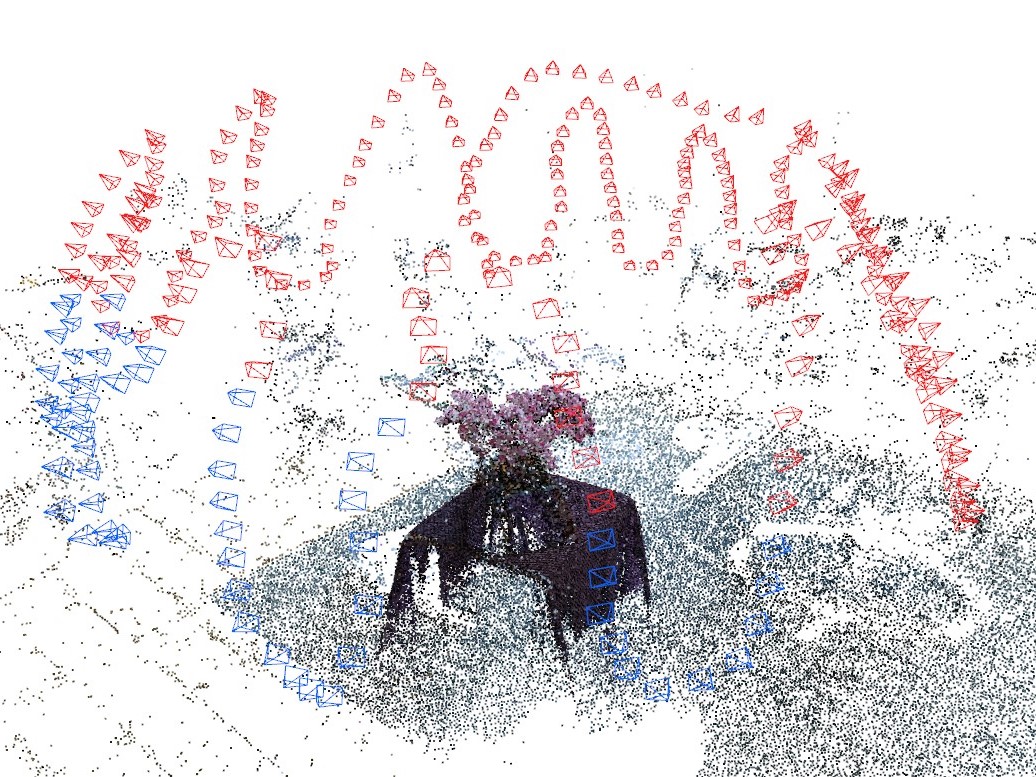} &
            \includegraphics[width=0.49\linewidth]{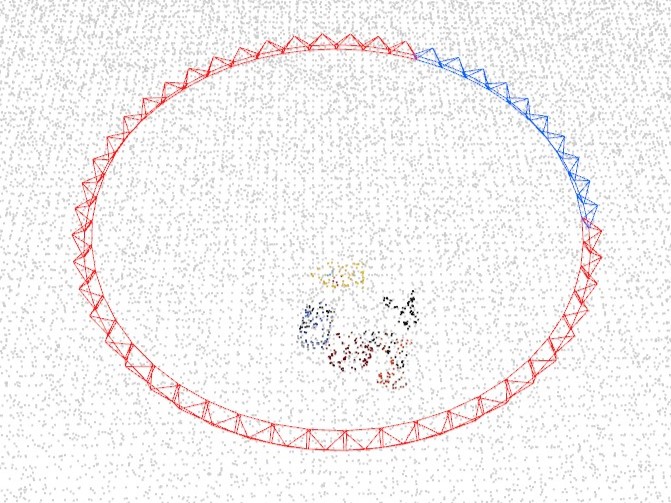} \\
            \textit{kitchen} & \textit{scene 3}\\
        \end{tabular}
    }
    \vspace{-1em}
    \caption{
    Visualization of the split of seen views and unseen views for evaluation, where the views with red cameras are unseen views and the views with blue cameras are seen views. The point cloud is for 3DGS training initialization. The left is from Mip360-\textit{bonsai} and the right is from \textit{scene 3} of constructed scenarios.
    }
    \label{fig:supply_view}
    \vspace{-1em}
\end{figure}

To evaluate the appearance quality of \ours, we selected scenes from the NVS task datasets, Mip360~\cite{mip360} and LERF~\cite{lerf}. Due to the specific requirements of the task, the chosen scenes were required to include images captured around significant objects to serve as ground truth, which posed challenges in dataset selection. Therefore, we also extracted some scenes from other task datasets, ToyDesk~\cite{toydesk} and 3DGS-CD \cite{gs-cd}. Ultimately, we obtained 11 scenes with images captured around objects, providing us with sufficient ground truth. Specifically, we selected the \textit{bonsai}, \textit{garden}, and \textit{kitchen} scenes from Mip360~\cite{mip360}, the \textit{donuts}, \textit{figurines}, \textit{show\_rack}, and \textit{teatime} scenes from LERF~\cite{lerf}, the \textit{scene1} and \textit{scene2} scenes from ToyDesk~\cite{toydesk}, and the \textit{Bench} and \textit{Desk} scenes from 3DGS-CD~\cite{gs-cd}.
For evaluation, we randomly select a start view and remove the $n$ nearest views in joint position–orientation space as unseen views (\ie, test views). Fig.~\ref{fig:supply_view} illustrates the resulting view splits on a real scenario and on a constructed scenario.

\subsection{Geometry}
To facilitate geometric evaluation in physically plausible and visually realistic environments, we construct a synthetic dataset tailored to multi-object 3D reconstruction tasks. Preliminary investigations reveal that existing datasets widely used in appearance-focused tasks, such as Mip360 and LERF, are unsuitable due to the lack of precise ground truth geometry. Likewise, datasets that focus solely on single isolated objects, such as DTU~\cite{jensen2014large}, BlenderMVS~\cite{yao2020blendedmvs}, and OmniObject~\cite{wu2023omniobject3d}, fail to capture the complex inter-object occlusions and collisions typical in real-world scenes. Although scene-level RGB-D datasets such as SUN RGB-D~\cite{sun}, and Replica~\cite{replica} offer realistic sensor data, they lack accurate and complete 3D ground truth and are thus excluded. Similarly, point cloud completion datasets like Redwood~\cite{redwood} are omitted due to the absence of accompanying RGB imagery, which is essential for our reconstruction objectives.

Due to these limitations, we build upon the YCB-Video~\cite{PoseCNN} dataset, which provides high-quality object meshes, to generate a set of synthetic scenes with controlled geometric and visual properties. For each scene, we first generate a horizontal base plane equipped with a collision volume to serve as the physical support. Multiple object meshes are then randomly instantiated within a bounded volume above the plane and assigned individual collision bodies. A rigid-body physics simulation is performed, allowing the objects to fall under gravity and interact naturally until a stable configuration is reached. After convergence, we simulate a camera moving along a circular trajectory around the scene center. From discrete camera poses, we render corresponding RGB images and depth maps, and record the associated camera extrinsics along with the complete ground truth scene mesh. This procedure ensures consistent, high-fidelity data suitable for quantitative geometric analysis under realistic multi-object occlusion and physical interaction.

\end{document}